%% file: main.tex
\documentclass[10pt,twocolumn,letterpaper]{article}

\usepackage[pagenumbers]{cvpr}

\input{macros}

\title{MyVLM: Personalizing VLMs for User-Specific Queries} 

\begin{document}

\author{Yuval Alaluf$^{*,1,2}$ \hspace{5mm}  Elad Richardson$^{2}$ \hspace{5mm}  Sergey Tulyakov$^{1}$ \hspace{5mm} Kfir Aberman$^{1}$ \hspace{5mm}  Daniel Cohen-Or$^{1,2}$ \\[6pt]
$^1$Snap Inc. \hspace{10mm} $^2$Tel Aviv University
\\[-12.5pt]
}

\twocolumn[{%
\vspace{-0.45cm}
\maketitle
\renewcommand\twocolumn[1][]{#1}%
\vspace{-0.175in}
\begin{center}
    \centering
    \includegraphics[width=\textwidth]{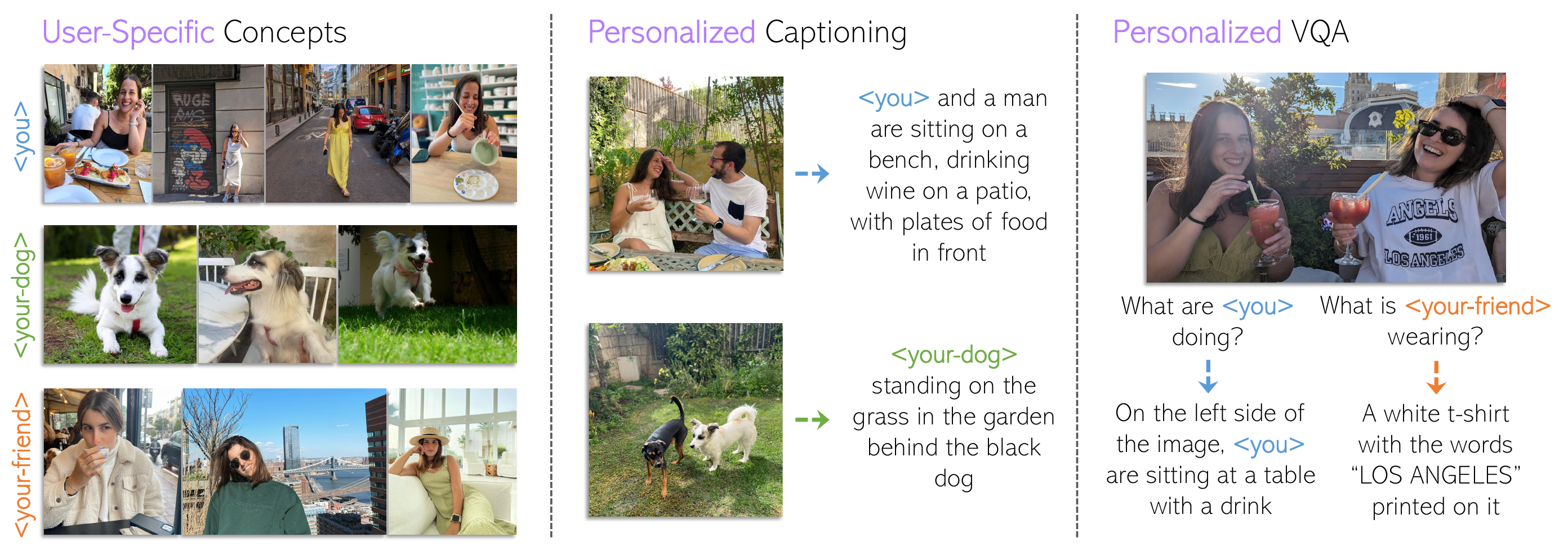}
    \vspace{-0.7cm}
    \captionof{figure}{
    Given a set of images depicting user-specific concepts such as \textcolor{blue}{$\langle$you$\rangle$}, \textcolor{darkgreen}{$\langle$your-dog$\rangle$} 
    and \textcolor{orange}{$\langle$your-friend$\rangle$}  (left), we teach a pretrained vision-language model (VLM) to understand and reason over these concepts. First, we enable the model to generate personalized captions incorporating the concept into its output text (middle). We further allow the user to ask subject-specific questions about these concepts, querying the model with questions such as ``What are \textcolor{blue}{$\langle$you$\rangle$} doing?'' or ``What is my \textcolor{orange}{$\langle$your-friend$\rangle$} wearing?'' (right). 
    }
    \label{fig:teaser}
\end{center}%
}]

\blfootnote{$^*$This research was performed while Yuval Alaluf was at Snap.}

\input{0-abstract}
\input{1-intro}
\input{2-related}

\input{3-method}
\input{4-results}

\input{5-conclusion}

\section*{Acknowledgements}
We would like to thank Assaf Ben-Kish, Or Patashnik, Moran Yanuka, Morris Alper, Yonatan Biton, and Yuwei Fang for their fruitful discussions and valuable input which helped improve this work.

{
    \small
    \bibliographystyle{ieee_fullname}
    \bibliography{main}
}

\clearpage
\newpage

\appendix
\appendixpage
\input{6-appendix}

\end{document}

%% file: macros.tex
\usepackage{booktabs}
\usepackage{multirow}
\usepackage[dvipsnames]{xcolor}
\usepackage{graphicx}
\usepackage{etoc}

\usepackage[toc,page,header]{appendix}
\usepackage{minitoc}

\definecolor{cvprblue}{rgb}{0.21,0.49,0.74}
\usepackage[pagebackref,breaklinks,colorlinks,citecolor=cvprblue]{hyperref}
\usepackage[capitalize]{cleveref}

\crefname{section}{Sec.}{Secs.}
\Crefname{section}{Section}{Sections}
\Crefname{table}{Table}{Tables}
\crefname{table}{Tab.}{Tabs.}
\crefname{equation}{Eq.}{Eqs.}
\Crefname{equation}{Eq.}{Eqs.}

\makeatletter
\def\blfootnote{\xdef\@thefnmark{}\@footnotetext}
\makeatother

\newenvironment{blockquote}{%
  \par%
  \medskip
  \leftskip=1.5em\rightskip=2em%
  \noindent\ignorespaces}{%
  \par\medskip}

\makeatletter
\DeclareRobustCommand\onedot{\futurelet\@let@token\@onedot}
\def\@onedot{\ifx\@let@token.\else.\null\fi\xspace}

\def\etal{\emph{et al}\onedot}

\def\etal{\emph{et al}\onedot}

\newcommand{\Sstar}{\textcolor{blue}{$S_*$}\;}

\definecolor{mydarkblue}{rgb}{0,0.08,1}
\definecolor{mydarkgreen}{rgb}{0.02,0.6,0.02}
\definecolor{mydarkorange}{rgb}{0.40,0.2,0.02}
\definecolor{mygreen}{RGB}{112,173,71}
\definecolor{myblue}{RGB}{91,155,213}
\definecolor{myorange}{RGB}{237,125,49}
\definecolor{myred}{RGB}{255,22,67}
\definecolor{honey}{RGB}{201,89,95}
\definecolor{amber}{RGB}{180,148,41}
\definecolor{darkblue}{RGB}{84,96,126}
\definecolor{darkgreen}{HTML}{008000}

%% file: 0-abstract.tex
\vspace*{-0.3cm}
\begin{abstract}
Recent large-scale vision-language models (VLMs) have demonstrated remarkable capabilities in understanding and generating textual descriptions for visual content. However, these models lack an understanding of \textit{user-specific} concepts. In this work, we take a first step toward the \textit{personalization} of VLMs, enabling them to learn and reason over user-provided concepts.  For example, we explore whether these models can learn to \textit{recognize} you in an image and \textit{communicate} what you are doing, tailoring the model to reflect \textit{your} personal experiences and relationships. To effectively recognize a variety of user-specific concepts, we augment the VLM with external concept heads that function as toggles for the model, enabling the VLM to identify the presence of specific target concepts in a given image. Having recognized the concept, we learn a new concept embedding in the intermediate feature space of the VLM. This embedding is tasked with guiding the language model to naturally integrate the target concept in its generated response. We apply our technique to BLIP-2 and LLaVA for personalized image captioning and further show its applicability for personalized visual question-answering. Our experiments demonstrate our ability to generalize to unseen images of learned concepts while preserving the model behavior on unrelated inputs. 
Project page: \url{https://snap-research.github.io/MyVLM/}.
\end{abstract}

%% file: 1-intro.tex
\vspace{-0.25cm}
\section{Introduction}~\label{sec:intro}
Large language models (LLMs)~\cite{zhao2023survey} have transformed human-computer interaction, offering users intuitive interfaces for interacting with textual information. The integration of vision into LLMs through vision-language models (VLMs)~\cite{yin2023survey} has further enhanced this interaction, enabling these models to ``see'' and reason over visual content. 
However, current VLMs possess \textit{generic} knowledge, lacking a personalized understanding of individual users.
For example, the VLM can easily recognize an image of \textit{\textbf{a}} dog but lacks the ability to understand that the depicted dog is \textit{\textbf{your}} personal dog.
This raises an intriguing question: can we equip these models with the ability to comprehend and utilize user-specific concepts, tailored specifically to \textit{\textbf{you}}? 
That is, can we ask the model questions about you, such as what \textit{\textbf{you}} are wearing or what \textit{\textbf{you}} are doing in the image? 
By personalizing these models, we can offer more meaningful interactions, better reflecting individual experiences and relationships.

Introducing personalized concepts into existing models poses significant challenges. Attempting to fine-tune these models for each user is computationally expensive and prone to catastrophic forgetting~\cite{ding2022dont,lee2019mixout}. 
In the context of LLMs, this has driven the development of model editing techniques designed to efficiently modify such large models~\cite{yao2023editing}. 
Yet, these methods only focus on altering the model's response to \textit{specific} user queries, for instance, editing the answer of ``Where is ECCV this year?'' from ``Tel Aviv'' to ``Milan''.

Successfully personalizing a VLM requires a deep understanding of how its visual and linguistic components interact. Intuitively, for a VLM to effectively respond to visual queries, it must not only \textit{recognize} and extract the relevant visual elements but also meaningfully \textit{communicate} them in its response. 
Introducing another layer of complexity to VLM personalization, we also find that the visual features extracted by pretrained VLMs are not expressive enough to effectively distinguish between semantically-similar objects.

To address these challenges, we propose augmenting the VLM with external heads that are trained to identify user-specific concepts within a scene. The signal from these heads is then used to add specific learnable vectors alongside the outputs of the vision encoder. In a sense, these learnable vectors are tasked with guiding the response generated by the language model to incorporate the matching personalized word in a way that is contextually accurate and aligned with the input image.
To train this concept vector, we utilize a small set of images ($3$-$5$) depicting the concept, each with a corresponding caption containing the personalized word.
We then optimize the concept embedding such that when given an image from the training set, appending the concept's embedding to the output of the vision encoder results in the VLM generating the corresponding personalized target caption.
To encourage the learnable embedding to remain in distribution with respect to the other image tokens, we incorporate an additional regularization over the attention assigned by the VLM to the concept embedding. 

Our personalization technique, named MyVLM, enables users to personalize a pretrained VLM without altering the original weights, preserving the model's general capabilities. 
Focusing on personalized image captioning, we apply MyVLM to both BLIP-2~\cite{li2023blip} and LLaVA~\cite{liu2023llava}, further demonstrating its applicability for visual-question answering, see~\Cref{fig:teaser}.
We show that MyVLM can effectively incorporate and contextualize personalized concepts, including specific objects and individuals, requiring only a few images of the concept.
We introduce and assess alternative baselines, highlighting our ability to better generalize to new instances of previously learned concepts.
To evaluate this new task, we introduce a new dataset containing various objects and individuals depicted in multiple contexts each with a corresponding personalized caption. The object dataset will be publicly available, aiming to facilitate further advancements in the personalization of VLMs. 

%% file: 2-related.tex
\section{Related Works}\label{sec:related}
\paragraph{\textbf{Vision-Language Models (VLMs).}}
The recent remarkable progress of large language models (LLMs)~\cite{chiang2023vicuna,chung2022scaling,touvron2023llama,touvron2023llama,brown2020language,alpaca}, has spurred efforts to equip them with the ability to reason over visual content~\cite{alayrac2022flamingo,yu2022coca,sun2023generative,li2020unimo,huang2023language,peng2023kosmos,bai2023qwen,wu2023visual,li2023otter,zhu2023minigpt,ye2023mplug,openai2023gpt4}.

A key area of research on VLMs focuses on leveraging frozen LLMs to align images and text within unified models that support both visual and language inputs. For instance, Flamingo~\cite{alayrac2022flamingo} fuses vision and language modalities using a cross-attention mechanism while keeping the vision encoder and language model fixed.
BLIP-2~\cite{li2023blip} introduces a Q-Former transformer to align visual features extracted from a fixed visual encoder with a large language model~\cite{chung2022scaling,zhang2022opt}. LLaVA~\cite{liu2023llava,liu2023improvedllava} and MiniGPT-4~\cite{zhu2023minigpt} employ instruction-tuned language models~\cite{wei2022finetuned,ouyang2022training,dai2023instructblip} and extract visual features from a pretrained visual encoder (e.g., CLIP~\cite{radford2021learning}). Specifically, LLaVA~\cite{liu2023llava} utilizes a simple linear layer to map the visual features to the input space of the language model. 

Recently, VLMs have been adopted for guiding various downstream tasks such as reinforcement learning~\cite{chen2024visionlanguage} and image generation~\cite{black2023training,richardson2023conceptlab}.
In this work, our focus is on personalizing VLMs, enabling them to reason over user-specific concepts. Importantly, our approach does not modify the original weights of the VLM, preserving its strong visual and linguistic priors.
We apply our method to BLIP-2~\cite{li2023blip} and LLaVA~\cite{liu2023llava}, demonstrating its effectiveness as a general framework applicable across various VLMs.

\vspace{-0.1cm}
\paragraph{\textbf{Personalization.}}
In the task of personalization, we aim to adapt a given model to capture new user-specific concepts. Personalization has been explored for a range of tasks including recommendation systems~\cite{amat2018artwork,benhamdi2017personalized} and object retrieval~\cite{cohen2022my,yeh2023meta,baldrati2023zero,karthik2023vision,saito2023pic2word}. PALAVRA~\cite{cohen2022my} optimizes a new token embedding within the input space of a text encoder to represent a new concept while Yeh~\etal~\cite{yeh2023meta} extend this for retrieving concepts in videos. Personalization has also been heavily studied in the context of image generation~\cite{gal2023image,ruiz2022dreambooth,kumari2022customdiffusion,rahman2023makeastory,gal2023encoder,alaluf2023neural,arar2023domainagnostic,li2023blipdiffusion,voynov2023p+,ye2023ip-adapter,wang2024instantid,nitzan2022mystyle}. Most relevant to our work are inversion-based approaches~\cite{gal2023image} where embeddings are optimized to capture the target concept. 

Another line of work focuses on personalizing image captioning models~\cite{wang2023user,chunseong2017attend,park2018towards,zeng2019automatic,shuster2019engaging}. Park~\etal~\cite{park2018towards} employ a memory network to store a user's active vocabulary and utilizes it to generate captions reflecting the user's personal writing style. More recently, Wang~\etal~\cite{wang2023user} employed a transformer to fuse visual features and text features encoding user-specific keywords. These features are then passed to a pretrained language model to generate personalized captions. 
Importantly, personalized captioning techniques focus on generating a specific \textit{writing style}. In contrast, we aim to teach the model to incorporate a new user-specific concept into a personalized textual output aligned with a given image.

\vspace{-0.2cm}
\paragraph{\textbf{Model Editing.}}
While modern machine learning systems excel in achieving state-of-the-art performance, their effectiveness can diminish post-deployment~\cite{balachandran2022correcting}, leading to hallucinations~\cite{benkish2023mocha,ji2023survey} and factual decay~\cite{sinitsin2020editable,i2023minimizing}.
Consequently, there is a growing need for model editing, which aims to make data-efficient modifications to a model's behavior while minimizing the impact on performance across other inputs.
In the context of language models, several approaches incorporate hypernetworks~\cite{ha2017hypernetworks} to predict edits for specific inputs~\cite{cao2021editing,mitchell2022serac,mitchell2022fast} or perform parameter-efficient model tuning~\cite{hu2021lora,meng2022locating,meng2022memit,li2023pmet}.
One particular area of interest is enabling a large set of edits within a single model~\cite{meng2022memit,hartvigsen2023aging}. Hartvigsen~\etal~\cite{hartvigsen2023aging} introduce a codebook within the language model's intermediate feature space, storing previously learned edits. For each new edit, a new key is added to the codebook, and its corresponding value is optimized such that the language model produces the desired output for the given query. 
Similar model editing techniques have been explored for generative image models~\cite{bau2020rewriting,gandikota2023erasing,kumari2023ablating,arad2023refact,tewel2023key} and multi-modal learning~\cite{cheng2023can}. 
Recently, Retrieval-Augmented Generation (RAG) has also emerged as an alternative approach for injecting knowledge into LLMs~\cite{gao2023retrieval,vu2023freshllms,lewis2020retrieval}.
We refer the reader to Yao~\etal~\cite{yao2023editing} for a comprehensive survey on model editing.

Our goal of personalizing VLMs necessitates a different approach from model editing. Model editing focuses on applying precise modifications to the model behavior (e.g., associating ``What is the capital of France?'' with ``Paris''). 
In contrast, personalization requires the model to adapt to new images of the concept, which may vary significantly (e.g., recognizing an individual across diverse settings).
Moreover, it is essential to disentangle the concept from its surroundings when teaching a model a new concept, such as separating an individual from the clothes they are wearing.
Finally, the VLM must not only identify the concept but also contextualize it within the generated response. For example, instead of simply outputting the concept identifier ``\textcolor{blue}{$S_*$}'', the model should produce a more descriptive response such as ``\Sstar sitting on a bench, drinking wine on a patio''.

%% file: 3-method.tex
\section{Method}\label{sec:method}
Our goal is to extend the capabilities of a vision-language model (VLM) by teaching it to generate personalized textual responses focusing on user-specific concepts.
We begin by outlining the specific families of VLM models considered in this work, namely BLIP-2~\cite{li2023blip} and LLaVA~\cite{liu2023llava}.
We then introduce our personalization technique, MyVLM, and demonstrate its application for both personalized captioning and visual question-answering.

\subsection{Preliminaries}
\paragraph{\textbf{BLIP-2.}}
The BLIP-2 model, introduced by Li~\etal~\cite{li2023blip}, is a VLM model that is built around three main components:  
(1) a pretrained ViT-L/14~\cite{dosovitskiy2020image} vision encoder, (2) a pretrained language model~\cite{chung2022scaling}, and (3) a trainable Querying Transformer (Q-Former) model tasked with bridging the vision-language modality gap. 
The Q-Former receives as input $32$ learnable query tokens, each of dimension $d=768$, and is composed of three types of layers: self-attention, cross-attention, and feed-forward layers. 
Most relevant to our work are the cross-attention layers, placed at every other transformer block. These blocks are designed to capture the interaction between the extracted image features and the learnable query tokens (as well as our learned concept representations). 
    
More specifically, at each cross-attention layer, the image features are first projected into a set of keys ($K$) and values ($V$) via learned linear projections. 
The intermediate representations of the $32$ learned query tokens are similarly projected into a set of attention queries $q_i$. 
For each query $q_i$, a weighted average is then computed over these representations, as given by:
\begin{equation}~\label{eq:attention}
\begin{split}
    A_{i} & = \text{softmax} \left ( \frac{q_i\cdot K^T}{\sqrt{d}} \right ) V.
\end{split}
\end{equation}
Intuitively, the probability defined by the softmax indicates the amount of information that will be passed from each image feature to each query token. 

\paragraph{\textbf{LLaVA.}}
Similar to BLIP, LLaVA~\cite{liu2023llava} seeks to connect a fixed vision encoder with a fixed language model, in this case, CLIP ViT-L/14~\cite{radford2021learning} and Vicuna~\cite{chiang2023vicuna} models, respectively. To do this, LLaVA follows a simpler architecture where a single linear layer is used to map the image features into the token embedding space of the language model.
This sequence of projected visual tokens is then fed directly to the language model, along with the encoded language instruction.

\input{figures/method}

\subsection{MyVLM}
We now turn to describe our approach to personalizing vision-language models for user-specific concepts. For simplicity, we describe MyVLM applied over the BLIP-2 model~\cite{li2023blip}, followed by a discussion of the adjustments necessary for integrating MyVLM with LLaVA~\cite{liu2023llava}. 
Given only a few images (${\sim}$3-5) of the specific concept and corresponding captions that contain the concept identifier \textcolor{blue}{$S_*$}, our objective is to augment the VLM with the ability to answer specific queries over new images depicting the concept.

Our technique is comprised of two key stages: first \textit{recognizing} the concept within the given scene, and then \textit{communicating} information about the concept to the language model. To achieve this, we introduce a \textit{concept head} designed to identify the presence of a personalized concept within an image. Then, a learned \textit{concept embedding}, representing an object or individual, is used to guide the LLM in incorporating the concept into its personalized textual response.
An overview of MyVLM is provided in~\Cref{fig:method}.

\vspace{-0.25cm}
\paragraph{\textbf{Recognizing.}}
To enable the pretrained VLM to reason over personalized concepts, we must first identify their presence in a given scene. A direct approach for doing so is to consider the feature space of the VLM's vision encoder. However, we empirically observe that the feature space of the frozen vision encoder is not expressive enough to visually distinguish the target concept from similar concepts (see~\Cref{sec:ablation_feature_spaces}). 
While one can potentially fine-tune the vision encoder itself to better recognize our object of interest, this 
may naturally harm its strong general knowledge and impact its ability to extract information about the entire image, which is also crucial for generating accurate responses.

Instead, we augment the VLM with a set of external \textit{concept heads}, with each head dedicated to recognizing a single personalized concept we wish to teach the model. 
These heads allow the model to identify the concepts of interest without hindering its ability to provide visual information about the entire scene depicted in the image. 
As the heads operate independently from the VLM model itself, we can support any specialized classification head to recognize our target concepts. Specifically, for identifying user-specific objects, we choose to employ a simple linear classifier trained over embeddings extracted from a pretrained CLIP model~\cite{radford2021learning,fang2023data}. 
To generate personalized outputs tailored to specific individuals, we utilize a pretrained face recognition network~\cite{deng2020retinaface,Deng_2022} as an additional concept head.
Importantly, defining a separate head for each concept provides additional flexibility, enabling one to naturally scale to additional concepts over time. %
Additional details on the construction of the concept heads are provided in~\Cref{sec:additional_details_training}.

\vspace{-0.25cm}
\paragraph{\textbf{Communicating.}}
Given the ability to \textit{recognize} our concept of interest, we now turn to describe our approach for teaching the VLM to \textit{communicate} responses about our target concepts. To do so, we learn a single concept embedding vector representing the concept within the intermediate feature space of the VLM. Intuitively, this embedding should guide the language model toward generating a text response incorporating the concept identifier that (1) is contextually correct and (2) aligns with both the provided image and language instruction.

To learn this embedding, we use a small set of images depicting the concept in various contexts, each with a corresponding target caption containing the concept identifier. 
For the identifier, we follow DreamBooth~\cite{ruiz2022dreambooth} and use an existing, uncommon word when personalizing outputs for objects and use a short name when personalizing individuals.
We find the concept embedding $e_*$ via direct optimization. The embedding $e_*$ is appended to the image features extracted from the frozen vision encoder and fed to the Q-Former network via the cross-attention layers.
The output of the Q-Former is then passed to the frozen language model that generates the predicted image caption. 
The optimization process aims to minimize the standard cross-entropy loss between the generated caption and the provided target caption. 

Our optimization can be defined as: 
\begin{equation}
\begin{split}
    e_* = \arg \min_e \sum_{i=1}^{N} \mathcal{L}_{CE} \left ( t_i, o(I_i, e) \right ),
\end{split}
\end{equation}
where $N$ is the number of training samples, $t_i$ represents our target caption of the $i$-th sample, and $o(I_i, e)$ is the generated output caption of the $i$-th image $I_i$, given the concept embedding $e$.
At inference, the embedding of a concept recognized by our concept heads is similarly appended to the output of the vision encoder.

\vspace{-0.15cm}
\paragraph{\textbf{Improving Generalization.}}
While the approach described above allows for generating personalized captions, we observe that directly appending the concept embedding to the image features may lead to unnatural captions being generated by the language model. This issue arises from two primary observations. 

First, within the cross-attention layers of the Q-Former, we observed that the vector norms of the key ($k_*$) and value ($v_*$) corresponding to the concept embedding were significantly larger compared to the norms of the frozen image features. This behavior was also previously observed in text-to-image personalized techniques~\cite{alaluf2023neural,tewel2023key}.
Therefore, before computing the cross-attention with the Q-Former query tokens, we normalize $k_*$ and $v_*$ to match the average norm of the original keys and values, denoted as $n_k$ and $n_v$, respectively. The modified key and value of our embedding are then given by:
\begin{equation}\label{eq:norm}
\begin{split}
    \hat{k}_{*} = \frac{k_{*}}{\lVert k_{*} \rVert} \cdot n_k  \qquad 
    \hat{v}_{*} = \frac{v_{*}}{\lVert v_{*} \rVert} \cdot n_v
\end{split}
\end{equation}

Second, in the attention weights computed in the Q-Former cross-attention layers (\Cref{eq:attention}), we observe that the concept token tended to dominate the attention distribution, causing the query tokens to no longer attend meaningfully to the image tokens. 
By failing to adequately attend to the original image tokens, the relevant visual information may no longer be passed to the language model, leading to a possible misalignment between the generated caption and the image.

To encourage a more balanced distribution of attention across all tokens, we introduce an $L2$ regularization over the attention probabilities assigned to the concept embedding by all $32$ Q-Former query tokens. 
That is, we compute: 
\begin{equation}~\label{eq:reg}
    \mathcal{L}_{reg} = \left \lVert \; \text{softmax} \left (  Q \cdot \hat{k}_*  \right ) \; \right \rVert_2^2.
\end{equation}
By encouraging the tokens to attend to the original image features, we found the outputs to be more coherent and aligned with the image (see~\Cref{sec:ablations_reg_augs}).

\input{figures/self_attention}

\vspace{-0.225cm}
\subsection{MyVLM over LLaVA}
\vspace{-0.075cm}
To apply MyVLM over LLaVA~\cite{liu2023llava} we make the following adjustments to the scheme presented above. 
First, we append the concept embeddings to the output of the linear projection rather than directly after the vision encoder. 
We find that this resulted in faster, more stable convergence.
Second, since LLaVA does not utilize a cross-attention mechanism, we omit the normalization of keys and values as presented in~\Cref{eq:norm}. Instead, we rescale the concept embedding such that its vector norm is equal to that of the [CLS] token outputted by the vision encoder.
Finally, we modify the attention-based regularization defined in~\Cref{eq:reg}. Here, we apply an L2 regularization that encourages low attention to be assigned from the other input tokens to the concept embedding, including from both the language tokens and from the other projected image tokens.

Interestingly, since our concept embedding is passed as input to the language model along with the other projected image features, we have a natural way to investigate whether our learned concept embeddings attend to meaningful regions within the input images. 
Specifically, we examine the self-attention layers of LLaVA's language model and visualize the attention weights assigned by the concept embedding to each of the image patches, as illustrated in~\Cref{fig:self_attention_visualization}. 
We believe that further exploration into the behavior of the concept embeddings within the attention layers could offer additional insights for extending the capabilities of MyVLM. We leave this exploration for future work.

\subsection{MyVLM for Additional Applications}

\paragraph{Personalized Vision Question-Answering}

For applying MyVLM for personalized visual question-answering, we follow a similar approach as introduced above, but modify the language instructions and target outputs used for defining our objective function. 

Observe that in personalized captioning, the language instruction passed to the language model when optimizing the concept embedding remains fixed. However, for visual question-answering, we are interested in generalizing to any question the user may ask over a given image. 
Therefore, we expand the set of instructions and targets used during the optimization process described above. 
Specifically, we define a set of $10$ pairs of questions and answers related to the target concept. 
For instance, we ask ``What color is \Sstar?'', ``Where is \Sstar located in the image?'', ``What is \Sstar wearing?'', etc. 
Then, at each optimization step, we randomly sample one question-answer pair to use for the current step. 
Intuitively, by optimizing the embedding vector through questions aimed specifically at the target concept, the embedding should better generalize to new questions the user may ask about the concept. 

\vspace{-0.3cm}
\paragraph{Personalized Referring Expression Comprehension.}
Next, we demonstrate the applicability of MyVLM for an additional personalized task: referring expression comprehension (REC)~\cite{qiao2020referring}, which involves localizing a target subject in a given image. To achieve this, we utilize MiniGPT-v2~\cite{chen2023minigptv2}, a recent VLM that can naturally handle various vision-language tasks by employing different task identifiers to define the language instructions passed to the language model. 
As MiniGPT-v2 shares the same architecture as LLaVA~\cite{liu2023llava}, we adopt the same training setup for learning our concept embeddings. 
Specifically, to optimize the concept embedding, we follow the same scheme as used for personalized captioning and use the instruction: 
\vspace{-0.1cm}
\begin{blockquote}
    \textit{``[caption] Please caption this image of \Sstar''}.
    \vspace{-0.1cm}
\end{blockquote}
\noindent During inference, to solve for REC we modify the language instruction to: 
\vspace{-0.1cm}
\begin{blockquote}
    \textit{``[refer] \Sstar in the image''},
    \vspace{-0.1cm}
\end{blockquote}
\noindent which returns the bounding box coordinates of the target subject within the provided image. We emphasize that this is achieved with only the captioning supervision during optimization. 
This builds on the inherent ability of the underlying VLM to solve for multiple tasks while highlighting that the learned embedding does indeed capture the semantic representation of the concept which the model can reuse for its different tasks.

%% file: figures/method.tex
\begin{figure*}[t]
    \centering
    \includegraphics[width=\textwidth]{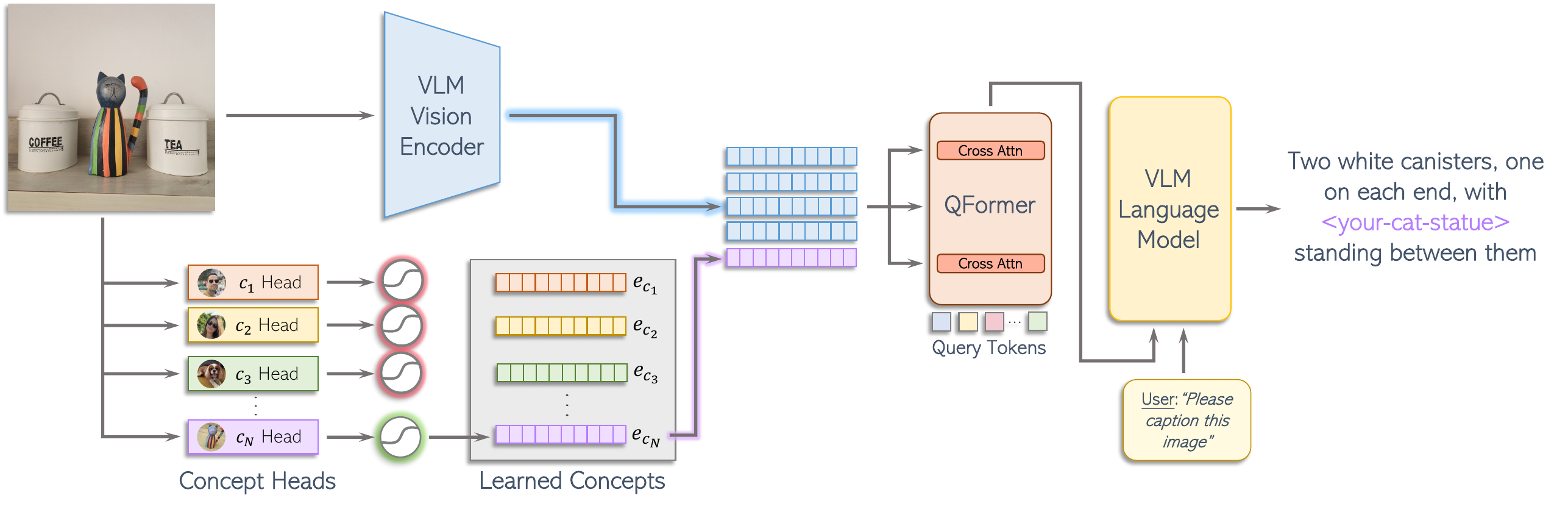} 
    \vspace{-0.5cm}
    \caption{
    \textbf{MyVLM overview}, applied over BLIP-2.
    Given an input image, we pass it through the frozen vision encoder of the VLM. In parallel, we pass the image through a set of learned \textit{concept heads}, each tasked with recognizing a single user-specific concept. 
    We append the \textit{concept embedding} of the identified concept to the extracted vision features.
    These features are then passed to the Q-Former via a set of cross-attention layers to extract relevant information from the image features and concept embedding.
    Given the Q-Former outputs and language instruction, the frozen LLM outputs a response incorporating the concept identifier while remaining aligned with the input. 
    } 
    \vspace{-0.2cm}
    \label{fig:method}
\end{figure*}

%% file: figures/self_attention.tex
\begin{figure}[t]
    \centering
    \addtolength{\belowcaptionskip}{-5pt}
    \renewcommand{\arraystretch}{1}
    \footnotesize
    \vspace{-0.1cm}
    \begin{tabular}{p{0.17\textwidth} p{0.17\textwidth} p{0.17\textwidth} p{0.17\textwidth}}

        \begin{center} \includegraphics[width=0.14\textwidth]{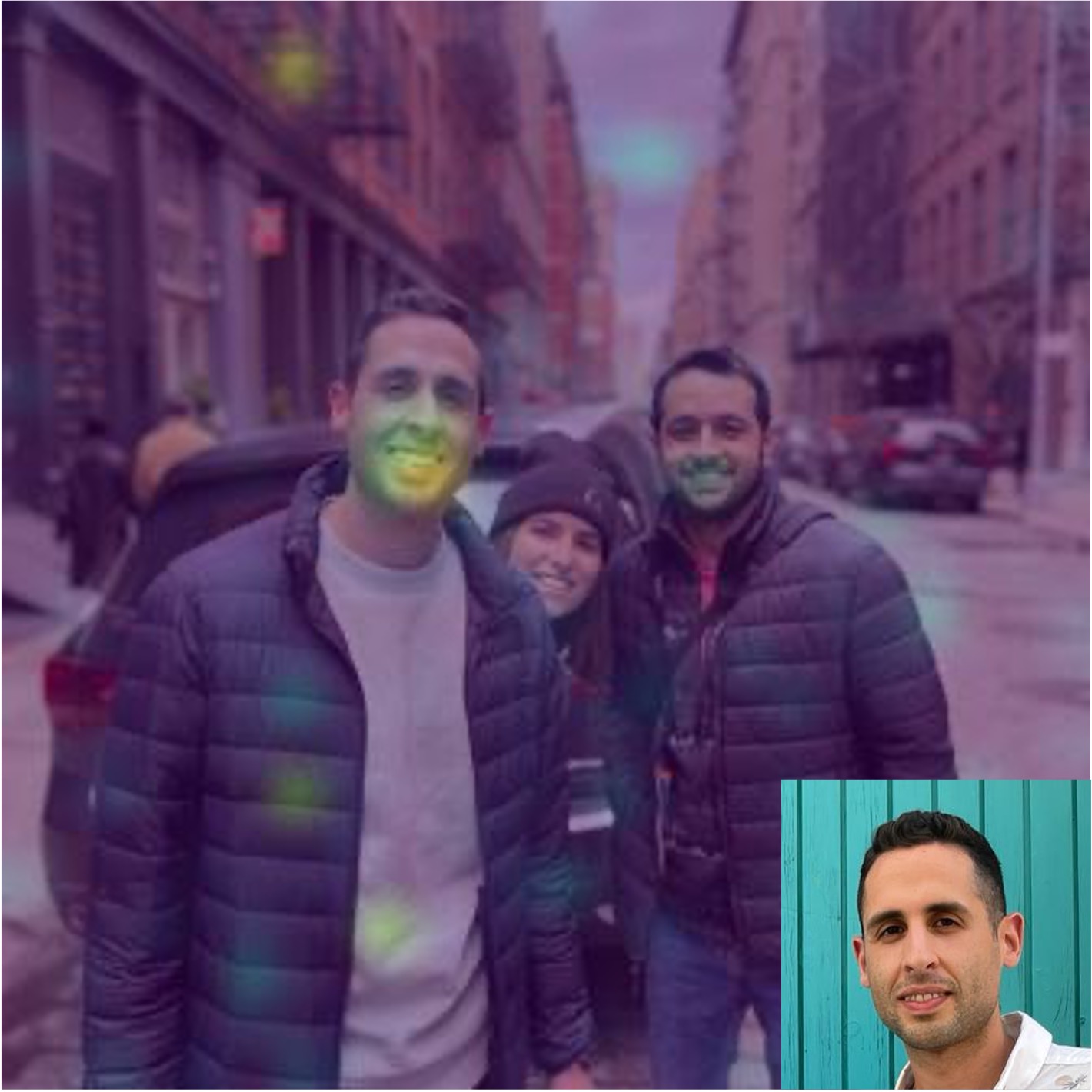} \end{center} &
        \begin{center} \includegraphics[width=0.14\textwidth]{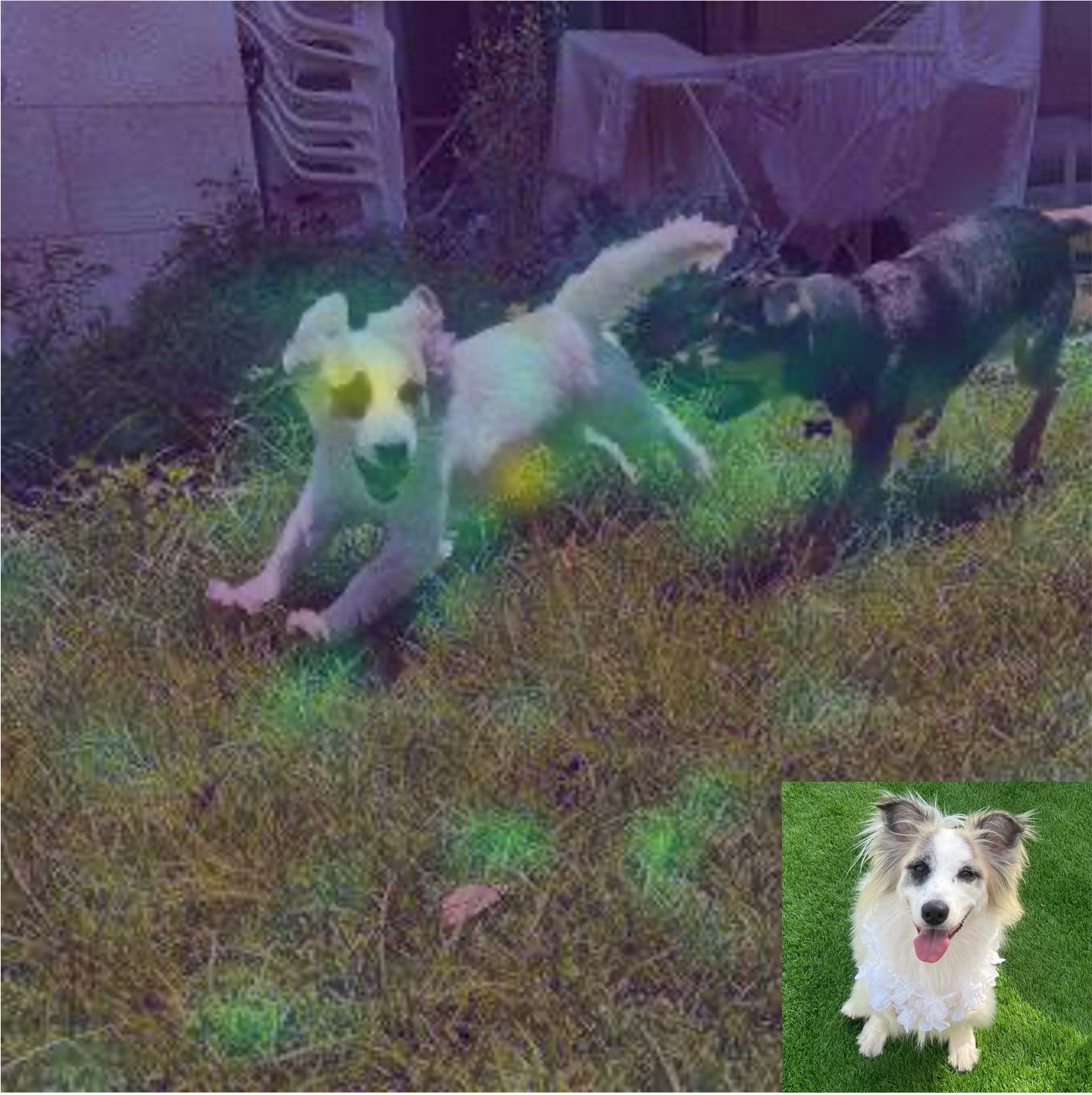} \end{center} \\[-0.85cm]

        \begin{center} \footnotesize  ``\textcolor{blue}{$S_*$}, dressed in a blue jacket and a green sweater...'' \end{center} &
        \begin{center} \footnotesize  ``\Sstar and a black dog running in a yard'' \end{center} \\[-0.85cm]
        
        \begin{center} \includegraphics[width=0.14\textwidth]{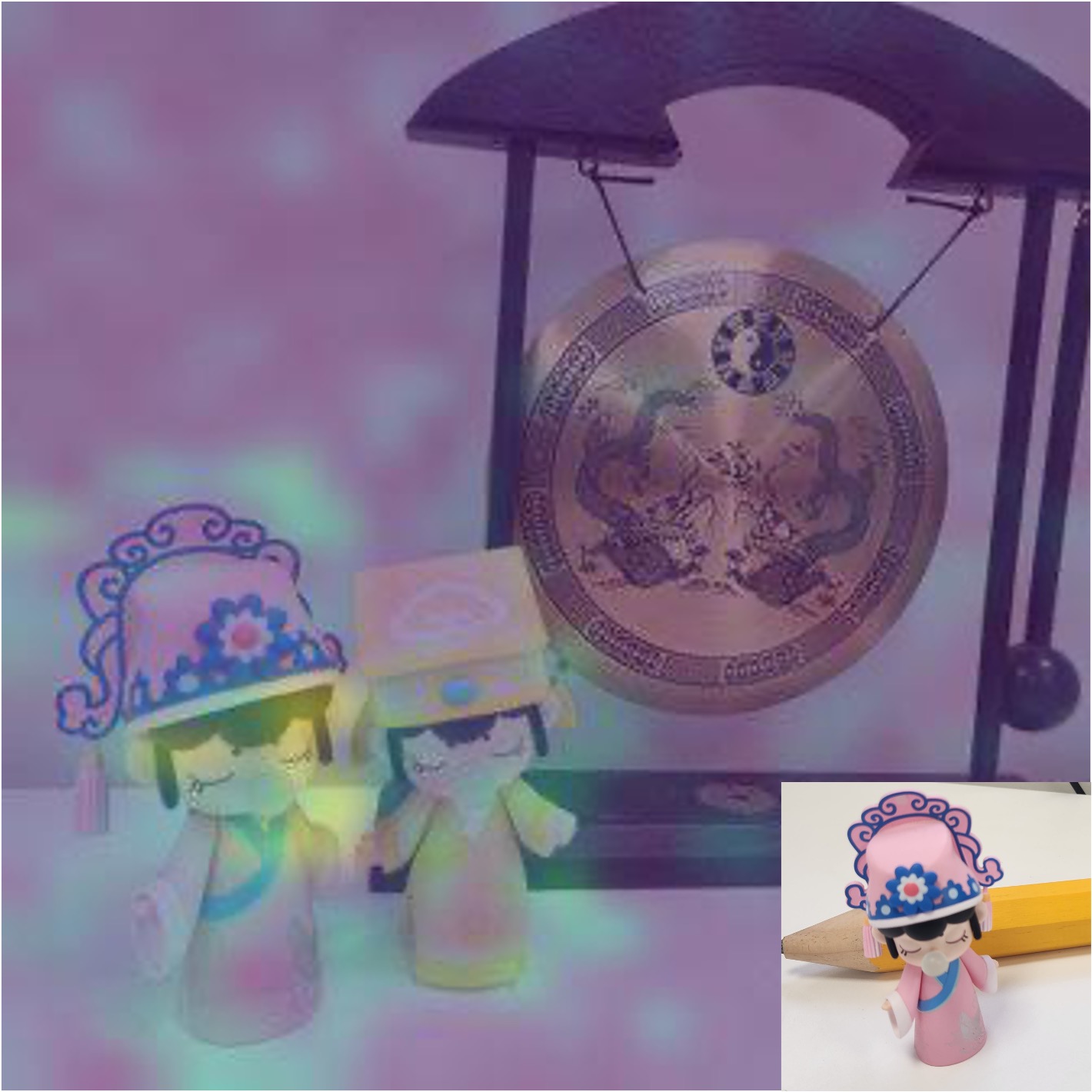} \end{center} &
        \begin{center} \includegraphics[width=0.14\textwidth]{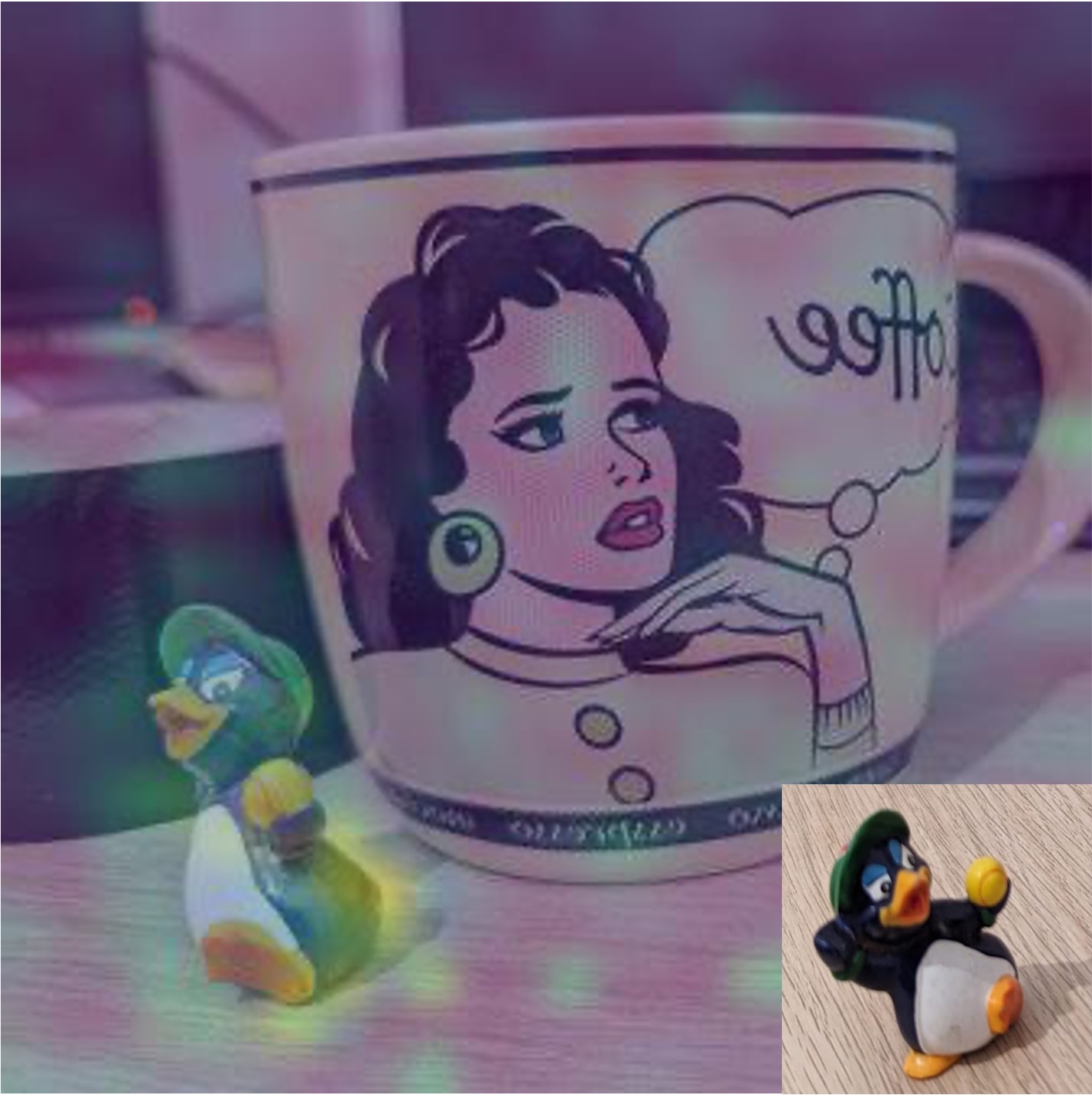} \end{center} \\[-0.85cm]
        
        \begin{center} \footnotesize ``\Sstar and a Chinese doll standing next to a gold gong...'' \end{center} &
        \begin{center} \footnotesize ``\Sstar is sitting next to a coffee mug with a cartoon character...'' \end{center}

    \end{tabular}
    \vspace{-0.5cm}
    \caption{
    \textbf{Self-attention visualization.} We examine the self-attention of LLaVA's language model to visualize the attention weights assigned from the concept embedding to each image feature. 
    As can be seen, the concept embedding attends to relevant regions within the images, assigning higher weights to areas where the concept is located.}
    \label{fig:self_attention_visualization}
    \vspace{-0.175cm}
\end{figure}

%% file: 4-results.tex
\section{Experiments}\label{sec:experiments}

\paragraph{\textbf{Dataset.}}
As there are no existing datasets for VLM personalization, we introduce a new dataset for evaluating this task. The dataset is split into two categories: objects and people. For objects, we curate a set of $29$ objects including various toys, statues, mugs, and pets. 
For each concept, we collected at least $10$ images containing the subject in diverse scenes alongside other objects and set against interesting backgrounds. 
For people, we collect images of $16$ individuals ranging from ages $25$ to $80$. Each individual is represented by a minimum of $15$ images, showcasing them in a range of scenarios, attire, and sometimes alongside other people in the same image.  
For each image, we wrote a corresponding personalized caption incorporating the concept identifier. 
Examples of each object are provided in~\Cref{sec:appendix_data}. 
The $29$ objects will be publicly available to facilitate further research into VLM personalization.

\vspace{-0.225cm}
\paragraph{\textbf{Evaluation Metrics.}}
In this work, we focus on quantitatively evaluating personalized image captioning, as data for this task is more readily available. We evaluate the personalized captions along two fronts. 
First, we measure recall and validate whether the concept identifier appears at least once in the generated caption. 
This evaluates both our ability to recognize the concept in new images and our ability to incorporate the concept in the output via its embedding.

Second, we assess the alignment of the generated caption with the input image and target caption, considering two metrics.
We first compute the CLIPScore~\cite{hessel2021clipscore} between the generated captions and input images. 
We additionally compute a sentence similarity measure, computing the average cosine similarity between sentence embeddings extracted from the target caption and the generated caption. 
For both, we replace the concept identifier with the concept's category. 
For example, \textcolor{Fuchsia}{$\langle$your-dog$\rangle$} is replaced with ``dog'' and \textcolor{orange}{$\langle$your-toy$\rangle$} with ``toy''. 
In~\Cref{sec:appendix_captioning}, we present standard captioning metrics, showing that MyVLM preserves the general captioning capabilities of the underlying VLM.

\input{figures/our_results}

\vspace{-0.225cm}
\paragraph{\textbf{Baselines.}}
Since there are currently no existing baselines focusing on generating personalized captions for a target concept, we introduce several alternative approaches for doing so. 
First, we generate captions using the frozen VLM model. Then, for each concept, we define a set of three keywords describing the concept, obtained using GPT-4V~\cite{openai2023gpt4} by providing it a cropped image of the concept. 
For people, we designate a single keyword per concept, either ``man'' or ``woman''. 
Given the caption generated by the VLM, we then search the caption for the keyword, and if found, we replace the keyword with the concept identifier. 

Additionally, we introduce an LLM-guided baseline. Here, given the captions generated by the frozen VLM, we pass the caption into a language model~\cite{jiang2023mistral} and ask it to integrate the concept identifier into the caption if one of the keywords is present. This approach offers a more flexible constraint, allowing the language model to more freely incorporate the concept into the caption. 

Finally, we compare MyVLM with GPT-4V~\cite{openai2023gpt4} by showing GPT-4V an image of the concept and its identifier and then asking it questions over new images.
Similarly, in~\Cref{sec:appendix_flamingo}, we quantitatively compare MyVLM to OpenFlamingo~\cite{awadalla2023openflamingo,alayrac2022flamingo}, which also supports interleaved image-text inputs.
Additional details on the baselines can be found in~\Cref{sec:appendix_data}.

\subsection{Personalized Captioning}
\paragraph{\textbf{Qualitative Evaluation.}}
In~\Cref{fig:our_results}, we present personalized captioning for various user-provided concepts generated by our method applied to LLaVA~\cite{liu2023llava}. Captions generated by MyVLM emphasize the target subject rather than offering a generic or abstract description of the entire scene, as generated by the original VLM. Moreover, MyVLM naturally integrates the concept identifier into the generated output while remaining aligned with the input image. 
In particular, even in scenes where multiple individuals are present in the image, MyVLM successfully focuses on the target identity when generating its caption. For instance, notice the man in the green sweater in the first column or the woman in the yellow dress in the third column. This is also evident when creating personalized captions for a user-provided object placed around numerous other objects in a scene. 
For instance, in the rightmost column, the original caption generated by LLaVA ignores the target ceramic mug entirely, whereas our personalized caption accurately communicates its location in the image. 
Additional personalized captioning results obtained over both BLIP~\cite{li2023blip} and LLaVA can be found in~\Cref{sec:additional_results}.

\paragraph{\textbf{Qualitative Comparison.}}
In~\Cref{fig:qualitative_comparisons}, we provide a visual comparison with our LLM-guided baseline. As can be seen, this baseline heavily relies on the original captions generated by the VLM. 
The baseline struggles when the target concept appears in the same image with another subject sharing the same keyword, resulting in an unnatural caption.
In contrast, MyVLM successfully identifies the target subject and generates captions that accurately contextualize the concept within its surroundings. Importantly, we do so when multiple subjects are present and when the concept comprises a small region of the image.

Next, we compare our method to GPT-4V in~\Cref{fig:comparison_gpt}. We provide it with an image of the target concept along with its identifier. We then ask it to caption images that may contain the concept. 
As can be seen, GPT-4V can generalize to new images of the concept. However, when presented with images of negative examples that have a similar textual description, GPT-4V misidentifies them as the target concept. For example, in the leftmost example, it incorrectly associates ``a cup with a blue eye design'' with the concept.
In contrast, MyVLM can distinguish between these hard negative examples and the target concepts. 

Interestingly, the fact that GPT-4V misidentifies visually distinct objects that share a similar textual description may hint that it heavily relies on the textual description of the object, even when prompted with an image of it. 
This emphasizes the advantage of \textit{learning} a dedicated embedding to represent our concept instead of relying solely on natural language, where describing our \textit{exact} target concept may be challenging.

\input{figures/comparisons}

\input{figures/comparison_gpt}

\paragraph{\textbf{Quantitative Comparison.}}
We now turn to quantitatively compare MyVLM with the alternative baselines. To provide a larger validation sample size, we perform bootstrapping without replacement over our constructed dataset. For each concept, we randomly sample five different training sets, each containing four images, and set the remaining images as the corresponding validation set. We then train MyVLM on each training set and generate captions for all validation images.
This results in a total of $2,430$ validation images, out of which $1,265$ contain user-specific objects, while the remaining images depict individuals.

We begin by measuring each baseline's ability to incorporate the concept identifier within the generated caption. Results are summarized in~\Cref{tb:quantitative_recall}.
As can be seen, for user-specific objects, trying to simply insert the concept identifier into the caption via a closed set of keywords is ineffective, with a notable gap in recall compared to MyVLM.
While incorporating an external language model greatly improves recall, MyVLM still outperforms the LLM-guided approach by $44\%$ when using BLIP-2 and $30\%$ for LLaVA. 
When considering individuals, although the keyword-replacement baseline and MyVLM achieve comparable results when applied over BLIP, MyVLM significantly outperforms both baselines when applied to LLaVA. The large gap to LLaVA appears to stem from the abstract-like captions generated by LLaVA, whereas BLIP-2 tends to generate simpler captions more likely to incorporate the predefined keywords.
This highlights the robustness of MyVLM to different VLM models, whereas the handcrafted baselines heavily rely on the captioning styles of the underlying VLM.

Next, we investigate MyVLM's performance when training the concept embedding using 4, 2, and only 1 image, where we evaluate all models over the same validation set. 
Results, averaged across all $45$ concepts, are presented in~\Cref{tb:quantitative_similarities}. 
In terms of recall, results over both BLIP-2 and LLaVA consistently improve when adding more training samples. Observe that even when trained using a single sample, MyVLM still outperforms all baselines by significant margins. 
We additionally compute the average similarities between our personalized captions and (1) the input images and (2) the target captions. 
As can be seen, adding additional training samples improves both the image similarity and text similarity, indicating improved generalization. 
This further highlights the effectiveness of MyVLM in generating personalized captions, even in challenging few-shot settings and across multiple VLM frameworks.

In~\Cref{sec:appendix_evals}, we provide additional ablation studies on the contribution of our augmentations and regularization techniques. We additionally explore the output space of the VLM vision encoder and validate the use of our concept heads, showing that they attain both high recall over new images of the target concept and high precision over negative samples, demonstrating our ability to support multiple concepts in a single VLM.

\input{tables/recall}
\input{tables/text_similarities}

\subsection{MyVLM for Additional Applications}

\paragraph{\textbf{Personalized Visual Question-Answering.}}
First, we demonstrate that MyVLM can be used for personalized visual question-answering. In~\Cref{fig:vqa}, we demonstrate results across several user-specific concepts. MyVLM correctly answers questions related to the target concept, even within scenes containing multiple individuals (columns one and two), and in scenes where the subject occupies a small area of the image (columns three and four). For instance, MyVLM not only correctly identifies that the dangling child toy is located in the refrigerator but also its precise location on the top shelf. This highlights that MyVLM can faithfully capture distinctive features associated with the target concept, allowing it to correctly identify and localize the concept in a new scene.

\vspace{-0.25cm}
\paragraph{\textbf{Personalized Referring Expression Comprehension.}}
In~\Cref{fig:rec}, we present personalized results for referring expression comprehension (REC) and captioning achieved by MyVLM using MiniGPT-v2~\cite{chen2023minigptv2}. As shown, MyVLM cannot only generate personalized captions but also pinpoint the concept within the image without any direct supervision on the localization task. Importantly, the ability of MiniGPT-v2 to accommodate multiple tasks through distinct task identifiers enables MyVLM to be extended naturally to additional personalized applications with minimal modifications. 

\input{figures/vqa}
\input{figures/rec}

%% file: figures/our_results.tex
\begin{figure*}[t]
    \centering
    \addtolength{\belowcaptionskip}{-5pt}
    \renewcommand{\arraystretch}{1}
    \footnotesize
    \vspace{-0.15cm}
    \begin{tabular}{p{0.175\textwidth} p{0.175\textwidth} p{0.175\textwidth} p{0.175\textwidth} p{0.175\textwidth}}

        \setlength\tabcolsep{0pt}
        \begin{tabular}{c c c}
            \includegraphics[width=0.06\textwidth]{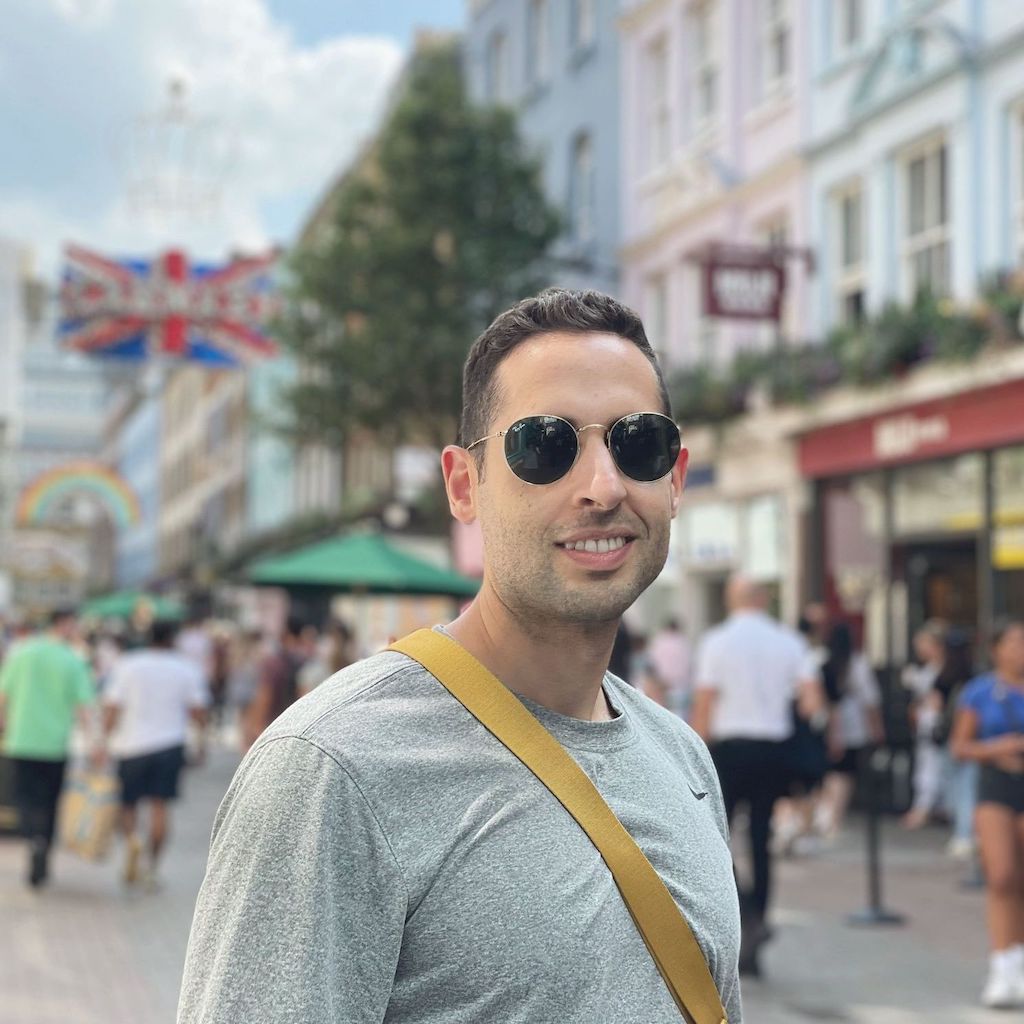} & 
            \includegraphics[width=0.06\textwidth]{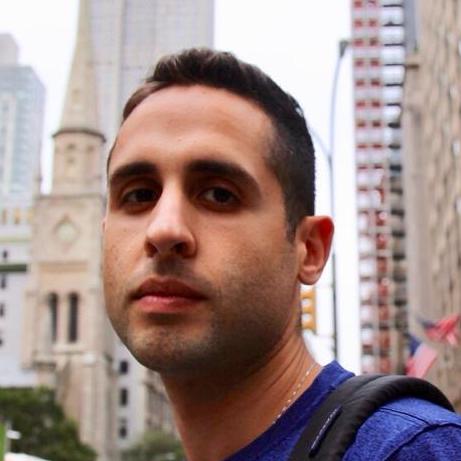} & 
            \includegraphics[width=0.06\textwidth]{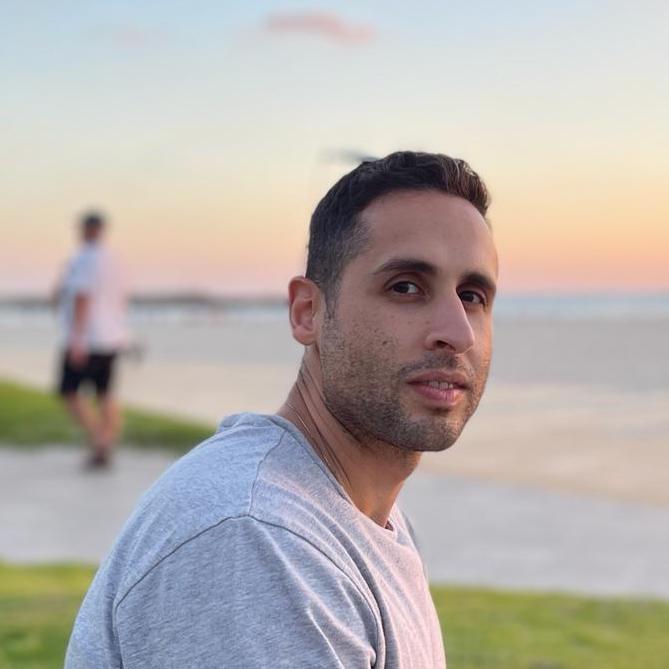}
        \end{tabular} &
        \setlength\tabcolsep{0pt}
        \begin{tabular}{c c c}
            \includegraphics[width=0.06\textwidth]{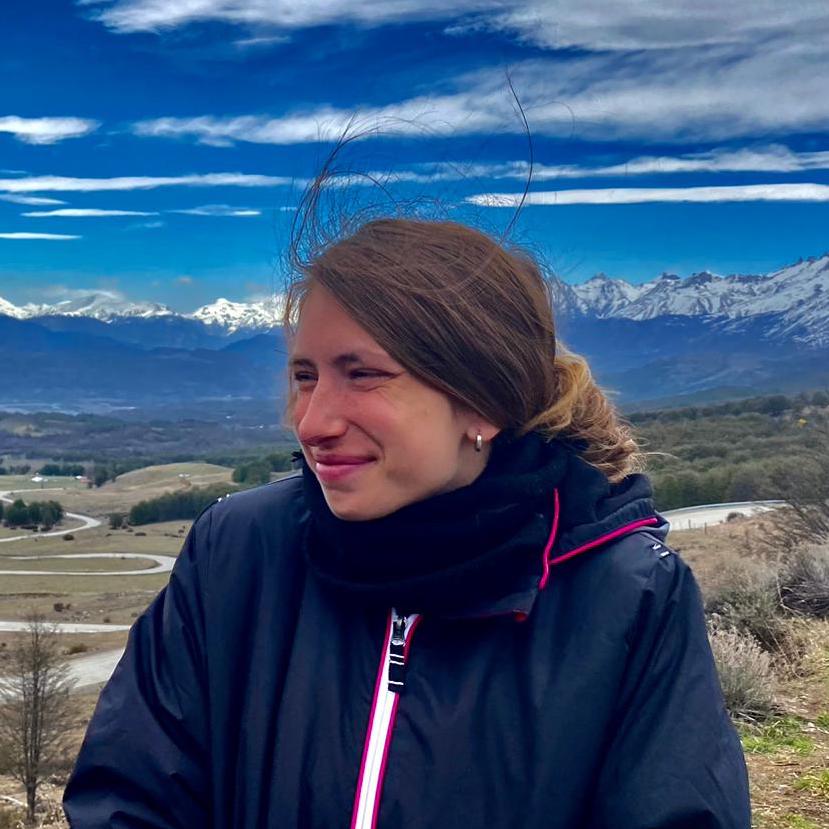} & 
            \includegraphics[width=0.06\textwidth]{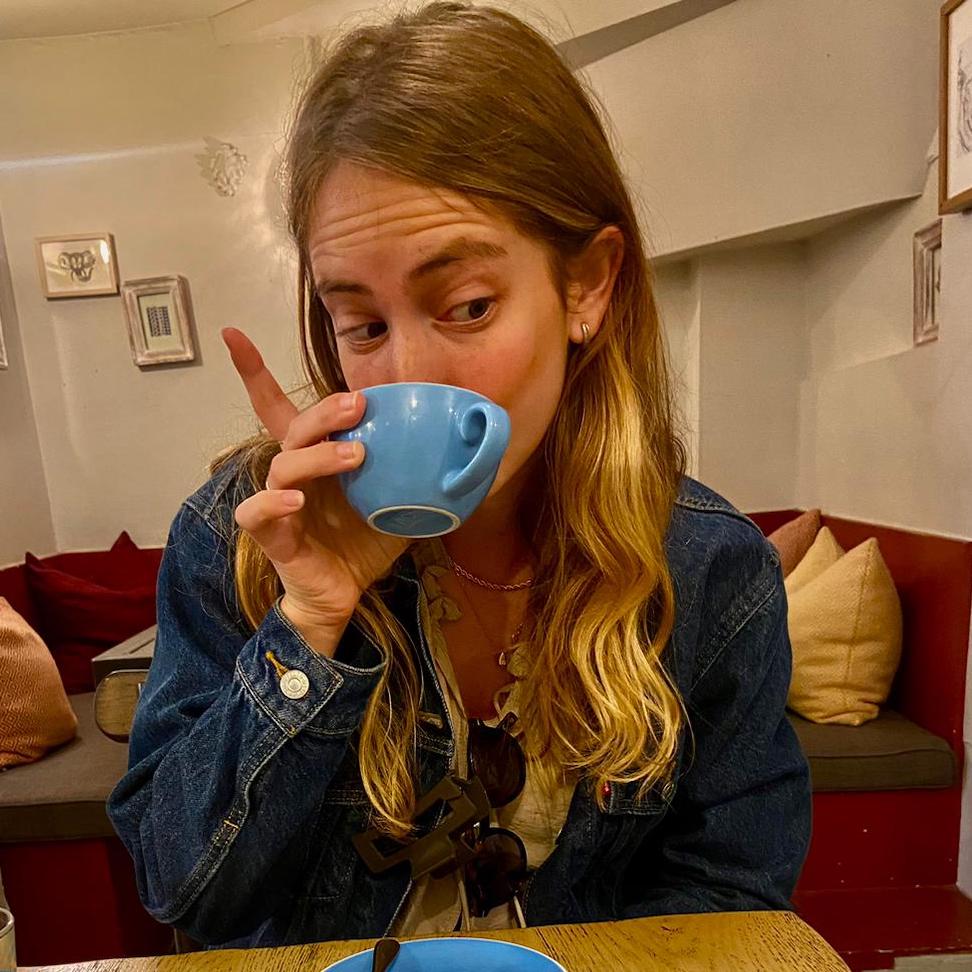} & 
            \includegraphics[width=0.06\textwidth]{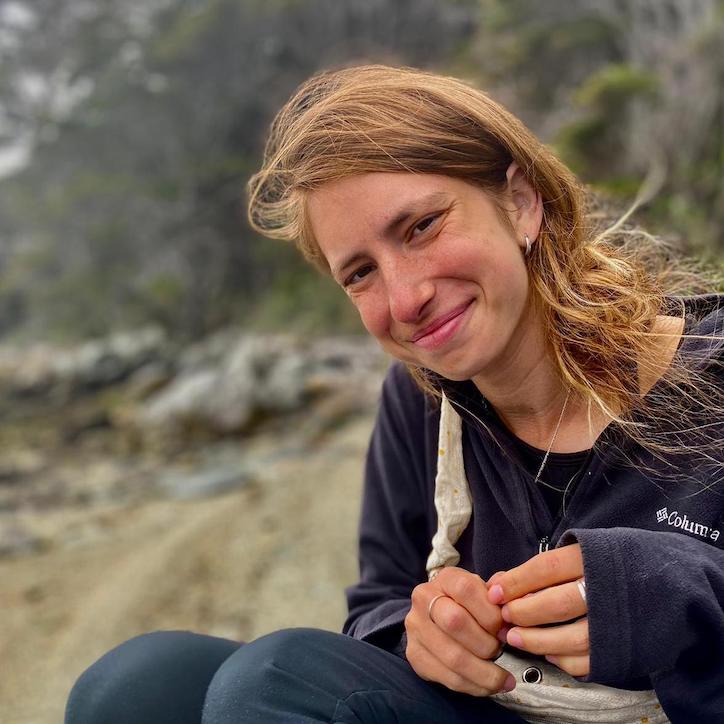}
        \end{tabular} &
        \setlength\tabcolsep{0pt}
        \begin{tabular}{c c c}
            \includegraphics[width=0.06\textwidth]{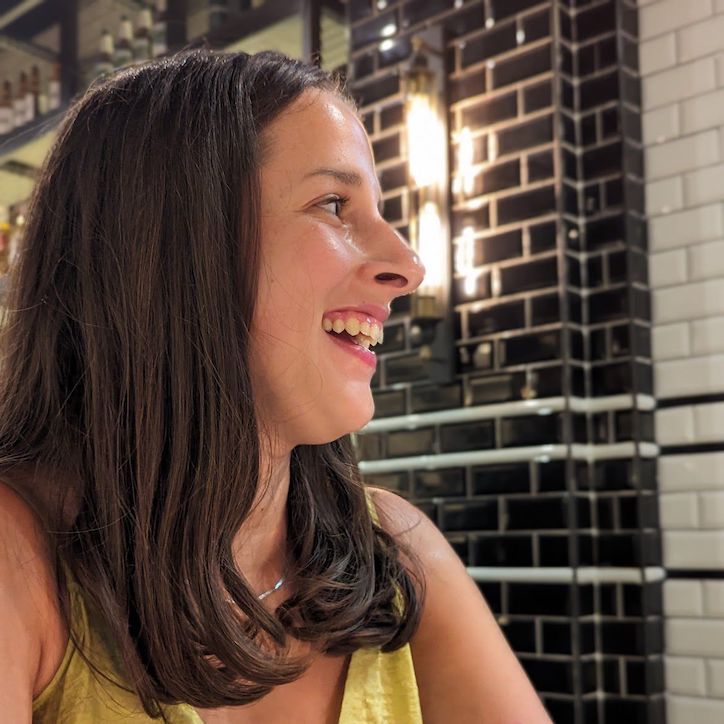} & 
            \includegraphics[width=0.06\textwidth]{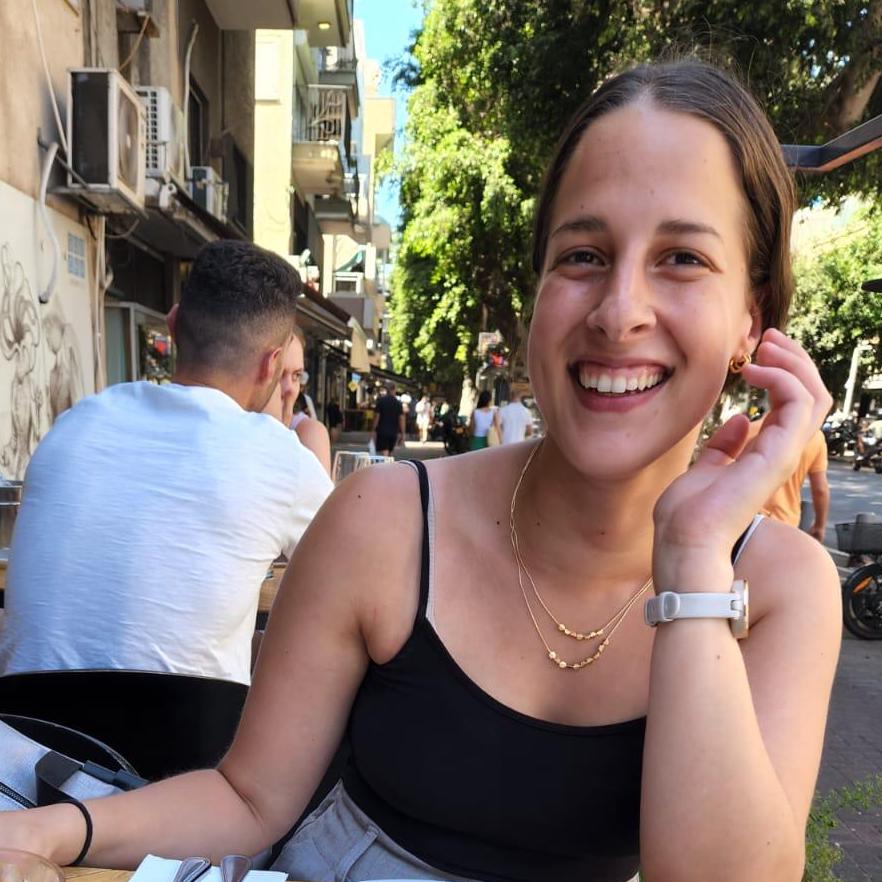} & 
            \includegraphics[width=0.06\textwidth]{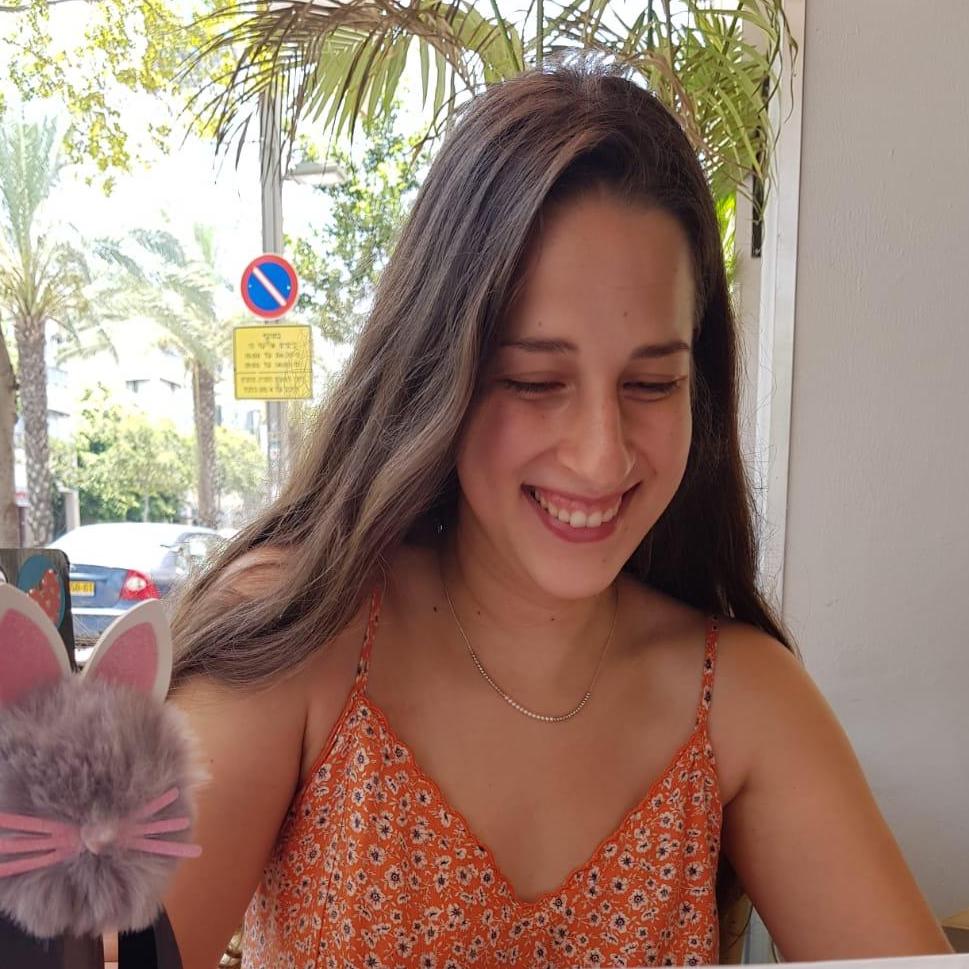}
        \end{tabular} &
        \setlength\tabcolsep{0pt}
        \begin{tabular}{c c c}
            \includegraphics[width=0.06\textwidth]{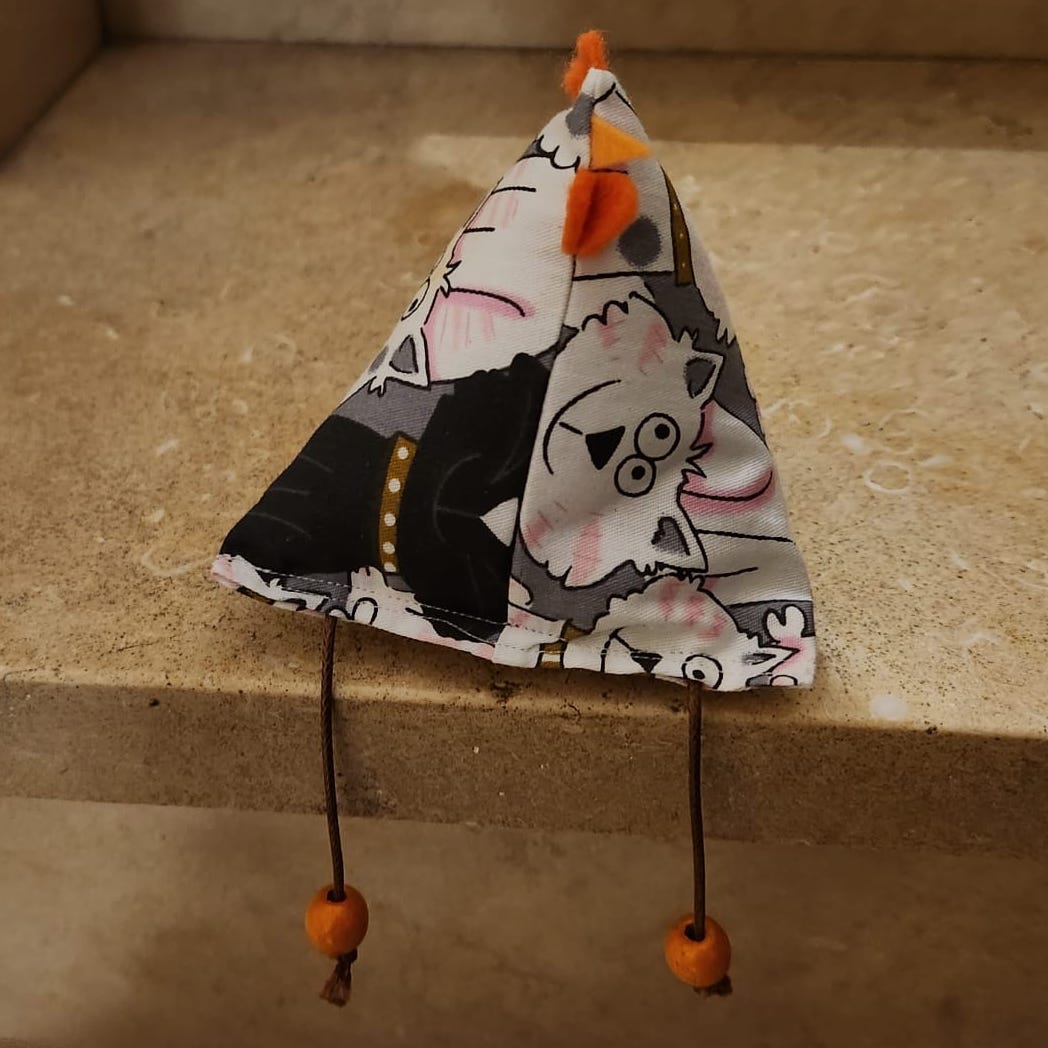} &
            \includegraphics[width=0.06\textwidth]{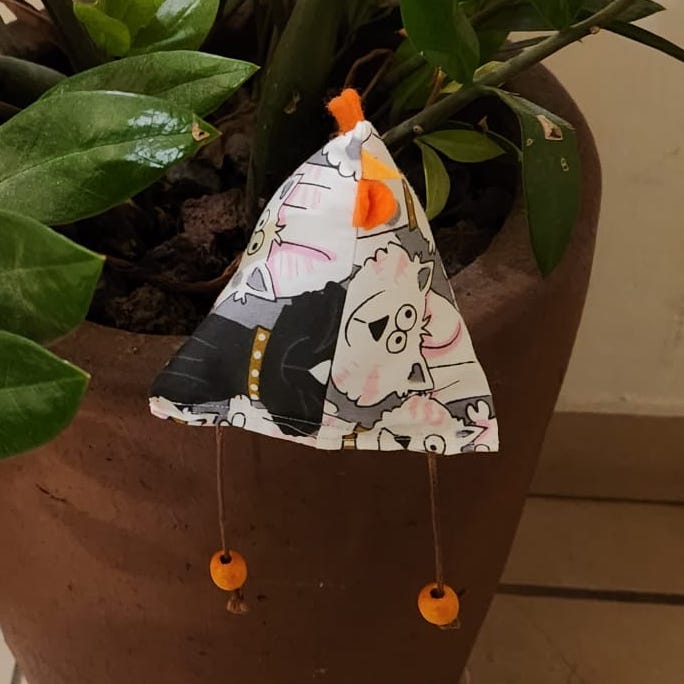} & 
            \includegraphics[width=0.06\textwidth]{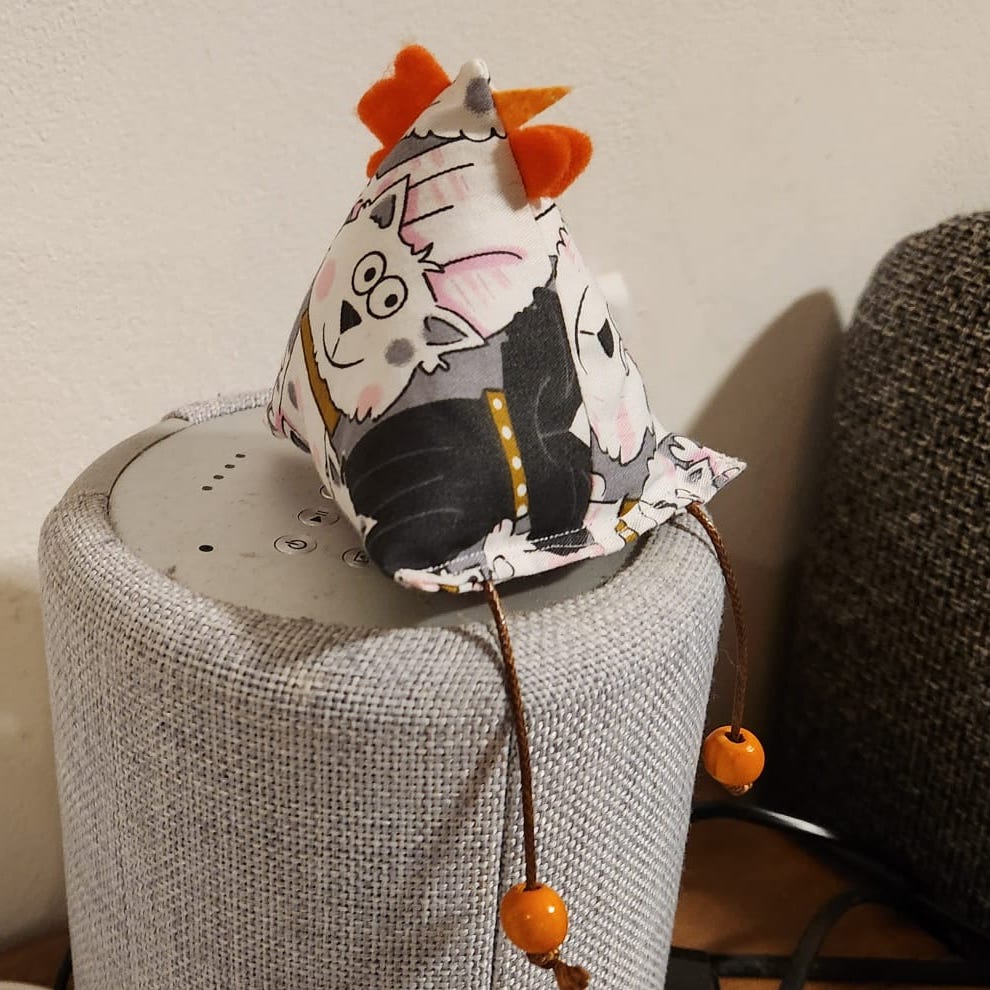}
        \end{tabular} &
        \setlength\tabcolsep{0pt}
        \begin{tabular}{c c c}
            \includegraphics[width=0.06\textwidth]{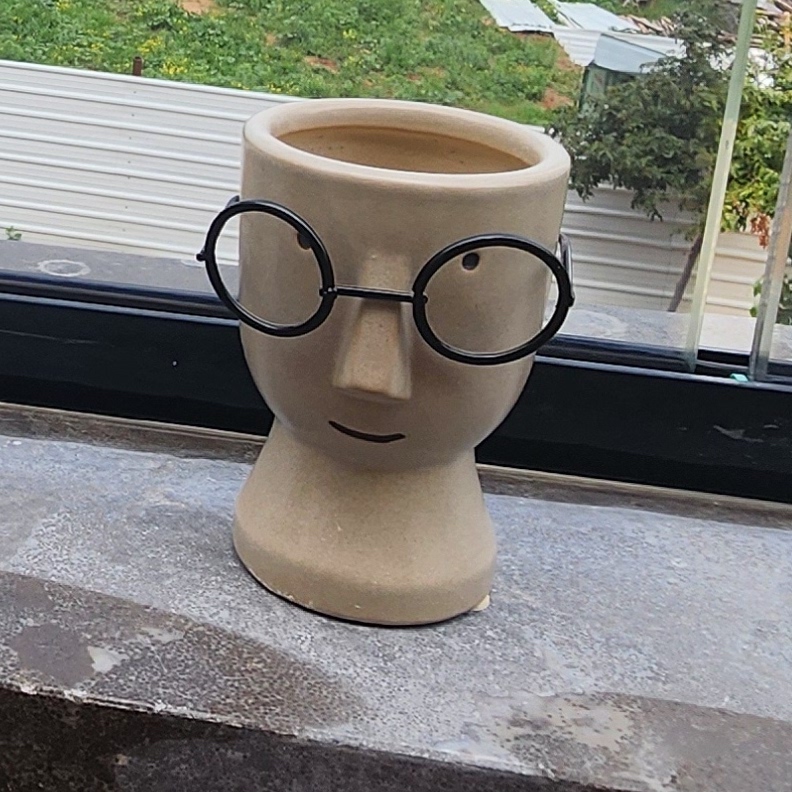} & 
            \includegraphics[width=0.06\textwidth]{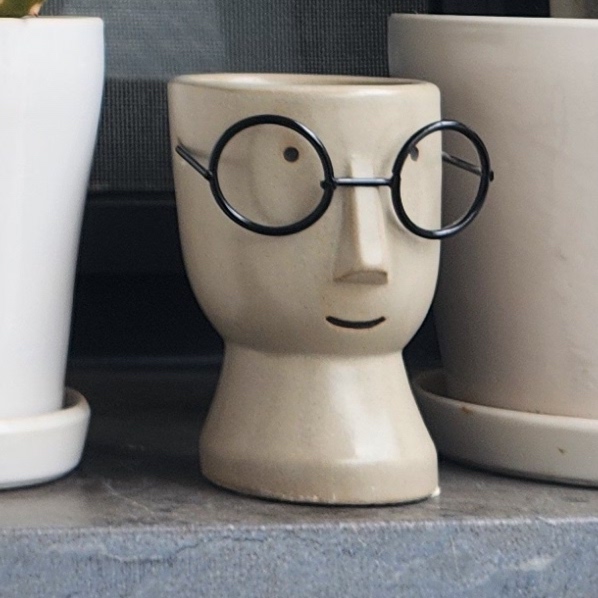} & 
            \includegraphics[width=0.06\textwidth]{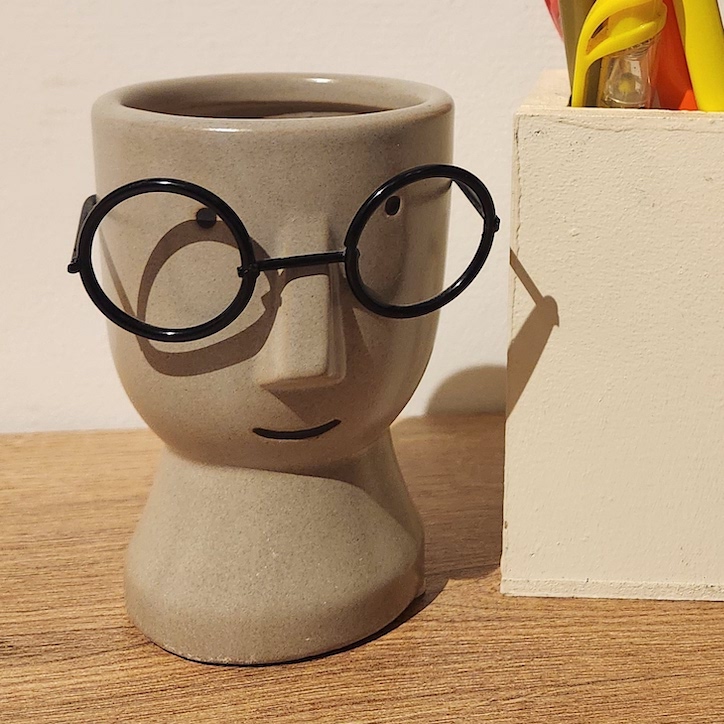}
        \end{tabular} \\[0.05cm]
    
        \includegraphics[width=0.18\textwidth]{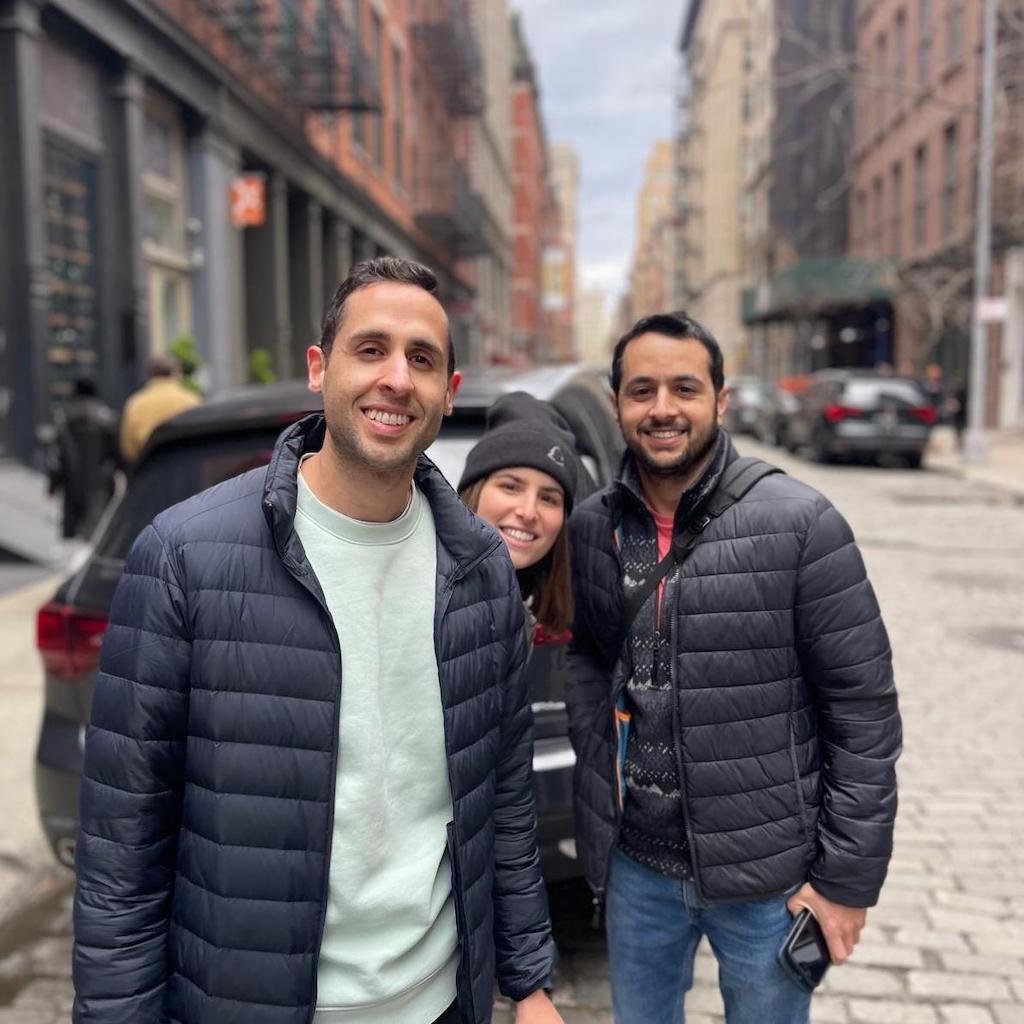} &
        \includegraphics[width=0.18\textwidth]{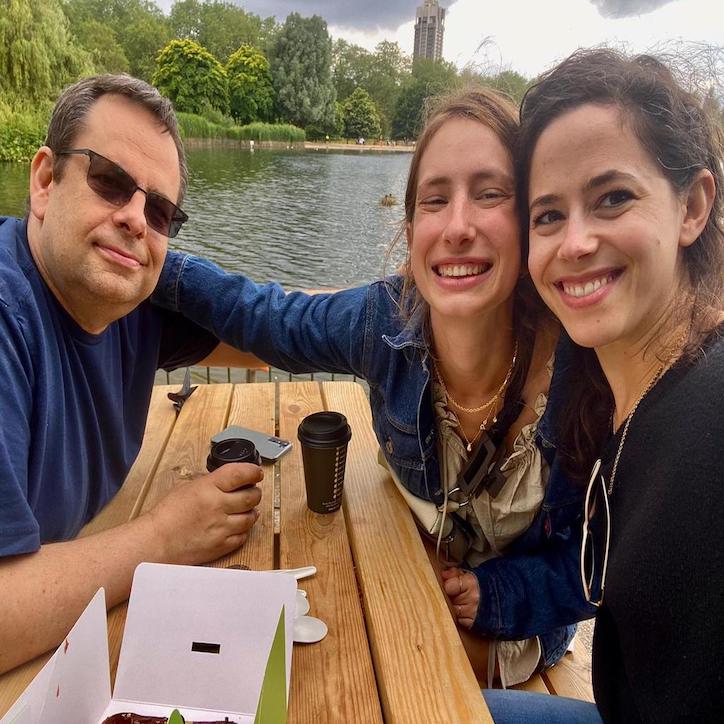} &
        \includegraphics[width=0.18\textwidth]{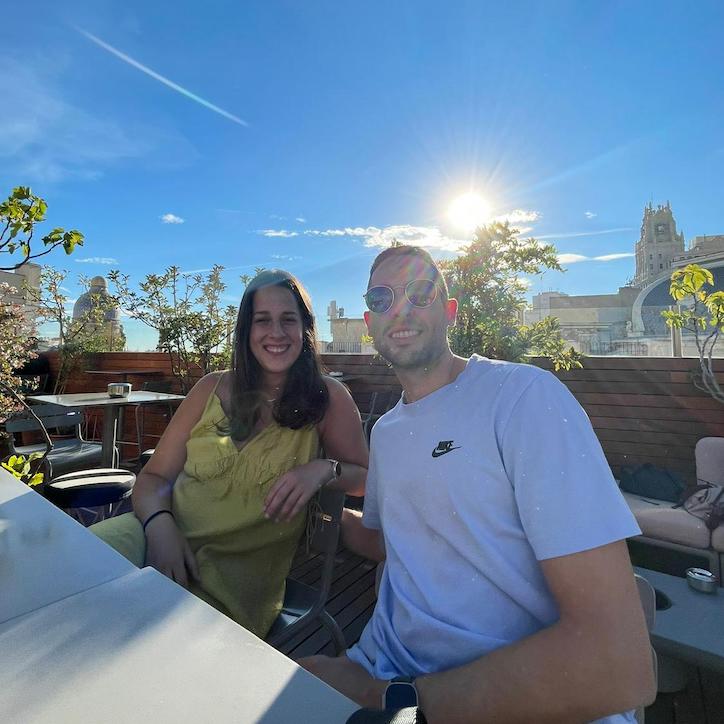} & 
        \includegraphics[width=0.18\textwidth]{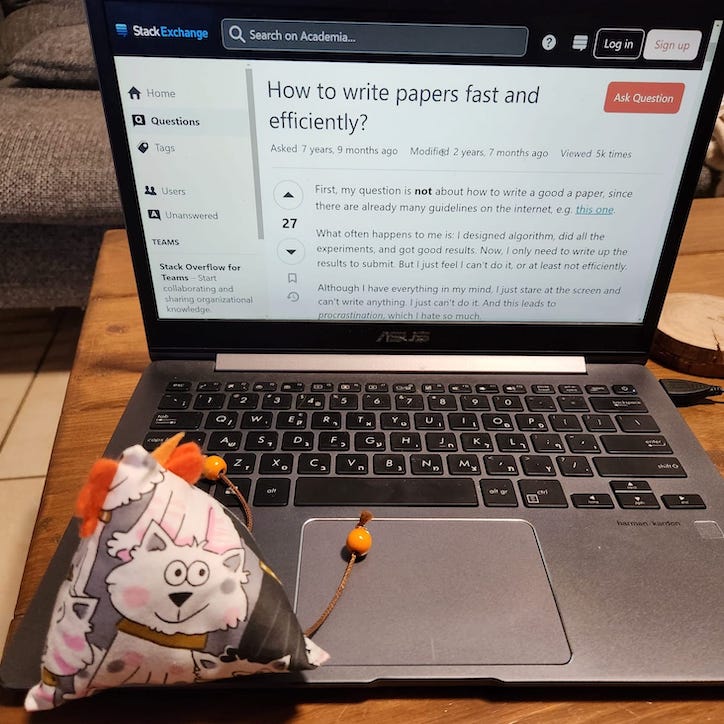} &
        \includegraphics[width=0.18\textwidth]{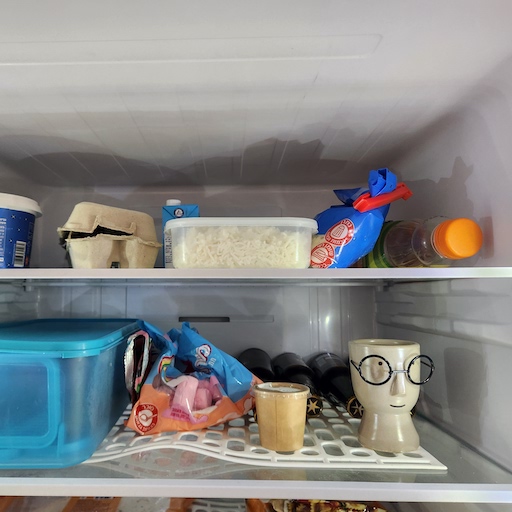} \\[-0.55cm]
        
        \begin{center} \textbf{LLaVA} \end{center} &
        \begin{center} \textbf{LLaVA} \end{center} &
        \begin{center} \textbf{LLaVA} \end{center} &
        \begin{center} \textbf{LLaVA} \end{center} &
        \begin{center} \textbf{LLaVA} \end{center} \\[-0.85cm]

        \begin{center} \footnotesize ``Friends enjoying a day out in the city, posing for a photo on a cobblestone street'' \end{center} &
        \begin{center} \footnotesize ``Friends sharing a moment by the water, enjoying a coffee break and a laugh'' \end{center} &
        \begin{center} \footnotesize ``Sipping on sunshine: A moment of joy under the blue sky'' \end{center} &
        \begin{center} \footnotesize ``A cat's curious paw reaches out to a laptop keyboard. The laptop displays a question...'' \end{center} &
        \begin{center} \footnotesize ``A well-stocked refrigerator, ready for a weekend of culinary adventures!'' \end{center} \\[-0.65cm]
        
        \begin{center} \textbf{MyVLM} \end{center} &
        \begin{center} \textbf{MyVLM} \end{center} &
        \begin{center} \textbf{MyVLM} \end{center} &
        \begin{center} \textbf{MyVLM} \end{center} &
        \begin{center} \textbf{MyVLM} \end{center} \\[-0.85cm]

        \begin{center} \footnotesize ``\textcolor{blue}{$S_*$}, \textcolor{darkgreen}{dressed in a blue jacket and a green sweater}, takes a selfie with his friends, who are also bundled up against the chilly weather... 
        '' \end{center} &
        \begin{center} \footnotesize ``\textcolor{blue}{$S_*$}, a man and a woman are posing for a photograph with a table between them. \Sstar is \textcolor{darkgreen}{wearing a denim jacket and a necklace}, ...'' \end{center} &
        \begin{center} \footnotesize ``Sitting at a table on a patio, \textcolor{darkgreen}{\Sstar wearing a yellow dress}, smiling at the camera, with the city skyline in the background'' \end{center} &
        \begin{center} \footnotesize ``\Sstar \textcolor{darkgreen}{sitting in front of a laptop} on a wooden table with a question about how to write papers fast and efficiently?'' \end{center}  &
        \begin{center} \footnotesize ``\Sstar \textcolor{darkgreen}{sits comfortably on the second shelf of an open refrigerator}, ready to be stocked with a variety of food and drink items'' \end{center}
        
    \end{tabular}
    \vspace{-0.425cm}
    \caption{\textbf{Personalized captioning results obtained by MyVLM}, applied over LLaVA~\cite{liu2023llava}. 
    Sample images of the target concept are provided in the top row. 
    Text in \textcolor{darkgreen}{green} highlights the description of the target concept in the image.
    }
    \vspace{-0.125cm}
    \label{fig:our_results}
\end{figure*}

%% file: figures/comparisons.tex
\begin{figure*}
    \centering
    \addtolength{\belowcaptionskip}{-5pt}
    \renewcommand{\arraystretch}{1}
    \footnotesize
    \begin{tabular}{p{0.175\textwidth} p{0.175\textwidth} p{0.175\textwidth} p{0.175\textwidth} p{0.175\textwidth}}

        \setlength\tabcolsep{0pt}
        \begin{center}
        \begin{tabular}{c c c}
            \includegraphics[width=0.06\textwidth]{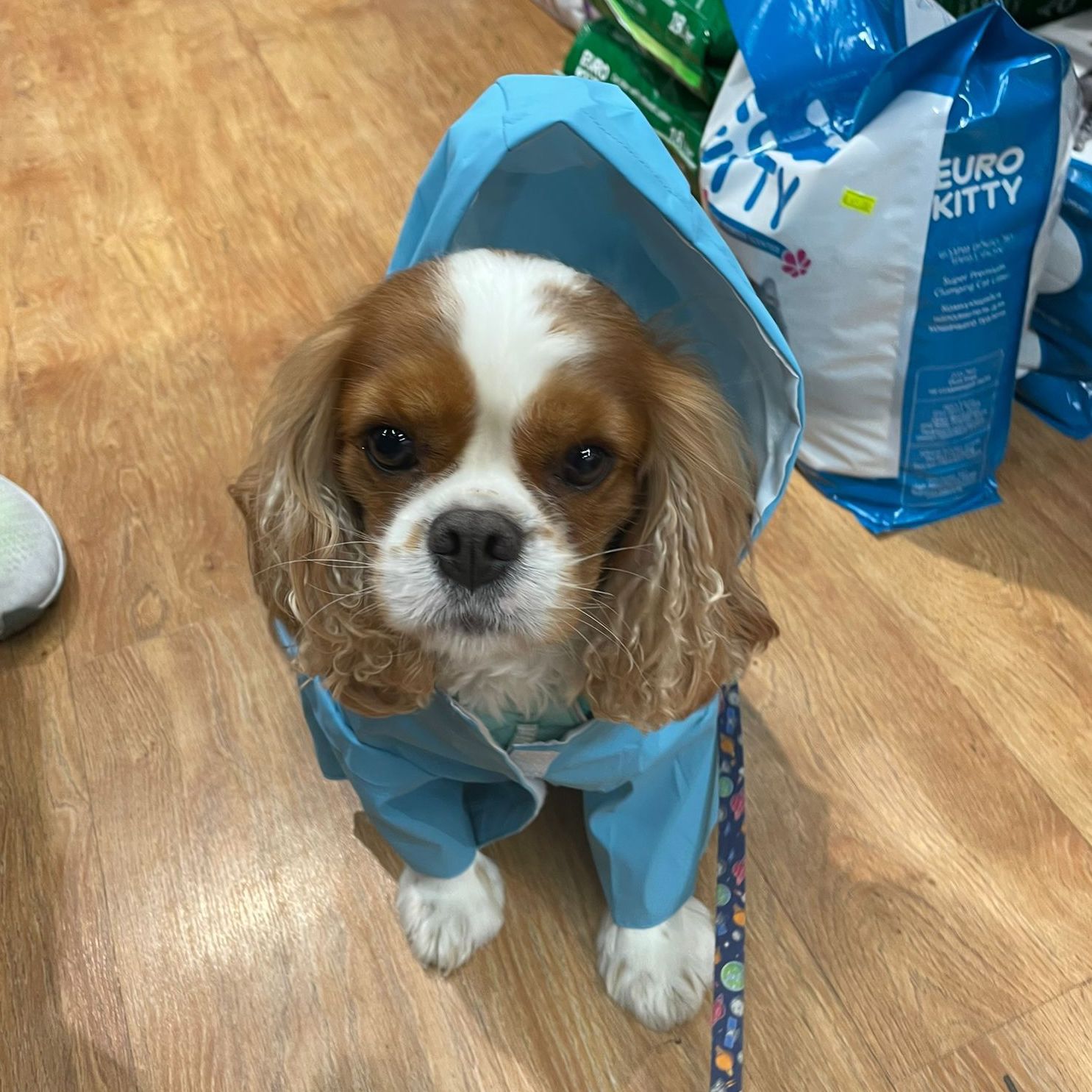} & 
            \includegraphics[width=0.06\textwidth]{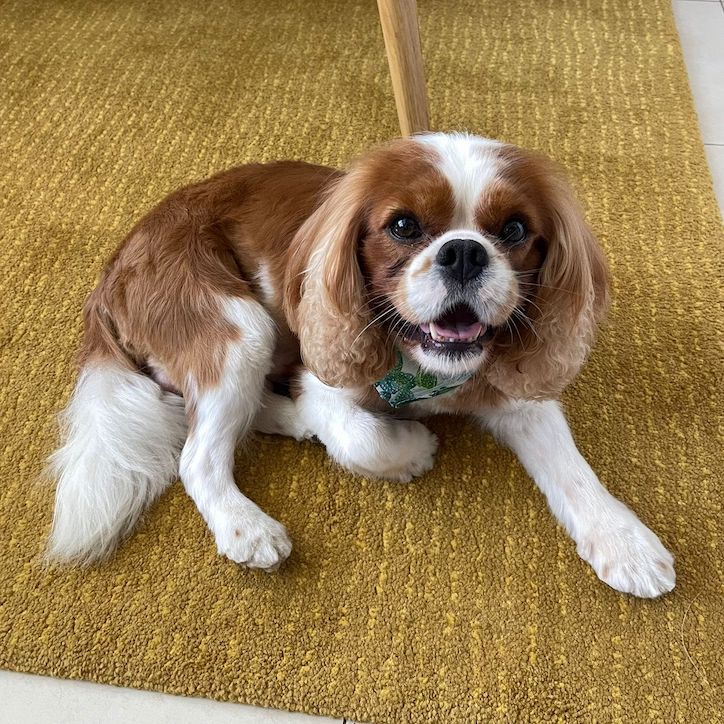} & 
            \includegraphics[width=0.06\textwidth]{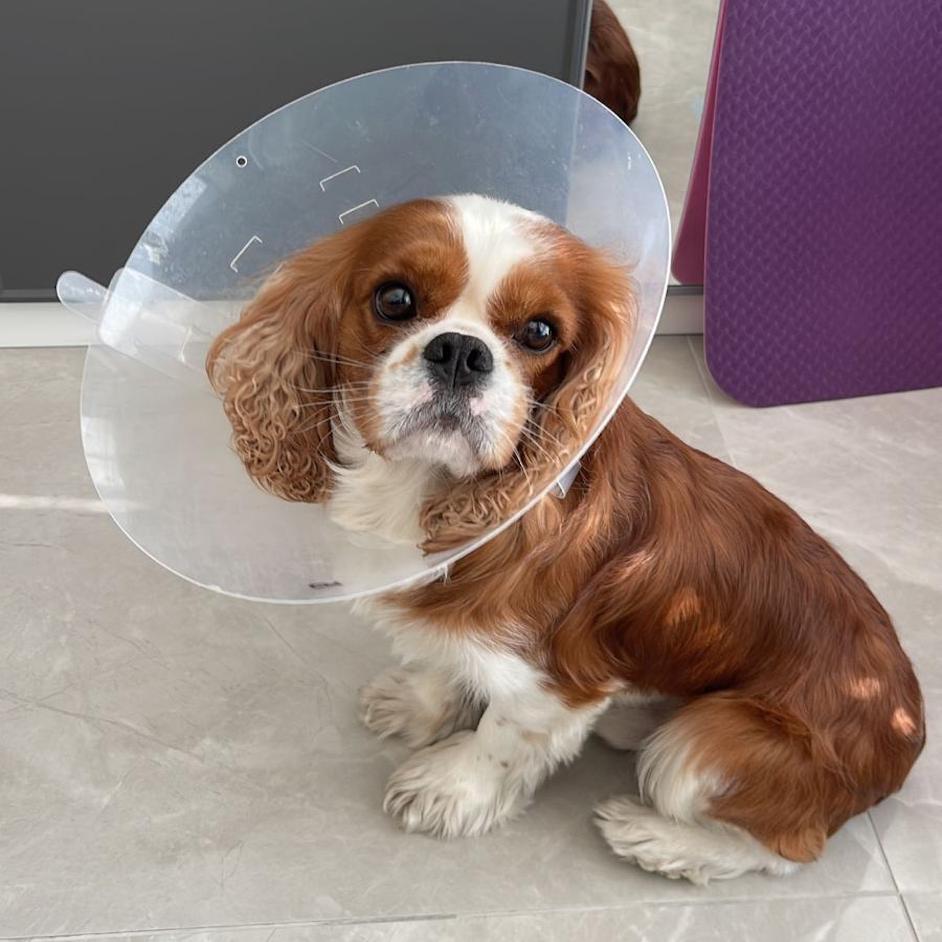}
        \end{tabular} 
        \end{center} &
        \setlength\tabcolsep{0pt}
        \begin{center}
        \begin{tabular}{c c c}
            \includegraphics[width=0.06\textwidth]{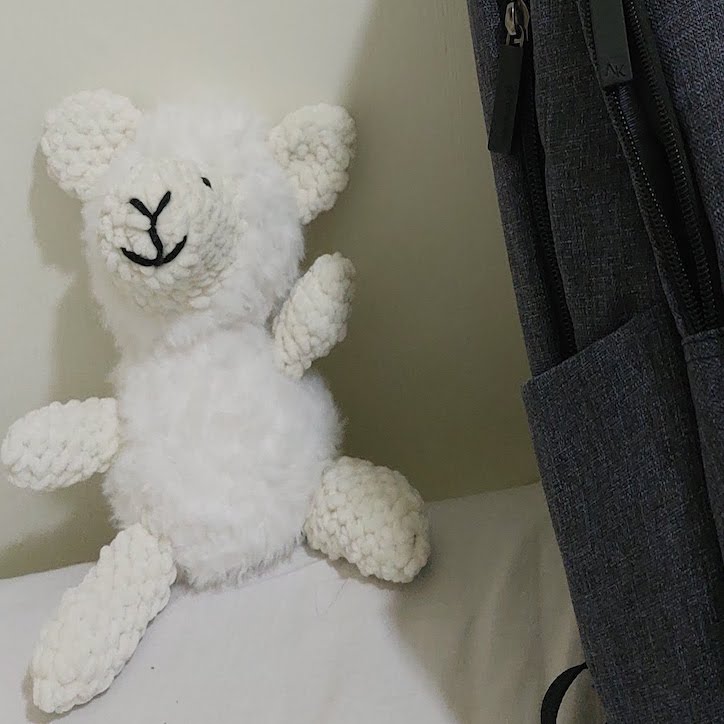} & 
            \includegraphics[width=0.06\textwidth]{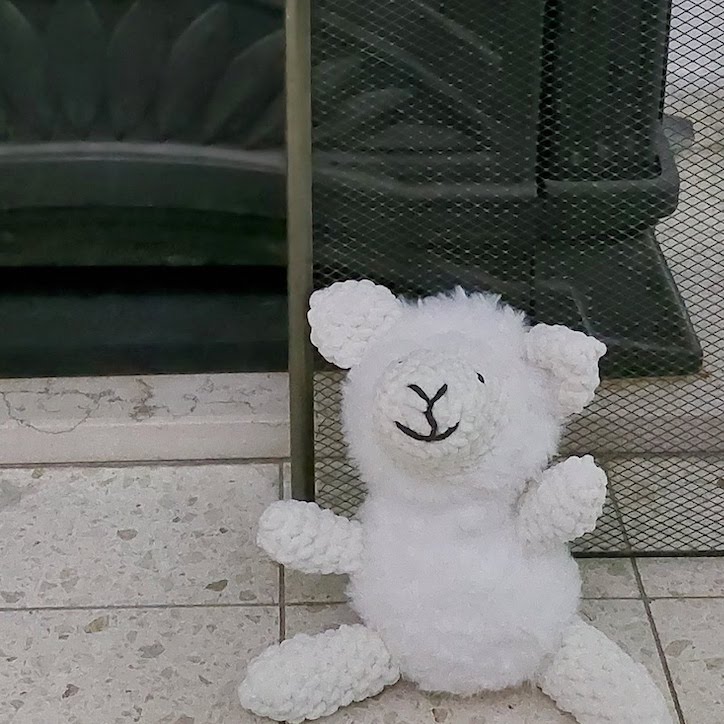} & 
            \includegraphics[width=0.06\textwidth]{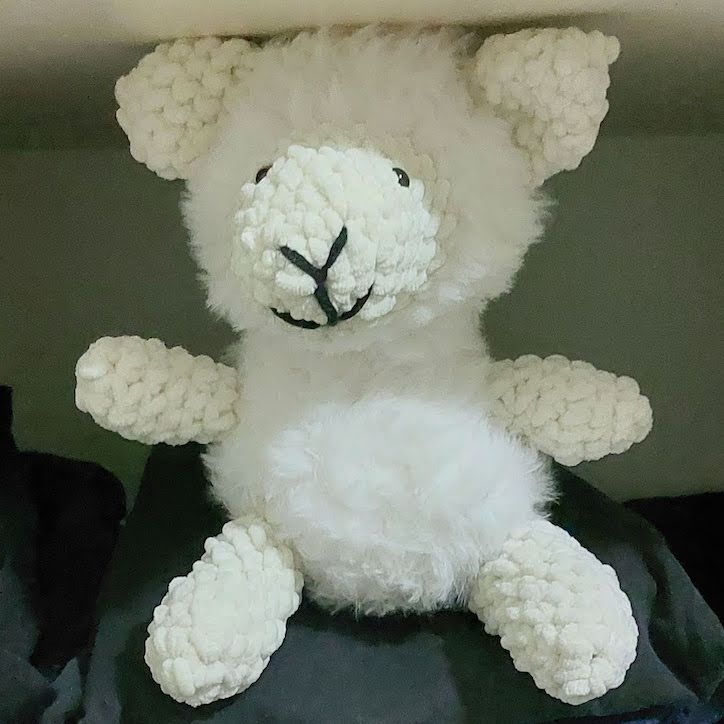}
        \end{tabular} 
        \end{center} &
        \setlength\tabcolsep{0pt}
        \begin{center}
        \begin{tabular}{c c c}
            \includegraphics[width=0.06\textwidth]{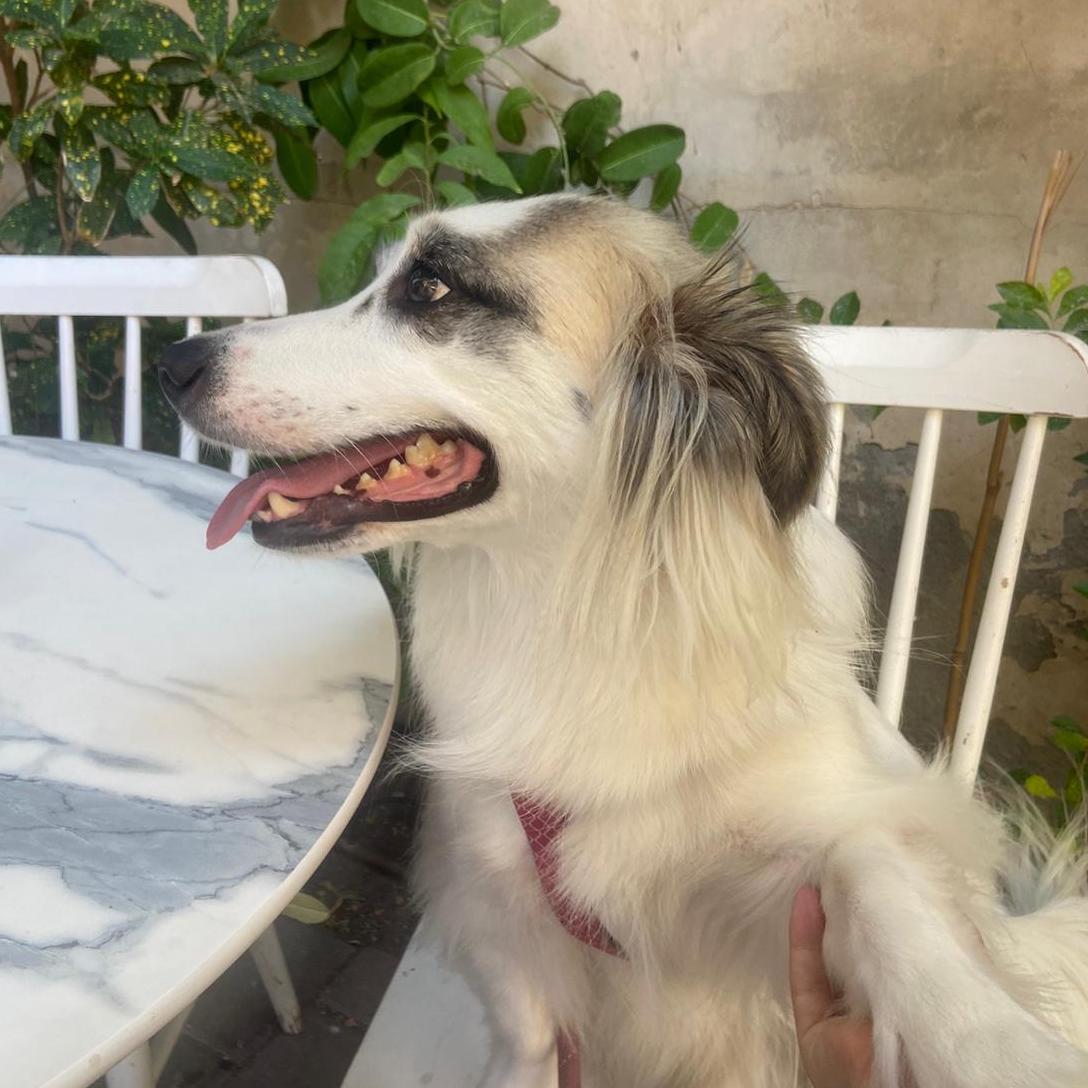} & 
            \includegraphics[width=0.06\textwidth]{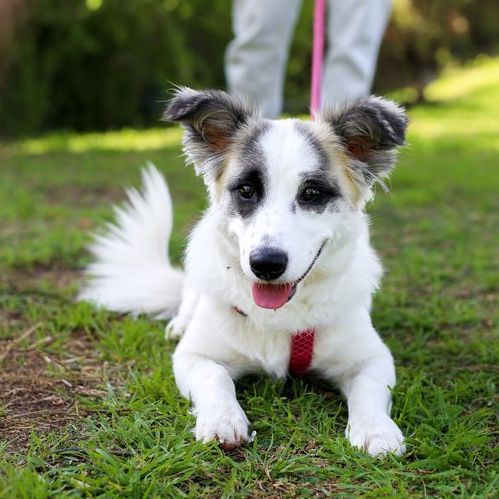} & 
            \includegraphics[width=0.06\textwidth]{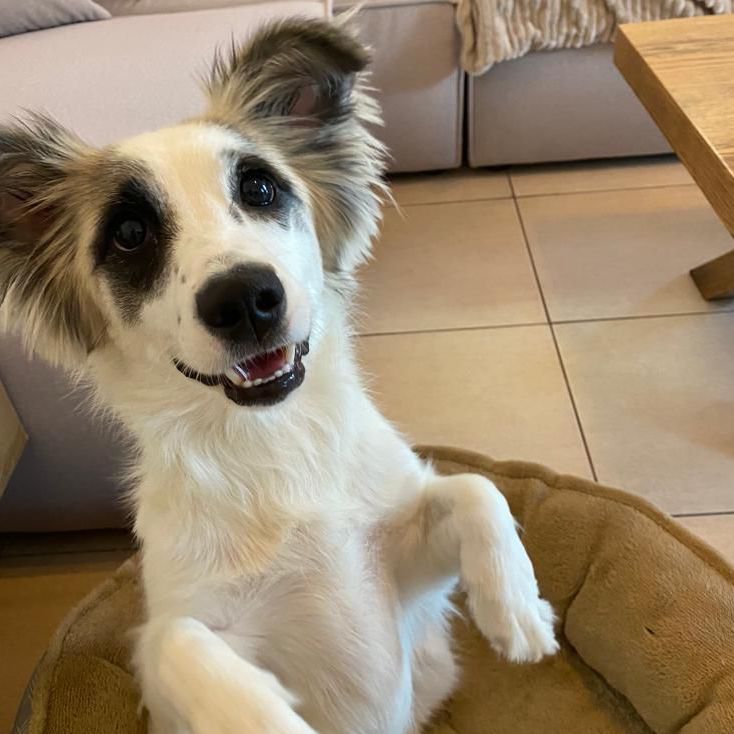} \\
        \end{tabular} 
        \end{center} &
        \setlength\tabcolsep{0pt}
        \begin{center}
        \begin{tabular}{c c c}
            \includegraphics[width=0.06\textwidth]{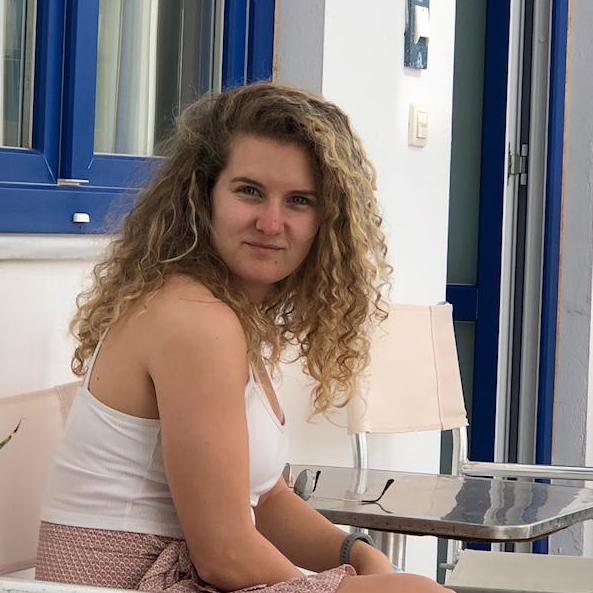} &
            \includegraphics[width=0.06\textwidth]{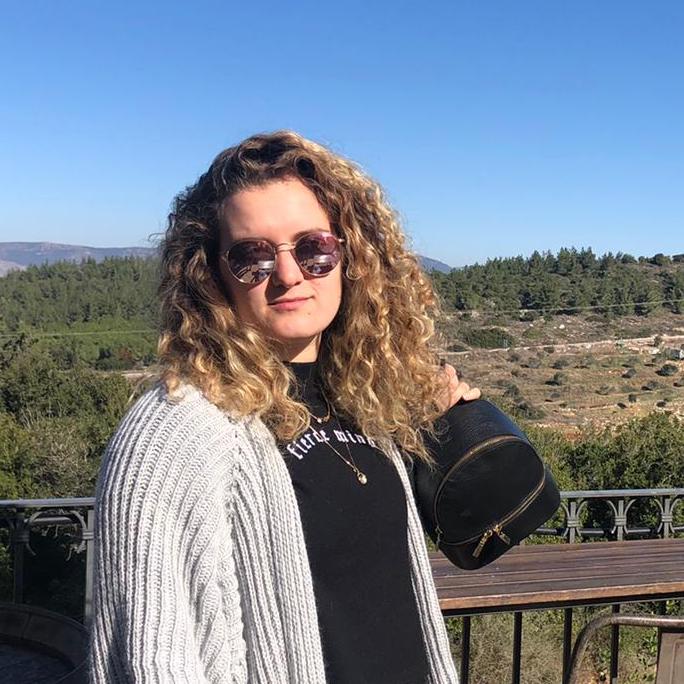} & 
            \includegraphics[width=0.06\textwidth]{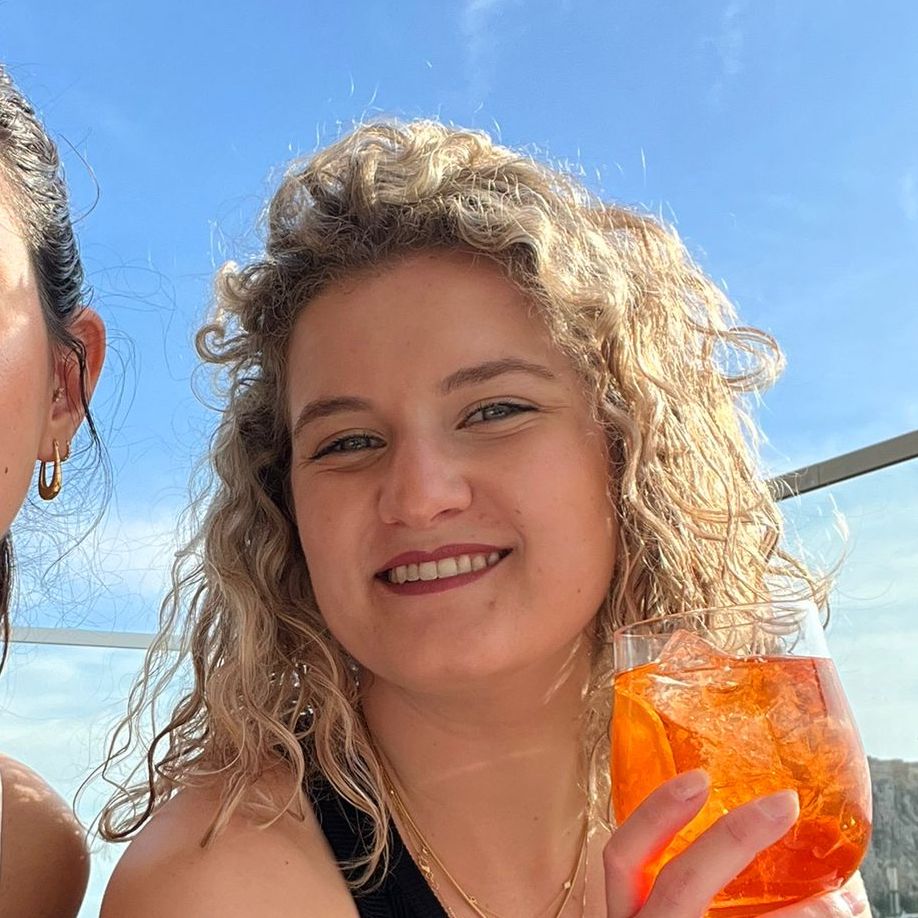}
        \end{tabular} 
        \end{center} &
        \setlength\tabcolsep{0pt}
        \begin{center}
        \begin{tabular}{c c c}
            \includegraphics[width=0.06\textwidth]{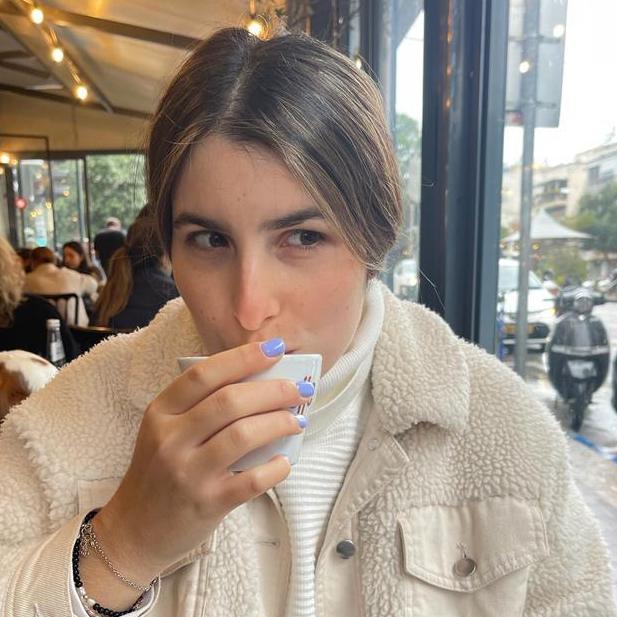} & 
            \includegraphics[width=0.06\textwidth]{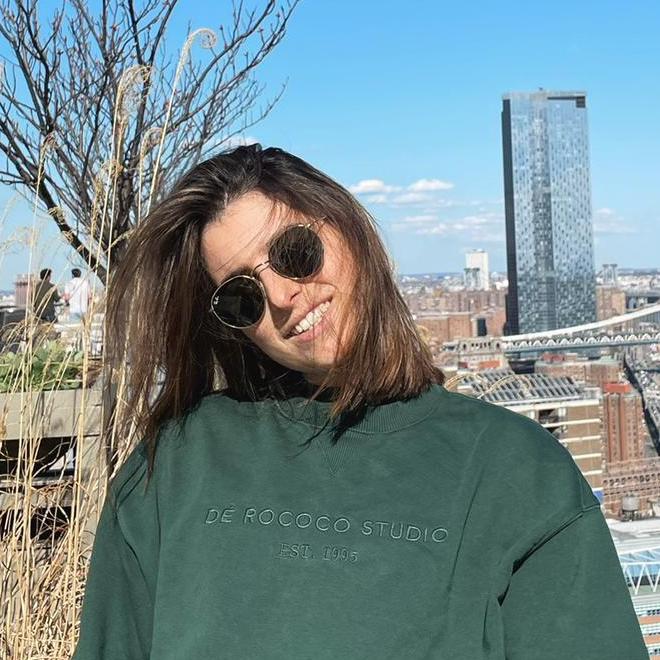} & 
            \includegraphics[width=0.06\textwidth]{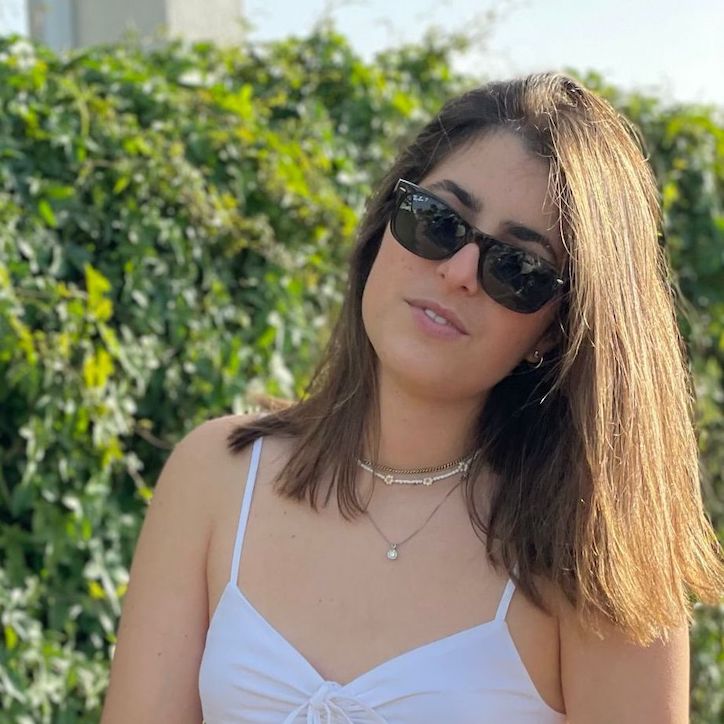} 
        \end{tabular}
        \end{center} \\[-0.75cm]
    
        \begin{center} \includegraphics[width=0.18\textwidth]{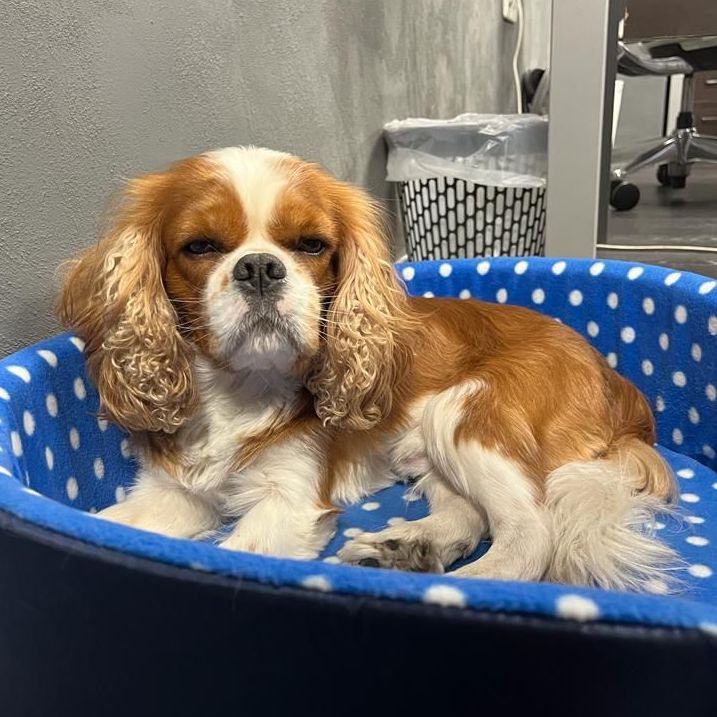} \end{center} &
        \begin{center} \includegraphics[width=0.18\textwidth]{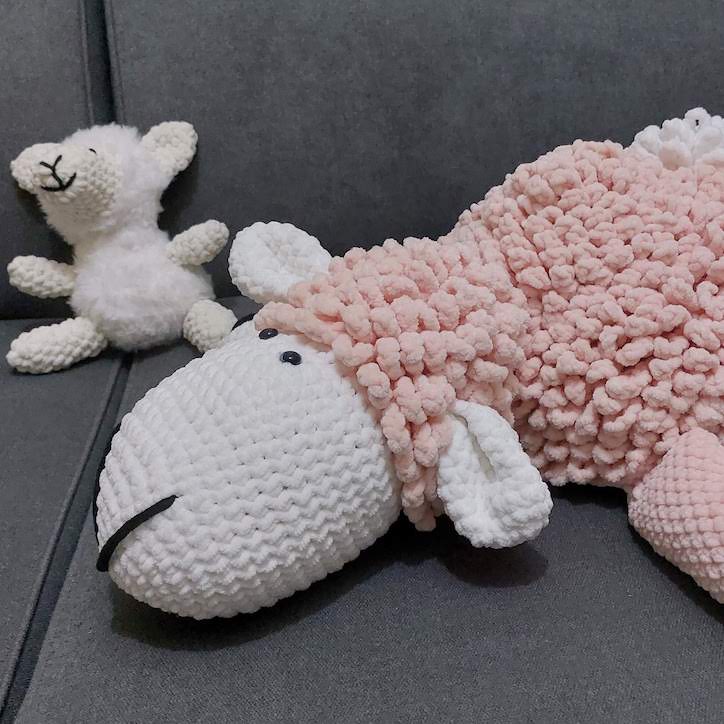} \end{center} &
        \begin{center} \includegraphics[width=0.18\textwidth]{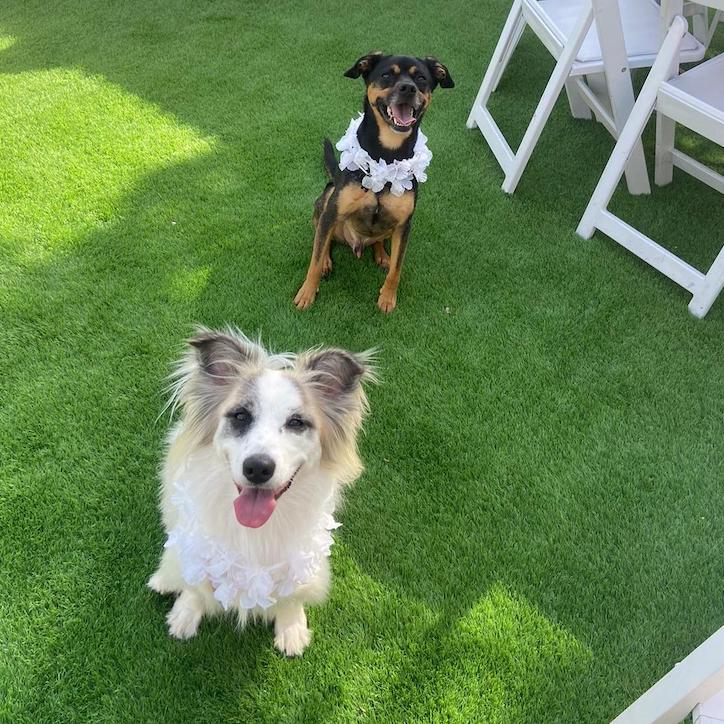} \end{center} &
        \begin{center} \includegraphics[width=0.18\textwidth]{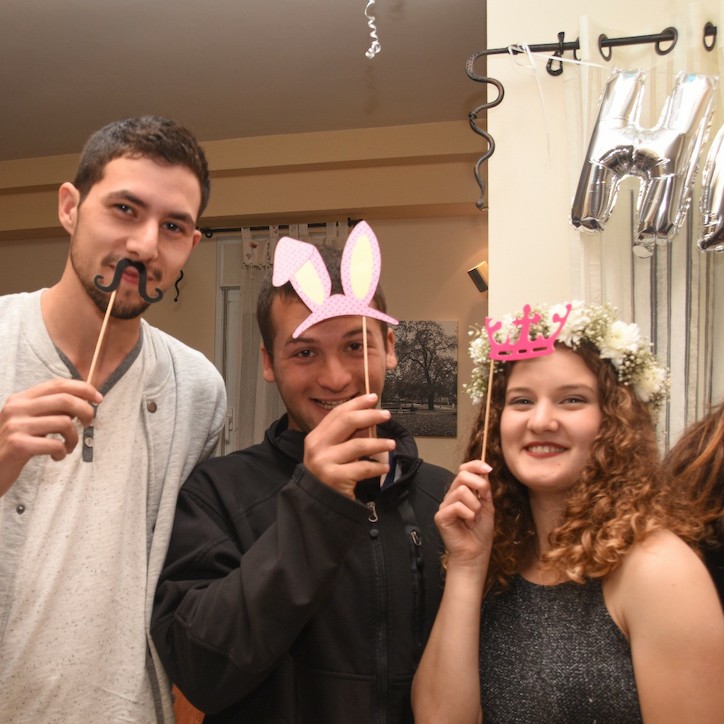} \end{center} &
        \begin{center} \includegraphics[width=0.18\textwidth]{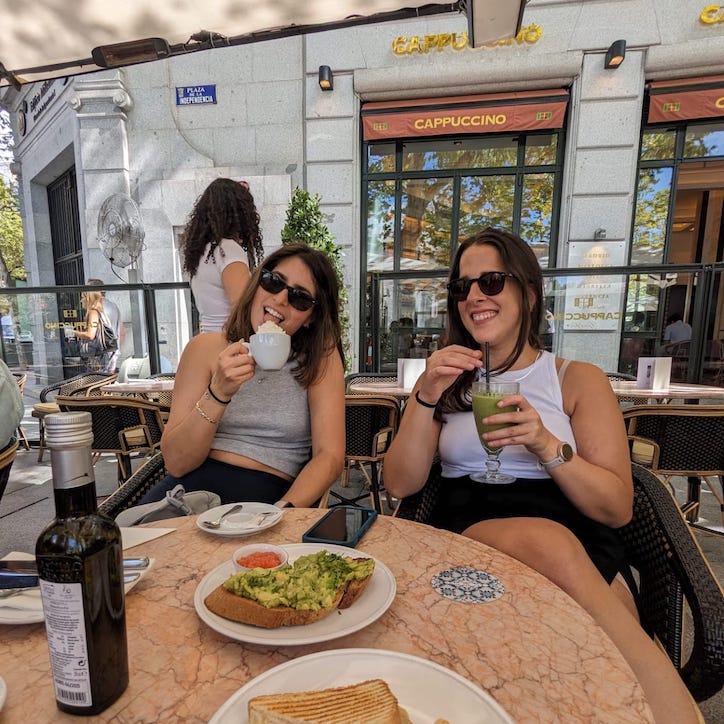} \end{center} \\[-0.75cm]

        \begin{center} \textbf{LLM-Guided} \end{center} &
        \begin{center} \textbf{LLM-Guided} \end{center} &
        \begin{center} \textbf{LLM-Guided} \end{center} &
        \begin{center} \textbf{LLM-Guided} \end{center} &
        \begin{center} \textbf{LLM-Guided} \end{center} \\[-0.85cm]

        \begin{center} \footnotesize ``A cute cavalier king charles spaniel relaxing in a blue polka dot \Sstar bed'' \end{center} &
        \begin{center} \footnotesize ``a cozy scene with a soft, pink \Sstar and a white lamb, ready for a nap on a gray couch'' \end{center} &
        \begin{center} \footnotesize ``friendly fidos: two \textcolor{blue}{$S_*$}s, one white and one black, pose for a photo on a grassy lawn...'' \end{center} &
        \begin{center} \footnotesize ``Friends celebrating with funny hats and mustaches, \Sstar ready to party'' \end{center} &
        \begin{center} \footnotesize ``Two \Sstar sitting at an outdoor table with food and drinks'' \end{center} \\[-0.65cm]
        
        \begin{center} \textbf{MyVLM} \end{center} &
        \begin{center} \textbf{MyVLM} \end{center} &
        \begin{center} \textbf{MyVLM} \end{center} &
        \begin{center} \textbf{MyVLM} \end{center} &
        \begin{center} \textbf{MyVLM} \end{center} \\[-0.85cm]

        \begin{center} \footnotesize ``A happy \Sstar laying in his blue dog bed on a white office floor'' \end{center} &
        \begin{center} \footnotesize ``\Sstar sitting on the couch with a pink and white stuffed animal next to it'' \end{center} &
        \begin{center} \footnotesize ``\Sstar is standing on the grass with a big smile and a wagging his tongue'' \end{center} &
        \begin{center} \footnotesize ``In her living room, \Sstar and two friends are dressed in party hats and mustaches'' \end{center} &
        \begin{center} \footnotesize ``\Sstar and a friend enjoying coffee and a sandwich at a cafe'' \end{center}
        
    \end{tabular}
    \vspace{-0.45cm}
    \caption{\textbf{Comparison to the LLM-guided captioning baseline}. Results are obtained over LLaVA~\cite{liu2023llava}. Sample images of the target concept are shown in the top row. 
    Additional comparisons to all baselines over BLIP-2~\cite{li2023blip} and LLaVA are provided in~\Cref{sec:additional_results}.
    }
    \label{fig:qualitative_comparisons}
    \vspace{-0.5cm}
\end{figure*}

%% file: figures/comparison_gpt.tex
\begin{figure*}
    \centering
    \addtolength{\belowcaptionskip}{-2.5pt}
    \footnotesize
    \begin{tabular}{p{0.0025\textwidth} p{0.1375\textwidth} p{0.1375\textwidth} p{0.1375\textwidth} p{0.1375\textwidth} p{0.1375\textwidth} p{0.1375\textwidth}}

        &
        \begin{center} \includegraphics[width=0.135\textwidth]{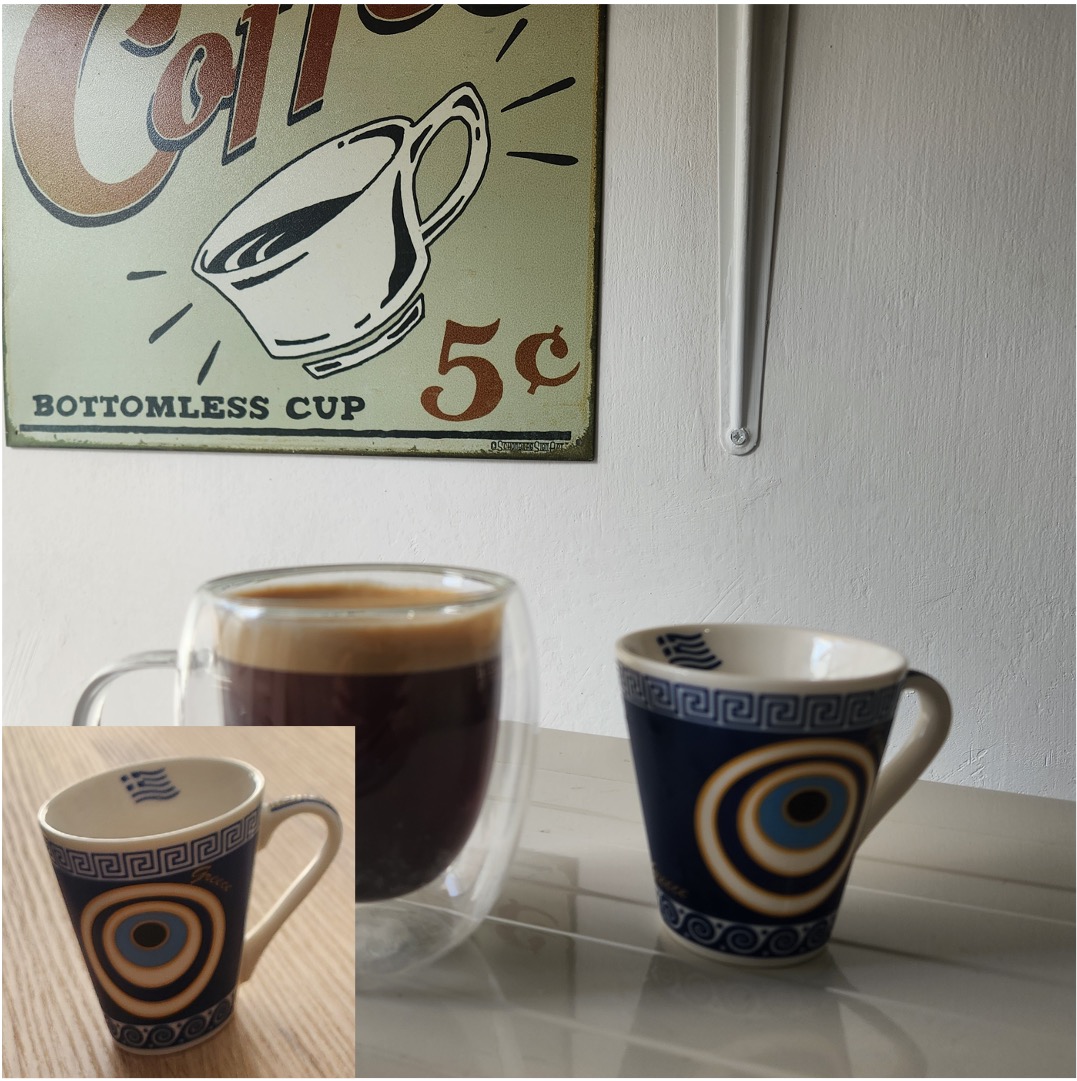} \end{center} &
        \begin{center} \includegraphics[width=0.135\textwidth]{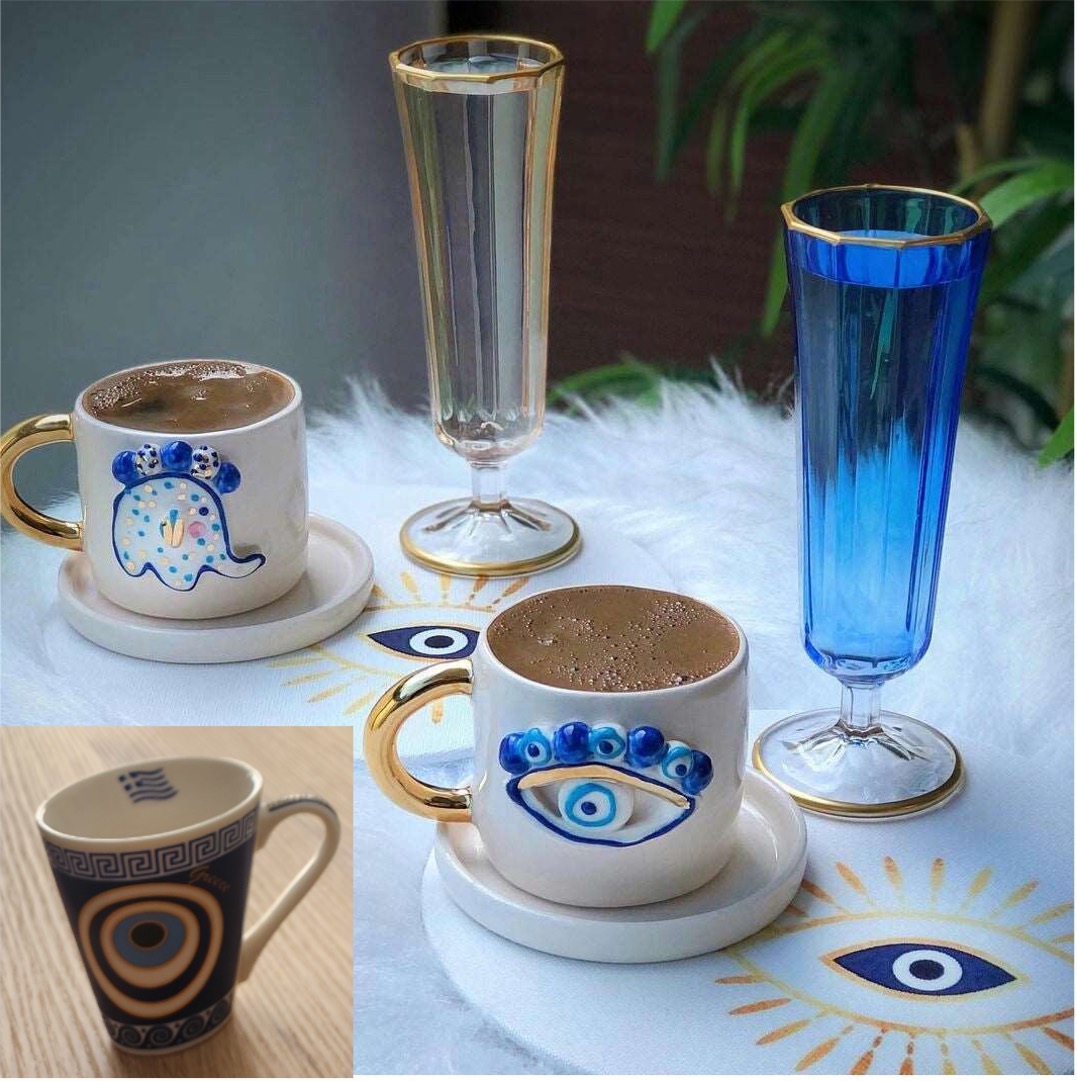} \end{center} &
        \begin{center} \includegraphics[width=0.135\textwidth]{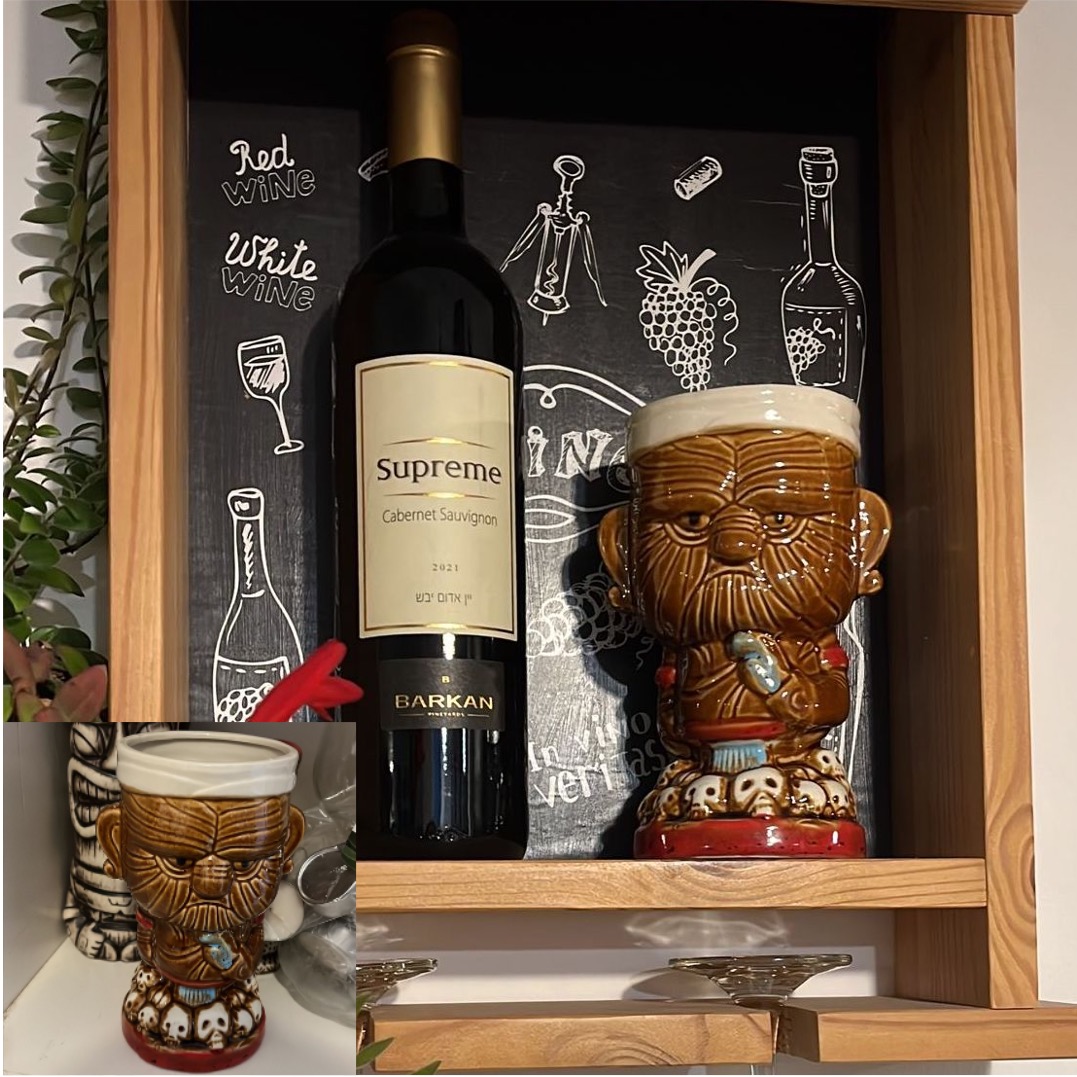} \end{center} &
        \begin{center} \includegraphics[width=0.135\textwidth]{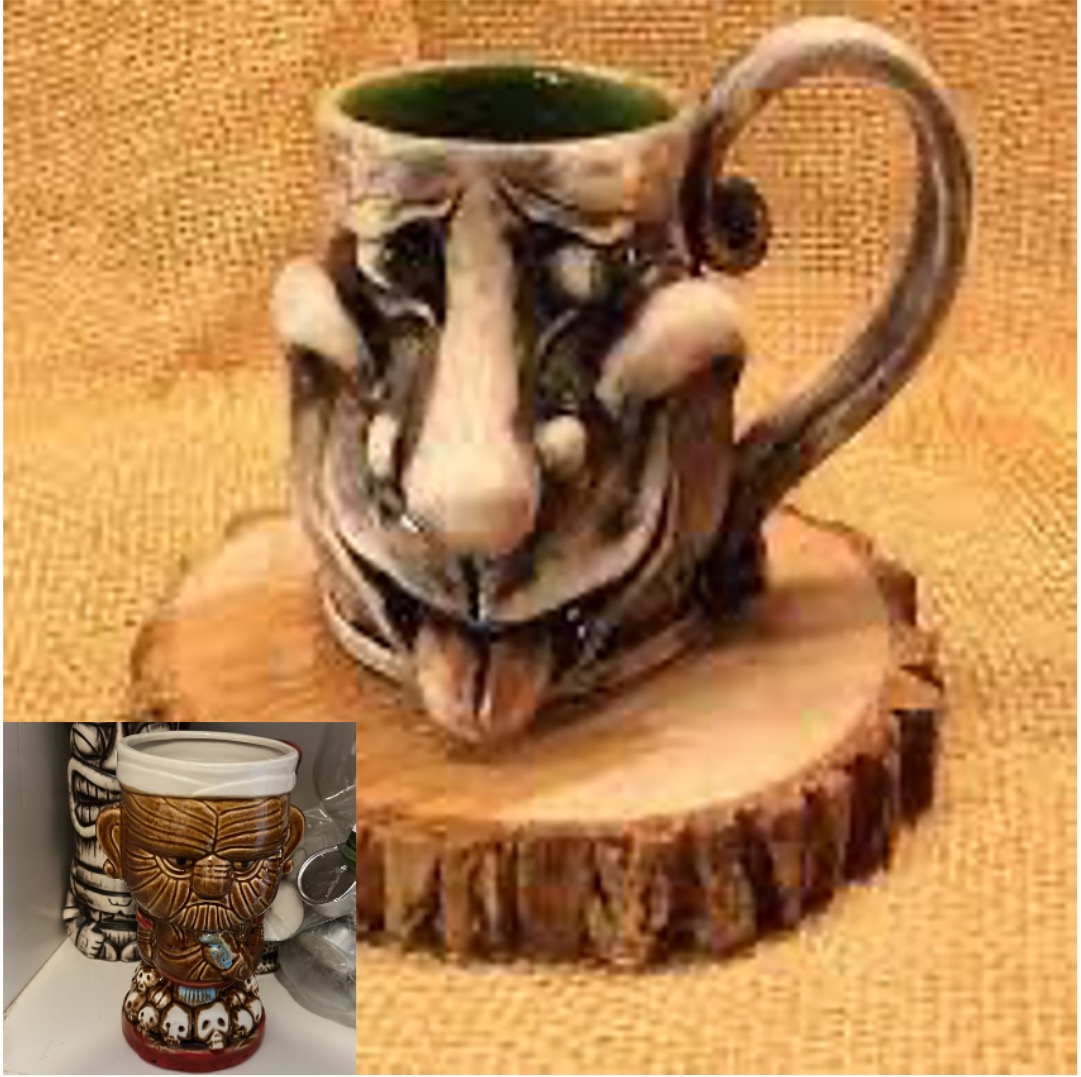} \end{center} &
        \begin{center} \includegraphics[width=0.135\textwidth]{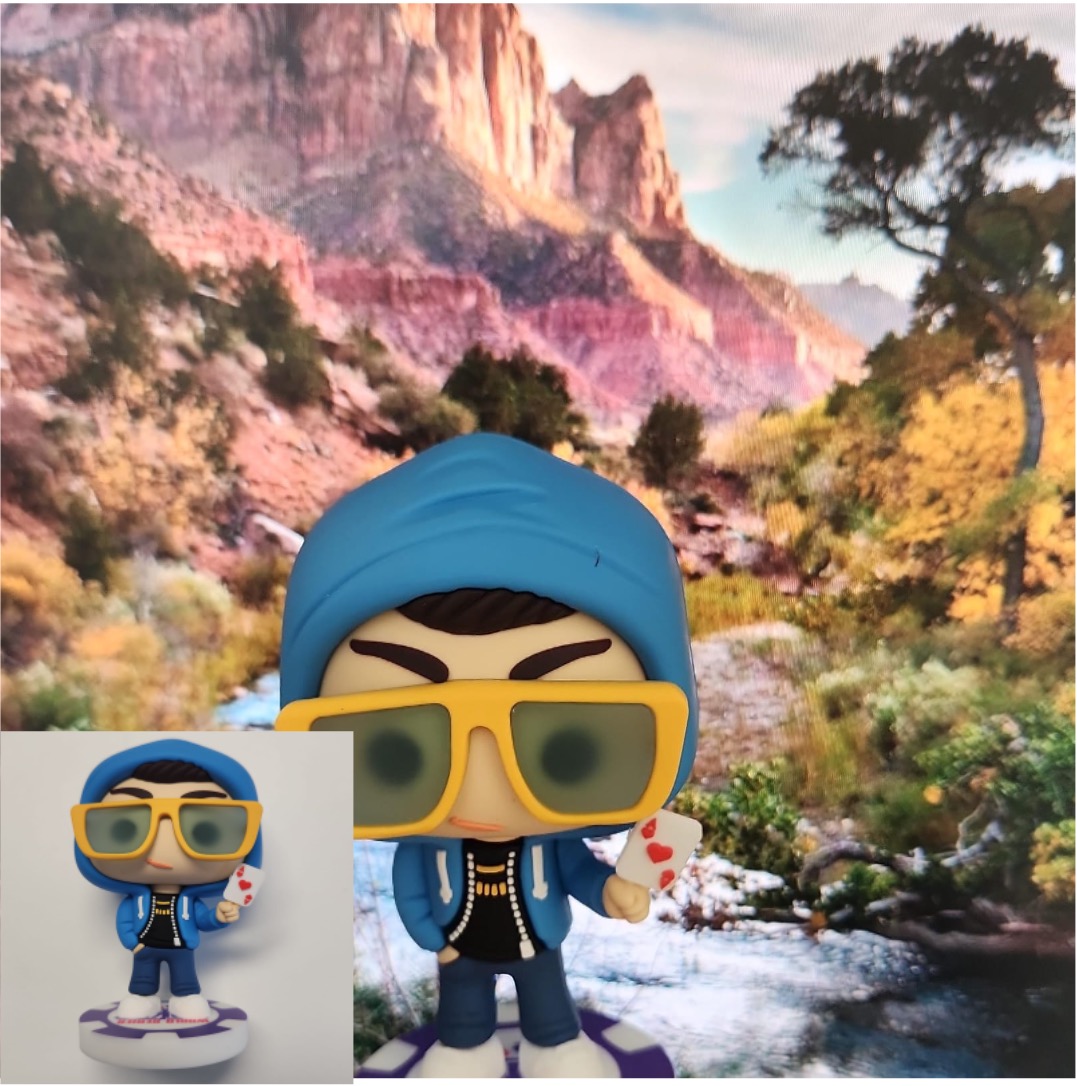} \end{center} &
        \begin{center} \includegraphics[width=0.135\textwidth]{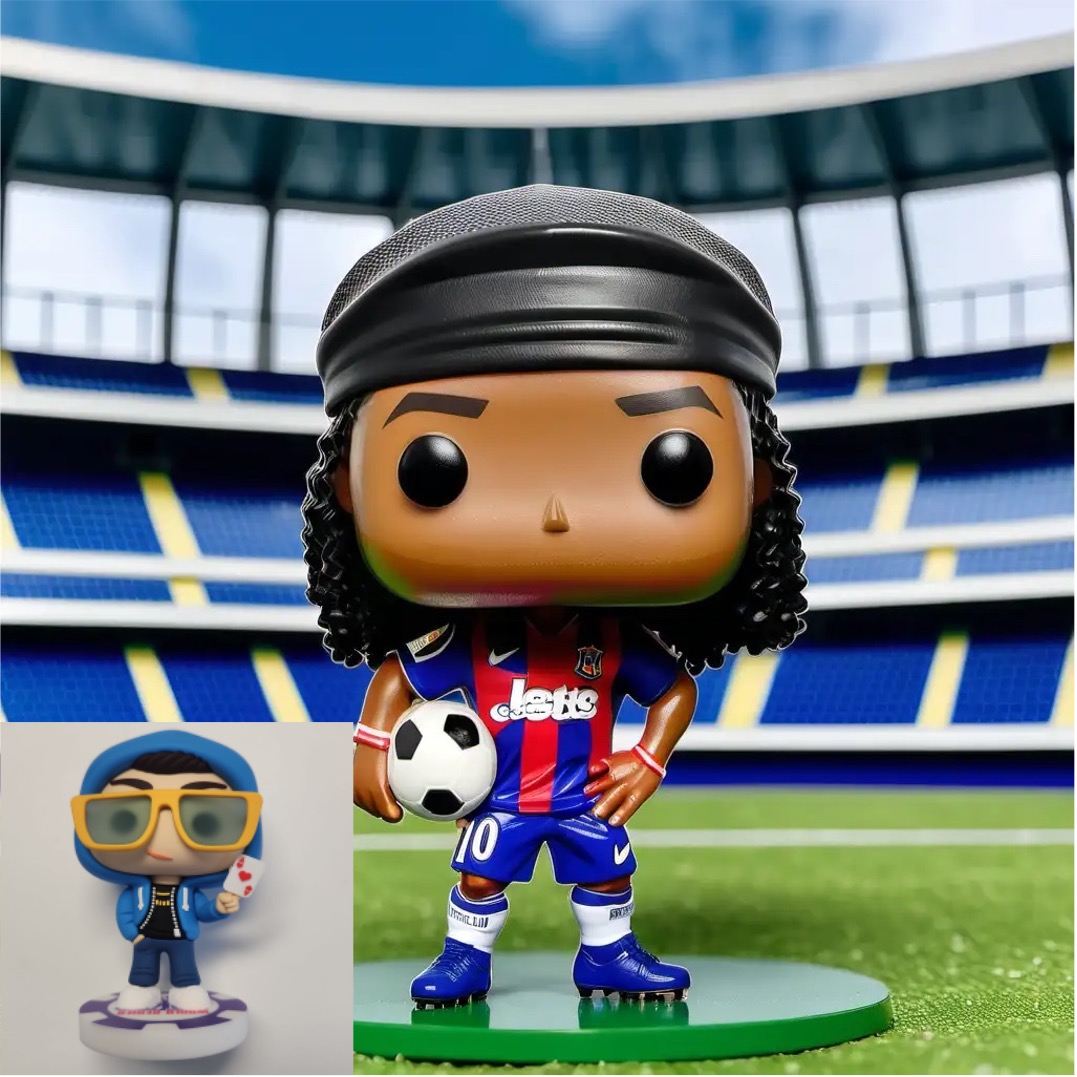} \end{center} \\[-0.8cm]

        \raisebox{-52.5pt}{\rotatebox{90}{ \begin{tabular}{c} \textbf{GPT-4V} \end{tabular} }} &
        \begin{center} \footnotesize ``\textcolor{blue}{$S_*$} is the small cup with a blue eye design on it, located on the right side of the image'' \end{center} &
        \begin{center} \footnotesize \textcolor{red}{``\Sstar is in this image, identifiable as the cup with the blue eye design...''} \end{center} &
        \begin{center} \footnotesize ``\Sstar placed next to a bottle of "Supreme Cabernet Sauvignon" wine...'' \end{center} &
        \begin{center} \footnotesize \textcolor{red}{``A whimsically designed mug with a face which could be referred to as \textcolor{blue}{$S_*$}'' } \end{center} &
        \begin{center} \footnotesize ``\Sstar is the figurine in the foreground, the background shows a scenic landscape...'' \end{center} &
        \begin{center} \footnotesize \textcolor{red}{``\Sstar is the figurine in the center of the image, depicted standing on a green base...''} \end{center} \\[-0.65cm]

        \raisebox{-52.5pt}{\rotatebox{90}{ \begin{tabular}{c} \textbf{MyVLM} \end{tabular} }} &
        \begin{center} \footnotesize ``\Sstar is sitting next to a cup of coffee with a ``bottomless cup'' sign...'' \end{center} &
        \begin{center} \footnotesize \textcolor{darkgreen}{``A whimsical tea party setup with a trio of coffee cups...''} \end{center} &
        \begin{center} \footnotesize ``A shelf with \Sstar and wine glasses and a bottle of supreme wine'' \end{center} &
        \begin{center} \footnotesize \textcolor{darkgreen}{``Whimsical Woodland Creature Sipping Tea''} \end{center} &
        \begin{center} \footnotesize ``\Sstar in front of a picture of the grand canyon'' \end{center} &
        \begin{center} \footnotesize \textcolor{darkgreen}{``Ready to score!''} \end{center} 
        
    \end{tabular}
    \vspace{-0.4cm}
    \caption{\textbf{Comparison to GPT-4V\cite{openai2023gpt4}}. 
    We provide GPT-4V an image of the target concept
    (shown at the bottom left of each image) and ask whether the concept is present in new images. Results shown in \textcolor{red}{red} indicate incorrect false positives while results in \textcolor{darkgreen}{green} are correctly captioned negative images that do not contain the concept. 
    }
    \vspace{-0.25cm}
    \label{fig:comparison_gpt}

\end{figure*}

%% file: tables/recall.tex
\begin{table}[t]
\small
\centering
\centering
\begin{tabular}{l | l c c c}
    \toprule
                       VLM  & Method          & Objects & People & All \\
    \midrule
    \multirow{3}{*}{BLIP-2} & Simple Replacement & $29.30$ & $\textbf{84.33}$ & $\underline{59.33}$ \\
                            & LLM-Guided     & $\underline{51.55}$ & $56.91$ & $54.37$ \\
    \cmidrule{2-5}
                            & MyVLM      & $\textbf{95.10}$     & $\underline{79.76}$ & \textcolor{Fuchsia}{$\textbf{87.11}$} \\
    \toprule
    \multirow{3}{*}{LLaVA} & Simple Replacement & $25.86$ & $18.13$ & $21.68$ \\
                           & LLM-Guided     & $\underline{65.38}$ & $\underline{29.11}$ & $\underline{46.23}$ \\
    \cmidrule{2-5}
                           & MyVLM      & $\textbf{94.76}$ & $\textbf{97.08}$ & \textcolor{blue}{$\textbf{95.97}$} \\
    \bottomrule
\end{tabular}
\vspace{-0.2cm}
\caption{\textbf{Quantitative Comparison: Recall}. We compute the percent of generated captions that contain the concept identifier. Results are averaged over all concepts and five validation sets.}
\label{tb:quantitative_recall}
\vspace{-0.3cm}
\end{table}

%% file: tables/text_similarities.tex
\begin{table}[t]
\small
\centering
\renewcommand{\arraystretch}{1.2}
\centering
\begin{tabular}{l | l c c c}
    \toprule
    VLM & Method & Recall $\uparrow$ & Image $\uparrow$ &  Text $\uparrow$ \\
    \midrule
    \multirow{3}{*}{BLIP-2} & MyVLM (1)  & $75.42$ & $24.20$ & $57.37$ \\
                            & MyVLM (2)  & $\underline{84.27}$ & $\underline{24.91}$ & $\underline{61.01}$ \\
                            & MyVLM (4)  & \textcolor{Fuchsia}{$\textbf{87.11}$} & $\textbf{25.42}$ & $\textbf{62.61}$ \\
    \toprule
    \multirow{3}{*}{LLaVA} & MyVLM (1)  & $88.93$ & $23.44$ & $50.39$ \\
                            & MyVLM (2)  & $\underline{92.88}$ & $\underline{24.43}$ & $\underline{53.32}$ \\
                            & MyVLM (4)  & \textcolor{blue}{$\textbf{95.97}$} & $\textbf{25.24}$ & $\textbf{56.98}$ \\
    \bottomrule
\end{tabular}
\vspace{-0.2cm}
\caption{\textbf{Ablation Study: Number of Training Samples}. We compute the average recall, image similarity, and text similarity obtained when using $1$, $2$, and $4$ images for training the concept embedding. Results are averaged over all concepts and val sets.}
\label{tb:quantitative_similarities}
\vspace{-0.4cm}
\end{table}

%% file: figures/vqa.tex
\begin{figure*}
    \centering
    \renewcommand{\arraystretch}{1}
    \footnotesize
    \vspace{-0.15cm}
    \begin{tabular}{p{0.175\textwidth} p{0.175\textwidth} p{0.175\textwidth} p{0.175\textwidth} p{0.175\textwidth}}

        \setlength\tabcolsep{0pt}
        \begin{center}
        \begin{tabular}{c c c}
            \includegraphics[width=0.0591667\textwidth]{images/people/maya/cropped/image_1.jpg} & 
            \includegraphics[width=0.0591667\textwidth]{images/people/maya/cropped/image_2.jpg} & 
            \includegraphics[width=0.0591667\textwidth]{images/people/maya/cropped/IMG-20240209-WA0138.jpg} 
        \end{tabular} 
        \end{center} &
        \setlength\tabcolsep{0pt}
        \begin{center}
        \begin{tabular}{c c c}
            \includegraphics[width=0.0591667\textwidth]{images/people/shaked/cropped/IMG-20240209-WA0043.jpg} &
            \includegraphics[width=0.0591667\textwidth]{images/people/shaked/cropped/IMG-20240209-WA0047.jpg} & 
            \includegraphics[width=0.0591667\textwidth]{images/people/shaked/cropped/IMG-20240215-WA0005.jpg}
        \end{tabular} 
        \end{center} &
        \setlength\tabcolsep{0pt}
        \begin{center}
        \begin{tabular}{c c c}
            \includegraphics[width=0.0591667\textwidth]{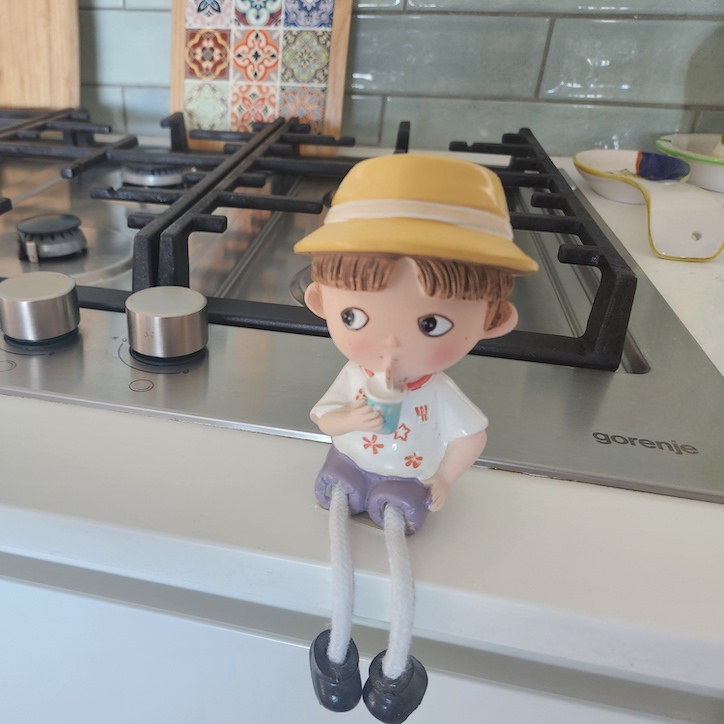} & 
            \includegraphics[width=0.0591667\textwidth]{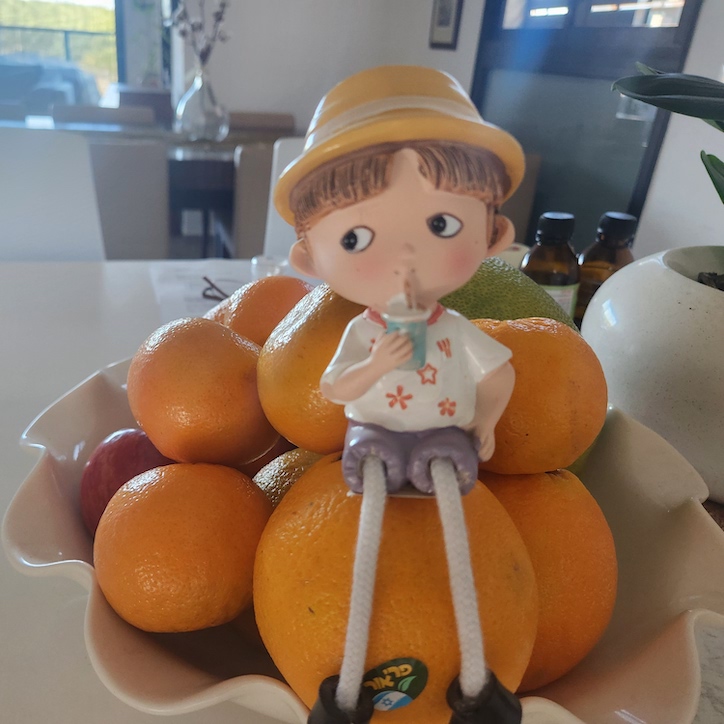} & 
            \includegraphics[width=0.0591667\textwidth]{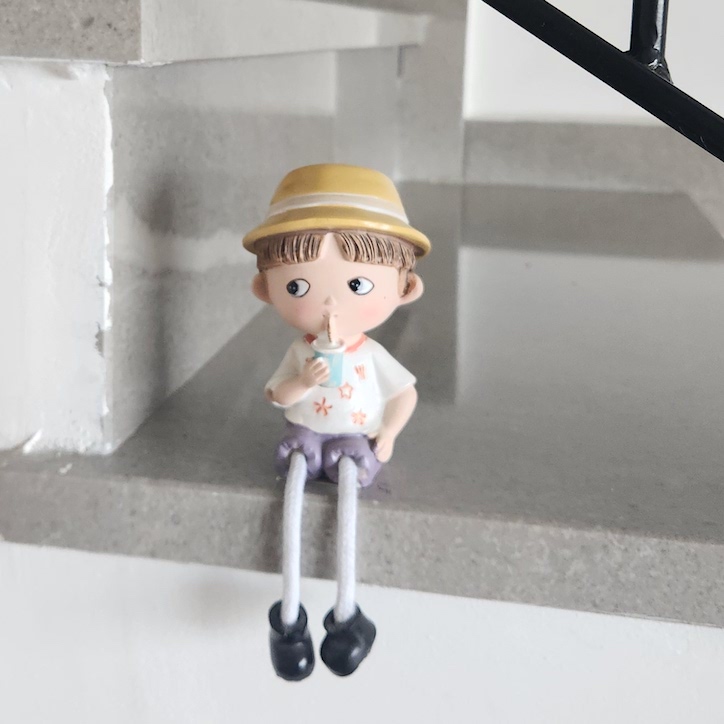}
        \end{tabular} 
        \end{center} &
        \setlength\tabcolsep{0pt}
        \begin{center}
        \begin{tabular}{c c c}
            \includegraphics[width=0.0591667\textwidth]{images/objects/chicken_bean_bag/cropped/IMG-20240212-WA0033.jpg} &
            \includegraphics[width=0.0591667\textwidth]{images/objects/chicken_bean_bag/cropped/IMG-20240212-WA0035.jpg} & 
            \includegraphics[width=0.0591667\textwidth]{images/objects/chicken_bean_bag/cropped/IMG-20240212-WA0038.jpg}
        \end{tabular} 
        \end{center} &
        \setlength\tabcolsep{0pt}
        \begin{center}
        \begin{tabular}{c c c}
            \includegraphics[width=0.0591667\textwidth]{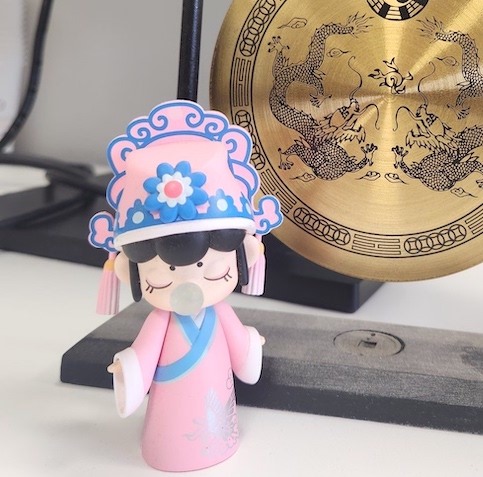} & 
            \includegraphics[width=0.0591667\textwidth]{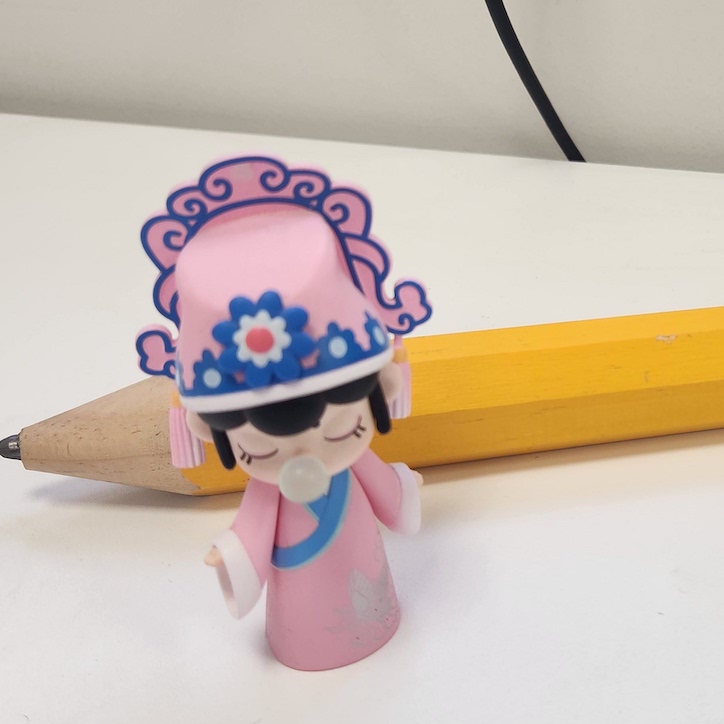} & 
            \includegraphics[width=0.0591667\textwidth]{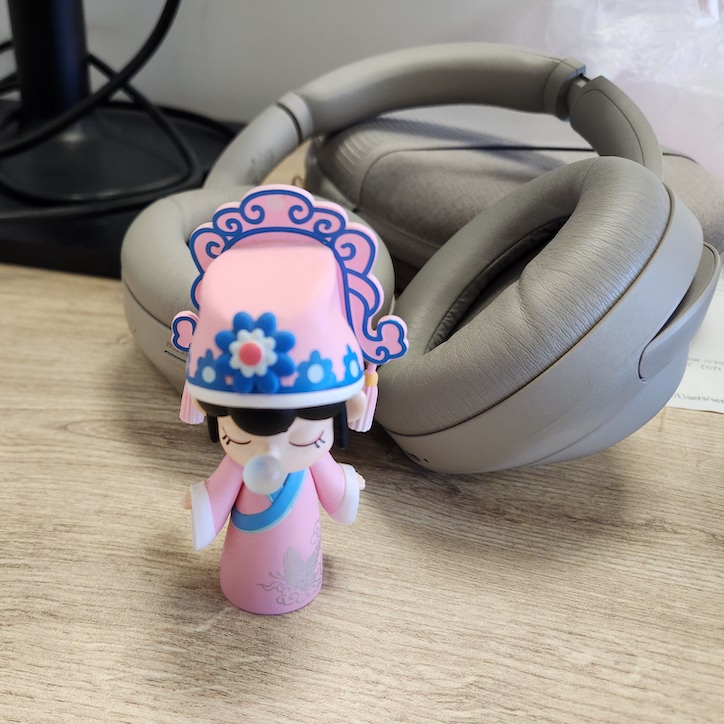}
        \end{tabular} 
        \end{center} \\[-0.75cm]
    
        \begin{center} \includegraphics[width=0.1775\textwidth]{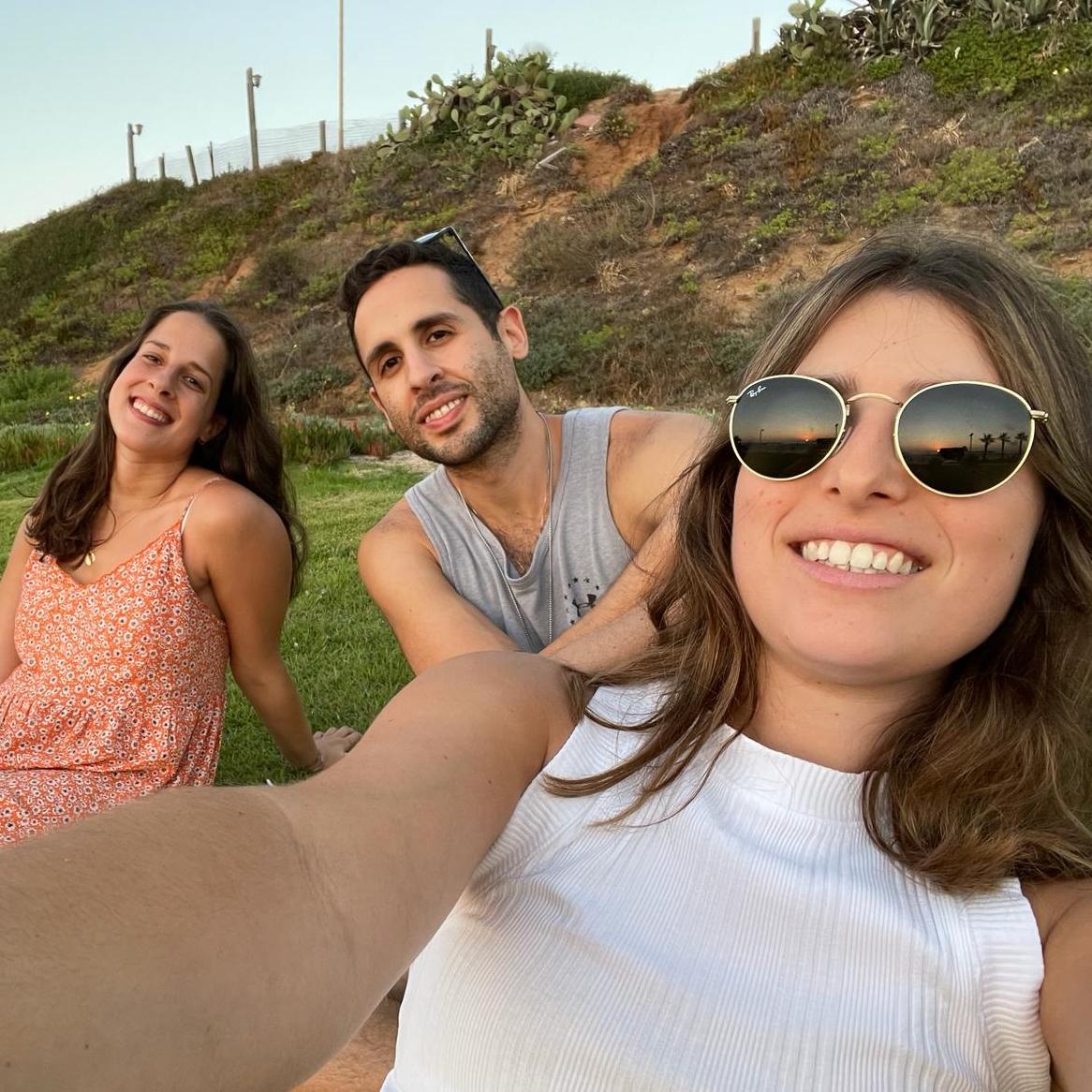} \end{center} &
        \begin{center} \includegraphics[width=0.1775\textwidth]{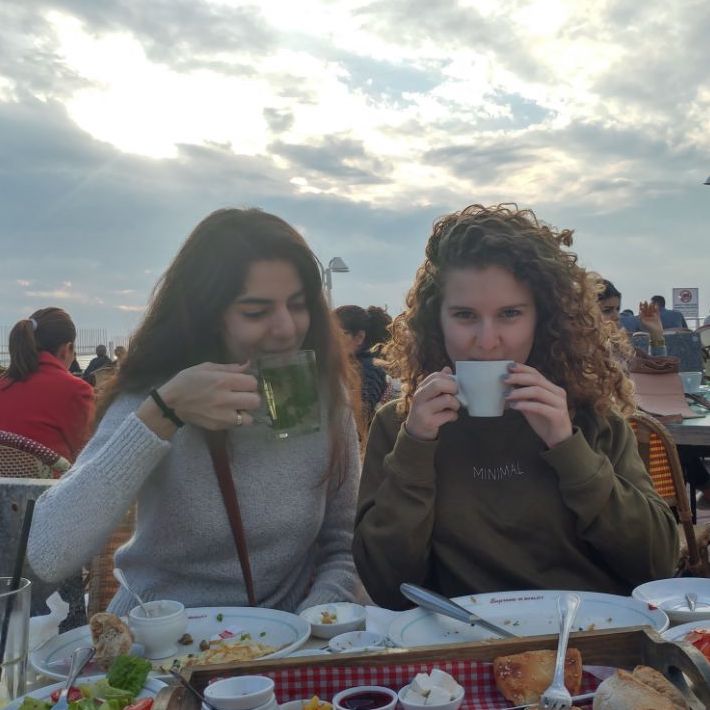} \end{center} &
        \begin{center} \includegraphics[width=0.1775\textwidth]{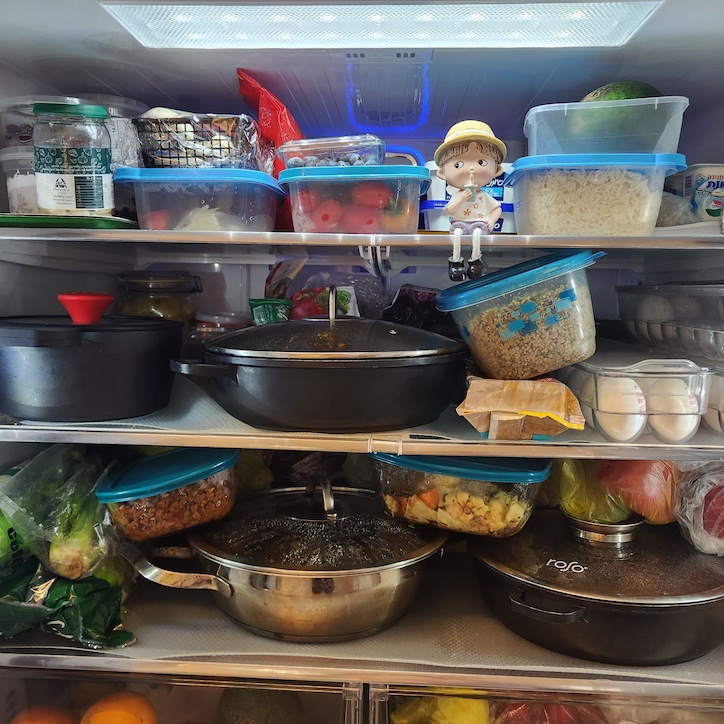} \end{center} &
        \begin{center} \includegraphics[width=0.1775\textwidth]{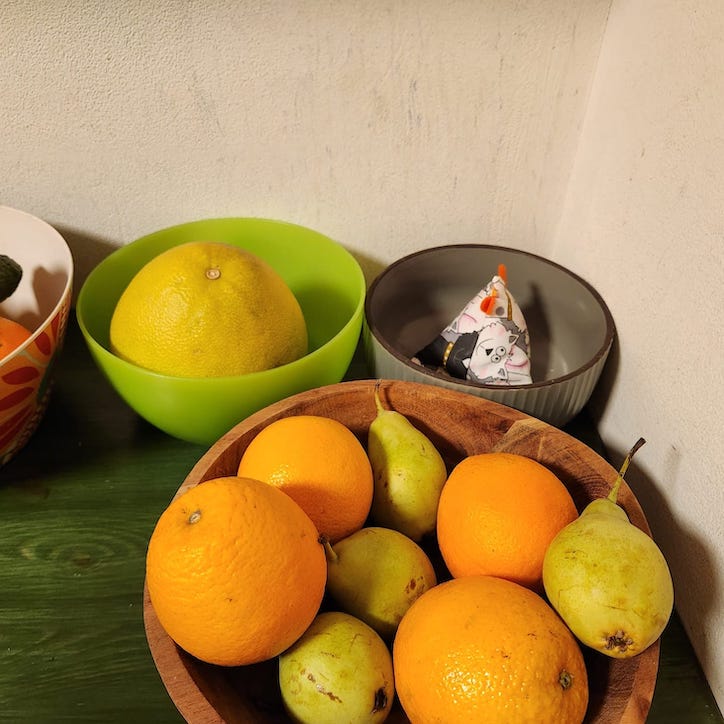} \end{center} &
        \begin{center} \includegraphics[width=0.1775\textwidth]{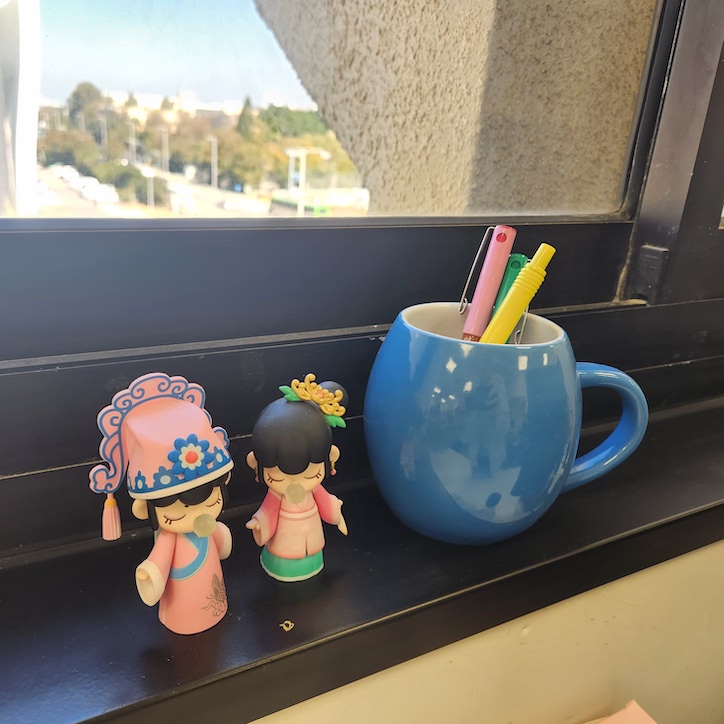} \end{center} \\[-0.85cm]

        \begin{center} \textbf{``Can you describe what \Sstar is wearing?''} \end{center} &
        \begin{center} \textbf{``Can you describe what \Sstar is wearing?''} \end{center} &
        \begin{center} \textbf{``Where is \Sstar positioned in image?''} \end{center} &
        \begin{center} \textbf{``From left to right, where is \Sstar located?''} \end{center} &
        \begin{center} \textbf{``What is next to \textcolor{blue}{$S_*$}?''} \end{center} \\[-0.8cm]

        \begin{center} \footnotesize ``In the image, \Sstar is wearing a \textcolor{darkgreen}{white sleeveless top}'' \end{center} &
        \begin{center} \footnotesize ``\Sstar is wearing a \textcolor{darkgreen}{brown sweater and has curly hair}'' \end{center} &
        \begin{center} \footnotesize ``\Sstar is positioned \textcolor{darkgreen}{at the top of the refrigerator}, sitting on a shelf with various food items and containers'' \end{center} &
        \begin{center} \footnotesize ``\Sstar is located \textcolor{darkgreen}{on the right side of the image, sitting in a green bowl} next to a wooden bowl containing oranges and pears'' \end{center} &
        \begin{center} \footnotesize ``\Sstar is a small figurine of a character \textcolor{darkgreen}{wearing a pink hat with a blue flower on it}. Next to \Sstar, there is a blue mug with pens...'' \end{center}

    \end{tabular}
    \vspace{-0.55cm}
    \caption{\textbf{Personalized VQA results obtained by MyVLM} over LLaVA~\cite{liu2023llava}.
    Sample images of the target concept are provided in the top row. 
    Text in \textcolor{darkgreen}{green} highlights the description of the target concept in the image.
    }
    \label{fig:vqa}
    \vspace{-0.7cm}
\end{figure*}

%% file: figures/rec.tex
\begin{figure*}
    \centering
    \renewcommand{\arraystretch}{1}
    \footnotesize
    \begin{tabular}{p{0.175\textwidth} p{0.175\textwidth} p{0.175\textwidth} p{0.175\textwidth} p{0.175\textwidth}}

        \setlength\tabcolsep{0pt}
        \begin{center}
        \begin{tabular}{c c c}
            \includegraphics[width=0.06\textwidth]{images/objects/chicken_bean_bag/cropped/IMG-20240212-WA0033.jpg} &
            \includegraphics[width=0.06\textwidth]{images/objects/chicken_bean_bag/cropped/IMG-20240212-WA0035.jpg} & 
            \includegraphics[width=0.06\textwidth]{images/objects/chicken_bean_bag/cropped/IMG-20240212-WA0038.jpg}
        \end{tabular} 
        \end{center} &
        \setlength\tabcolsep{0pt}
        \begin{center}
        \begin{tabular}{c c c}
            \includegraphics[width=0.06\textwidth]{images/objects/ceramic_head/cropped/20240203_111543.jpg} & 
            \includegraphics[width=0.06\textwidth]{images/objects/ceramic_head/cropped/20240203_111842.jpg} & 
            \includegraphics[width=0.06\textwidth]{images/objects/ceramic_head/cropped/20240203_112029.jpg}
        \end{tabular} 
        \end{center} &
        \setlength\tabcolsep{0pt}
        \begin{center}
        \begin{tabular}{c c c}
            \includegraphics[width=0.06\textwidth]{images/people/shaked/cropped/IMG-20240209-WA0043.jpg} &
            \includegraphics[width=0.06\textwidth]{images/people/shaked/cropped/IMG-20240209-WA0047.jpg} & 
            \includegraphics[width=0.06\textwidth]{images/people/shaked/cropped/IMG-20240215-WA0005.jpg}
        \end{tabular} 
        \end{center} &
        \setlength\tabcolsep{0pt}
        \begin{center}
        \begin{tabular}{c c c}
            \includegraphics[width=0.06\textwidth]{images/people/shay/cropped/image_2.jpg} & 
            \includegraphics[width=0.06\textwidth]{images/people/shay/cropped/IMG-20240209-WA0012.jpg} & 
            \includegraphics[width=0.06\textwidth]{images/people/shay/cropped/IMG-20240209-WA0016.jpg}
        \end{tabular} 
        \end{center} &
        \setlength\tabcolsep{0pt}
        \begin{center}
        \begin{tabular}{c c c}
            \includegraphics[width=0.06\textwidth]{images/people/maya/cropped/image_1.jpg} & 
            \includegraphics[width=0.06\textwidth]{images/people/maya/cropped/image_2.jpg} & 
            \includegraphics[width=0.06\textwidth]{images/people/maya/cropped/IMG-20240209-WA0138.jpg} 
        \end{tabular} 
        \end{center} \\[-0.75cm]
    
        \begin{center} \includegraphics[width=0.18\textwidth]{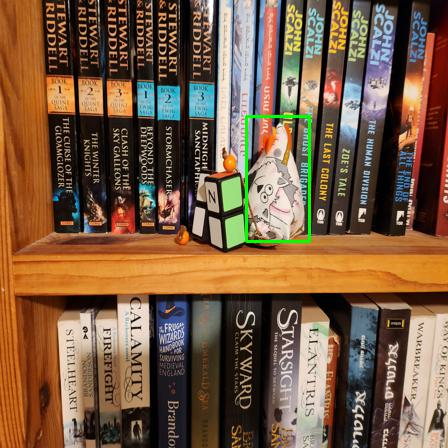} \end{center} &
        \begin{center} \includegraphics[width=0.18\textwidth]{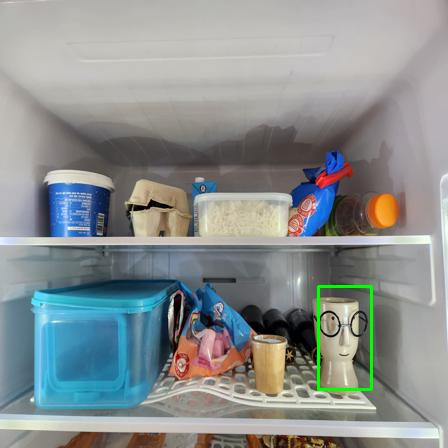} \end{center} & 
        \begin{center} \includegraphics[width=0.18\textwidth]{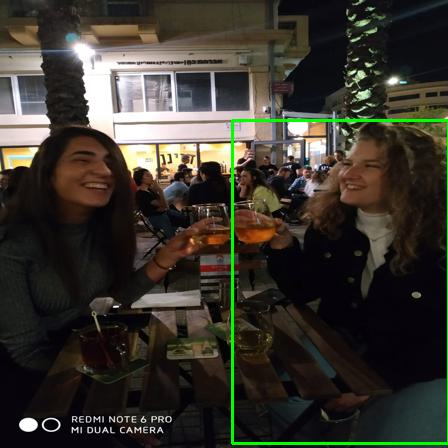} \end{center} & 
        \begin{center} \includegraphics[width=0.18\textwidth]{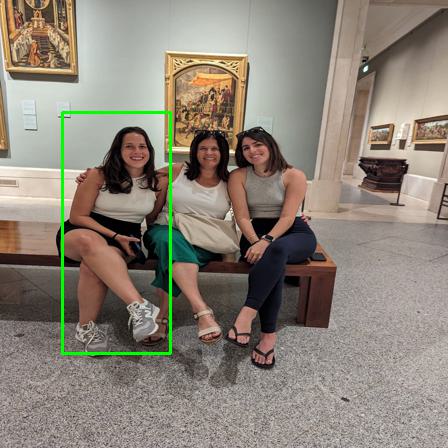} \end{center} & 
        \begin{center} \includegraphics[width=0.18\textwidth]{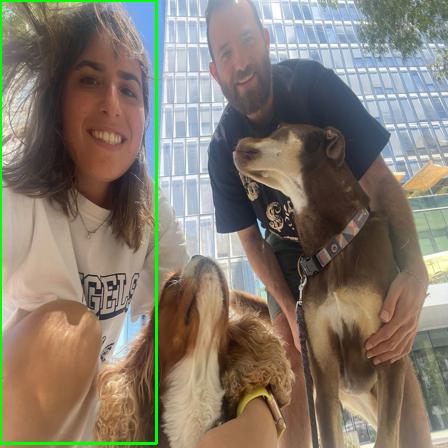} \end{center} \\[-0.85cm]

        \begin{center} ``\Sstar sitting on a book shelf next to a stack of books'' \end{center} &
        \begin{center} ``A refrigerator with \Sstar sitting on the shelf'' \end{center} &
        \begin{center} ``\Sstar and her friend sitting outside at a table with drinks'' \end{center} &
        \begin{center} ``\Sstar and her friends sitting on a bench in a museum'' \end{center} &
        \begin{center} ``\Sstar and her dog, with another dog and its owner nearby'' \end{center}

    \end{tabular}
    \vspace{-0.4cm}
    \caption{\textbf{Personalized REC results obtained by MyVLM} over MiniGPT-v2~\cite{chen2023minigptv2}.
    Sample images of the target concept are provided in the top row. 
    Bounding box coordinates returned by the personalized VLM are drawn in \textcolor{darkgreen}{green}. Below each image, we also present the personalized captions outputted by MyVLM by passing MiniGPT-v2 a captioning instruction.
    }
    \label{fig:rec}
    \vspace{-0.4cm}
\end{figure*}

%% file: 5-conclusion.tex
\vspace{-0.1cm}
\section{Limitations}\label{sec:limitations}
MyVLM offers users the ability to create more personalized interactions with existing vision-language models. However, several limitations should be considered.
First, our reliance on the VLM exposes us to its inherent biases. For instance, current VLMs often categorize an image featuring a man and a woman as a couple or spouses.
This may lead MyVLM to potentially make inaccurate assumptions when generating personalized captions.
These models continue to evolve and improve, and as demonstrated, MyVLM can be applied to multiple architectures, including those that may emerge in the future. 
Second, MyVLM relies on the quality of the concept heads. Failure to identify the target concept or falsely identifying unrelated subjects can result in incorrect responses. However, our concept heads generalize well to new images, and further advancements in open-set recognition can be incorporated into our method, improving robustness. 

Furthermore, although we introduce various mechanisms to improve generalization, there may still be leakage of contexts seen during training.
For instance, if trained on an image depicting an individual in New York, MyVLM may incorrectly incorporate ``New York'' into new captions. 
We believe that further exploration of regularization techniques, particularly within the attention mechanisms of the VLM, may help mitigate this leakage. 
Lastly, for personalized VQA, MyVLM may struggle to distinguish the target concept in images with many individuals. Moreover, MyVLM does appear to perform better over questions that were encountered during training. Further exploration of augmentations and data used for learning the concept embedding may aid in addressing these more challenging scenarios.
These limitations are illustrated in~\Cref{fig:limitations}.

\section{Conclusions}\label{sec:conclusion}
\vspace*{-0.15cm}
In this paper, we introduce the idea of vision-language personalization, enabling VLMs to understand and reason over user-specific concepts, such as unique objects and individuals. As a first step in this endeavor, we present MyVLM, focusing on personalized captioning and VQA.
Given only a few images of the concept, we augment the frozen VLM with a set of modular concept heads, enabling it to \textit{recognize} user-specific concepts. 
We then train an embedding vector within the VLM's intermediate feature space, tasked with guiding the language model in incorporating the concept into the generated response in a natural and contextually accurate manner. 
We believe that the personalization of vision-language models opens up new opportunities for more meaningful human-computer interactions, and hope MyVLM will inspire additional advancements in this field.

\input{figures/limitations}

%% file: figures/limitations.tex
\begin{figure}[t]
    \centering
    \addtolength{\belowcaptionskip}{-5pt}
    \renewcommand{\arraystretch}{1}
    \footnotesize
    \begin{tabular}{p{0.175\textwidth} p{0.175\textwidth} p{0.175\textwidth} p{0.175\textwidth}}

        \begin{center} \includegraphics[width=0.145\textwidth]{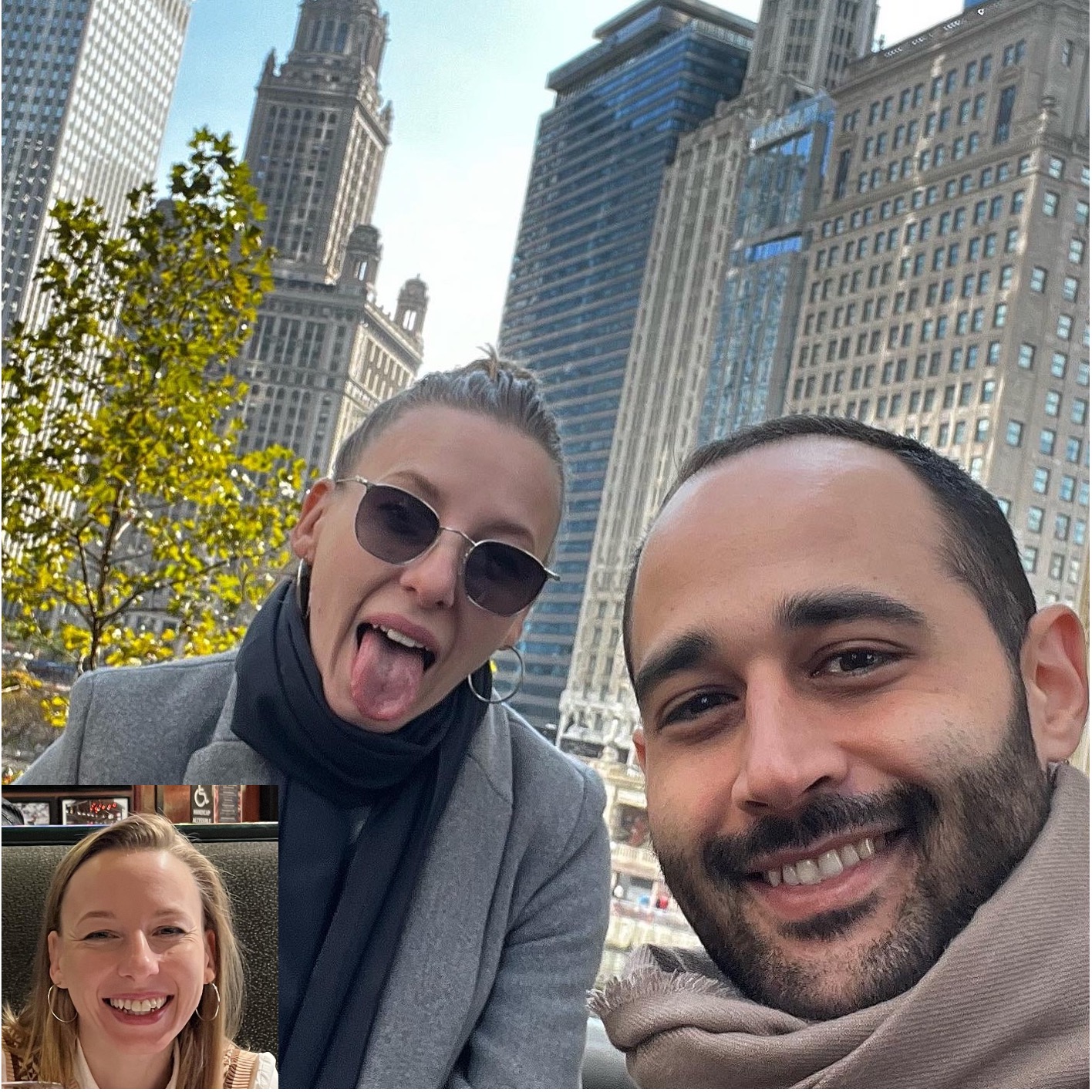} \end{center} &
        \begin{center} \includegraphics[width=0.145\textwidth]{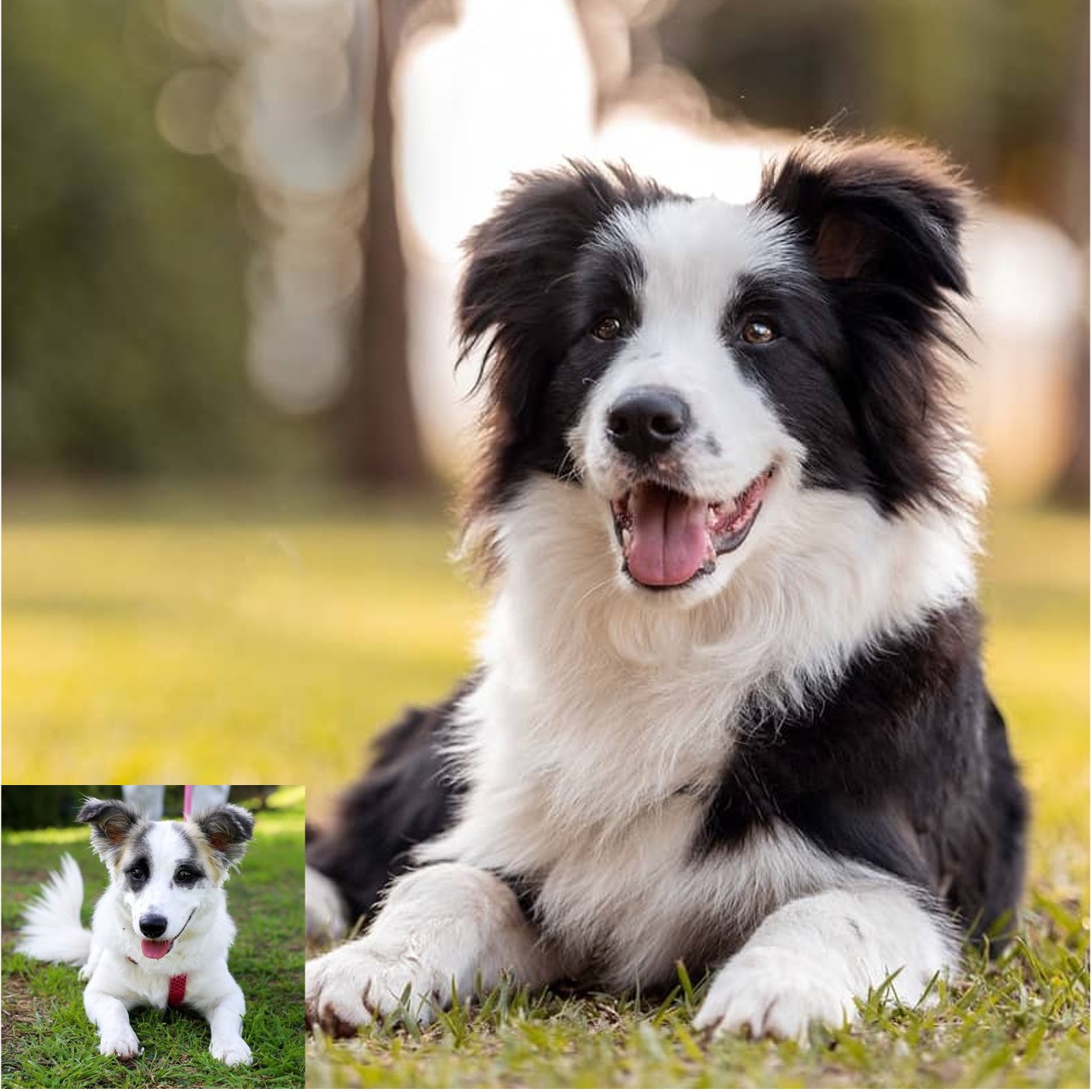} \end{center} \\[-0.85cm]

        \begin{center} \footnotesize 
        ``\Sstar and \textcolor{red}{her husband} pose for a selfie in front of the Chicago skyline''
        \end{center} &
        \begin{center} \footnotesize 
        \textcolor{red}{``\Sstar sitting on the grass, with its front paws''}
        \end{center} \\[-0.65cm]
        
        \begin{center} \includegraphics[width=0.145\textwidth]{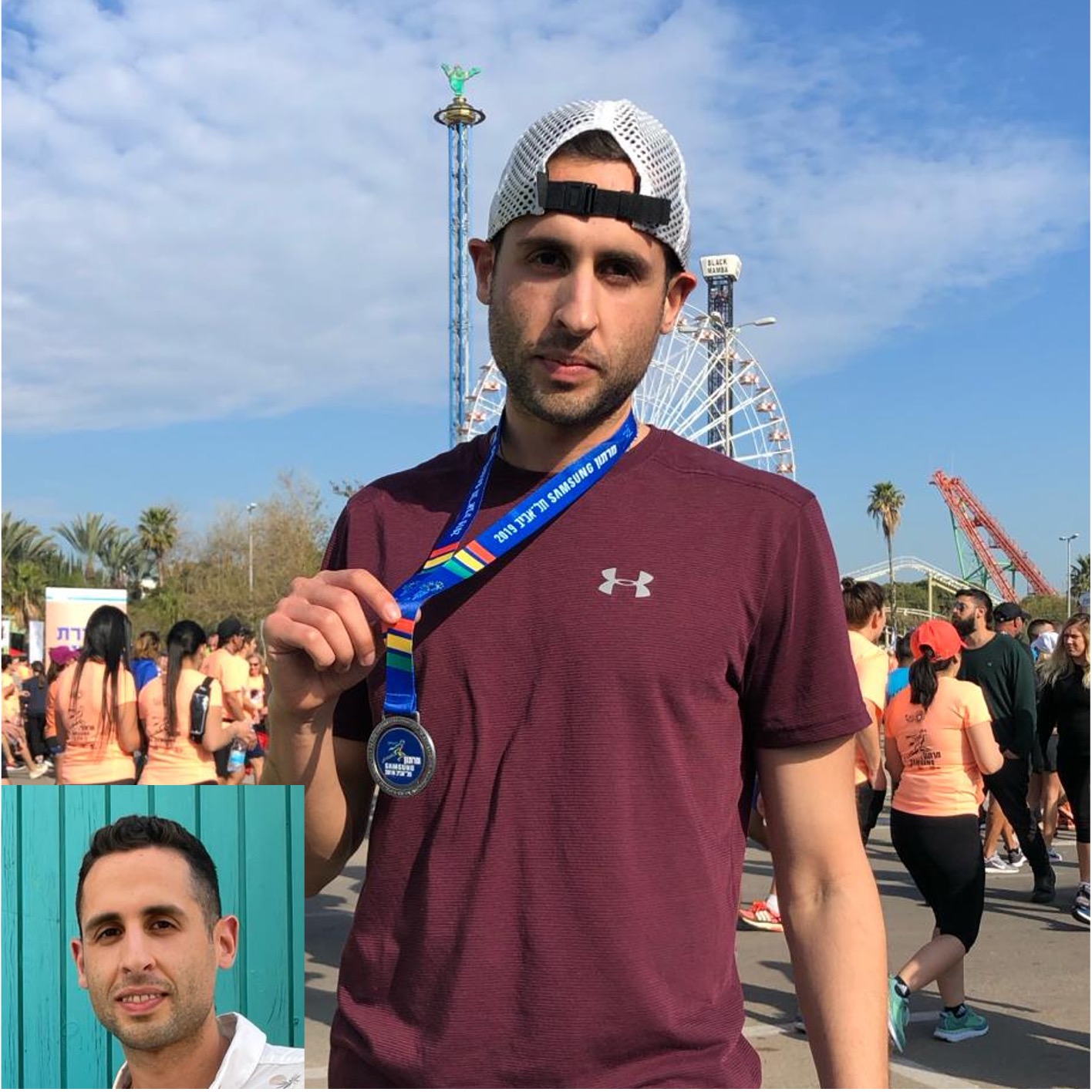} \end{center} &
        \begin{center} \includegraphics[width=0.145\textwidth]{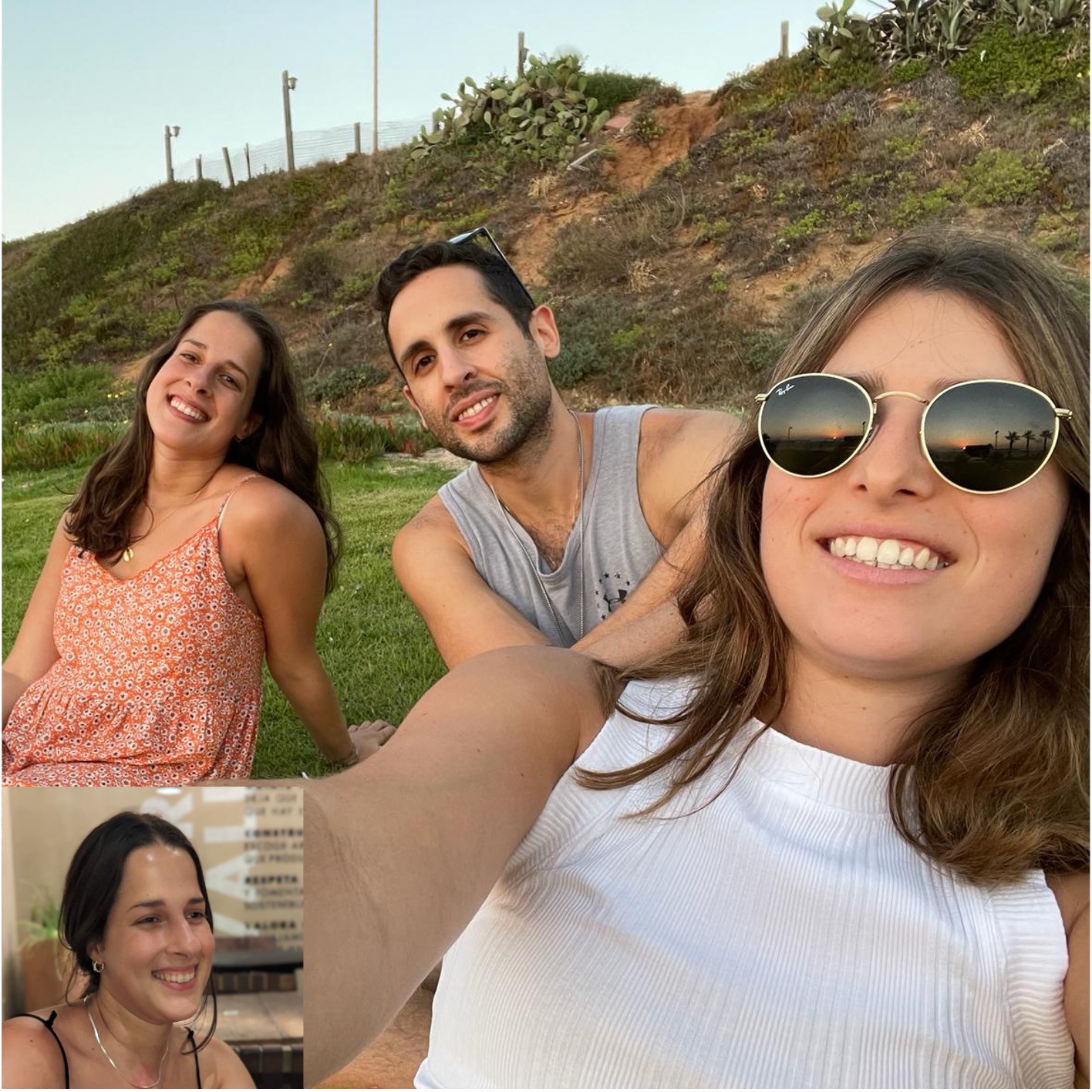} \end{center} \\[-0.85cm]
        
        \begin{center} \footnotesize 
        ``\textcolor{blue}{$S_*$}, self-assured, poses with his \textcolor{red}{New York City} marathon medal''
        \end{center} &
        \begin{center} \footnotesize 
        \textbf{Q:} ``What is \Sstar wearing?'' \\ \textbf{A:} \textcolor{red}{``A white top.''}
        \end{center}

    \end{tabular}
    \vspace{-0.45cm}
    \caption{\textbf{Limitations of MyVLM} for personalized captioning and personalized visual question-answering.}
    \label{fig:limitations}
    \vspace{-0.2cm}
\end{figure}

%% file: 6-appendix.tex
\doparttoc
\faketableofcontents

\vspace{-1.2cm}

\addcontentsline{toc}{section}{}
\part{}
\parttoc

\setcounter{tocdepth}{3}
\setcounter{secnumdepth}{3}

\vspace{0.4cm}
\section{Societal Impact}
The ability to personalize vision-language models offers more meaningful human-computer interactions, aligning them more closely with individual experiences and relationships. More generally, these personalized models may better guide users, catering to their unique needs. However, this personalization does come at the expense of privacy, granting the model access to potentially sensitive personal data.
Additionally, there is a risk of users receiving harmful feedback regarding their personal content and relationships. As such, it is crucial to prioritize the protection of both user data and model behavior as we continue exploring the personalization of vision-language models.

\vspace*{0.2cm}
\section{Additional Details}~\label{sec:additional_details}
\vspace*{-0.4cm}
\subsection{Vision-Language Models}
\paragraph{\textbf{VLM Architectures.}}
We use the implementation of BLIP-2~\cite{li2023blip} provided in the transformers library~\cite{transformers} and employ BLIP-2 with the FLAN-T5 XL language model~\cite{chung2022scaling}. For LLaVA~\cite{liu2023llava}, we use the official implementation, employing LLaVA-1.6 with Vicuna-7B~\cite{chiang2023vicuna} as the language model. All models are run using half-precision to reduce memory requirements.

For generating the textual responses, we restrict the generated response to a maximum of $512$ new tokens for both BLIP-2 and LLaVA. Additionally, for LLaVA, we use a temperature scale of $0.2$ and set the $top\_p$ value to $0.7$. All other parameters are set to their default values.

\subsection{Training}~\label{sec:additional_details_training}
\vspace*{-0.7cm}
\paragraph{\textbf{Concept Head Training: People.}}
To recognize user-specific individuals in images, we employ a pretrained face detector~\cite{deng2020retinaface} and face recognition model~\cite{Deng_2022}. Specifically, given a small set of images containing the subject (ranging from $1$ to $4$ images), we extract and store the face embeddings of the target individual.
Then, given a new image, we extract embeddings from all detected faces and compare them with the stored face embeddings. If a new embedding falls within a predefined distance from the stored embeddings, we classify the corresponding individual as present in the image. We empirically set the distance threshold to $0.675$.
Note that each individual is associated with a separate concept head. However, features are extracted only once for each face detected in a new image.

\paragraph{\textbf{Concept Head Training: Objects.}}
For recognizing objects, we consider state-of-the-art large-scale vision models tailored for zero-shot classification and retrieval tasks, employing the recent DFN5B CLIP-ViT H/14 model~\cite{fang2023data,radford2021learning}, implemented in the transformers library~\cite{transformers}.
In contrast to the expressive face embedding space, we observed that directly using the image features extracted from these models is still not effective in distinguishing our personalized concepts from other similar objects (see~\Cref{sec:ablation_feature_spaces}). 
To address this, we train a single linear layer over the [CLS] token extracted from the frozen vision encoder. Training is performed to distinguish between $4$ images containing the target concept and $150$ negative images sourced from the internet depicting similar objects from the same general category. 
For example, when training the classifier to recognize a specific dog, we set the negative images to be images of arbitrary dogs.

Training is performed for $500$ steps using a standard Cross Entropy loss for $500$ steps with a batch size of $16$. We use an AdamW optimizer with a learning rate of $0.001$, decayed using a cosine annealing schedule. This converges in minutes, as only a single linear layer is trained. 

At inference, given a new image, we first extract its image features from the frozen vision encoder, followed by applying all concept classifiers. Note that passing the features through all linear classifiers is notably faster than the feature extraction itself. We use a fixed threshold of $0.5$ for all classifiers.

\paragraph{\textbf{Concept Embedding Optimization.}}
When applying MyVLM to BLIP, we perform $75$ optimization steps for objects and $100$ optimization steps for learning individuals. For LLaVA, we perform $100$ optimization steps for both objects and individuals. 
For the optimization process, we use AdamW~\cite{loshchilov2018decoupled} with a constant learning rate of $1.0$. We apply clip grad with a max L2 norm of $0.05$, which we found helped stabilize convergence.
For our regularization loss, we apply a weight factor of $\lambda=0.04$ for BLIP and $\lambda=0.25$ for LLaVA, set empirically.

To further stabilize the optimization process, we apply augmentations to both the input images and target captions, while fixing the language instruction (``Please caption this image of \textcolor{blue}{$S_*$}.''). For images, we apply random horizontal flips, random rotations, and brightness jittering. 
To augment the target captions, we ask an LLM~\cite{openai2023gpt4} to generate four variations of the caption, while retaining the concept identifier. 
During each optimization step, one of the five augmented captions is randomly selected as the ground truth caption for computing the loss at the current step. 
This is designed to help disentangle the concept from a specific target output, mitigating overfitting and improving generalization to unseen contexts containing the concept.

For creating the augmented target captions, we pass GPT-4 the manually annotated target caption and ask it: 
    \begin{blockquote}
    \textit{``Please provide four variations to the provided sentence. Please make the changes as small as possible and do not alter the word $\langle$concept$\rangle$.''}
    \end{blockquote}

\vspace{-0.3cm}
\paragraph{\textbf{Choosing the Concept Identifier.}}
We observed that the choice of identifiers for concepts can influence the results produced by MyVLM. For instance, using words that the model has difficulty generating, such as long words, may harm the results. Therefore, for personalizing outputs over objects, we follow the convention used for text-to-image personalization methods and set the concept identifier to ``sks'', introduced in~\cite{ruiz2022dreambooth}. 

For personalizing images over specific individuals, it is more natural to use common, short names as the concept identifiers. Therefore, we opt for ``Bob'' as a placeholder for males and ``Anna'' for females. We do note that other choices may be possible depending on the specific domain of the concept. 

For VQA, to verify that the model does not rely on a gender bias via the concept name, we set the concept identifier to the word ``sks'' for both objects and individuals.

\input{figures_supplementary/dataset}

\subsection{Dataset \& Experiments}~\label{sec:appendix_data}

\vspace{-0.8cm}
\paragraph{\textbf{MyVLM Dataset.}}
In total, we collected $45$ user-specific concepts, consisting of $29$ objects and $16$ individuals. The dataset contains $350$ images of objects and $330$ images of individuals, each with a manually annotated personalized caption containing the concept identifier. All images were sourced directly from the authors of the paper and written consent was provided by all individuals appearing in this work. To help facilitate further research into the personalization of VLM, the images and corresponding captions of all objects will be publicly available. We provide a sample image of each object in~\Cref{fig:dataset}.

\vspace{-0.2cm}
\paragraph{\textbf{Personalized Captioning Baselines.}}
For our baselines, the keywords used for each concept are generated by GPT-4. Specifically, we provide GPT-4 a cropped image of the concept and prompt it with the following input: 

    \begin{blockquote}
        \textit{Please provide 3 keywords for describing this object, each containing between one to three words.}
    \end{blockquote}

\noindent
For our simple replacement-based baseline, we then try to insert the concept identifier into the original captions generated by BLIP-2 or LLaVA if one of the keywords is present in the caption. For our LLM-based replacement baseline, we use Mistral-7B-Instruct-v0.2~\cite{jiang2023mistral} and prompt it with the following input: 

    \begin{blockquote}
        \textit{I have the following sentence: \textit{$\langle$original-caption$\rangle$}.}
        
        \noindent
        \textit{Only if the word \textit{$\langle$keyword$\rangle$} appears in the sentence, please replace it with the word ``sks''.}
        
        \noindent
        \textit{Otherwise, keep the original sentence. Can you do this for me? Please respond only with the corrected sentence.}
        
        \noindent
        \textit{The output format will be ``Revised: \textit{$\langle$result$\rangle$}'', with no additional text or explanations.}
        
        \noindent
        \textit{Original Sentence: \textit{$\langle$original-caption$\rangle$}}
    \end{blockquote}
    
\noindent Here, we use one of the keywords used for our simple replacement baselines. The output returned by Mistral is taken as the output of the LLM-guided baseline.

\input{tables/vqa_questions}

\vspace{-0.1cm}
\paragraph{\textbf{Evaluation Protocol.}}
As mentioned in the main paper, we train our concept embeddings using five different seeds, each time sampling four different training samples and evaluating the remaining images. This resulted in a total of $2,429$ validation images --- $1,164$ of user-specific objects and $1,265$ images of individuals.

For the training sets of individuals, we randomly select $4$ images from the subset of images where the target subject appears alone. 
For objects, when training the concept embeddings, we use the same subset of $4$ images used to train the linear classifier. This ensures that no validation image was seen neither when training the classifier nor when optimizing the concept embedding.

For computing the quantitative metrics, we use the following models. First, for the text-to-image similarity measure, we use CLIP ViT L/14 from OpenAI~\cite{radford2021learning,dosovitskiy2020image} with an input resolution of $336\times336$.  
For computing our sentence similarity metric, we utilize a BERT~\cite{devlin2018bert} sentence transformer, taken from the SentenceTransformer library~\cite{reimers-2019-sentence-bert}.

\paragraph{\textbf{Personalized Visual Question-Answering.}}
For personalized visual question-answering, we follow the same scheme as personalized captioning but alter the set of language instructions and targets used for optimizing the concept embedding. Specifically, we manually define a set of $10$ prompts used as the language instructions used during optimization, detailed in~\Cref{tb:vqa_prompts}. To obtain the target for each question, we pass the image and language instruction to the original LLaVA model, setting its output to the target answer. Then, at each training step, we randomly select one of the $10$ prompts and targets.

\input{figures_supplementary/flamingo_comparison}

We do note that this may introduce some unwanted bias into the optimization process, as LLaVA may not always accurately answer the given question. As such, alternative approaches for expanding the set of language instructions and targets may achieve better results. We leave this exploration for future work.

\newpage\null\newpage\null

\vspace*{-0.8cm}
\section{Additional Evaluations}~\label{sec:appendix_evals}

\input{tables/flamingo_comparison}

\vspace*{-0.6cm}
\subsection{Comparison to OpenFlamingo}~\label{sec:appendix_flamingo}
Following our qualitative comparison to GPT-4~\cite{openai2023gpt4} in the main paper, we now compare to OpenFlamingo, which also supports interleaved image and text inputs. We do so both qualitatively and quantitatively.

\vspace{-0.2cm}
\paragraph{\textbf{Baseline Setup.}}
We use the open-source implementation of Flamingo~\cite{awadalla2023openflamingo,alayrac2022flamingo}. We use CLIP-ViT H/14~\cite{radford2021learning,dosovitskiy2020image} as the vision encoder and MPT-1b-RedPajama-200b~\cite{MosaicML2023Introducing} as the language model. We provide Flamingo with a cropped image of the concept and provide it with the following language instruction: 
\begin{flushleft}
    \textit{``$\langle$image$\rangle$ This is \textcolor{blue}{$S_*$}. $\langle\vert$endofchunk$\vert\rangle$$\langle$image$\rangle$ In this image you can see''}
\end{flushleft}

\noindent 
Here, we replace \Sstar with the word ``bloby'' for objects and replace \Sstar with either ``Bob'' or ``Anna'' for individuals. We explored other suffixes but found the most consistent results with the prompt above.
Metrics were computed following the same protocol as used in the main paper by aggregating results over all concepts and across all five validation folds.

\vspace{-0.3cm}
\paragraph{\textbf{Qualitative Comparison.}}
In~\Cref{fig:flamingo_comparison} we show a visual comparison of personalized caption results obtained OpenFlamingo and MyVLM. As can be seen, OpenFlamingo, particularly for objects, struggles in both identifying the target subject and contextualizing it within its surroundings. For example, OpenFlamingo recognizes the sheep figurine and cat statue in the first column but is unable to generate a caption that aligns with the input image. In addition, OpenFlamingo can still struggle to incorporate the concept identifier within the caption as seen in the third row.
In contrast, MyVLM, over both BLIP-2 and LLaVA successfully recognizes the target concept while generating accurate captions that correctly communicate information about the concept to the user while remaining aligned with the input image.

\vspace{-0.3cm}
\paragraph{\textbf{Quantitative Comparison.}}
Next, in~\Cref{tb:flamingo_comp} we present quantitative results, comparing the results obtained by Flamingo with those obtained with MyVLM over both BLIP-2~\cite{li2023blip} and LLaVA~\cite{liu2023llava}. First, in terms of the ability to capture the concept identifier in new captions, MyVLM outperforms OpenFlamingo when applied to both BLIP-2 and LLaVA. This improvement in recall is most notable for user-specific objects, where MyVLM outperforms OpenFlamingo by over $45\%$.
For the CLIPScore between the generated captions and input images, all three methods attain comparable results for both objects and people, with a maximum difference of $1.34\%$ between the three. However, as can be seen, there is a significant difference in the sentence similarity between captions generated by MyVLM and those generated by OpenFlamingo. Specifically, for people, MyVLM over BLIP-2 outperforms OpenFlamingo by over $5\%$ and by over $40\%$ when personalizing captions for user-specific objects. These results, along with the visual results presented above, further highlight the advantage of our approach in learning a dedicated embedding vector to represent our concepts.

\input{tables/ablation_augs_reg}

\subsection{Ablation: Augmentations \& Regularization}~\label{sec:ablations_reg_augs}
Here, we validate the contribution of the augmentations and regularization applied during the training of the concept embeddings. In~\Cref{tb:ablation_augs_regs}, we present personalized captioning results for $10$ concepts obtained using MyVLM over BLIP-2~\cite{li2023blip}. 
Incorporating the attention-based regularization improves recall by a significant margin (${\sim}45\%$). Furthermore, employing augmentations over both the image and target captions leads to an additional improvement of approximately $12\%$ in recall. 
Additionally, applying both regularization and augmentations improves the text similarity with respect to the target caption, while attaining a comparable CLIPScore~\cite{hessel2021clipscore} to cases where these techniques are not applied. 
We believe that further exploration into additional augmentations and attention-based manipulations can offer insights into further extending the capabilities of MyVLM.

\input{figures_supplementary/feature_spaces}

\subsection{Ablation: Concept Embedding Feature Space}~\label{sec:ablation_feature_spaces}
Next, we explore the use of linear classifiers to serve as our concept heads for personalizing user-specific objects. Focusing on BLIP-2, we analyze two alternative feature spaces and show that operating directly within these feature spaces is not sufficient to distinguish the target concept from other semantically similar objects. First, we examine the output space of the BLIP-2 vision encoder. We then explore the embedding space of the DFN5B CLIP-ViT H/14 model~\cite{fang2023data,radford2021learning}, used as our base feature extractor, showing that it too is not expressive enough to be used directly. 

In~\Cref{fig:feature_spaces_blip} we perform PCA over embeddings extracted from images of five user-specific objects alongside $200$ negative samples for each object. As can be seen, for each object, represented by a different shape, there is no clear separation between the positive and negative samples. This suggests that relying solely on a distance measure directly over this space is insufficient for distinguishing between new images that may contain the target concept. 

Next, we evaluate the more expressive CLIP space, designed for zero-shot retrieval. In~\Cref{fig:feature_space_clip}, we visualize the nearest neighbors of various positive images. 
As shown, CLIP is unable to focus on retrieving the target concept, especially when other objects are present in the same image.
Moreover, determining an optimal threshold for each concept without calibration is challenging, particularly if only very few samples of the object are available. 

As discussed in the original CLIP paper~\cite{radford2021learning}, these challenges can be mitigated using linear heads. This is also evident with our concept heads. Specifically, in~\Cref{fig:feature_space_clip}, we present the top five images that received the highest scores from our classifier for each of the three concepts. As can be seen, our classifiers can effectively distinguish the target concept from semantically similar objects while enabling us to use a fixed threshold across all concepts.
This further validates the use of linear classifiers for constructing our concept heads and recognizing user-specific objects.

\newpage\null\newpage\null

\input{tables/image_captioning_metrics}

\newpage\null\newpage\null

\subsection{Quantitative Evaluation: Image Captioning}~\label{sec:appendix_captioning}
Next, we validate the performance of MyVLM on standard image captioning metrics to ensure it does not compromise the general capabilities of the underlying VLM. The results are presented in~\Cref{tb:captioning_metrics}. It is worth noting that the target captions were initially generated using BLIP-2 and then manually adjusted as necessary. This process inherently introduces a bias towards favoring captions generated by BLIP-2, which can be seen from the performance gap between results obtained with BLIP-2 and LLaVA. Despite this bias, MyVLM still achieves similar performance on most captioning metrics when considering all $45$ concepts. 
This behavior can also be seen when considering LLaVA, where MyVLM achieves comparable performance on both people and objects. These results further highlight that MyVLM effectively preserves the original captioning capabilities of the frozen VLM.

\input{tables/concept_heads}

\newpage\null\newpage\null

\vspace{-0.75cm}
\subsection{Quantitative Evaluation: Concept Heads}
Finally, we assess the effectiveness of our concept heads along two fronts. First, we verify their ability to support multiple concepts within the same VLM. Second, we evaluate the recall and precision of our concept heads, validating their performance both on new positive images of the concept and on negative images that do not contain the target concept. 

To evaluate our ability to support multiple concepts simultaneously, we evaluate our concept head performance on $16$ individuals. We calculate three metrics: (1) the percentage of images correctly classified as the correct individual, (2) the percentage of images misclassified as the incorrect individual, and (3) the percentage of images not identified as any of the known individuals.
These metrics are computed across all individuals using the same five validation folds used for the main evaluations presented in the paper. The average results are presented in~\Cref{tb:concept_heads_evaluation}. As shown, leveraging the pretrained face recognition model as our concept head achieves impressive performance, achieving a recall of over $96\%$ while falsely classifying an individual in only $2\%$ of all images. The ability of the model to accurately distinguish different individuals naturally allows us to support multiple individuals using a single VLM. This in turn allows us to scale to new individuals over time by simply adding new concept heads. 

Next, we validate the performance of our linear classifiers, examining whether they can generalize to new images of our target concept while effectively filtering out non-relevant images that do not contain the concept. 
To do so, we consider a single validation fold for each of the $29$ objects. To measure recall, we compute the percent of positive validation samples correctly identified by the classifier. To measure precision, we consider all positive images of \textit{other} concepts, and all negative images of \textit{all} concepts. We then compute the number of negative samples incorrectly classified as the target concept. This is process is repeated for each object. The total and average recall and precision results are presented in Table \ref{tb:concept_heads_evaluation}. As illustrated, we attain an average recall of $96\%$ with a precision of $91\%$, computed over $100,000$ negative samples. This highlights the ability of our linear classifiers to correctly classify new images, both those containing our concept and those that do not.

\section{Additional Qualitative Results}~\label{sec:additional_results}
In the remainder of this document, we provide additional results and comparisons, as follows: 
\begin{enumerate}
    \item In~\Cref{fig:supplementary_our_results_blip,fig:supplementary_our_results_blip_2}, we provide additional personalized captioning results obtained by MyVLM over BLIP-2~\cite{li2023blip}.
    \item In~\Cref{fig:supplementary_our_results_llava,fig:supplementary_our_results_llava_2}, we present additional personalized captioning results of MyVLM over LLaVA~\cite{liu2023llava}.
    \item In~\Cref{fig:supplementary_qualitative_comparisons_blip}, we provide additional comparisons over BLIP-2 with our alternative captioning baselines, both the simple replacement technique and the LLM-guided approach. 
    \item In~\Cref{fig:supplementary_qualitative_comparisons_llava,fig:supplementary_qualitative_comparisons_llava_2}, we present additional visual comparisons to both baselines, applied over LLaVA.
    \item In~\Cref{fig:our_results_blip_llava,fig:our_results_blip_llava_2}, we show personalized captioning obtained by MyVLM over both BLIP-2 and LLaVA on the same set of images, highlighting MyVLM's applicability to both architectures.
    \item In~\Cref{fig:vqa_supp,fig:vqa_llava_sup_2}, we show additional personalized visual question-answering results obtained by MyVLM applied over LLaVA.
    \item Finally, in~\Cref{fig:rec_supplementary}, we present additional personalized referring expression comprehension and captioning results obtained by MyVLM applied over MiniGPT-v2~\cite{chen2023minigptv2}.
\end{enumerate}

\input{figures_supplementary/our_results_blip}
\input{figures_supplementary/our_results_blip_2}
\input{figures_supplementary/our_results_llava}
\input{figures_supplementary/our_results_llava_2}
\input{figures_supplementary/comparisons_blip}
\input{figures_supplementary/comparisons_llava}
\input{figures_supplementary/comparisons_llava_2}
\input{figures_supplementary/our_results_blip_llava}
\input{figures_supplementary/our_results_blip_llava_2}
\input{figures_supplementary/vqa_llava}
\input{figures_supplementary/vqa_llava_2}
\input{figures_supplementary/rec}

%% file: figures_supplementary/dataset.tex
\begin{figure*}[h!]
    \centering
    \addtolength{\belowcaptionskip}{-12.5pt}
    \small
    \begin{tabular}{c c c c c}

        \\ \\ \\ \\ \\ \\
        
        \includegraphics[width=0.15\linewidth]{images/objects/billy_dog/IMG-20240131-WA0026.jpg} &
        \includegraphics[width=0.15\linewidth]{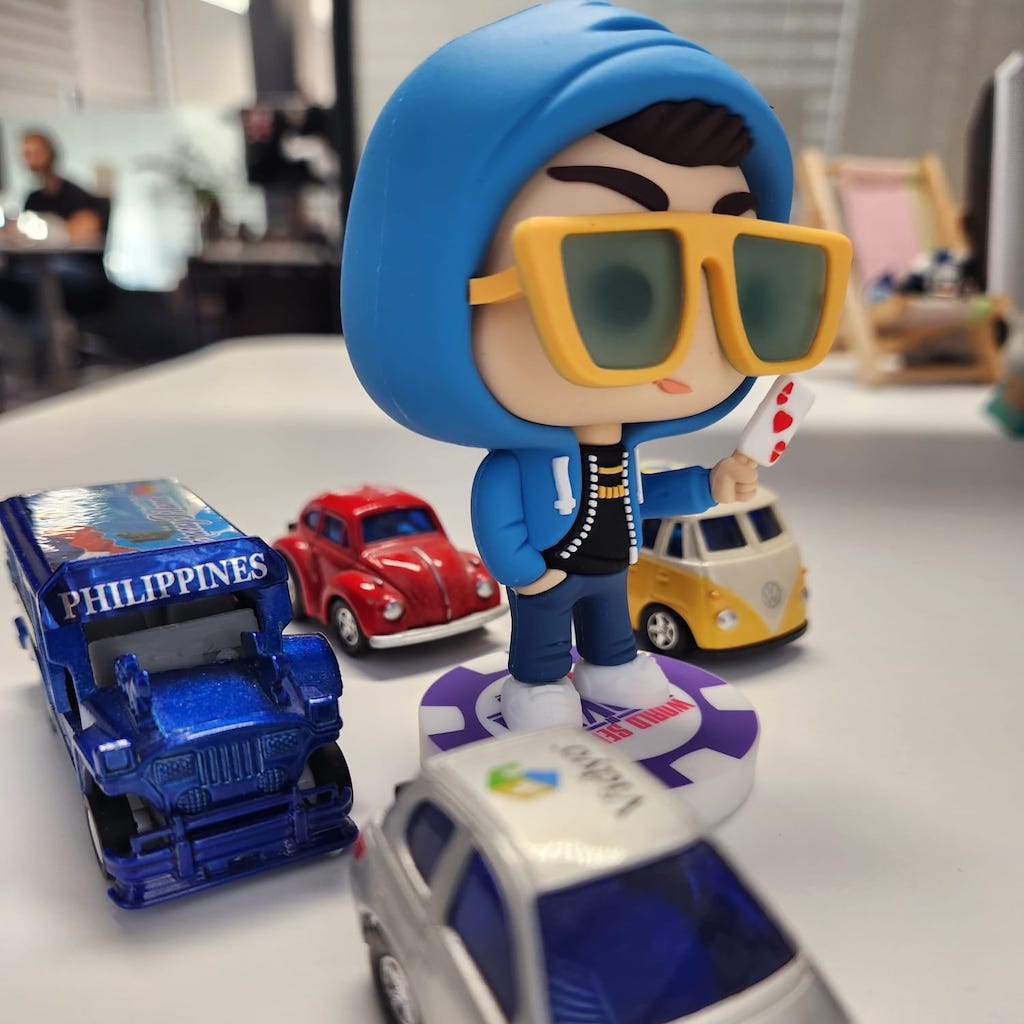} &
        \includegraphics[width=0.15\linewidth]{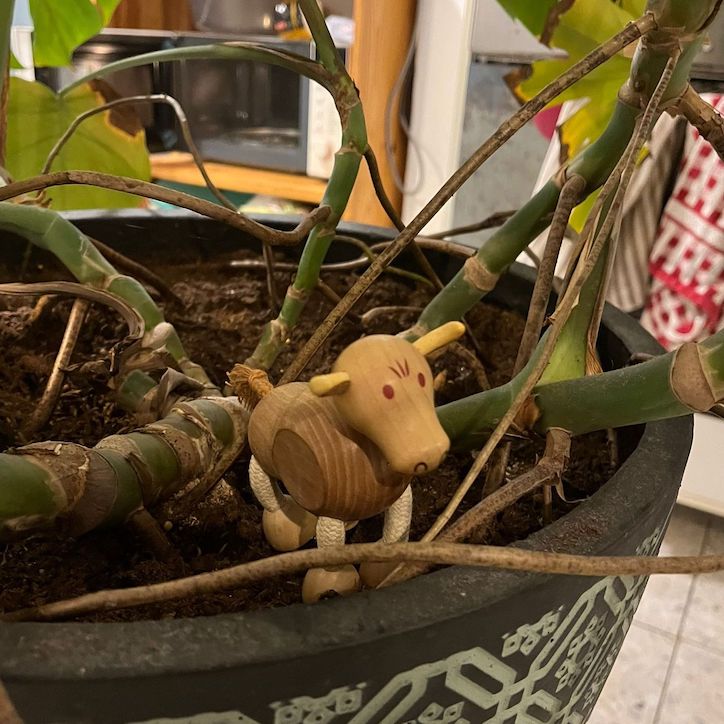} &
        \includegraphics[width=0.15\linewidth]{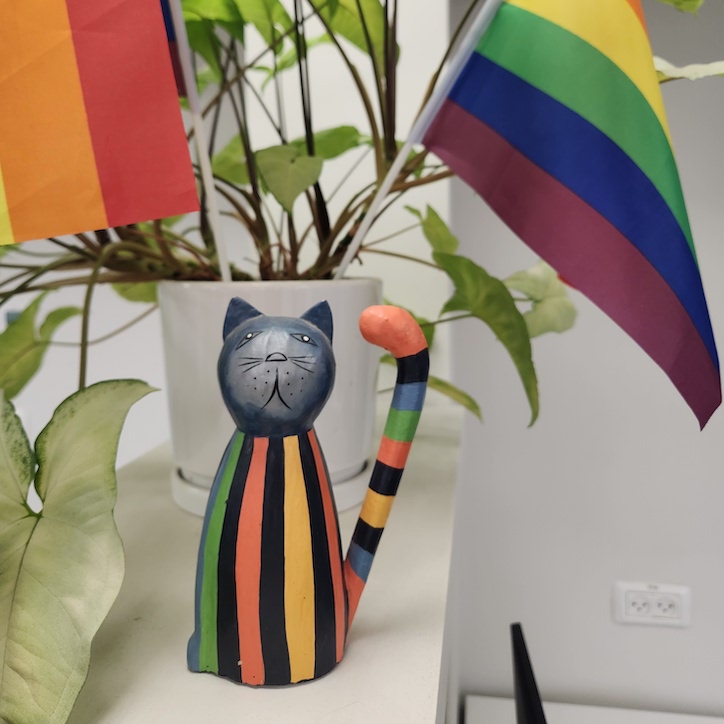} &
        \includegraphics[width=0.15\linewidth]{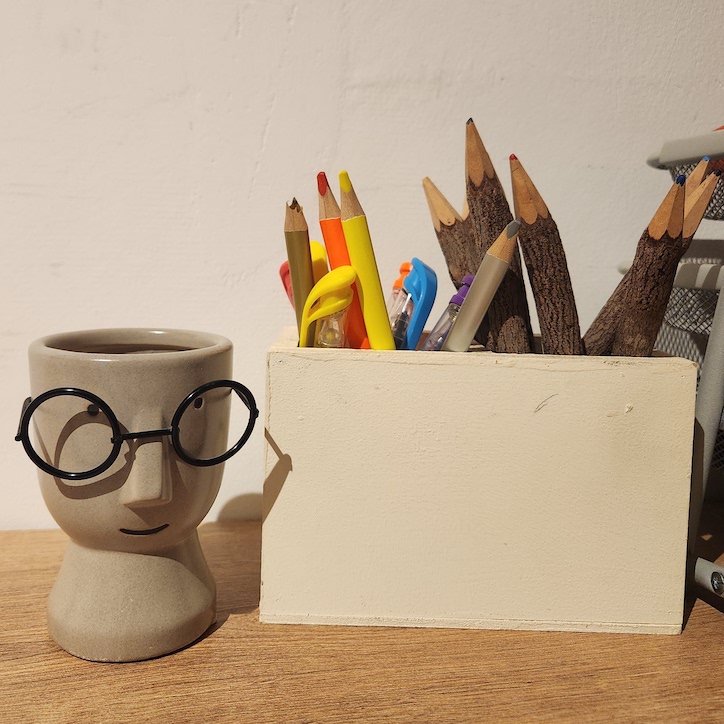} \\

        Billy Dog & Boy Funko Pop & Bull & Cat Statue & Ceramic Head \\

        \includegraphics[width=0.15\linewidth]{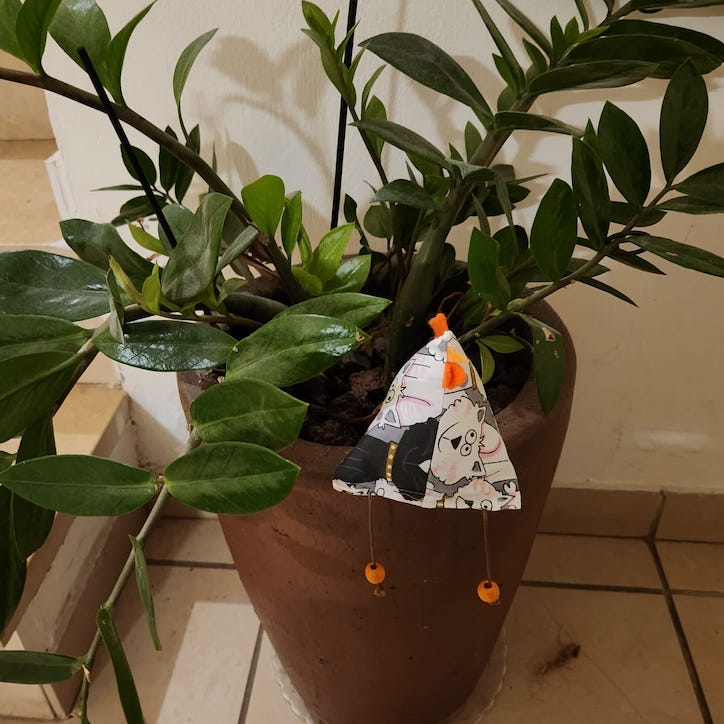} &
        \includegraphics[width=0.15\linewidth]{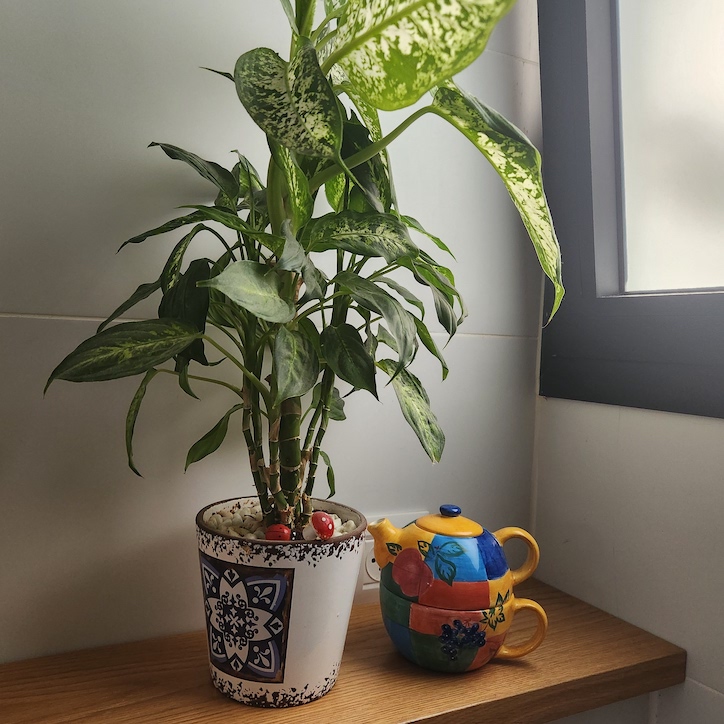} &
        \includegraphics[width=0.15\linewidth]{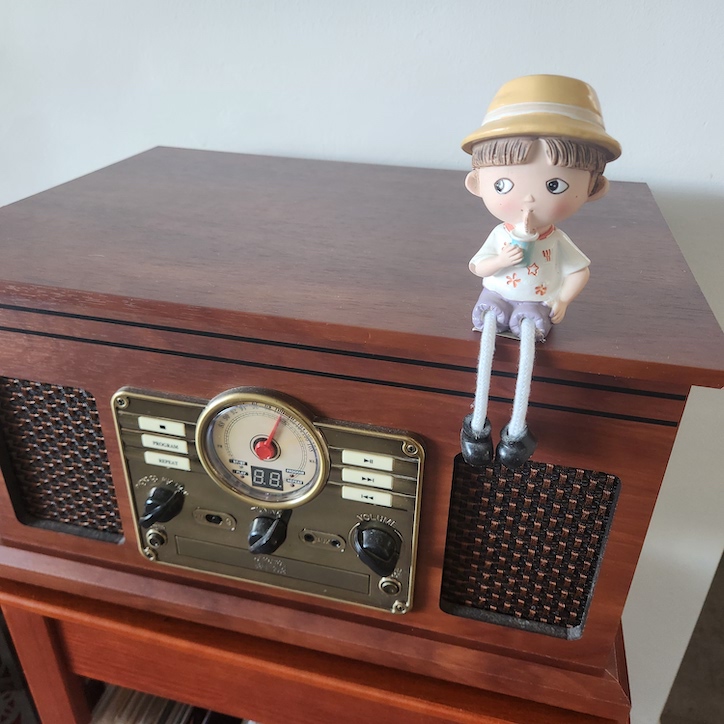} &
        \includegraphics[width=0.15\linewidth]{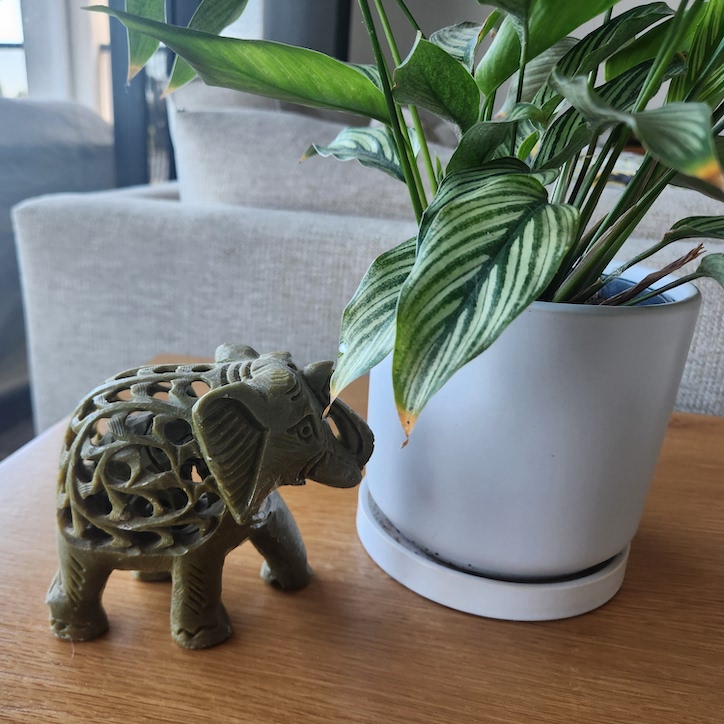} &
        \includegraphics[width=0.15\linewidth]{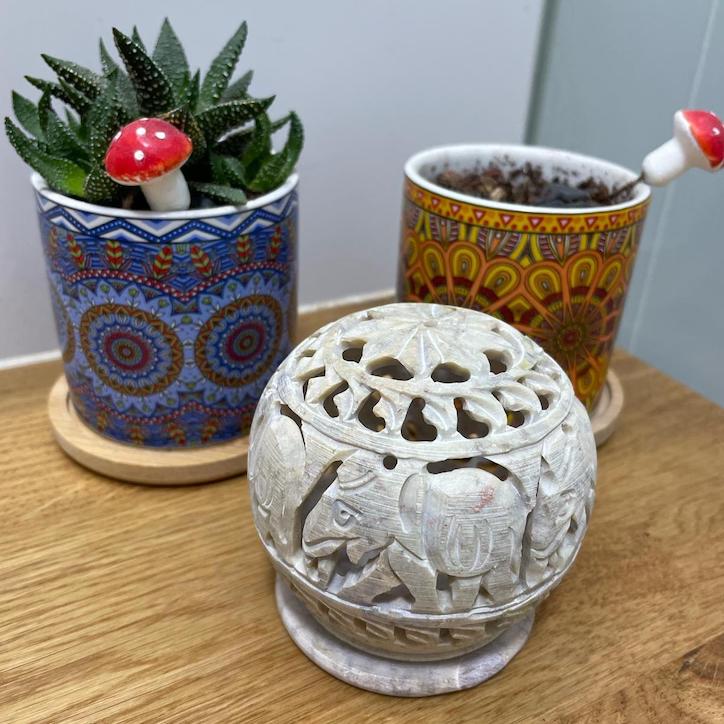} \\

        Chicken Bean Bag & Colorful Teapot & Dangling Child & Elephant & Elephant Sphere \\

        \includegraphics[width=0.15\linewidth]{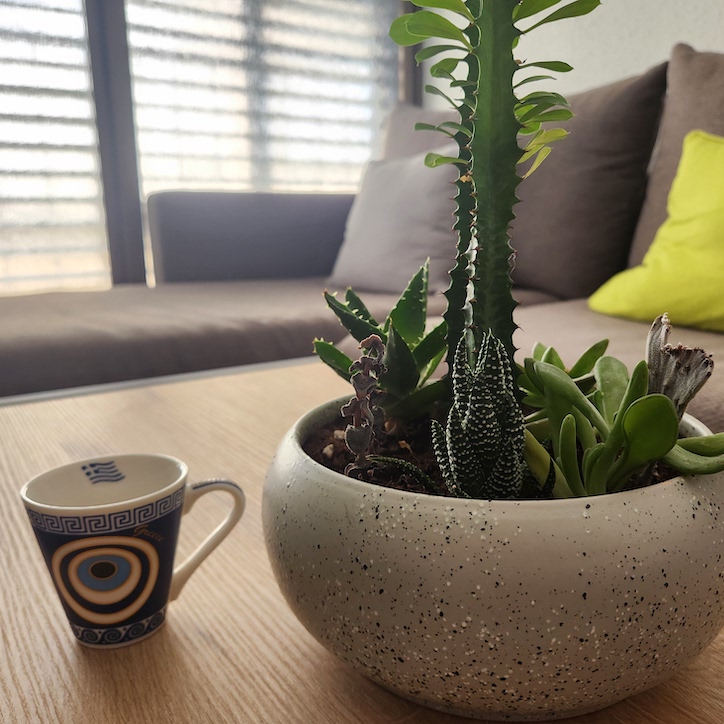} &
        \includegraphics[width=0.15\linewidth]{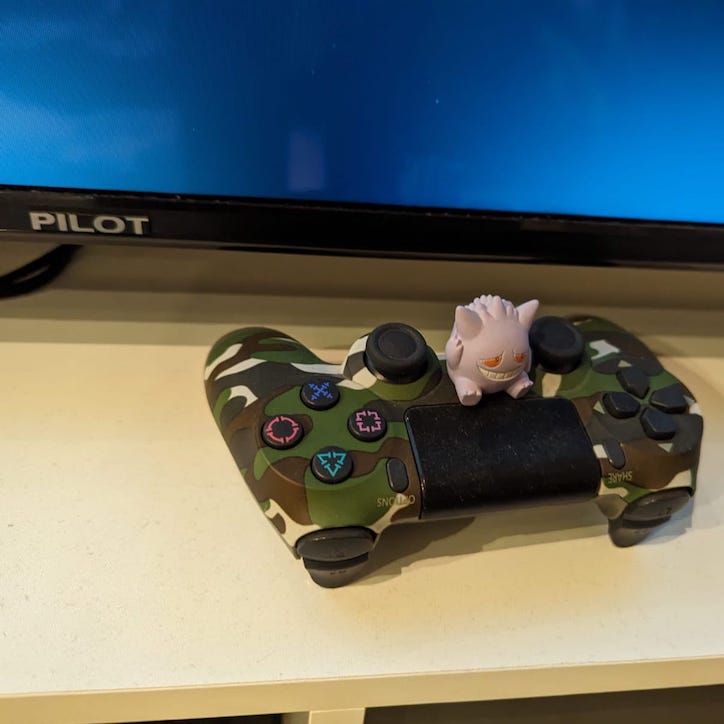} &
        \includegraphics[width=0.15\linewidth]{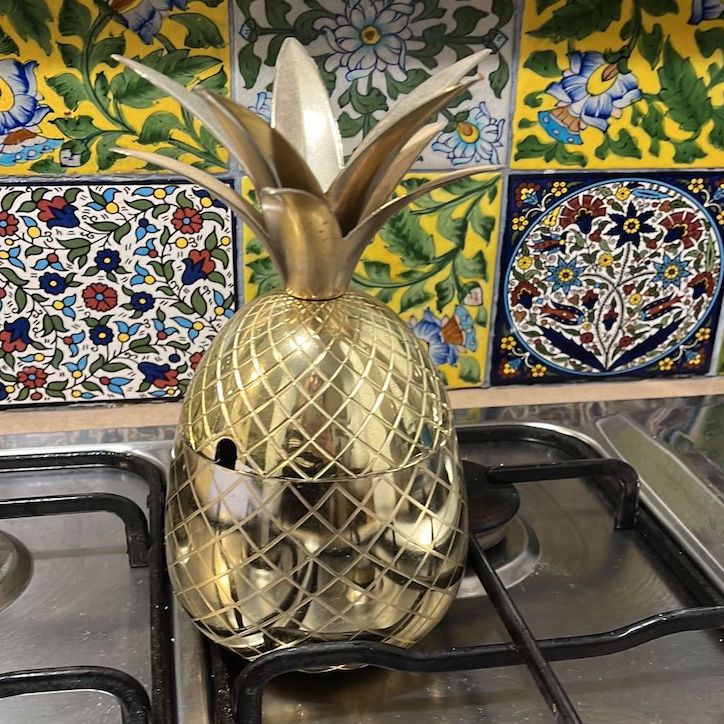} &
        \includegraphics[width=0.15\linewidth]{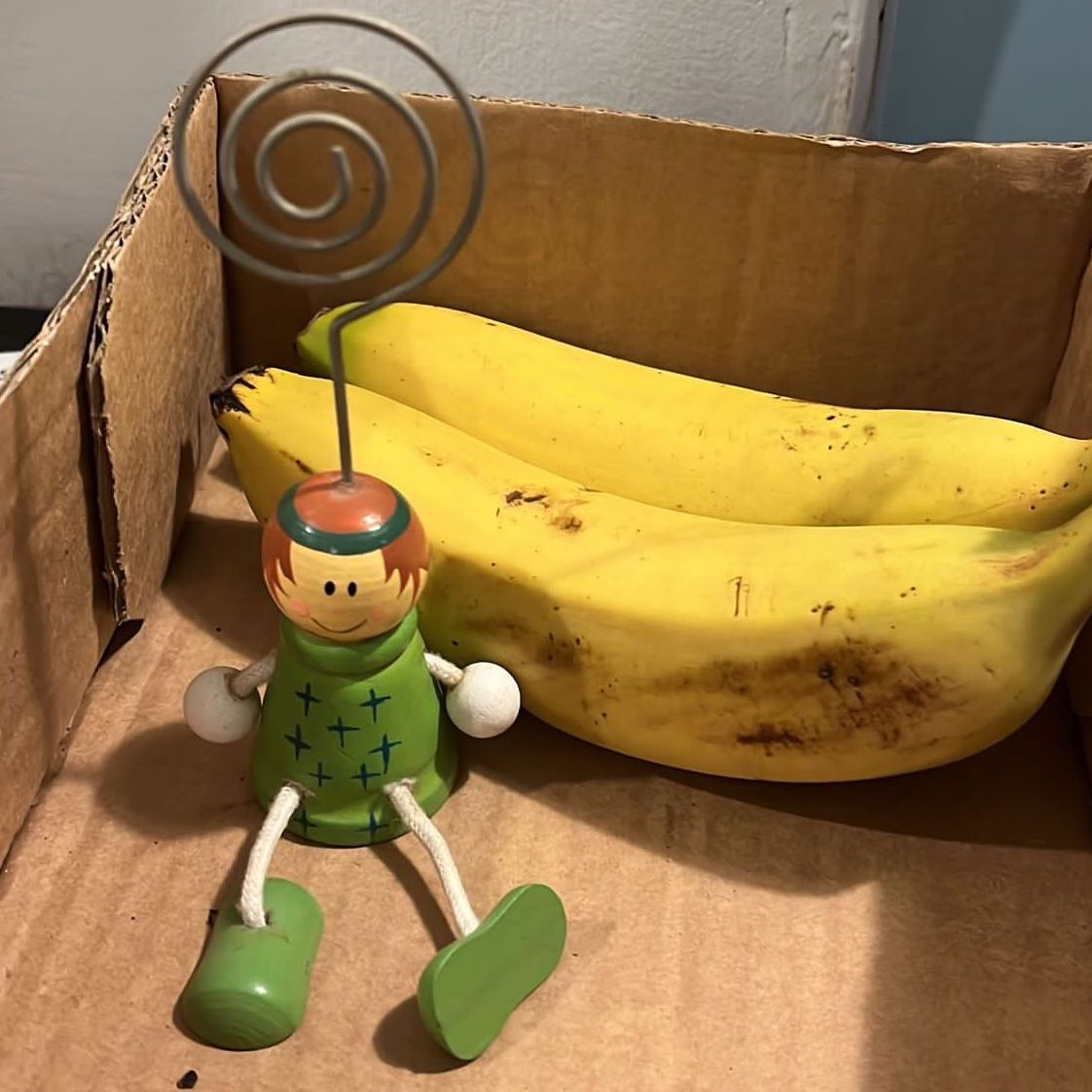} &
        \includegraphics[height=0.15\linewidth,width=0.15\linewidth]{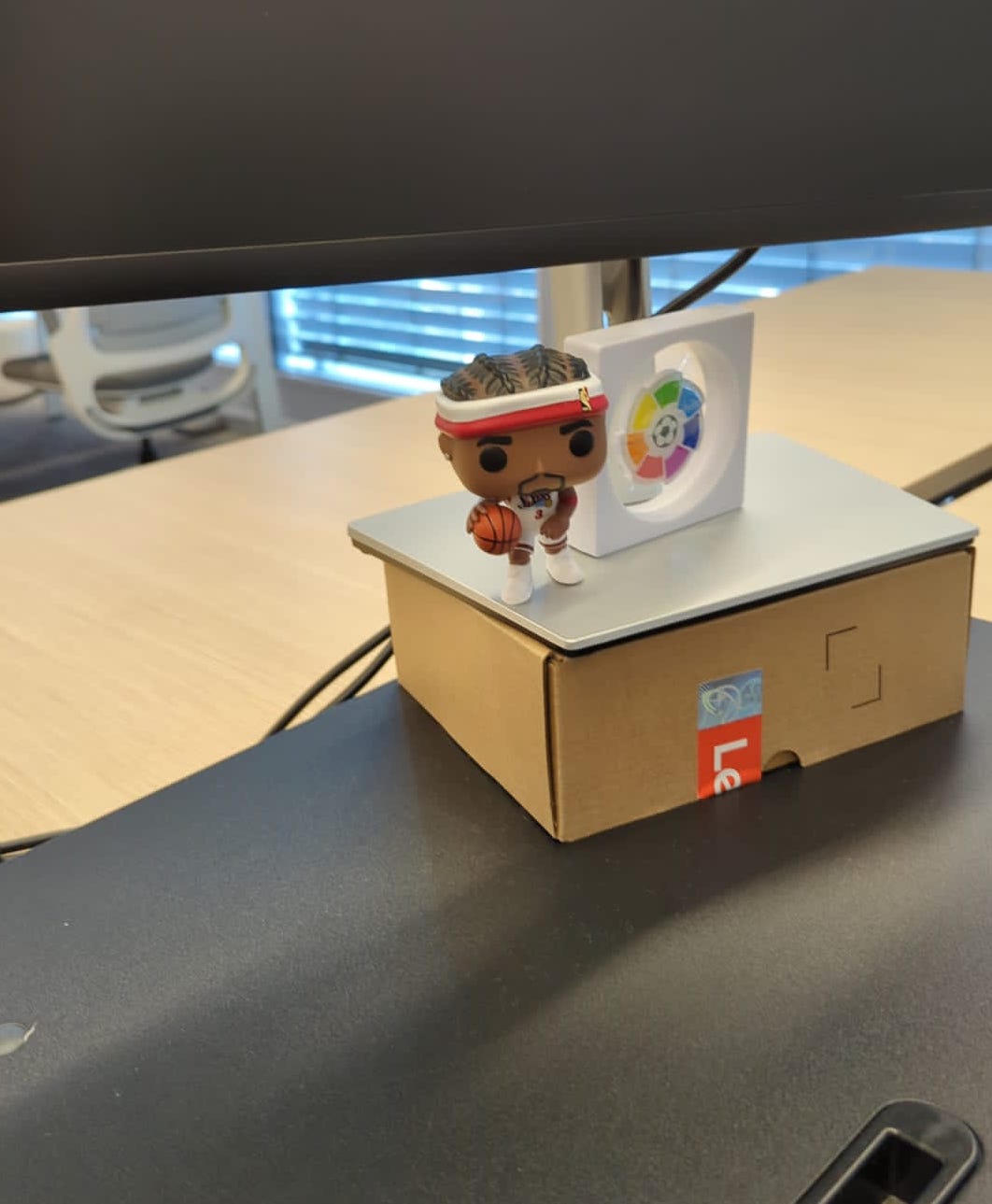} \\

        Espresso Cup & Gengar Toy & Gold Pineapple & Green Doll & Iverson Funko Pop \\

        \includegraphics[width=0.15\linewidth]{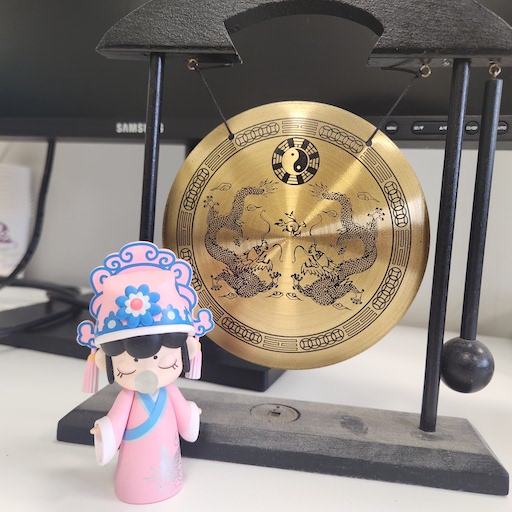} &
        \includegraphics[width=0.15\linewidth]{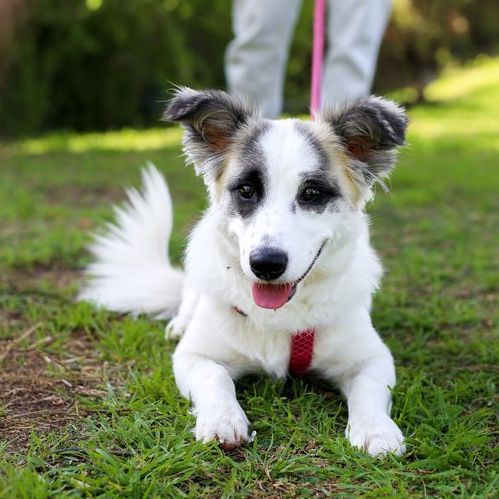} &
        \includegraphics[width=0.15\linewidth]{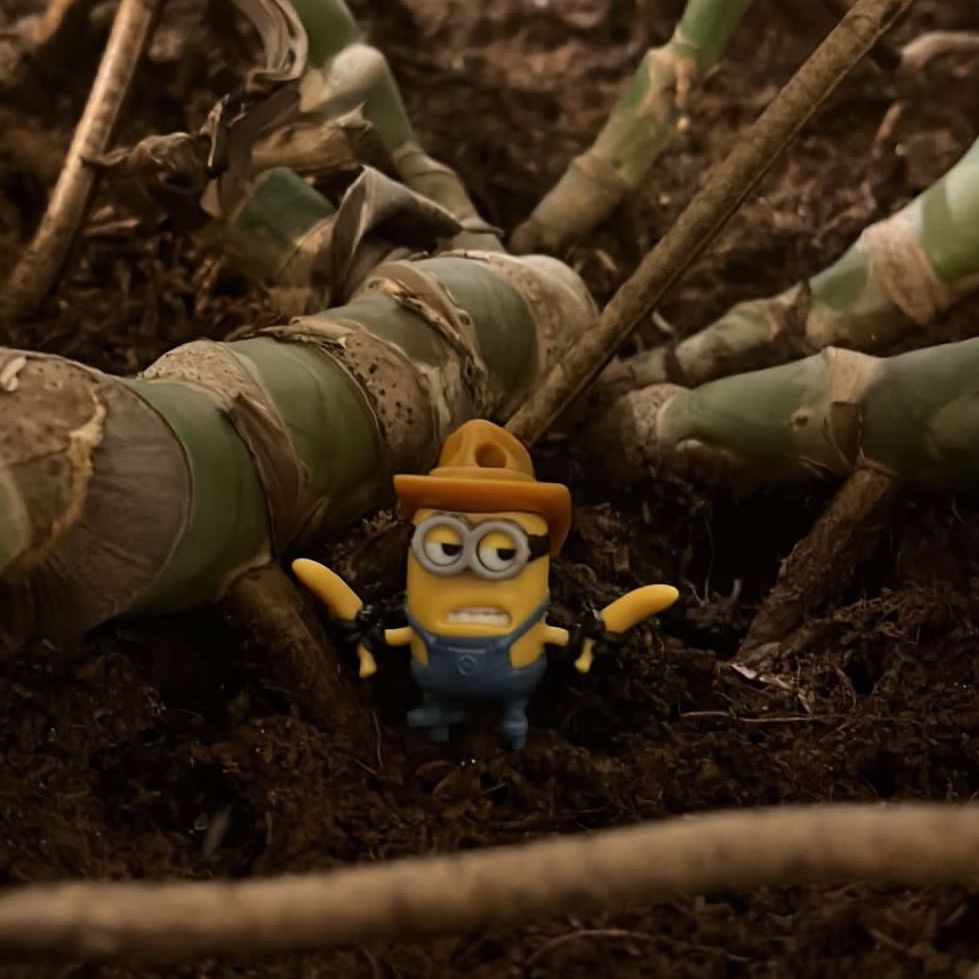} &
        \includegraphics[width=0.15\linewidth]{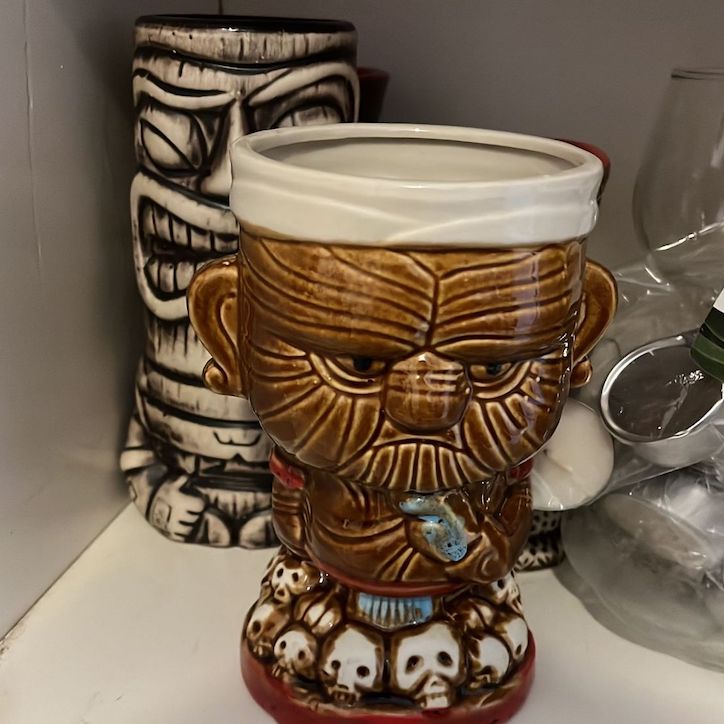} &
        \includegraphics[width=0.15\linewidth]{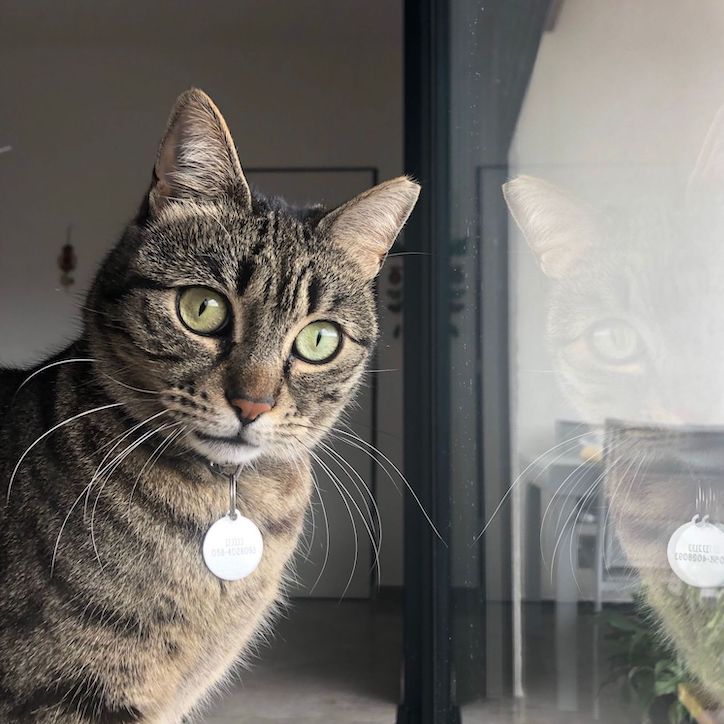} \\

        Asian Doll & Maeve Dog & Minion Toy & Skulls Mug & Cat \\

        \includegraphics[width=0.15\linewidth]{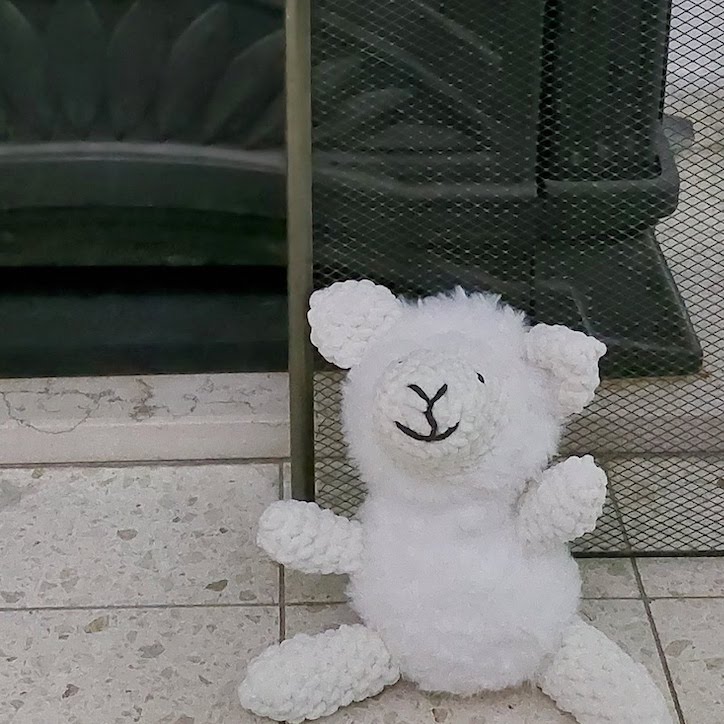} &
        \includegraphics[width=0.15\linewidth]{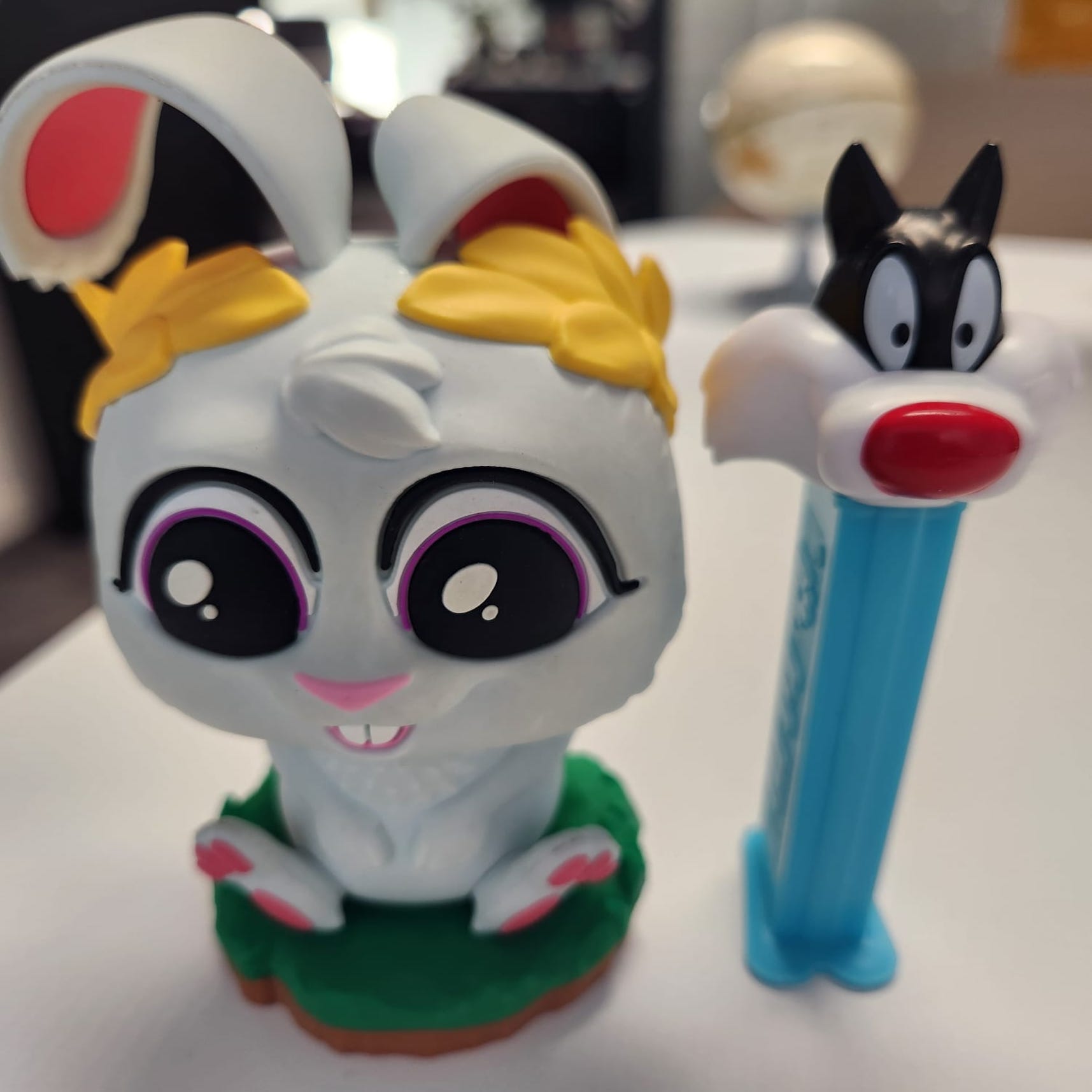} &
        \includegraphics[width=0.15\linewidth]{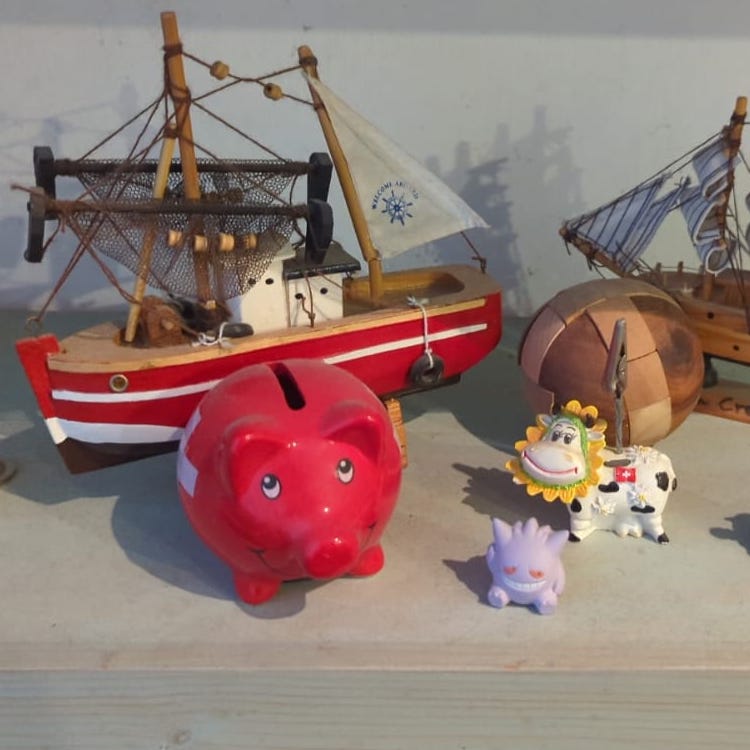} &
        \includegraphics[width=0.15\linewidth]{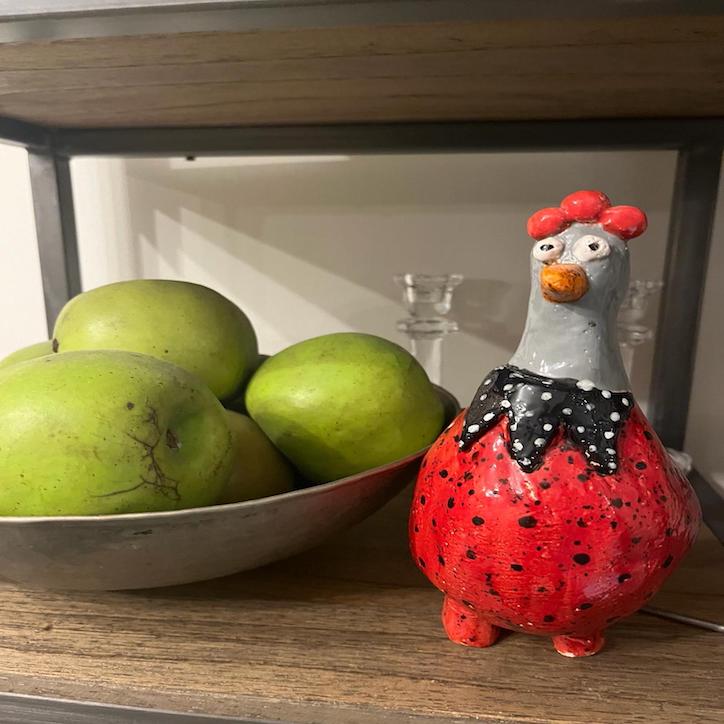} &
        \includegraphics[width=0.15\linewidth]{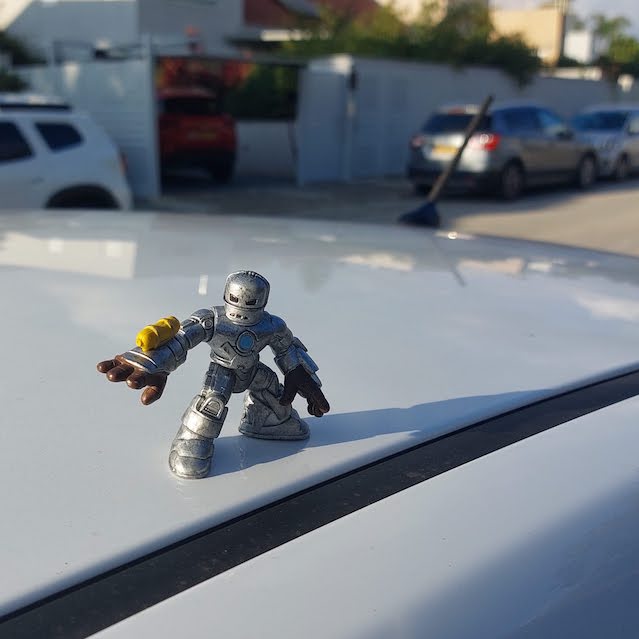} \\

        Sheep Plush & Rabbit Funko Pop & Red Piggy Bank & Red Chicken & Robot Toy \\
        
        \includegraphics[width=0.15\linewidth]{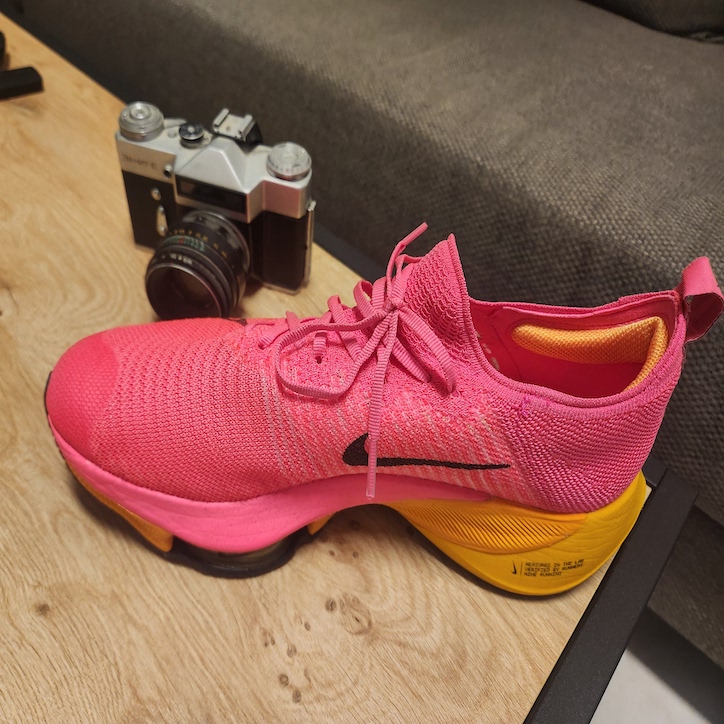} & 
        \includegraphics[width=0.15\linewidth]{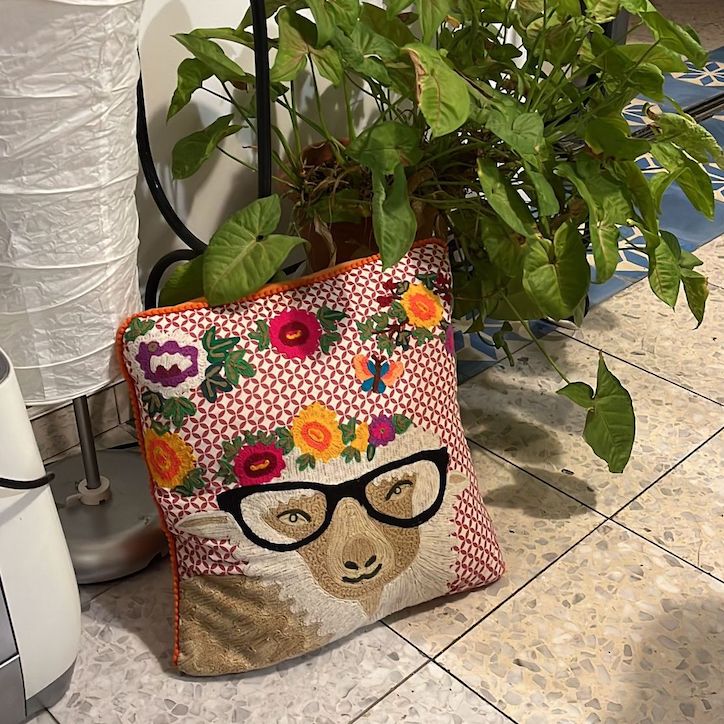} &
        \includegraphics[width=0.15\linewidth]{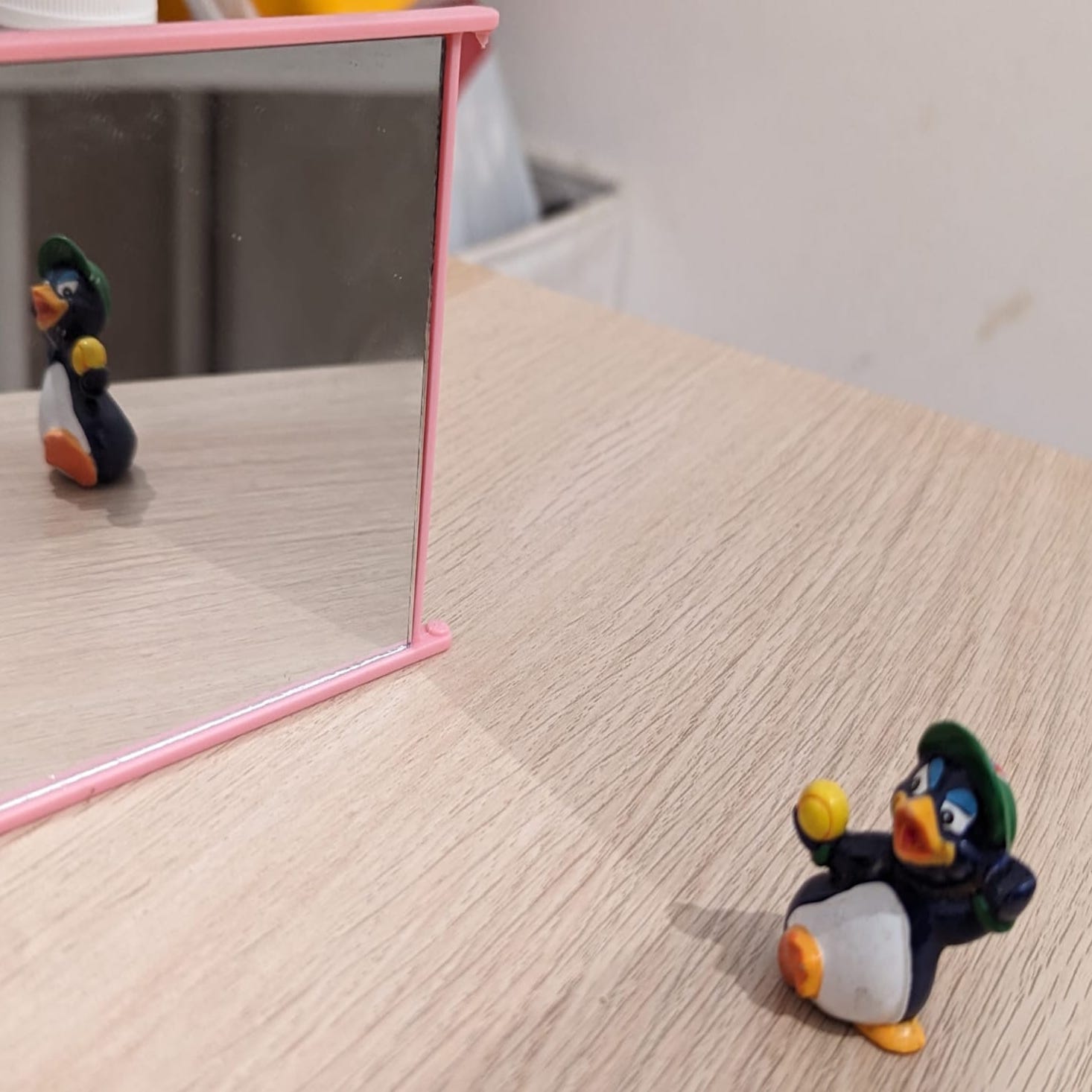} & 
        \includegraphics[width=0.15\linewidth]{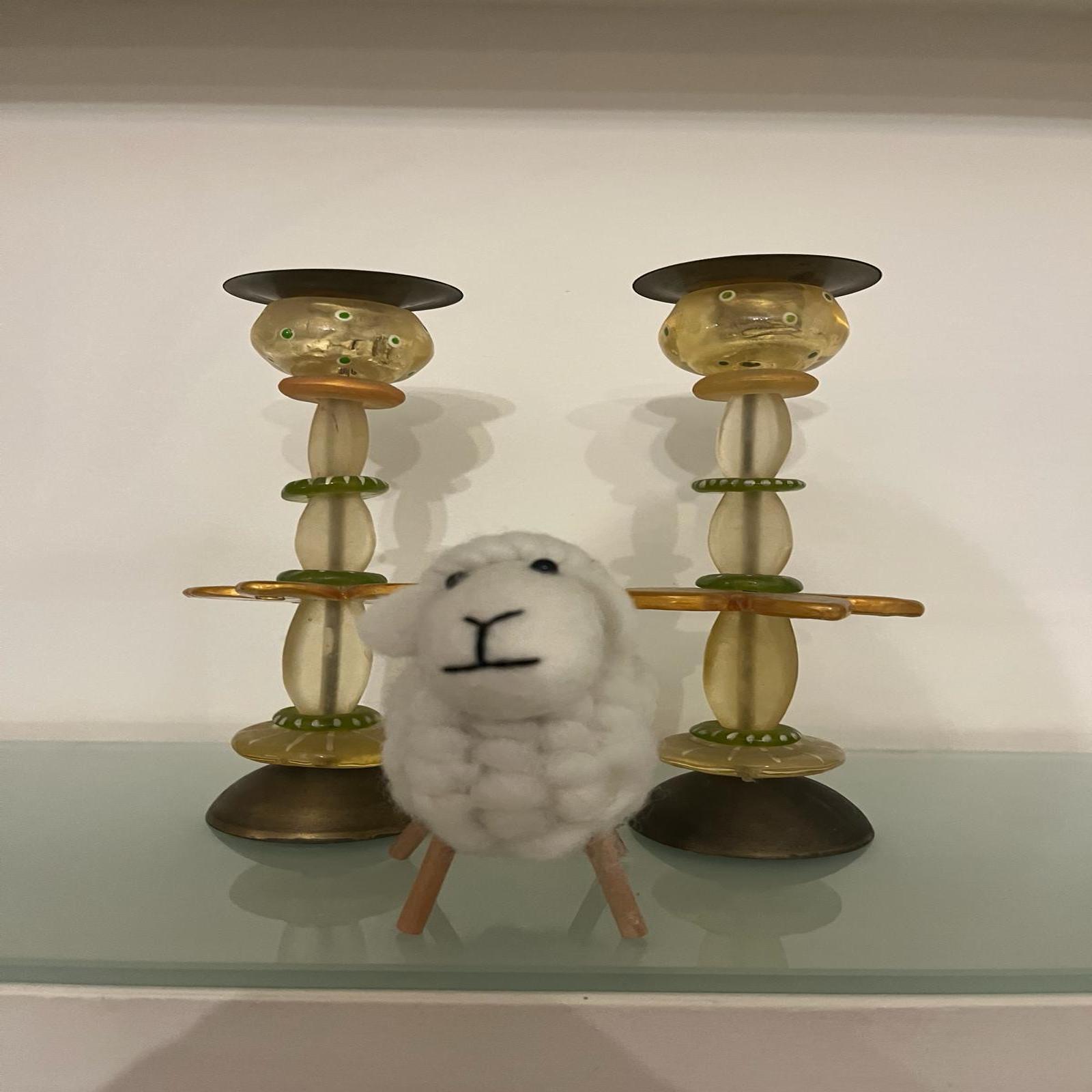} & \\

        Running Shoes & Sheep Pillow & Small Penguin Toy & Sheep Toy 

        \\ \\
        
    \end{tabular}
    \vspace{-0.2cm}
    \caption{\textbf{MyVLM Dataset.} Example images for each object in our constructed dataset.}
    \label{fig:dataset}
    \vspace{1cm}
\end{figure*}

%% file: tables/vqa_questions.tex
\begin{table*}
\small
\centering
\setlength{\tabcolsep}{3pt}   
\renewcommand{\arraystretch}{1.2}
\caption{A list of the $10$ language instructions used when optimizing the concept embedding for personalized visual question-answering.\\[-0.3cm]}
{\small
\begin{tabular}{p{0.475\linewidth} | p{0.475\linewidth}}
    \toprule
    \textbf{Objects} & \textbf{People} \\
    \midrule

    What color is $\langle$concept$\rangle$? & What is $\langle$concept$\rangle$ wearing in the image? \\
    Where is $\langle$concept$\rangle$ in the image? & What color shirt is $\langle$concept$\rangle$ wearing? \\
    Where is $\langle$concept$\rangle$ positioned in the image? & What is $\langle$concept$\rangle$ doing in the image? \\
    Does $\langle$concept$\rangle$ appear to be the main subject of the image? & Where is $\langle$concept$\rangle$ in the image? \\
    What objects is $\langle$concept$\rangle$ interacting with in the image? & Can you describe what $\langle$concept$\rangle$ is wearing? \\
    How would you describe the texture of $\langle$concept$\rangle$ in the image? & From left to right, where is $\langle$concept$\rangle$ positioned in the image? \\
    What types of materials is $\langle$concept$\rangle$ be made of? & What kind of hair does $\langle$concept$\rangle$ have? \\
    Is $\langle$concept$\rangle$ large or small in the image? & What is the expression on $\langle$concept$\rangle$ face? \\
    Is $\langle$concept$\rangle$ close to the camera or far away? & Is there anything unique about $\langle$concept$\rangle$'s appearance? \\
    Please caption this image of $\langle$concept$\rangle$ & Please caption this image of $\langle$concept$\rangle$ \\
    \bottomrule
\end{tabular}
}
\label{tb:vqa_prompts}
\vspace{-0.3cm}
\end{table*}

%% file: figures_supplementary/flamingo_comparison.tex
\begin{figure*}
    \centering
    \renewcommand{\arraystretch}{1}
    \footnotesize
    \vspace{-0.2cm}
    \begin{tabular}{p{0.175\textwidth} p{0.175\textwidth} p{0.175\textwidth} p{0.175\textwidth} p{0.175\textwidth}}

        \setlength\tabcolsep{0pt}
        \begin{center}
        \begin{tabular}{c c c}
            \includegraphics[height=0.06\textwidth,width=0.06\textwidth]{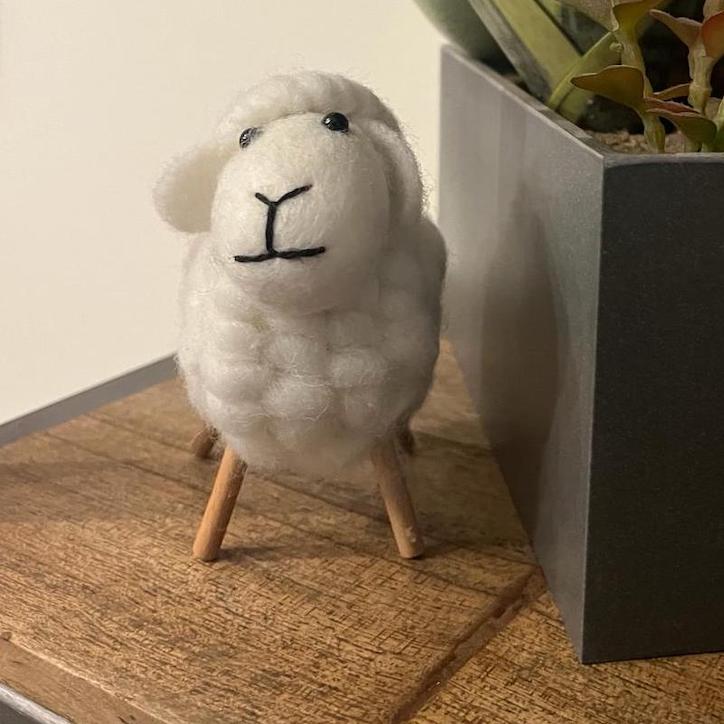} &
            \includegraphics[height=0.06\textwidth,width=0.06\textwidth]{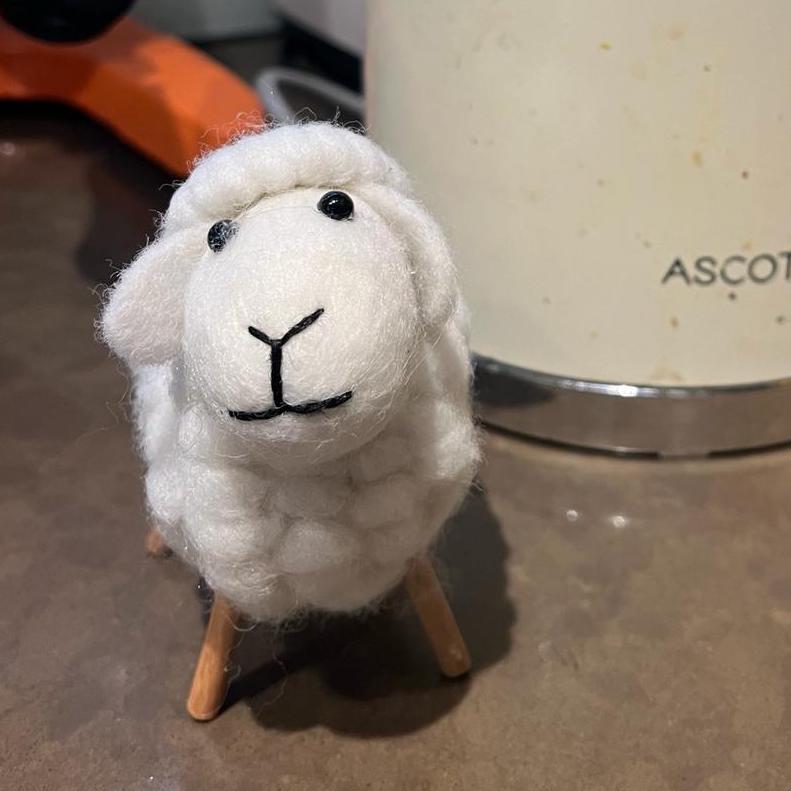} & 
            \includegraphics[height=0.06\textwidth,width=0.06\textwidth]{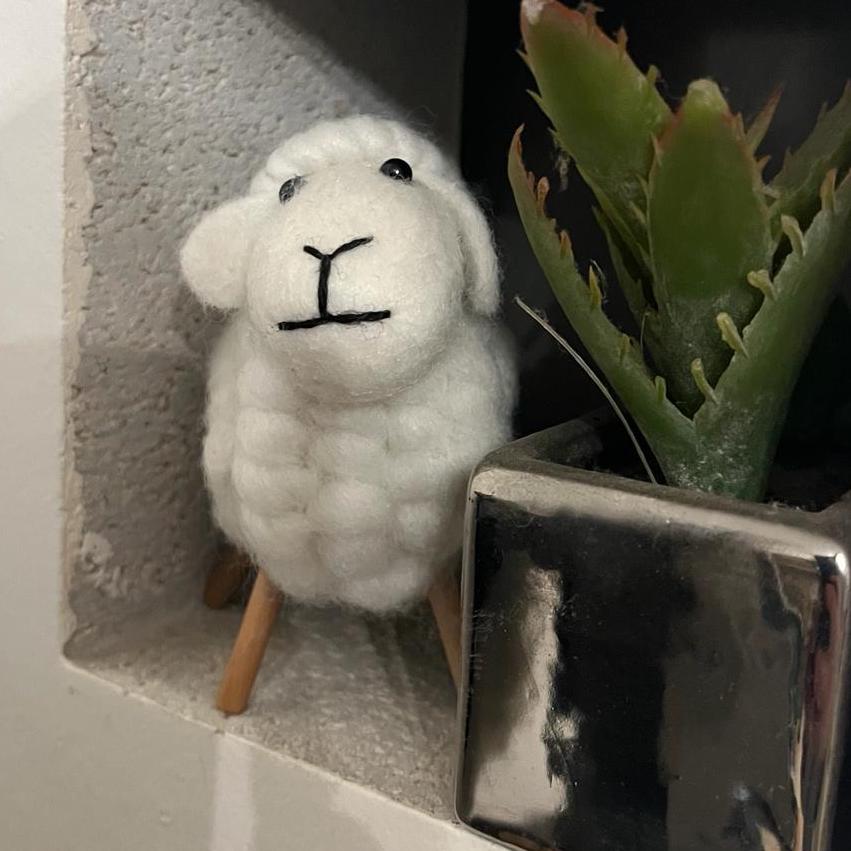}
        \end{tabular} 
        \end{center} &
        \setlength\tabcolsep{0pt}
        \begin{center}
        \begin{tabular}{c c c}
            \includegraphics[height=0.06\textwidth,width=0.06\textwidth]{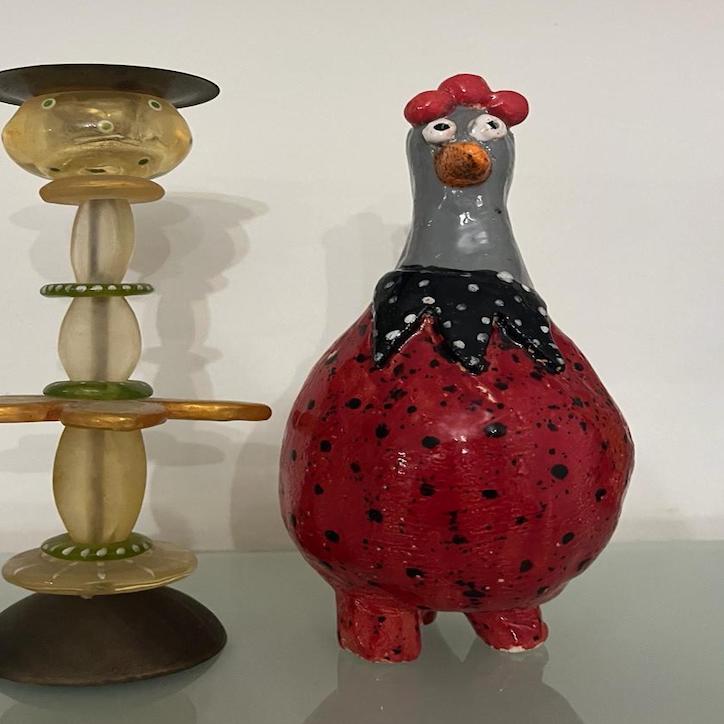} & 
            \includegraphics[height=0.06\textwidth,width=0.06\textwidth]{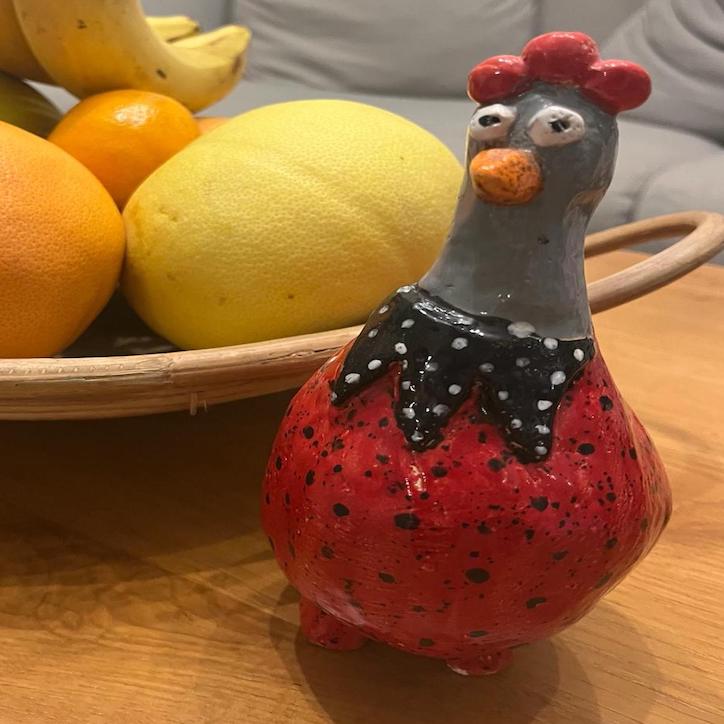} & 
            \includegraphics[height=0.06\textwidth,width=0.06\textwidth]{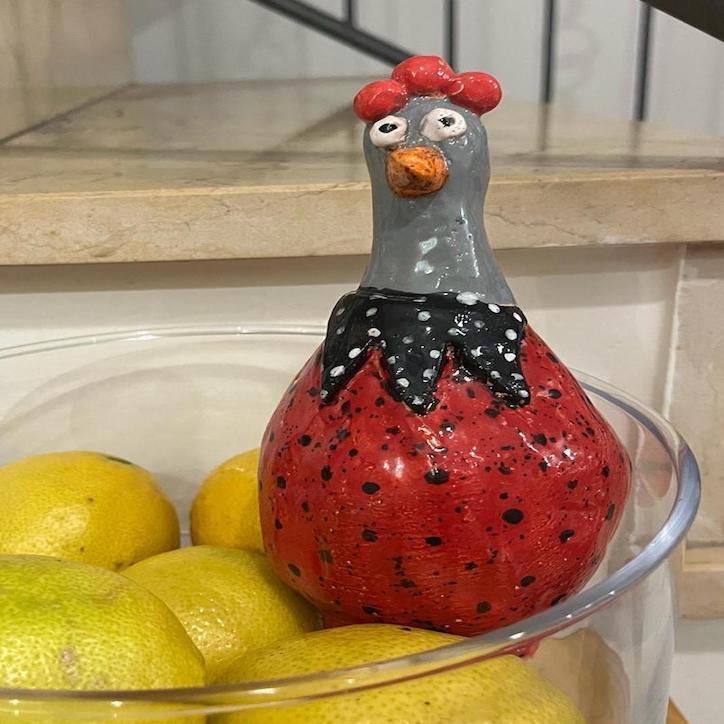} 
        \end{tabular} 
        \end{center} &
        \setlength\tabcolsep{0pt}
        \begin{center}
        \begin{tabular}{c c c}
            \includegraphics[height=0.06\textwidth,width=0.06\textwidth]{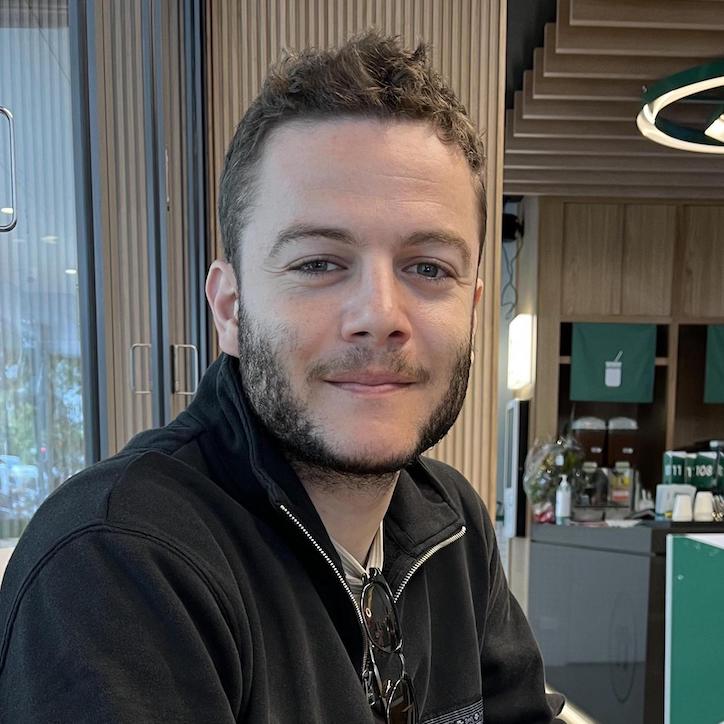} &
            \includegraphics[height=0.06\textwidth,width=0.06\textwidth]{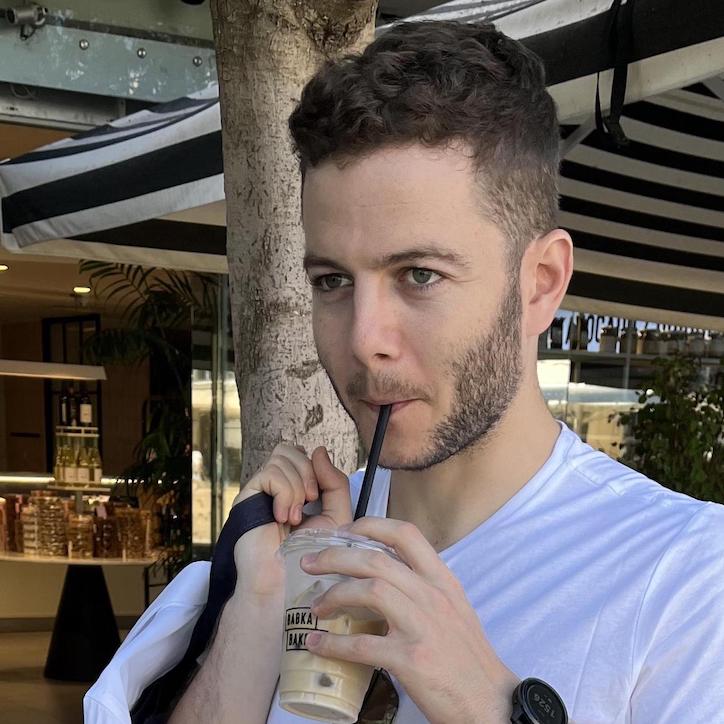} &
            \includegraphics[height=0.06\textwidth,width=0.06\textwidth]{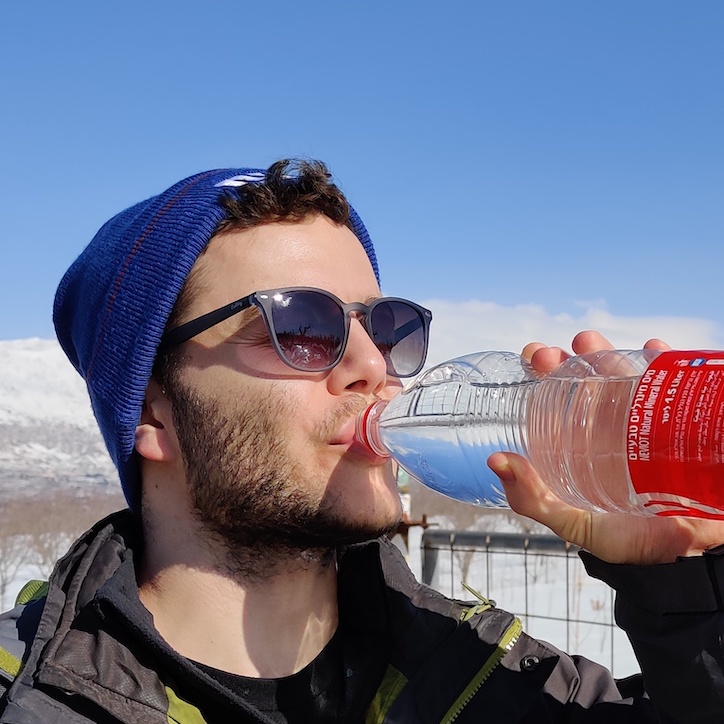}
        \end{tabular} 
        \end{center} &
        \setlength\tabcolsep{0pt}
        \begin{center}
        \begin{tabular}{c c c}
            \includegraphics[height=0.06\textwidth,width=0.06\textwidth]{images/people/shay/cropped/image_2.jpg} & 
            \includegraphics[height=0.06\textwidth,width=0.06\textwidth]{images/people/shay/cropped/IMG-20240209-WA0012.jpg} & 
            \includegraphics[height=0.06\textwidth,width=0.06\textwidth]{images/people/shay/cropped/IMG-20240209-WA0016.jpg}
        \end{tabular} 
        \end{center} &
        \setlength\tabcolsep{0pt}
        \begin{center}
        \begin{tabular}{c c c}
            \includegraphics[height=0.06\textwidth,width=0.06\textwidth]{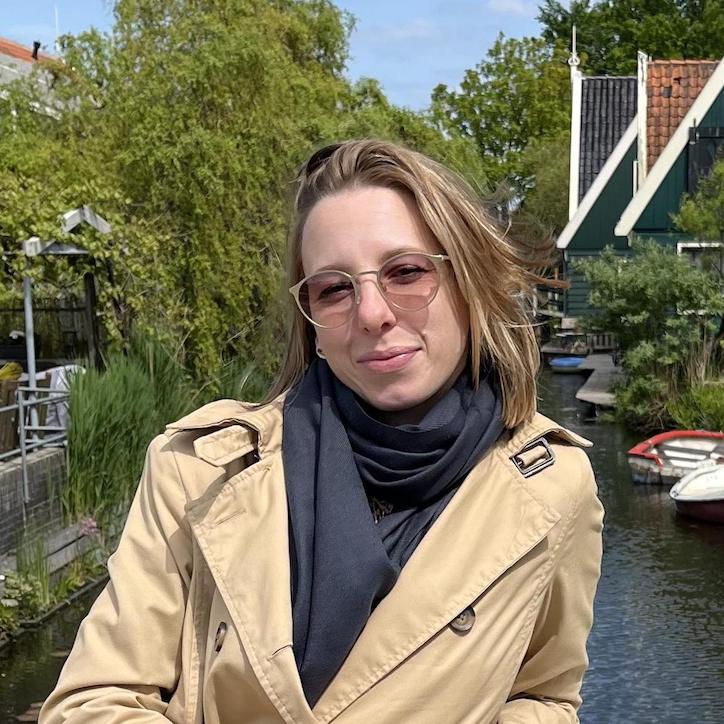} &
            \includegraphics[height=0.06\textwidth,width=0.06\textwidth]{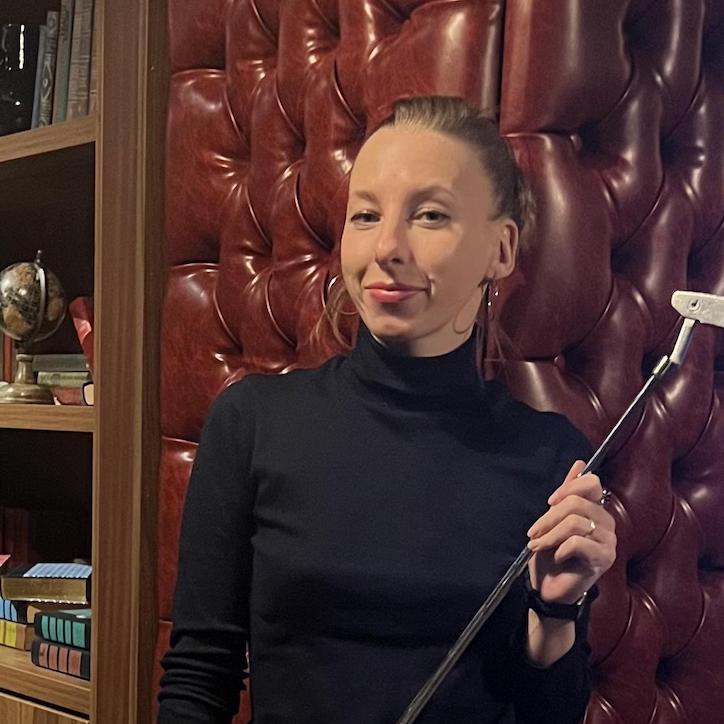} & 
            \includegraphics[height=0.06\textwidth,width=0.06\textwidth]{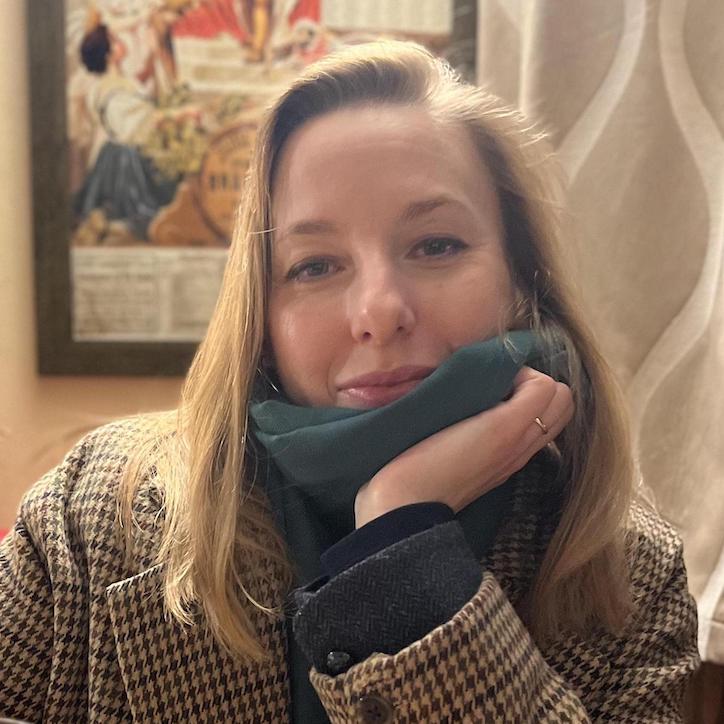}
        \end{tabular} 
        \end{center} \\[-0.8cm]
    
        \begin{center} \includegraphics[width=0.18\textwidth]{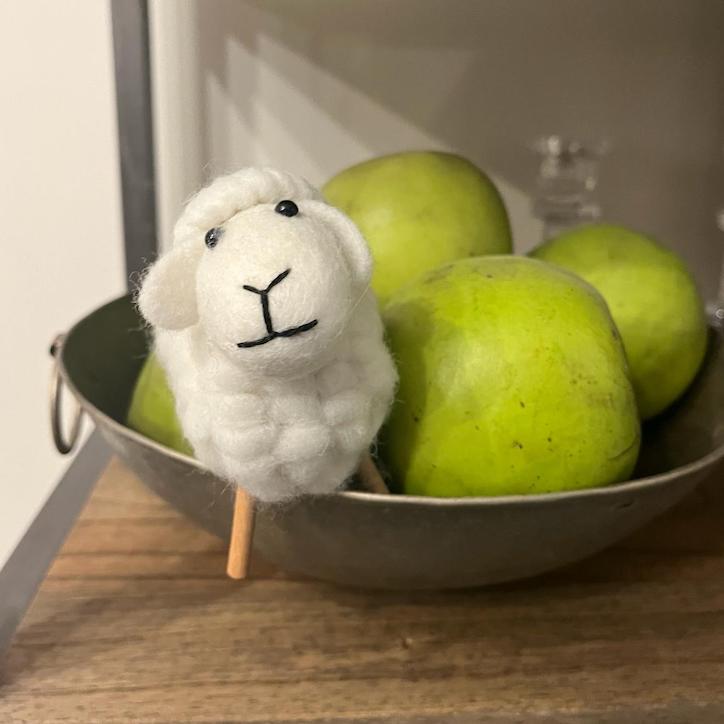} \end{center} &
        \begin{center} \includegraphics[width=0.18\textwidth]{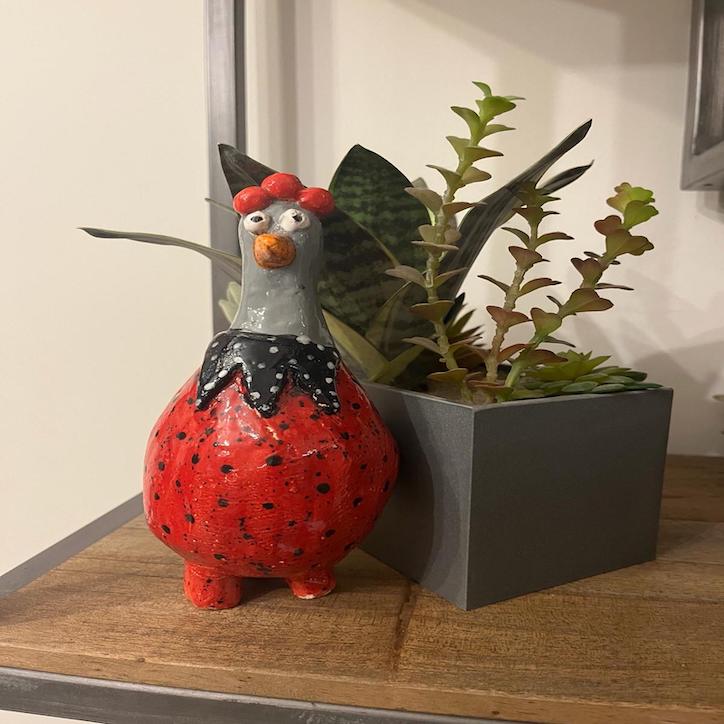} \end{center} &
        \begin{center} \includegraphics[width=0.18\textwidth]{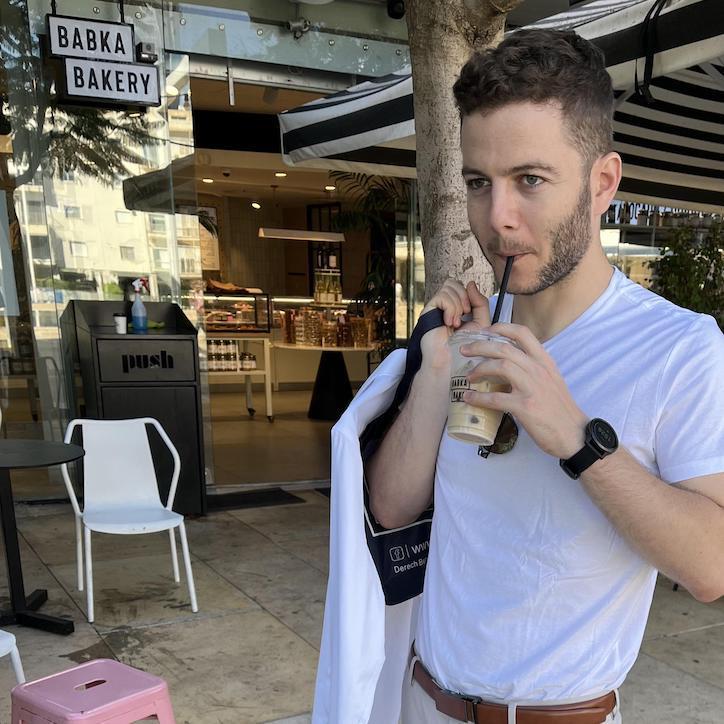} \end{center} &
        \begin{center} \includegraphics[width=0.18\textwidth]{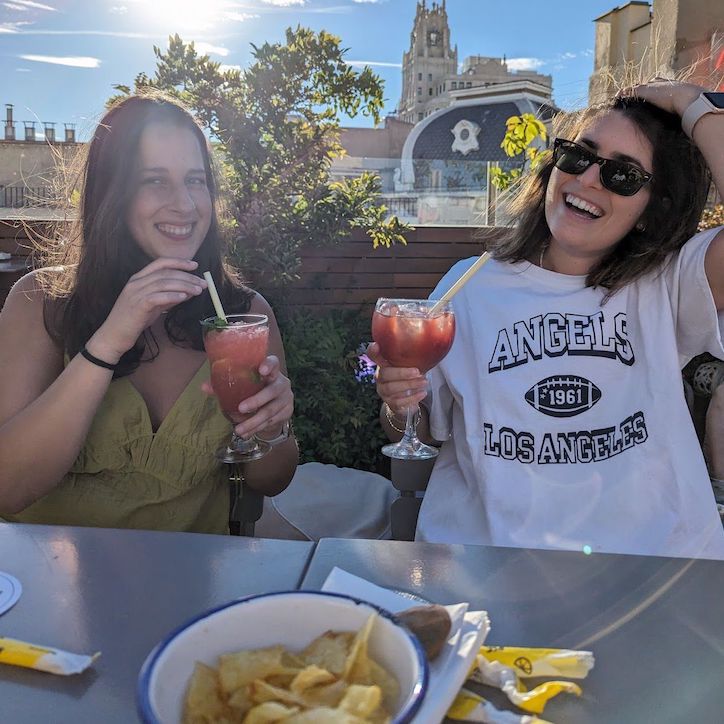} \end{center} &
        \begin{center} \includegraphics[width=0.18\textwidth]{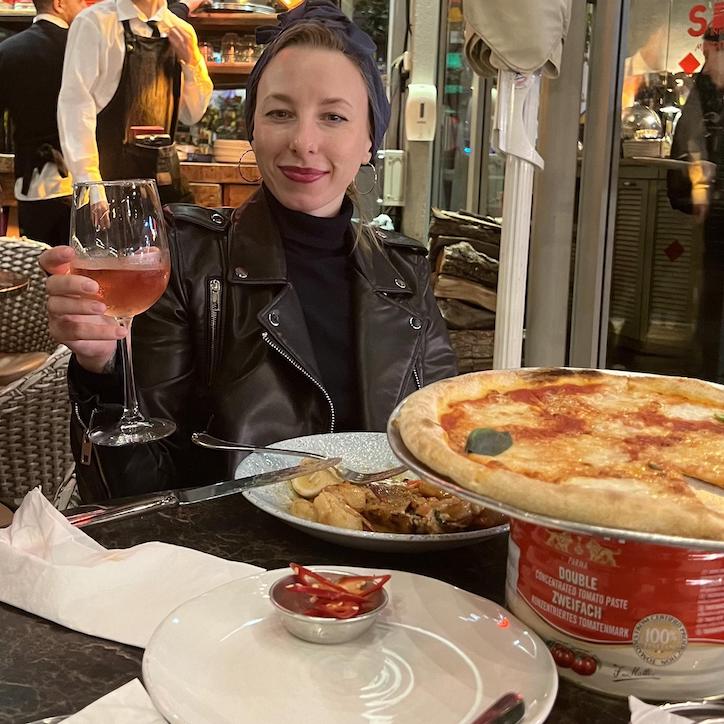} \end{center} \\[-0.8cm]

        \begin{center} \textbf{OpenFlamingo} \end{center} &
        \begin{center} \textbf{OpenFlamingo} \end{center} &
        \begin{center} \textbf{OpenFlamingo} \end{center} &
        \begin{center} \textbf{OpenFlamingo} \end{center} &
        \begin{center} \textbf{OpenFlamingo} \end{center} \\[-0.8cm]

        \begin{center} \footnotesize ``A sheep with a \Sstar'' \end{center} &
        \begin{center} \footnotesize ``A red rooster with a black hat and a red bow tie'' \end{center} &
        \begin{center} \footnotesize ``He is wearing a white t-shirt and khaki pants'' \end{center} &
        \begin{center} \footnotesize ``\Sstar and her friend enjoying a drink at a rooftop bar in Barcelona'' \end{center} &
        \begin{center} \footnotesize ``\Sstar with a glass of wine in her hand and a slice of pizza in her hand'' \end{center} \\[-0.65cm]
        
        \begin{center} \textbf{MyBLIP-2} \end{center} &
        \begin{center} \textbf{MyBLIP-2} \end{center} &
        \begin{center} \textbf{MyBLIP-2} \end{center} &
        \begin{center} \textbf{MyBLIP-2} \end{center} &
        \begin{center} \textbf{MyBLIP-2} \end{center} \\[-0.8cm]

        \begin{center} \footnotesize ``\Sstar sitting in a bowl of green apples'' \end{center} &
        \begin{center} \footnotesize ``\Sstar sitting next to a plant on a shelf'' \end{center} &
        \begin{center} \footnotesize ``\textcolor{blue}{$S_*$}, wearing white shirts and pants, is enjoying a beverage outdoors.'' \end{center} &
        \begin{center} \footnotesize ``\textcolor{blue}{$S_*$}, with a glass of wine and a strawberry margarita, at a restaurant in Madrid'' \end{center} &
        \begin{center} \footnotesize ``\textcolor{blue}{$S_*$}, wearing a black leather jacket, is enjoying a glass of wine at an Italian restaurant'' \end{center} \\[-0.8cm]
        
    \end{tabular}

    \begin{tabular}{p{0.175\textwidth} p{0.175\textwidth} p{0.175\textwidth} p{0.175\textwidth} p{0.175\textwidth}}

        \setlength\tabcolsep{0pt}
        \begin{center}
        \begin{tabular}{c c c}
            \includegraphics[height=0.06\textwidth,width=0.06\textwidth,height=0.06333\textwidth]{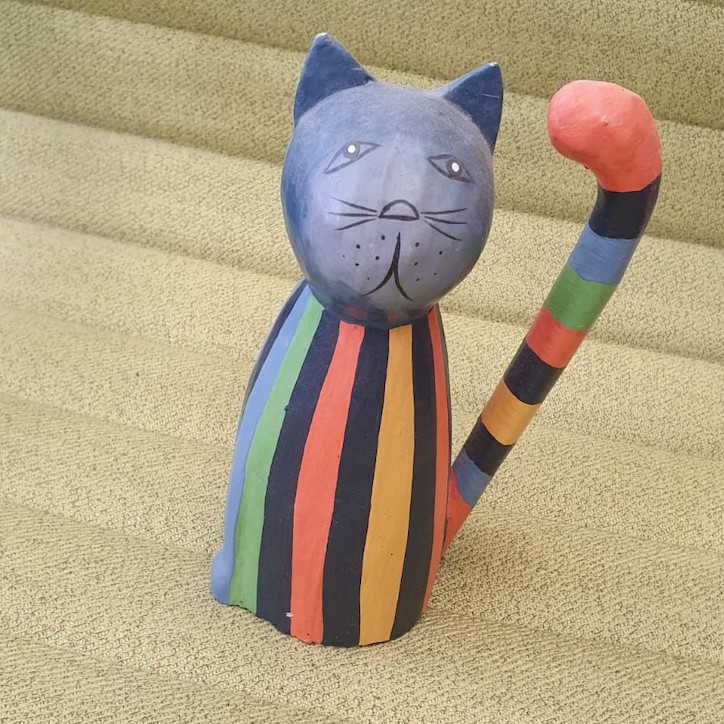} & 
            \includegraphics[height=0.06\textwidth,width=0.06\textwidth,height=0.06333\textwidth]{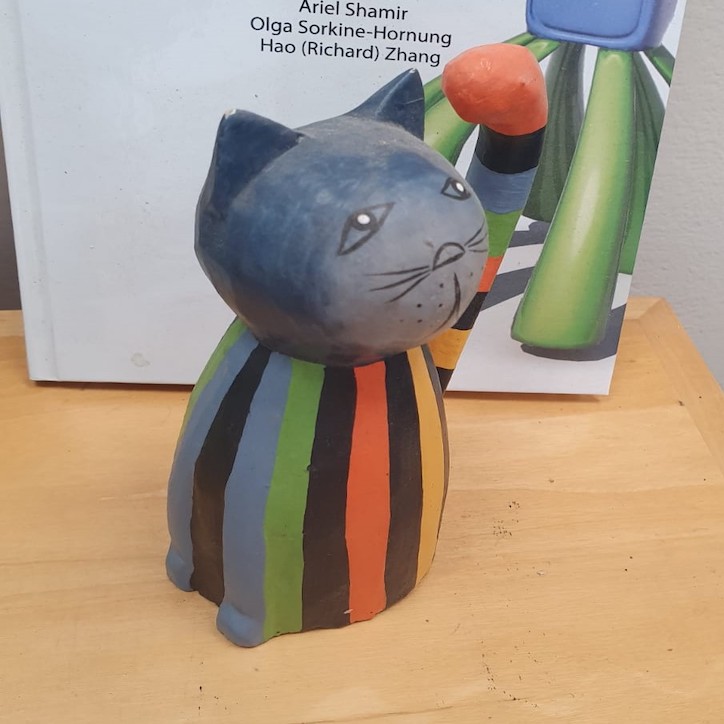} & 
            \includegraphics[height=0.06\textwidth,width=0.06\textwidth,height=0.06333\textwidth]{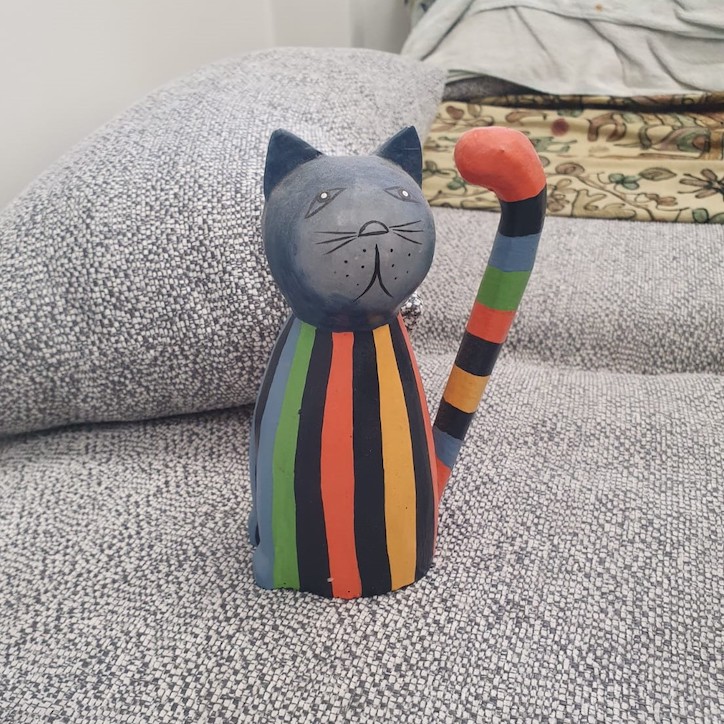}
        \end{tabular} 
        \end{center} &
        \setlength\tabcolsep{0pt}
        \begin{center}
        \begin{tabular}{c c c}
            \includegraphics[height=0.06\textwidth,width=0.06\textwidth,height=0.06333\textwidth]{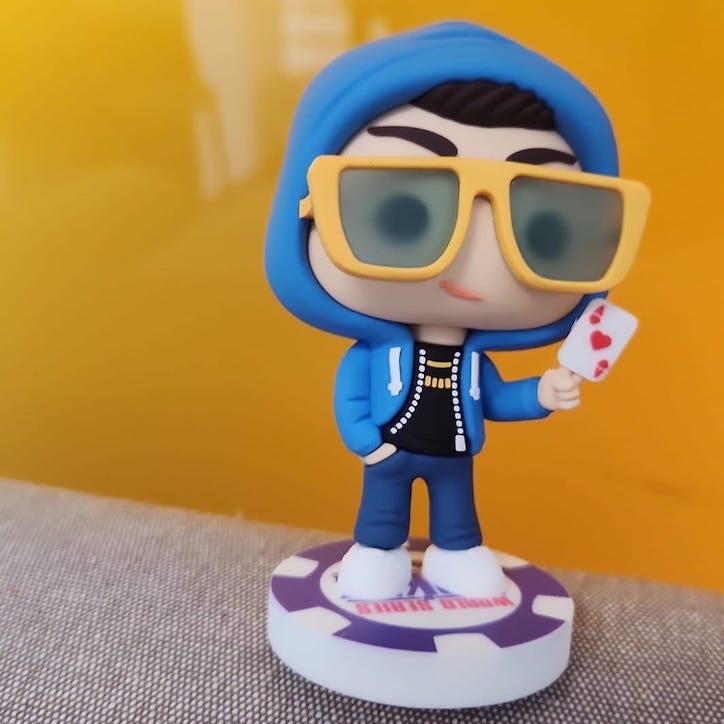} & 
            \includegraphics[height=0.06\textwidth,width=0.06\textwidth,height=0.06333\textwidth]{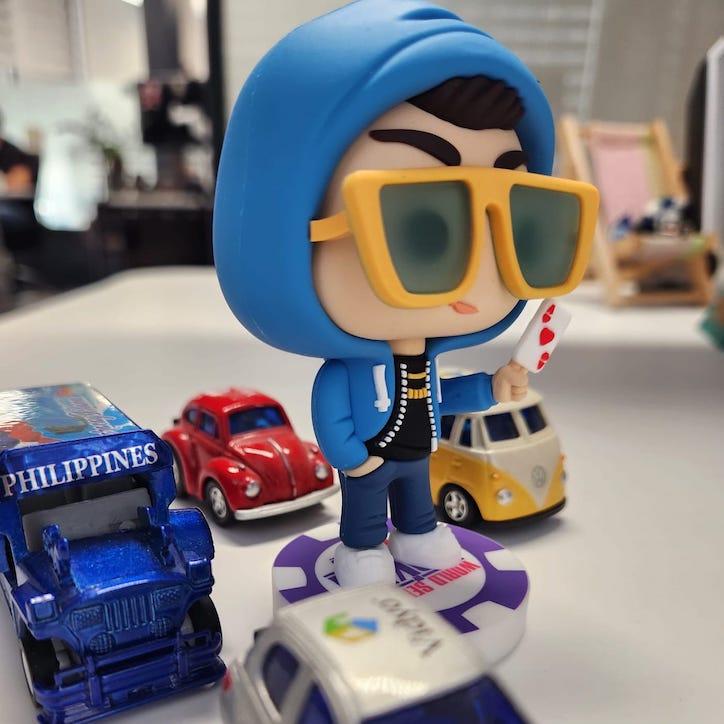} & 
            \includegraphics[height=0.06\textwidth,width=0.06\textwidth,height=0.06333\textwidth]{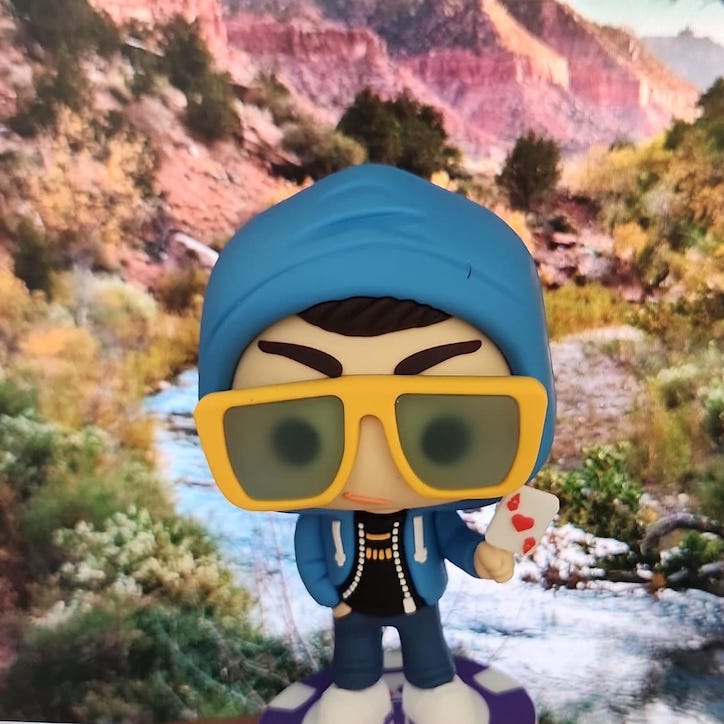} 
        \end{tabular} 
        \end{center} &
        \setlength\tabcolsep{0pt}
        \begin{center}
        \begin{tabular}{c c c}
            \includegraphics[height=0.06\textwidth,width=0.06\textwidth]{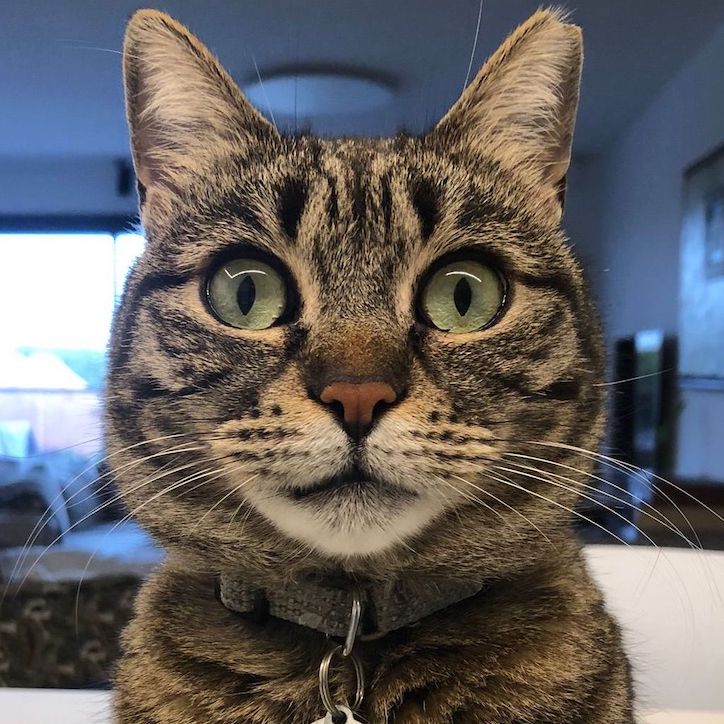} & 
            \includegraphics[height=0.06\textwidth,width=0.06\textwidth]{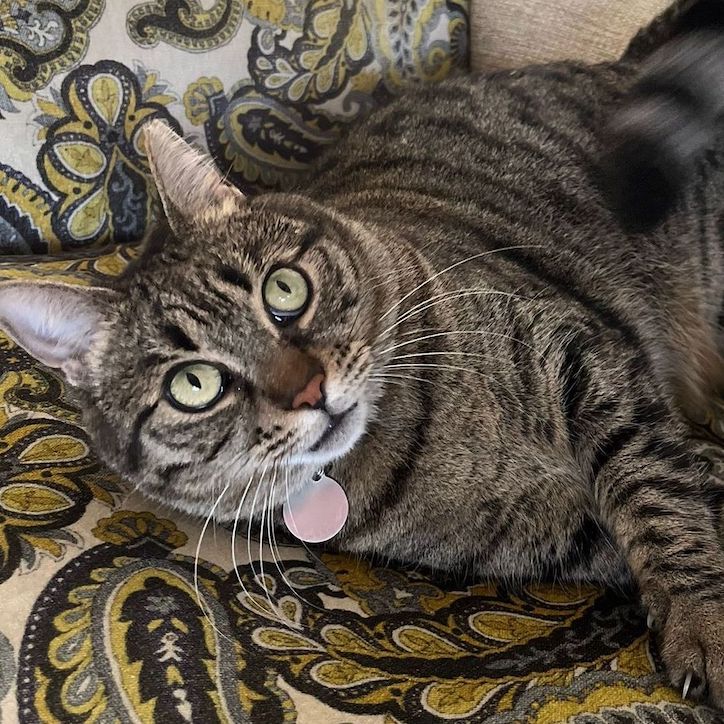} & 
            \includegraphics[height=0.06\textwidth,width=0.06\textwidth]{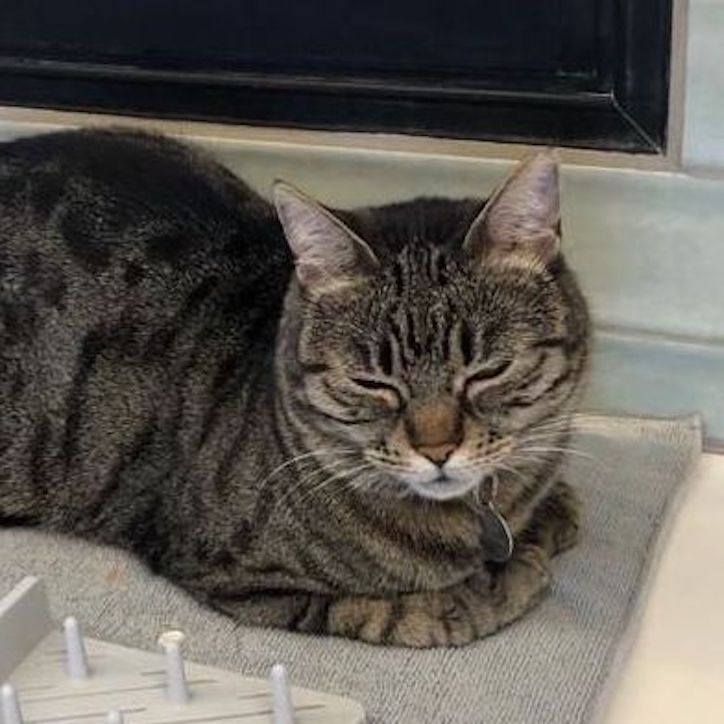}
        \end{tabular} 
        \end{center} &
        \setlength\tabcolsep{0pt}
        \begin{center}
        \begin{tabular}{c c c}
            \includegraphics[height=0.06\textwidth,width=0.06\textwidth]{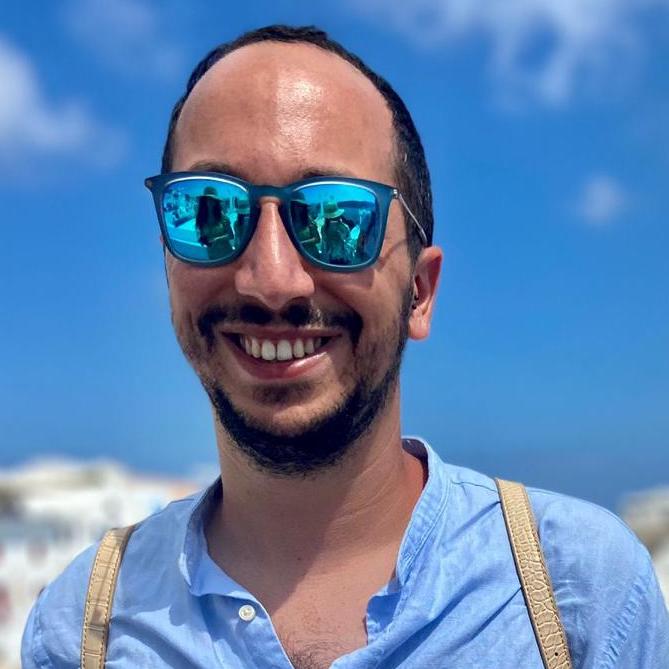} & 
            \includegraphics[height=0.06\textwidth,width=0.06\textwidth]{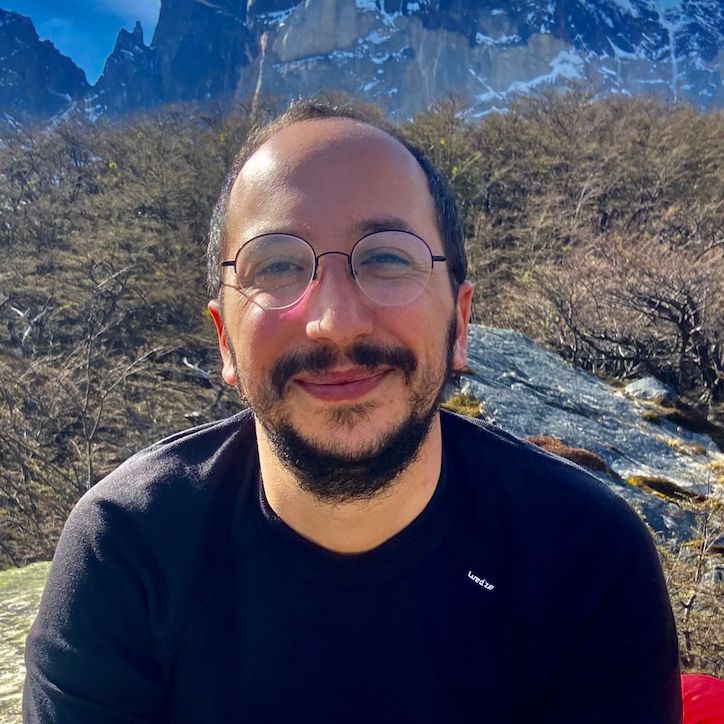} & 
            \includegraphics[height=0.06\textwidth,width=0.06\textwidth]{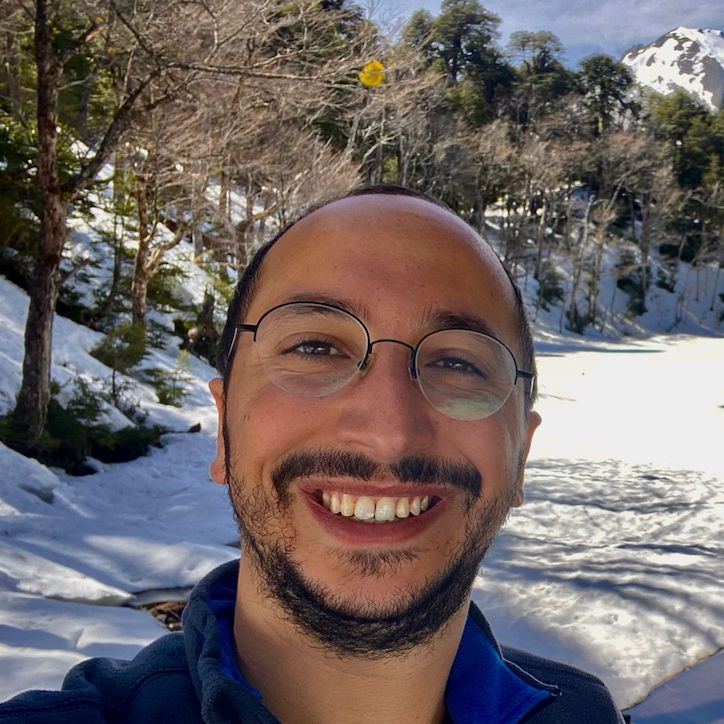}
        \end{tabular} 
        \end{center} &
        \setlength\tabcolsep{0pt}
        \begin{center}
        \begin{tabular}{c c c}
            \includegraphics[height=0.06\textwidth,width=0.06\textwidth]{images/people/assaf/cropped/IMG_0133.jpg} &
            \includegraphics[height=0.06\textwidth,width=0.06\textwidth]{images/people/assaf/cropped/IMG_0936.jpg} &
            \includegraphics[height=0.06\textwidth,width=0.06\textwidth]{images/people/assaf/cropped/IMG_20220130_104352.jpg}
        \end{tabular} 
        \end{center} \\[-0.8cm]
    
        \begin{center} \includegraphics[width=0.18\textwidth]{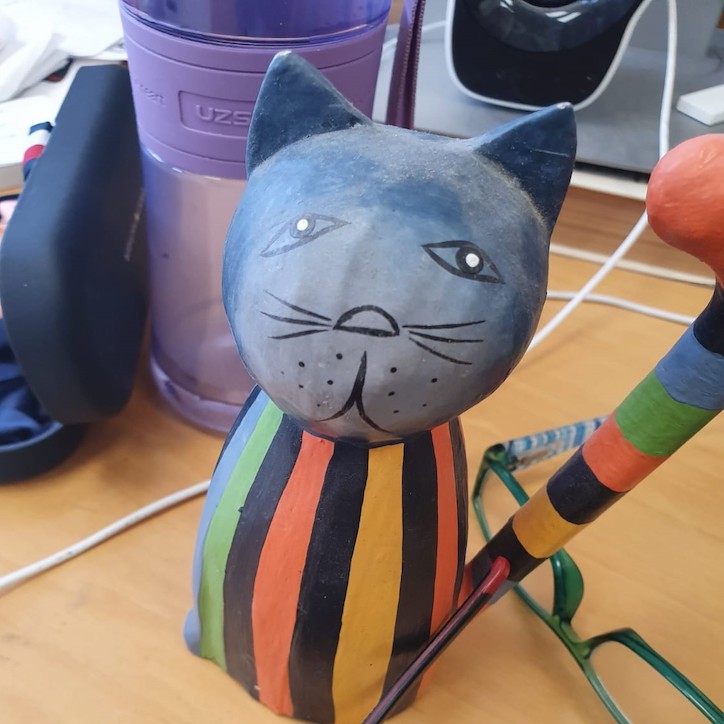} \end{center} &
        \begin{center} \includegraphics[width=0.18\textwidth]{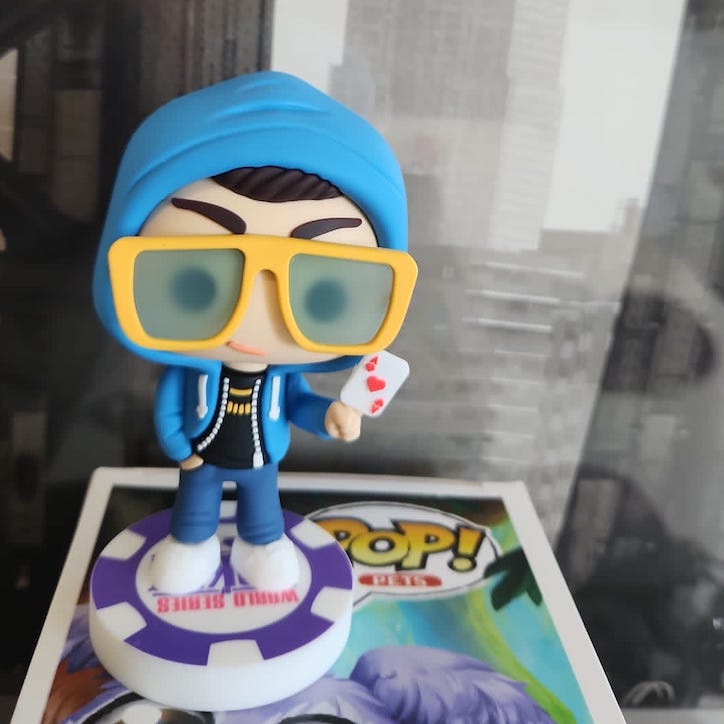} \end{center} &
        \begin{center} \includegraphics[width=0.18\textwidth]{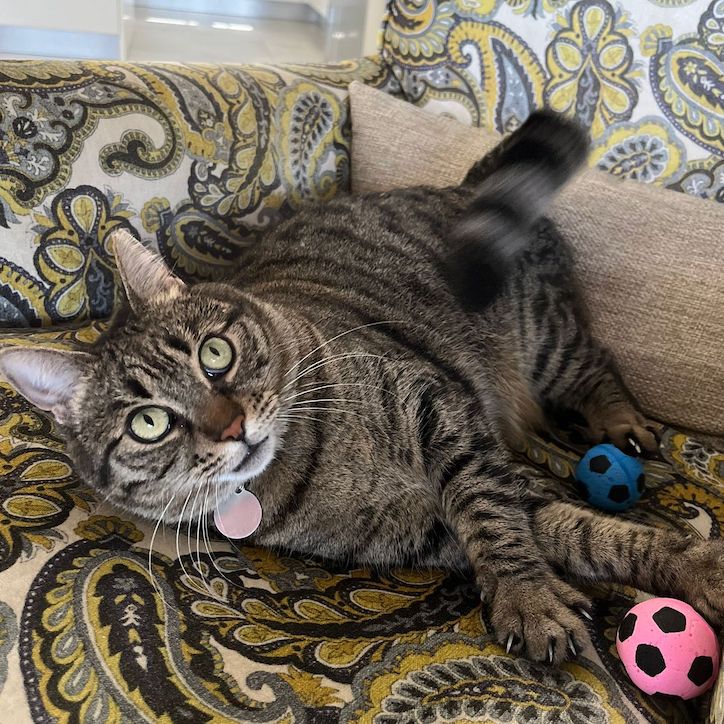} \end{center} &
        \begin{center} \includegraphics[width=0.18\textwidth]{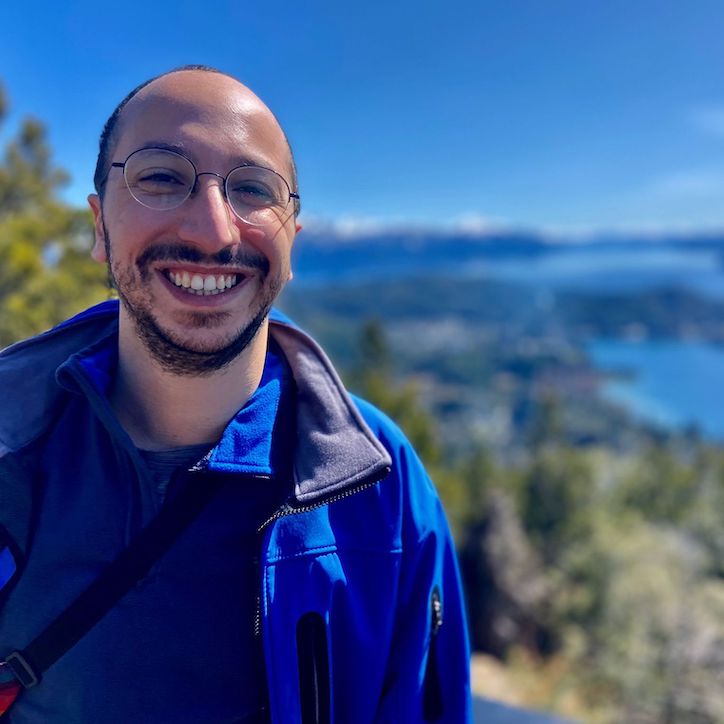} \end{center} &
        \begin{center} \includegraphics[height=0.18\textwidth,width=0.18\textwidth]{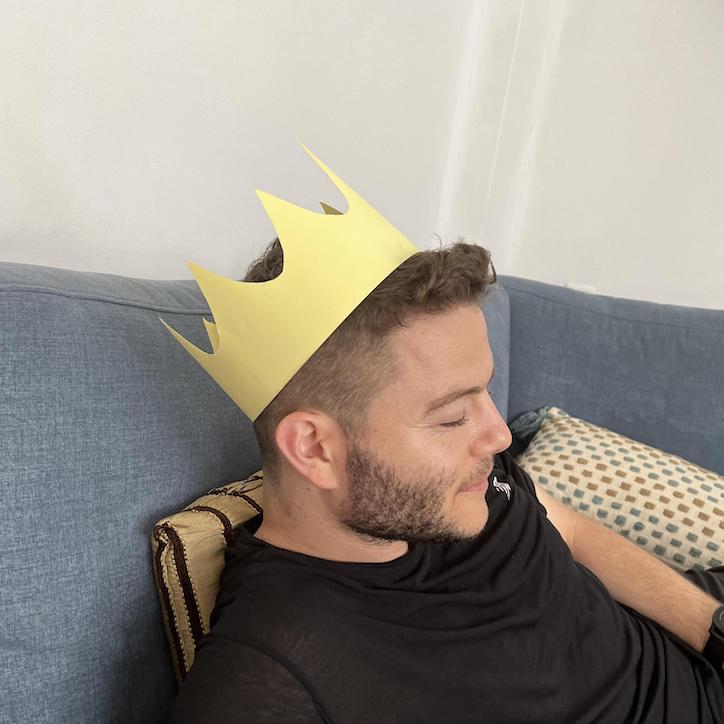} \end{center} \\[-0.8cm]

        \begin{center} \textbf{OpenFlamingo} \end{center} &
        \begin{center} \textbf{OpenFlamingo} \end{center} &
        \begin{center} \textbf{OpenFlamingo} \end{center} &
        \begin{center} \textbf{OpenFlamingo} \end{center} &
        \begin{center} \textbf{OpenFlamingo} \end{center} \\[-0.8cm]

        \begin{center} \footnotesize ``A \Sstar, which is a cat made of wood'' \end{center} &
        \begin{center} \footnotesize ``A \Sstar holding a deck of cards'' \end{center} &
        \begin{center} \footnotesize ``A kitty playing with a ball'' \end{center} &
        \begin{center} \footnotesize ``that \Sstar is wearing a blue jacket, a blue shirt, and a blue hat'' \end{center} &
        \begin{center} \footnotesize ``A crown on his head'' \end{center} \\[-0.65cm]
        
        \begin{center}\textbf{MyLLaVA}\end{center} &
        \begin{center}\textbf{MyLLaVA}\end{center} &
        \begin{center}\textbf{MyLLaVA}\end{center} &
        \begin{center}\textbf{MyLLaVA}\end{center} &
        \begin{center}\textbf{MyLLaVA}\end{center} \\[-0.8cm]

        \begin{center} \footnotesize ``\Sstar sitting on a cluttered desk, surrounded by various items including a purple water bottle, a pair of glasses...'' \end{center} &
        \begin{center} \footnotesize ``\Sstar standing in front of a skyscraper in the city.'' \end{center} &
        \begin{center} \footnotesize ``\Sstar laying on a couch and playing with two balls, one pink and one blue. The kitten is wearing a collar around its neck'' \end{center} &
        \begin{center} \footnotesize ``\Sstar with a smile on a mountain top, before a beautiful lake, with a blue sky in the background'' \end{center} &
        \begin{center} \footnotesize ``In his bedroom, \Sstar is wearing a yellow paper crown on his head. He is sitting on a blue couch, looking relaxed...'' \end{center} 
        
    \end{tabular}

    \vspace{-0.4cm}
    \caption{Comparison to OpenFlamingo for personalized captioning. We show results of MyVLM over BLIP-2 (top) and LLaVA (bottom).}
    \label{fig:flamingo_comparison}
    \vspace{-0.35cm}
\end{figure*}

%% file: tables/flamingo_comparison.tex
\begin{table*}[t]
\small
\centering
\caption{\textbf{Quantitative Comparison: OpenFlamingo~\cite{awadalla2023openflamingo,alayrac2022flamingo}}. We compute the average recall, text-to-image similarity, and text-to-text similarity obtained over all $16$ individuals and $29$ objects. Results are averaged across all five validation sets.}
\vspace{-0.3cm}
\centering
\begin{tabular}{l l c c c}
    \toprule
    Data & Model & Recall $\uparrow$ & Text Similarity $\uparrow$ & Image Similarity $\uparrow$ \\
    \midrule
    \multirow{3}{*}{People} & OpenFlamingo     & $74.81$             & $\underline{43.72}$ & $\textbf{24.33}$    \\
                            & MyVLM + BLIP-2   & $\underline{79.76}$ & $\textbf{48.99}$    & $22.99$             \\
                            & MyVLM + LLaVA    & $\textbf{97.08}$    & $43.58$             & $\underline{23.06}$ \\
    \midrule
    \midrule
    \multirow{3}{*}{Objects} & OpenFlamingo     & $49.77$             & $34.12$             & $\underline{27.65}$ \\
                             & MyVLM + BLIP-2   & $\textbf{95.10}$    & $\textbf{77.71}$    & $\textbf{28.12}$    \\
                             & MyVLM + LLaVA    & $\underline{94.76}$ & $\underline{71.49}$ & $27.60$             \\
    \bottomrule
\end{tabular}
\label{tb:flamingo_comp}
\vspace{-0.15cm}
\end{table*}

%% file: tables/ablation_augs_reg.tex
\begin{table}[t]
\small
\centering
\caption{\textbf{Ablation Study: Regularization \& Augmentations}. We compute the average recall, text-to-image similarity, and text-to-text similarity obtained over $5$ objects and $5$ individuals with and without our augmentations and regularization techniques. Results are obtained over BLIP-2 and averaged across all validation sets.}
\vspace{-0.3cm}
\centering
\begin{tabular}{l c c c}
    \toprule
     & Recall $\uparrow$ & Text Sim. $\uparrow$ & Image Sim. $\uparrow$ \\
    \midrule
    w/o Aug. \& Reg. & $25.88$             & $\underline{56.32}$  & $\textbf{24.76}$    \\
    w/o Aug.         & $\underline{72.77}$ & $55.03$              & $24.00$             \\
    MyVLM            & $\textbf{84.87}$    & $\textbf{58.68}$     & $\underline{24.65}$ \\
    \bottomrule
\end{tabular}
\label{tb:ablation_augs_regs}
\vspace{-0.1cm}
\end{table}

%% file: figures_supplementary/feature_spaces.tex
\begin{figure*}[t]
    \centering
    \setlength{\tabcolsep}{0.3pt}
    \renewcommand{\arraystretch}{0.3}
    \addtolength{\belowcaptionskip}{-10pt}
    \includegraphics[width=0.8\textwidth]{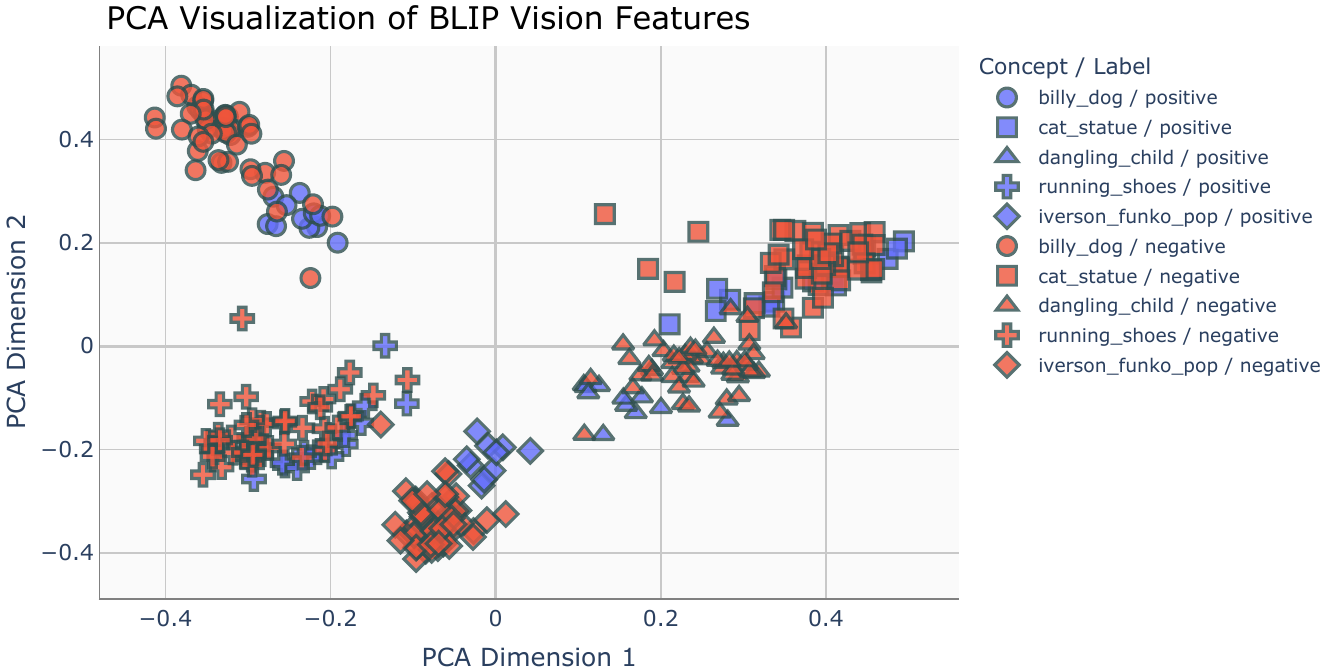}
    \caption{\textbf{PCA Visualization of the output space of the BLIP-2 vision encoder.} We project the [CLS] token embeddings extracted from all positive and $200$ negative images of five different objects, each shown using a different shape. 
    As shown, these embeddings are not well-separated enough to effectively distinguish between positive and negative samples of the target object.}
    \label{fig:feature_spaces_blip}
\end{figure*}

\begin{figure*}[t]
    \centering
    \setlength{\tabcolsep}{0.3pt}
    \renewcommand{\arraystretch}{0.7}
    \addtolength{\belowcaptionskip}{-5pt}
    {\small
    \hspace{0.05cm}
        
    \begin{tabular}{c c c c c c}

        \includegraphics[width=0.12\textwidth]{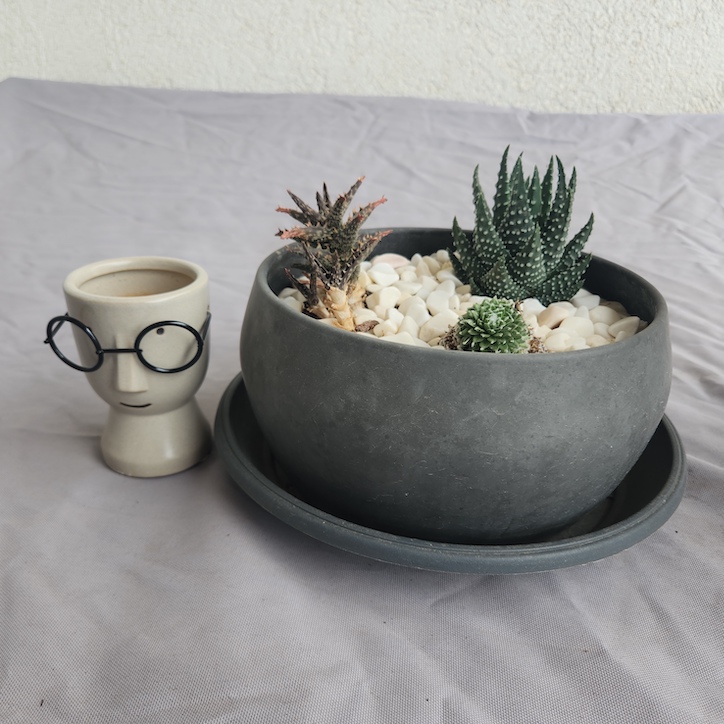} &
        \includegraphics[width=0.12\textwidth]{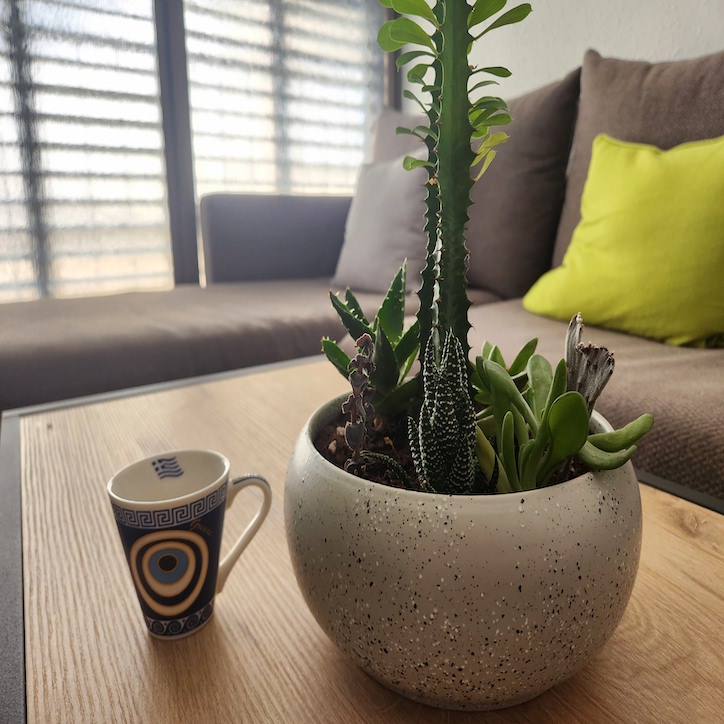} &
        \includegraphics[width=0.12\textwidth]{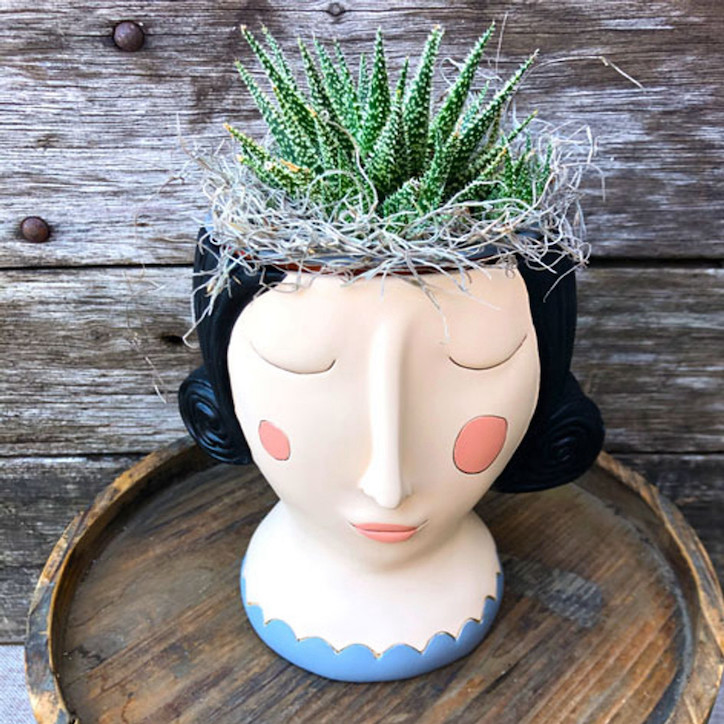} &
        \includegraphics[width=0.12\textwidth]{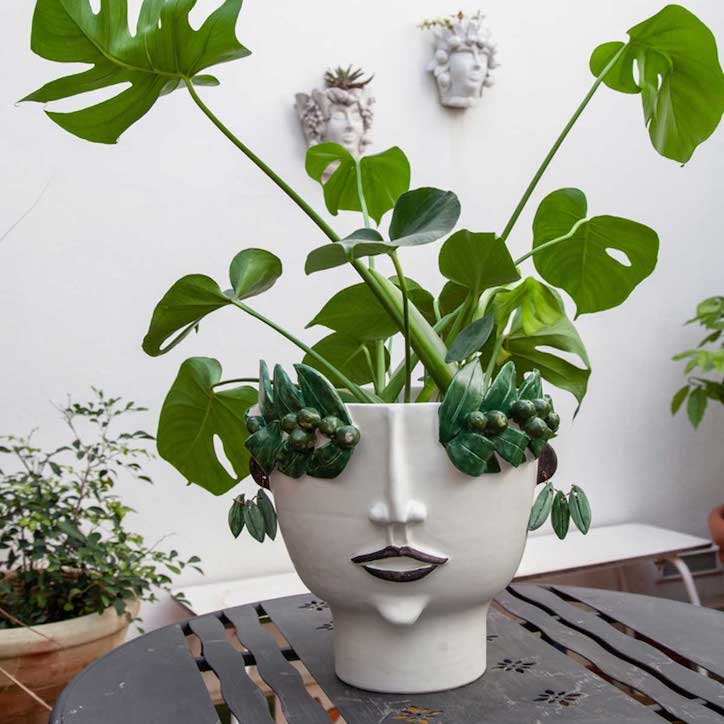} &
        \includegraphics[width=0.12\textwidth]{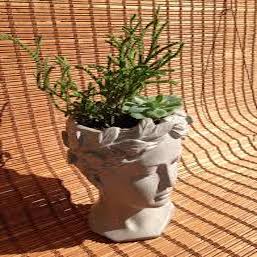} &
        \includegraphics[width=0.12\textwidth]{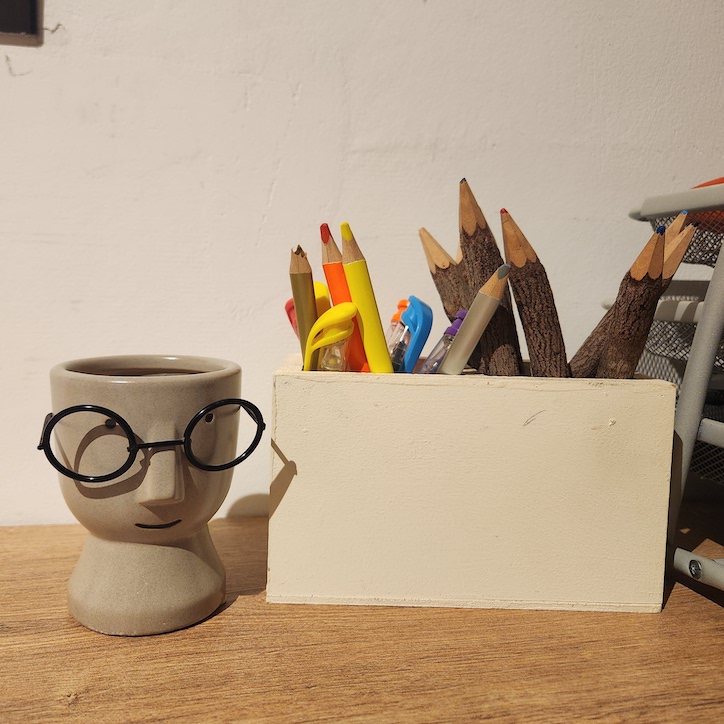} \\

        & \textcolor{red}{0.714} & \textcolor{red}{0.689} & \textcolor{red}{0.686} & \textcolor{red}{0.680} & \textcolor{darkgreen}{0.673} \\ \\

        \raisebox{0.75cm}{\begin{tabular}{c} Ceramic \\ Head \end{tabular}} & 
        \includegraphics[width=0.12\textwidth]{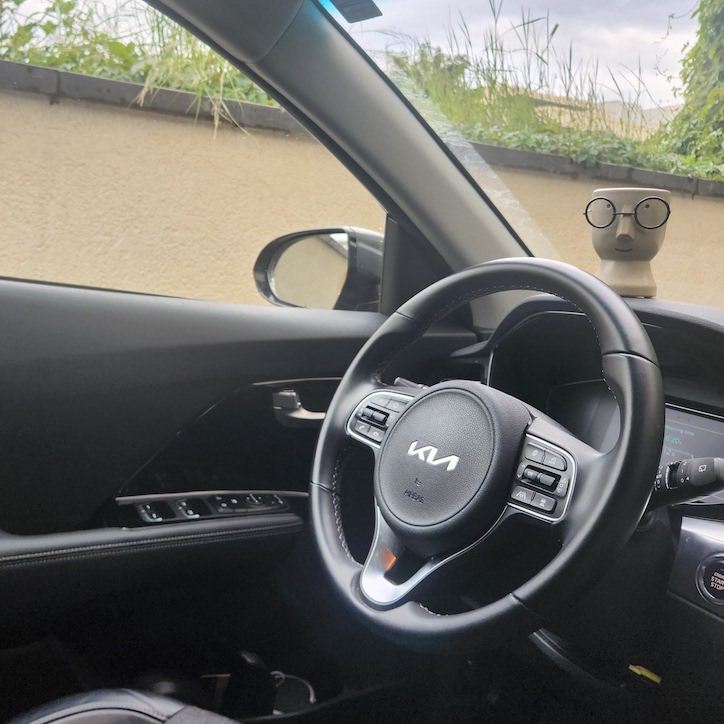} &
        \includegraphics[width=0.12\textwidth]{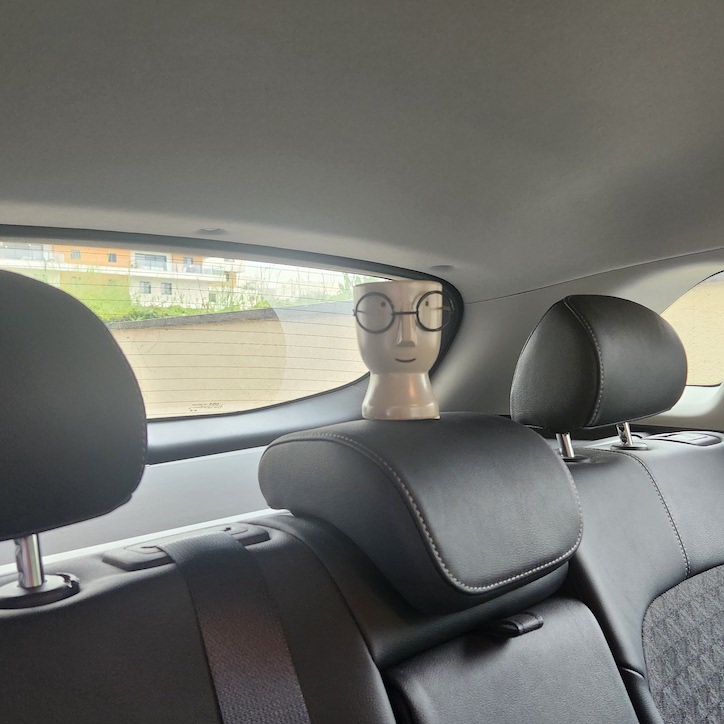} &
        \includegraphics[width=0.12\textwidth]{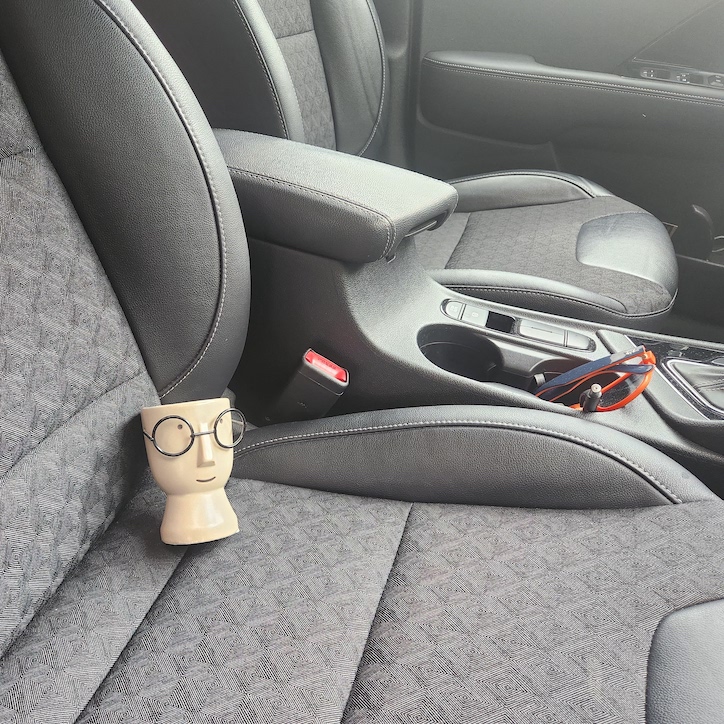} &
        \includegraphics[width=0.12\textwidth]{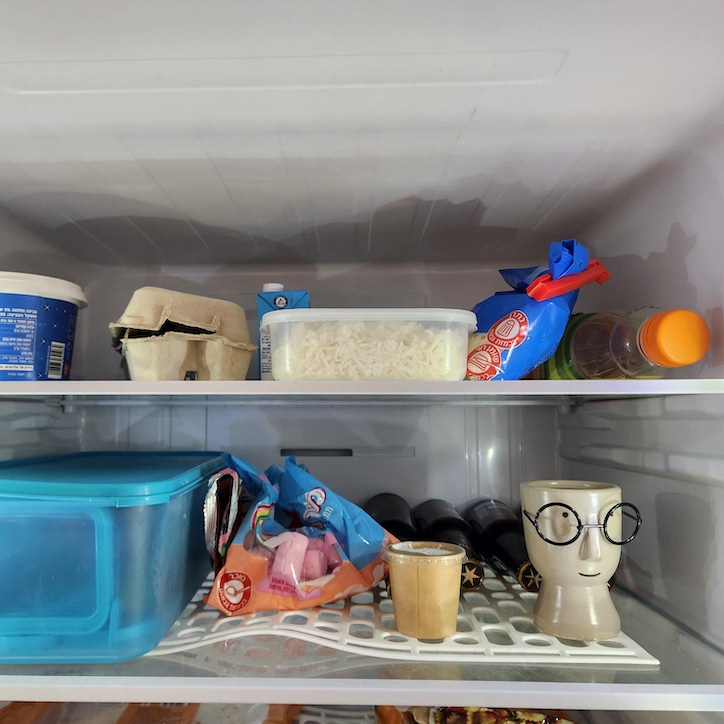} &
        \includegraphics[width=0.12\textwidth]{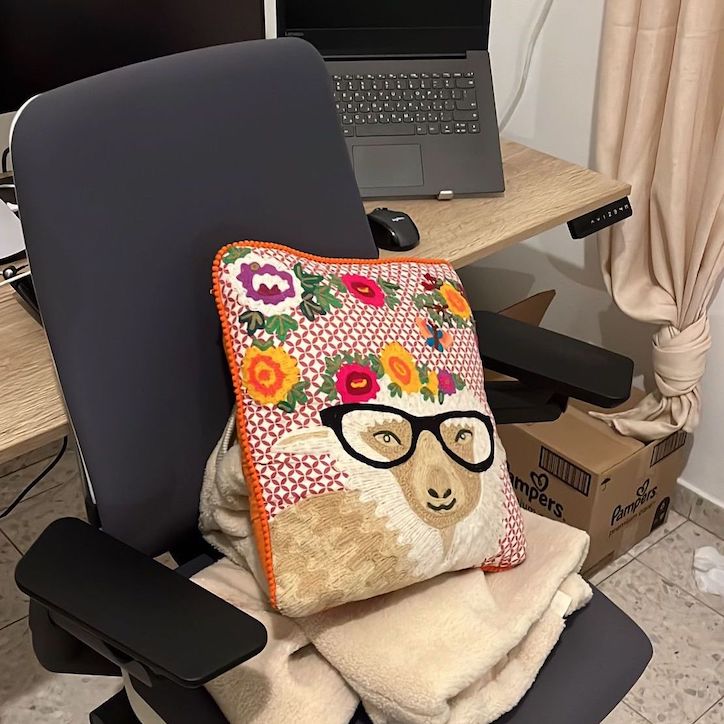} \\
        
        & \textcolor{darkgreen}{$0.958$} & \textcolor{darkgreen}{$0.946$} & \textcolor{darkgreen}{$0.944$} & \textcolor{darkgreen}{$0.891$} & \textcolor{red}{$0.889$} \\ \\
 
        \includegraphics[width=0.12\textwidth]{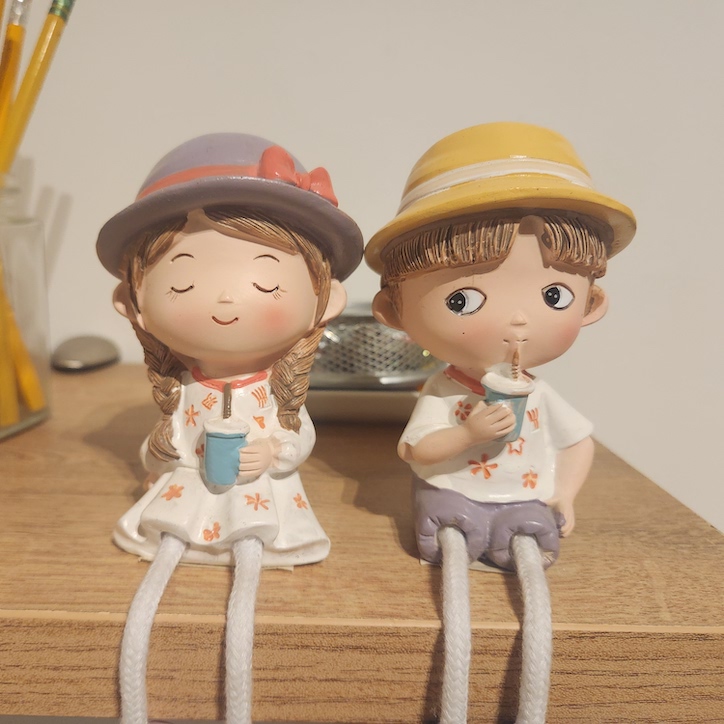} &
        \includegraphics[width=0.12\textwidth]{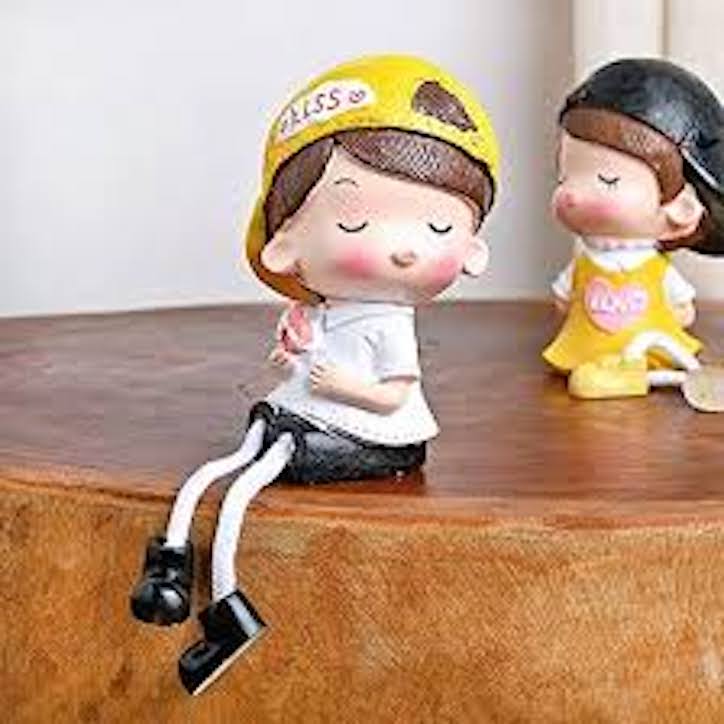} &
        \includegraphics[width=0.12\textwidth]{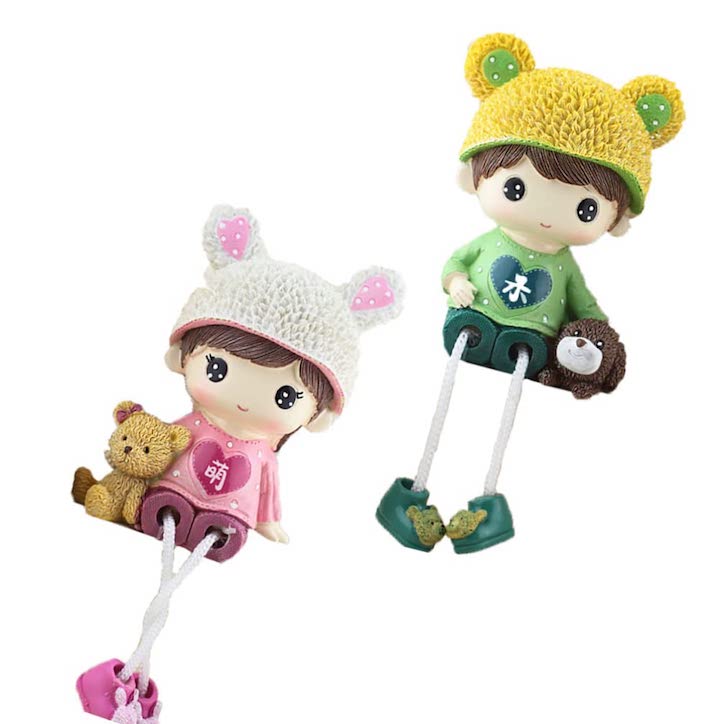} &
        \includegraphics[width=0.12\textwidth]{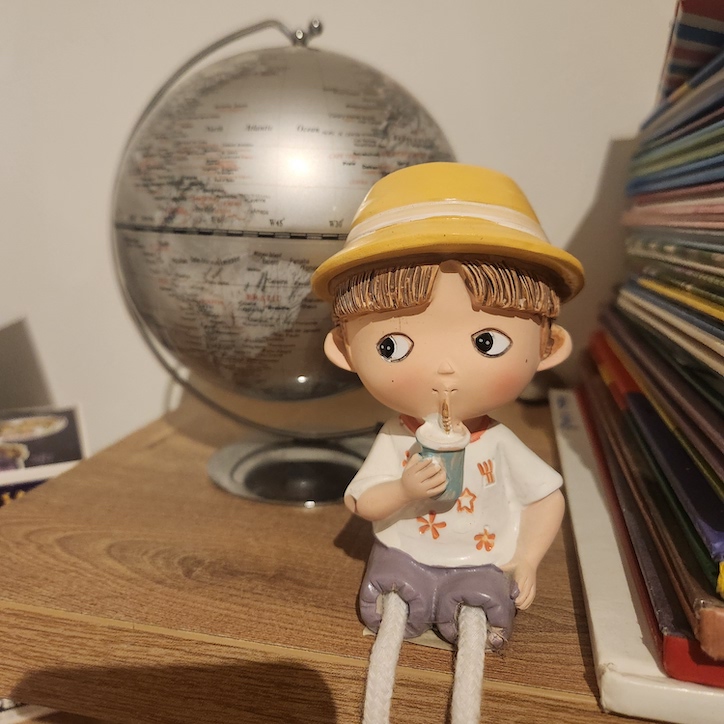} &
        \includegraphics[width=0.12\textwidth]{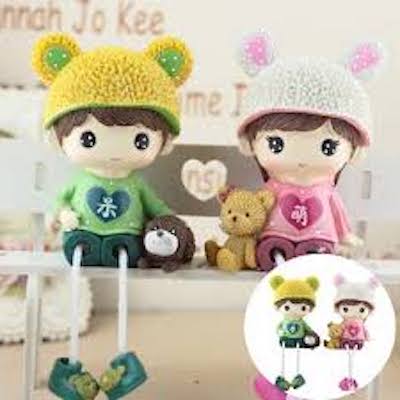} &
        \includegraphics[width=0.12\textwidth]{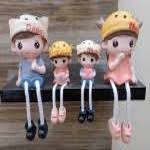} \\

        & \textcolor{red}{0.799} & \textcolor{red}{0.786} & \textcolor{darkgreen}{0.771} & \textcolor{red}{0.770} & \textcolor{red}{0.741} \\ \\

        \raisebox{0.75cm}{\begin{tabular}{c} Dangling \\ Child \end{tabular}} & 
        \includegraphics[width=0.12\textwidth]{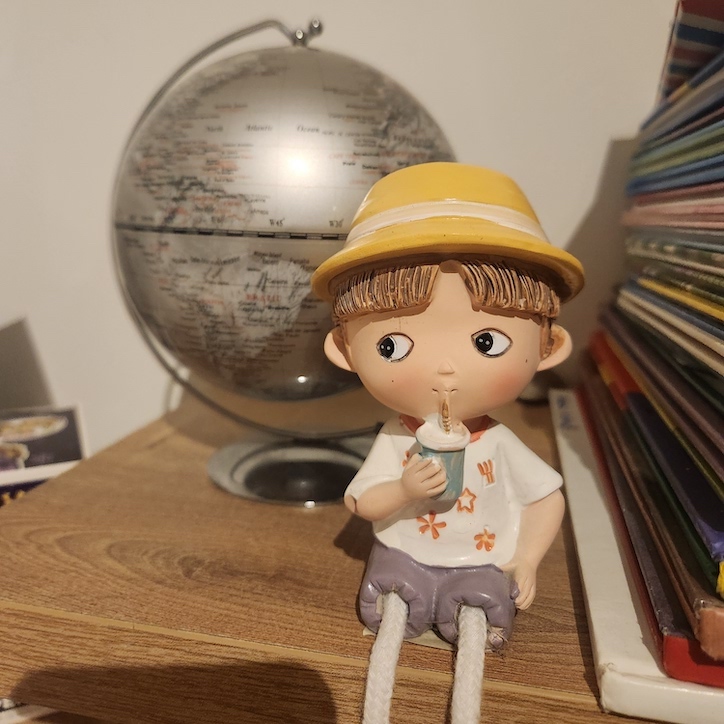} &
        \includegraphics[width=0.12\textwidth]{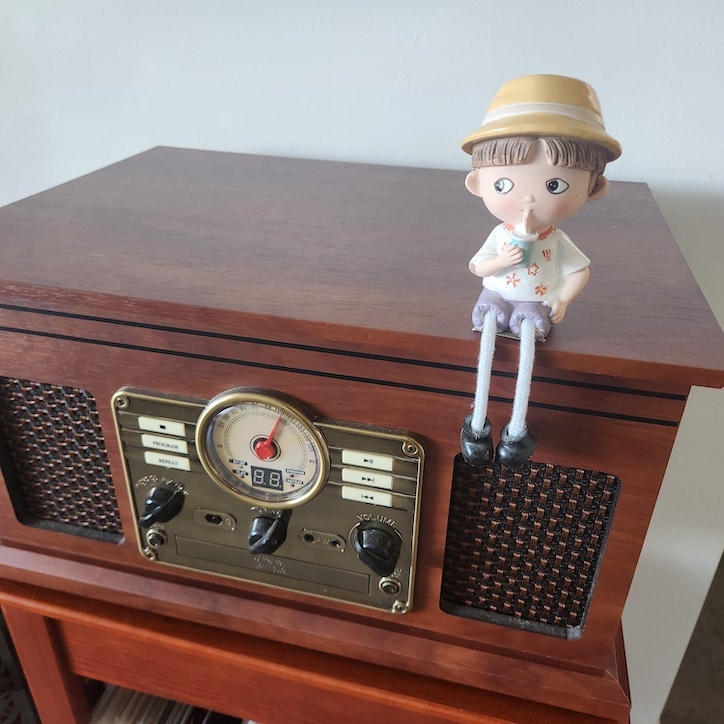} &
        \includegraphics[width=0.12\textwidth]{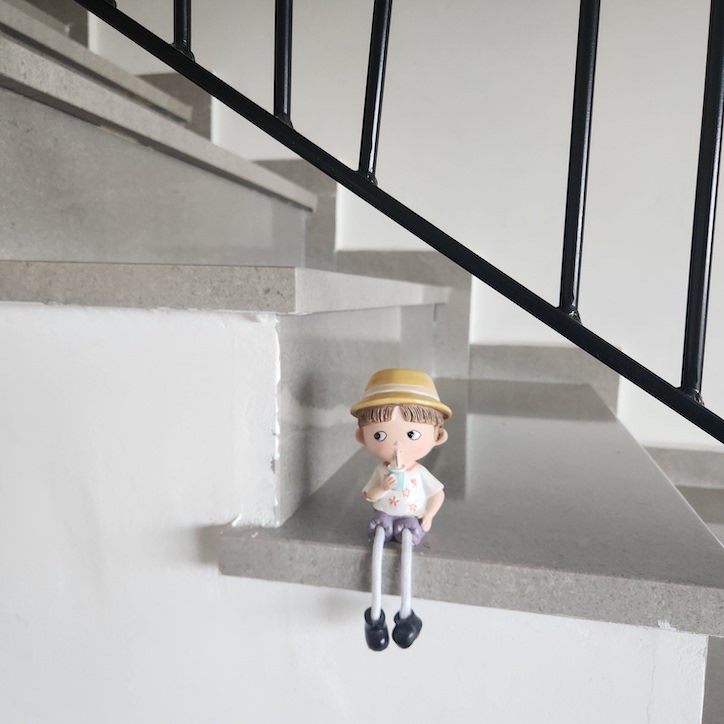} &
        \includegraphics[width=0.12\textwidth]{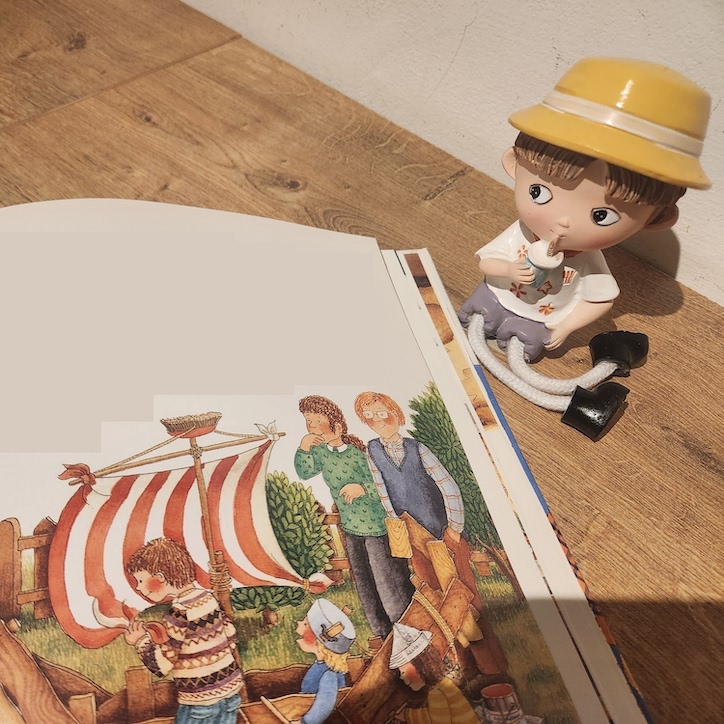} &
        \includegraphics[width=0.12\textwidth]{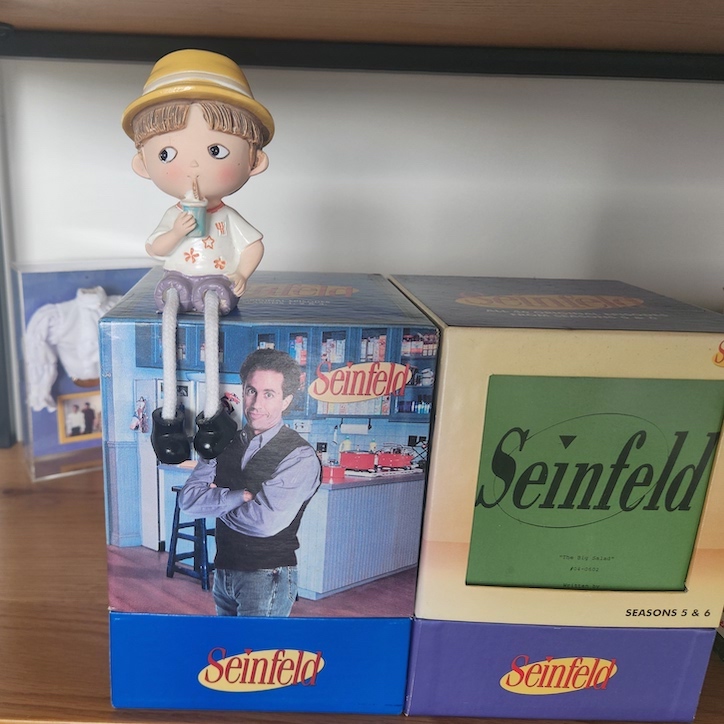} \\
        
        & \textcolor{darkgreen}{$0.957$} & \textcolor{darkgreen}{$0.923$} & \textcolor{darkgreen}{$0.906$} & \textcolor{darkgreen}{$0.876$} & \textcolor{darkgreen}{$0.871$} \\ \\

        \includegraphics[width=0.12\textwidth]{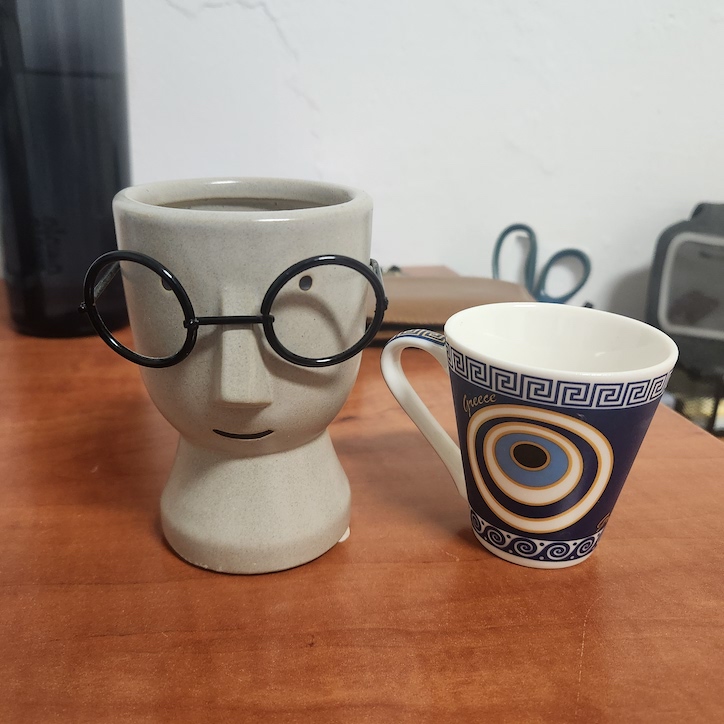} &
        \includegraphics[width=0.12\textwidth]{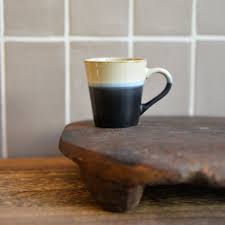} &
        \includegraphics[width=0.12\textwidth]{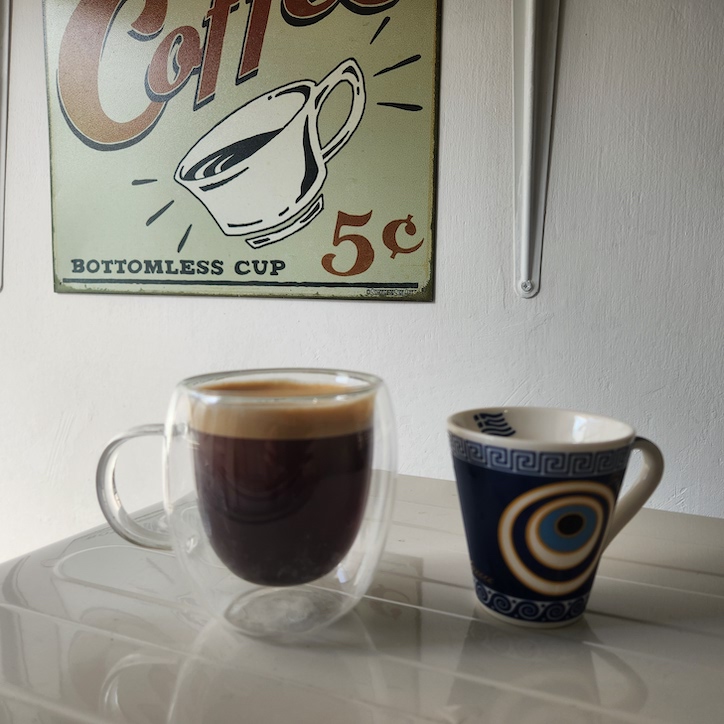} &
        \includegraphics[width=0.12\textwidth]{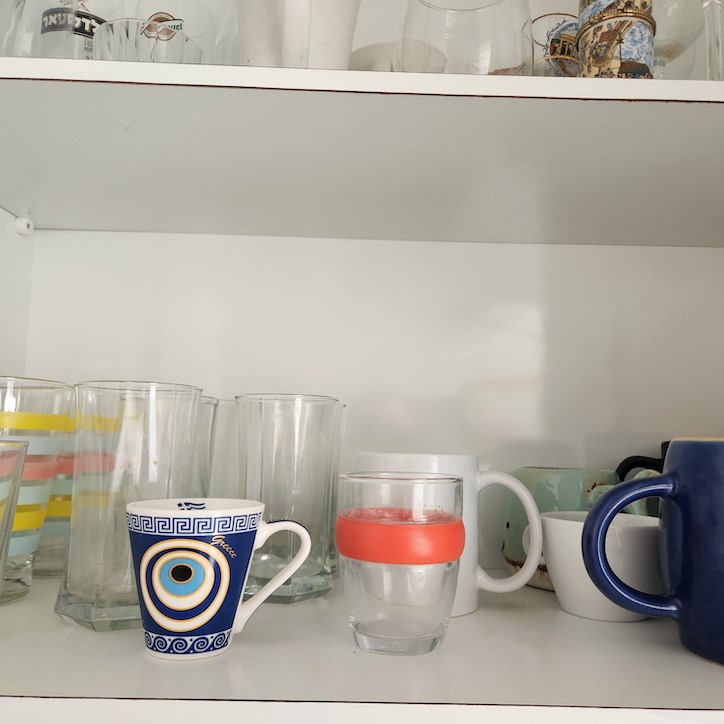} &
        \includegraphics[width=0.12\textwidth]{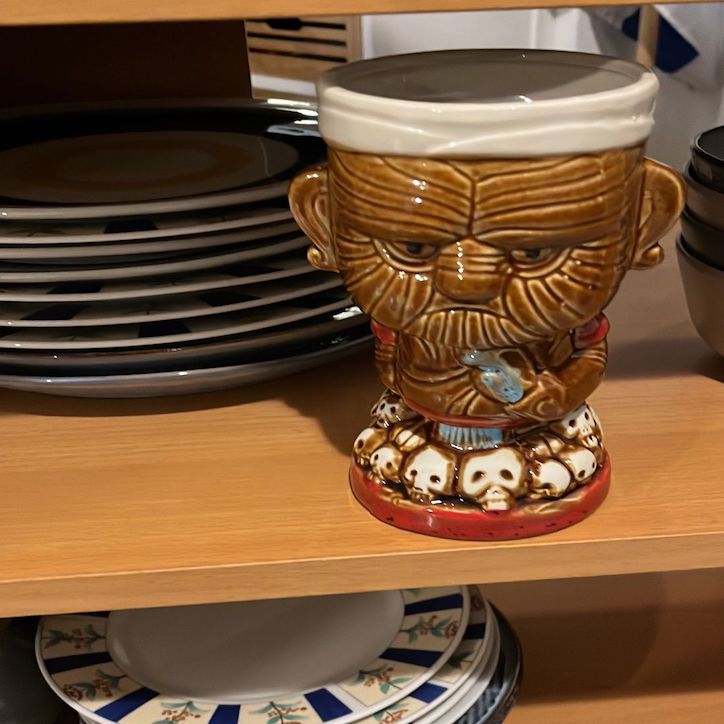} &
        \includegraphics[width=0.12\textwidth]{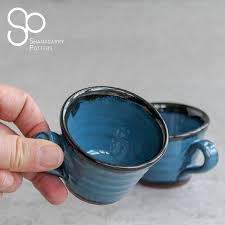} \\

        & \textcolor{red}{0.652} & \textcolor{darkgreen}{0.592} & \textcolor{darkgreen}{0.585} & \textcolor{red}{0.585} & \textcolor{red}{0.582} \\ \\

        \raisebox{0.75cm}{\begin{tabular}{c} Espresso \\ Cup \end{tabular}} & 
        \includegraphics[width=0.12\textwidth]{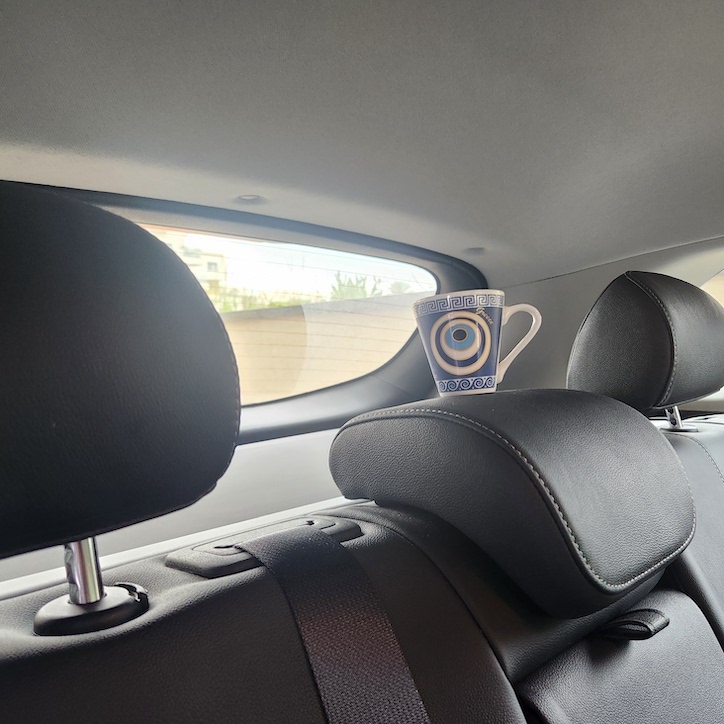} &
        \includegraphics[width=0.12\textwidth]{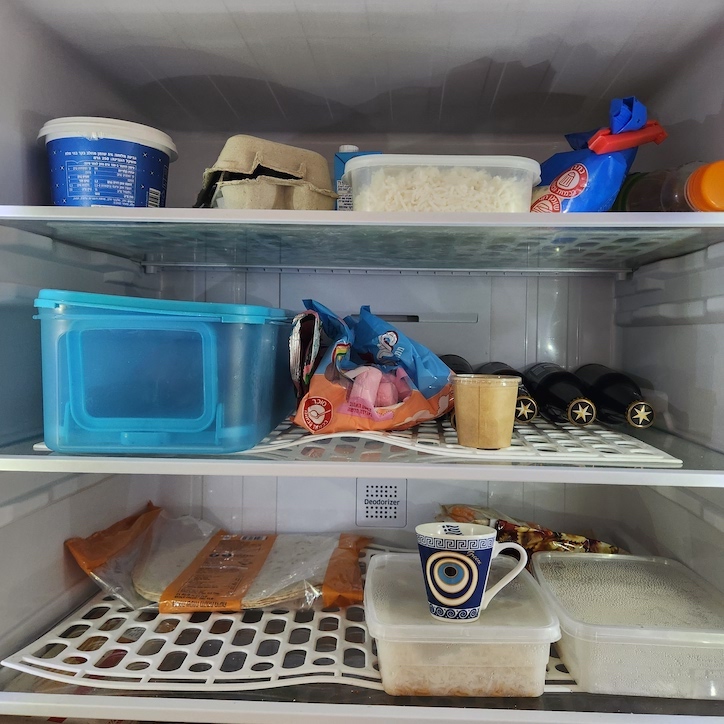} &
        \includegraphics[width=0.12\textwidth]{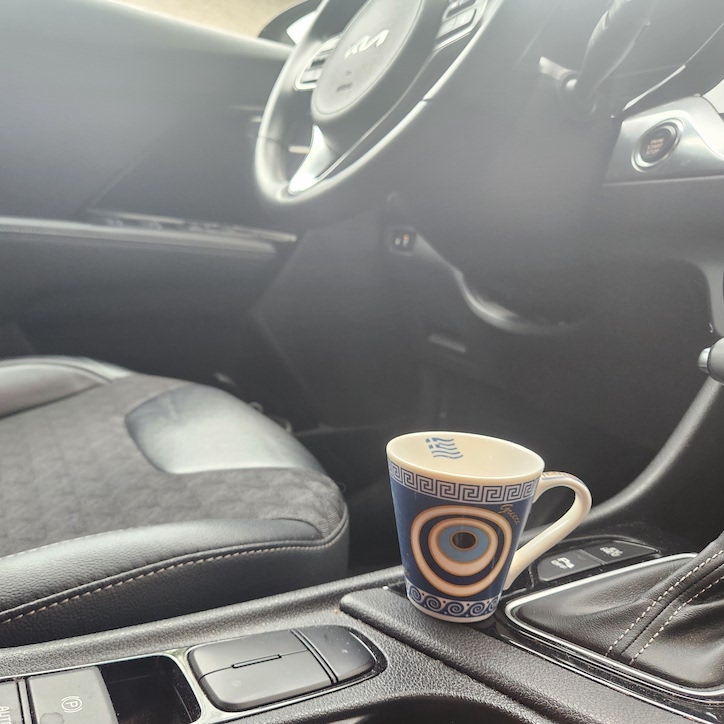} &
        \includegraphics[width=0.12\textwidth]{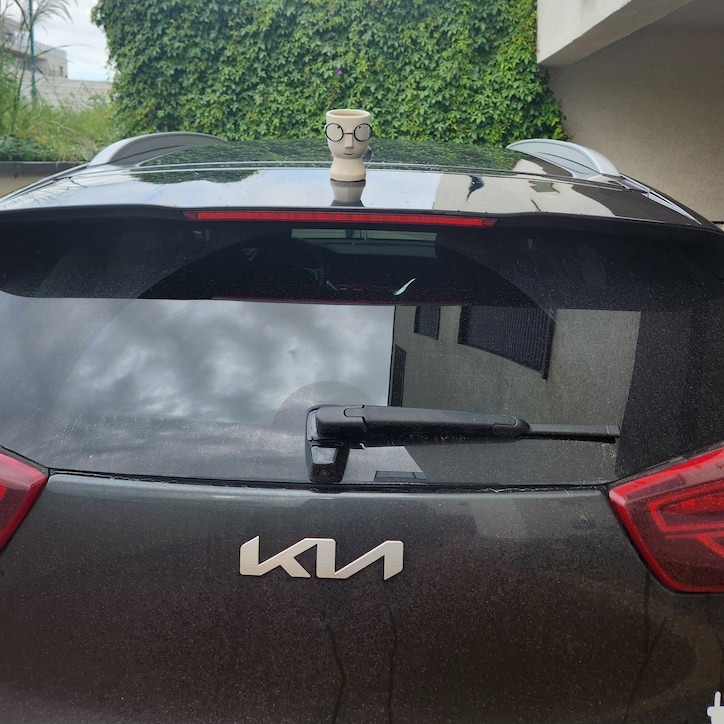} &
        \includegraphics[width=0.12\textwidth]{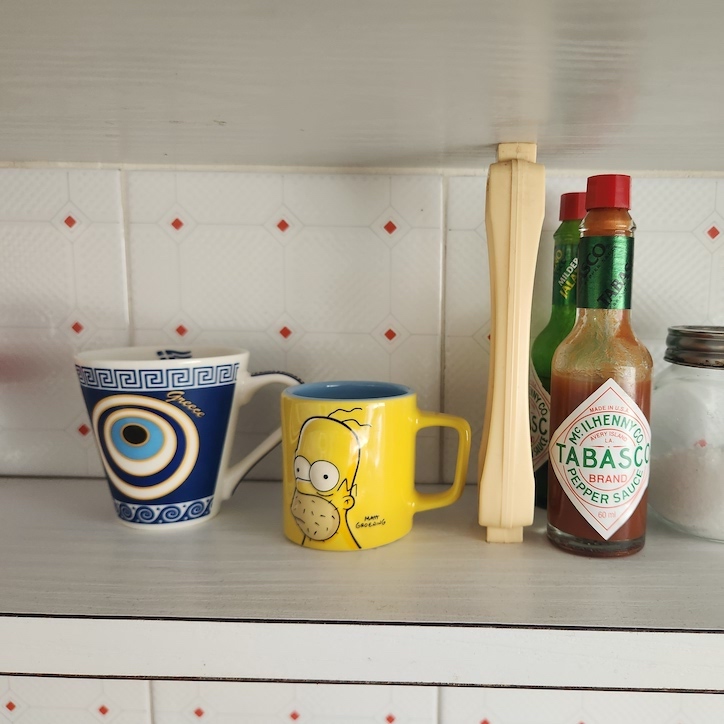} \\
        
        & \textcolor{darkgreen}{$0.896$} & \textcolor{darkgreen}{$0.890$} & \textcolor{darkgreen}{$0.865$} & \textcolor{darkgreen}{$0.843$} & \textcolor{darkgreen}{$0.834$}

    \end{tabular}
    }
    
\vspace{-0.1cm}
\caption{\textbf{Ablation Study: The CLIP Space.} 
For each concept, we visualize the $5$ nearest neighbors of the query image shown to the left within the CLIP embedding space. The nearest neighbors often include both negative samples of the target object and positive samples of other objects, making it challenging to directly operate within the space. In the second row of each concept, we visualize the five images that received the highest scores from our corresponding concept head. As shown, our linear classifier is effective in distinguishing the target concept from negative samples.}
\label{fig:feature_space_clip}
\vspace{-0.3cm}
\end{figure*}

%% file: tables/image_captioning_metrics.tex
\begin{table*}[t]
\small
\centering
\setlength{\tabcolsep}{9pt}
\caption{\textbf{Quantitative Metrics: Standard Image Captioning Metrics}. We compute standard image captioning metrics over personalized captions generated by MyVLM, trained with $4$ images. For each image, we use all $5$ augmented captions as the set of ground truth captions. Results are obtained over all $5$ validation folds and averaged over all concepts.}
\vspace{-0.25cm}
\centering
\begin{tabular}{l l c c c c c c c c}
    \toprule
    Dataset                  & Method & B1  & B2  &  B3 &  B4 &  CIDEr & METEOR & ROUGE\_L & SPICE \\
    \midrule                            
     \multirow{2}{*}{People}  & BLIP-2 & $0.69$ & $0.63$ & $0.58$ & $0.53$ & $2.21$ & $0.31$	& $0.63$ & $0.27$ \\
                              & MyVLM  & $0.53$ & $0.40$ & $0.30$ & $0.23$ & $1.06$ & $0.21$ & $0.44$ & $0.15$ \\
     \midrule                       
     \multirow{2}{*}{Objects} & BLIP-2 & $0.63$ & $0.51$ & $0.43$ & $0.36$ & $1.53$ & $0.26$ & $0.55$ & $0.23$ \\
                              & MyVLM  & $0.64$ & $0.50$ & $0.38$ & $0.28$ & $1.44$ & $0.28$ & $0.56$ & $0.26$ \\
     \midrule
     \multirow{2}{*}{All}     & BLIP-2 & $0.66$ & $0.57$ & $0.51$ & $0.45$ & $1.89$ & $0.28$ & $0.59$ & $0.25$ \\
                              & MyVLM  & $0.59$ & $0.45$ & $0.34$ & $0.26$ & $1.28$ & $0.25$ & $0.50$ & $0.20$ \\
    \bottomrule
\end{tabular}
\vspace{-0.2cm}
\begin{center} \textbf{BLIP-2} \end{center}

\begin{tabular}{l l c c c c c c c c}
    \toprule
    Dataset                  & Method & B1  & B2  &  B3 &  B4 &  CIDEr & METEOR & ROUGE\_L & SPICE \\
    \midrule
     \multirow{2}{*}{People}  & LLaVA & $0.27$ & $0.14$	& $0.08$ & $0.04$ & $0.18$ & $0.11$ & $0.24$ & $0.06$ \\
                              & MyVLM  & $0.28$ & $0.19$	& $0.13$ & $0.09$ & $0.39$ & $0.18$ & $0.34$ & $0.11$ \\
     \midrule
     \multirow{2}{*}{Objects} & LLaVA & $0.26$ & $0.15$	& $0.09$ & $0.05$ & $0.15$ & $0.16$ & $0.27$ & $0.11$ \\
                              & MyVLM  & $0.36$ & $0.26$	& $0.19$ & $0.13$ & $0.73$ & $0.26$ & $0.44$ & $0.21$ \\
     \midrule
     \multirow{2}{*}{All}     & LLaVA & $0.26$ & $0.15$	& $0.08$ & $0.05$ & $0.17$ & $0.13$ & $0.26$ & $0.09$ \\
                              & MyVLM  & $0.32$ & $0.22$	& $0.15$ & $0.11$ & $0.58$ & $0.22$ & $0.39$ & $0.16$ \\
    \bottomrule
\end{tabular}
\vspace{-0.2cm}
\begin{center} \textbf{LLaVA} \end{center}

\label{tb:captioning_metrics}
\end{table*}

%% file: tables/concept_heads.tex
\begin{table*}[h!]
\small
\centering
\caption{\textbf{Concept Head Evaluations.} Left: we measure the recall and classification rate over $16$ individuals using our face recognition network used for defining our concept head. Right: we compute the average recall and precision of our linear classifiers over our $29$ user-specific objects.}
\vspace{-0.4cm}
\begin{minipage}{.48\linewidth}
\centering
\begin{tabular}{c c c}
    \toprule
    Recall & False Positive Rate & Missed Rate \\
    \midrule
    $96.39\%$ & $2.33\%$ & $1.28\%$ \\
    \bottomrule \\[-0.5cm]
\end{tabular}
\begin{center} \textbf{People} \end{center}
\end{minipage}%
\hspace{0.2cm}
\begin{minipage}{.48\linewidth}
\centering
\setlength{\tabcolsep}{4pt}
\begin{tabular}{l c c c c}
    \toprule
    &            Correctly Classified & Total Samples & Percent Correct  \\
    \midrule
    \textbf{Positives}  & $226$    & $234$     & $96.58\%$ \\
    \textbf{Negatives}  & $95.724$ & $105,328$ & $90.88\%$ \\
    \bottomrule \\[-0.5cm]
\end{tabular}
\begin{center} \textbf{Objects} \end{center}
\end{minipage}
\label{tb:concept_heads_evaluation}
\end{table*}

%% file: figures_supplementary/our_results_blip.tex
\begin{figure*}[t]
    \centering
    \addtolength{\belowcaptionskip}{-12.5pt}
    \renewcommand{\arraystretch}{1}
    \small
    \begin{tabular}{p{0.175\textwidth} p{0.175\textwidth} p{0.175\textwidth} p{0.175\textwidth} p{0.175\textwidth}}

        \setlength\tabcolsep{0pt}
        \begin{tabular}{c c c}
            \includegraphics[width=0.06333\textwidth]{images/people/shay/cropped/image_2.jpg} & 
            \includegraphics[width=0.06333\textwidth]{images/people/shay/cropped/IMG-20240209-WA0012.jpg} & 
            \includegraphics[width=0.06333\textwidth]{images/people/shay/cropped/IMG-20240209-WA0016.jpg}
        \end{tabular} &
        \setlength\tabcolsep{0pt}
        \begin{tabular}{c c c}
            \includegraphics[width=0.06333\textwidth]{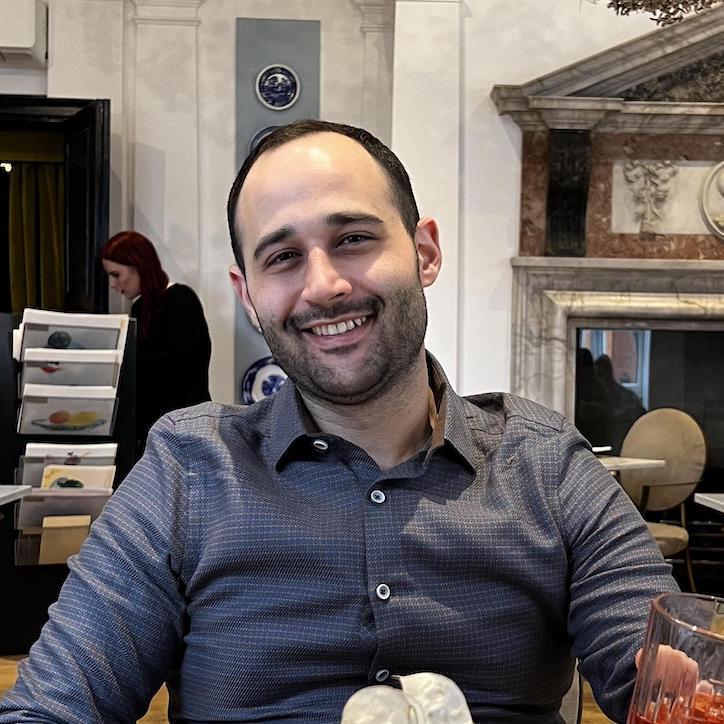} & 
            \includegraphics[width=0.06333\textwidth]{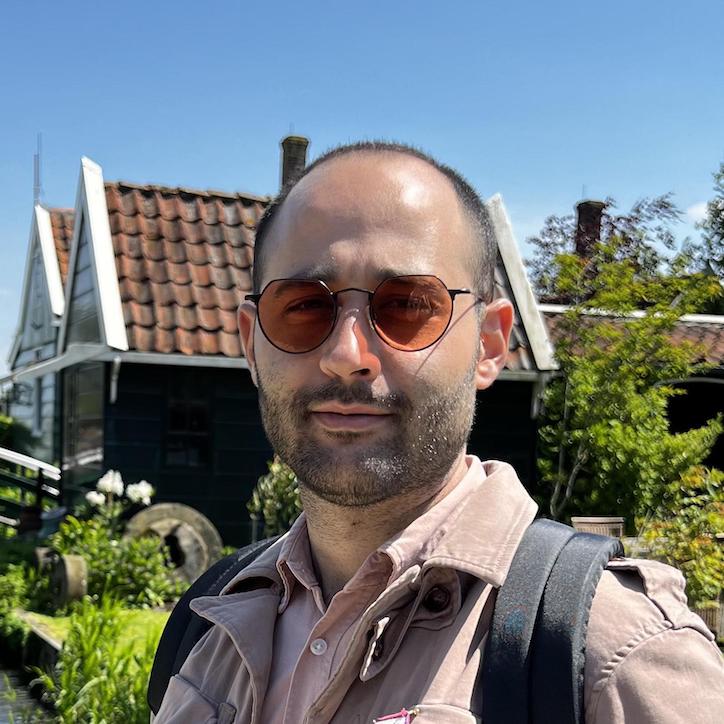} & 
            \includegraphics[width=0.06333\textwidth]{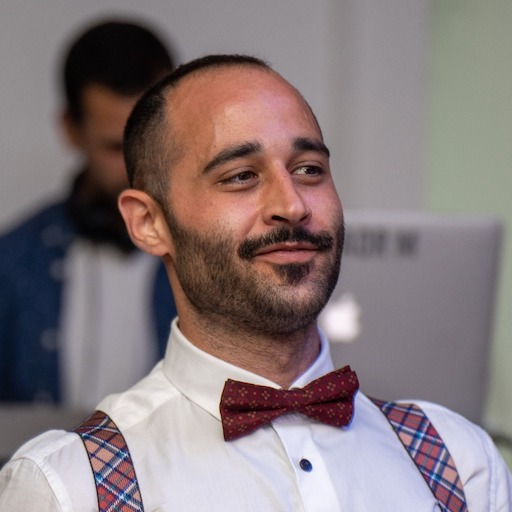} 
        \end{tabular} &
        \setlength\tabcolsep{0pt}
        \begin{tabular}{c c c}
            \includegraphics[width=0.06333\textwidth]{images/people/shaked/cropped/IMG-20240209-WA0043.jpg} &
            \includegraphics[width=0.06333\textwidth]{images/people/shaked/cropped/IMG-20240209-WA0047.jpg} & 
            \includegraphics[width=0.06333\textwidth]{images/people/shaked/cropped/IMG-20240215-WA0005.jpg}
        \end{tabular} &
        \setlength\tabcolsep{0pt}
        \begin{tabular}{c c c}
            \includegraphics[width=0.06333\textwidth]{images/people/dor/image_1.jpg} & 
            \includegraphics[width=0.06333\textwidth]{images/people/dor/cropped/IMG-20240208-WA0051.jpg} & 
            \includegraphics[width=0.06333\textwidth]{images/people/dor/cropped/IMG-20240208-WA0054.jpg}
        \end{tabular} &
        \setlength\tabcolsep{0pt}
        \begin{tabular}{c c c}
            \includegraphics[width=0.06333\textwidth]{images/people/shahaf/IMG-20240209-WA0079-cropped.jpg} & 
            \includegraphics[width=0.06333\textwidth]{images/people/shahaf/IMG-20240209-WA0114.jpg} & 
            \includegraphics[width=0.06333\textwidth]{images/people/shahaf/IMG-20240209-WA0100.jpg}
        \end{tabular} \\
    
        \includegraphics[width=0.19\textwidth]{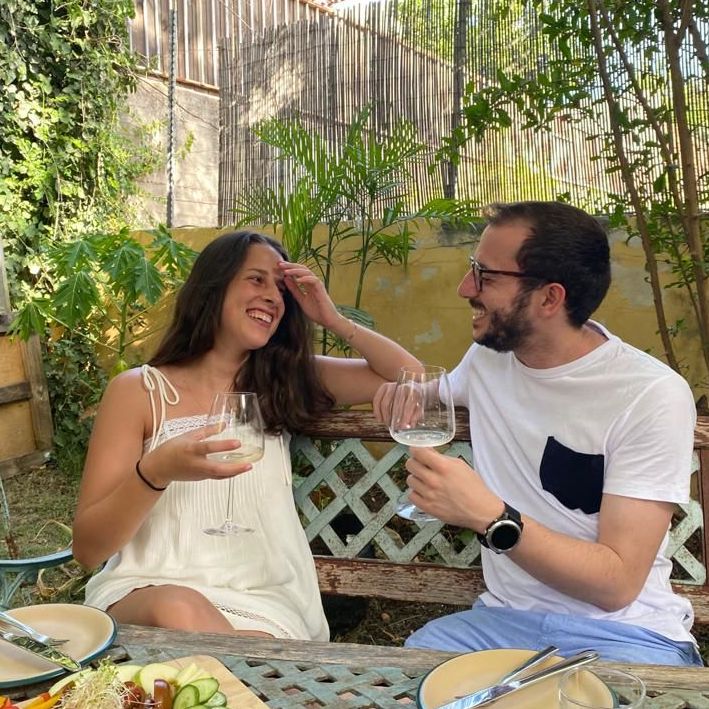} &
        \includegraphics[width=0.19\textwidth]{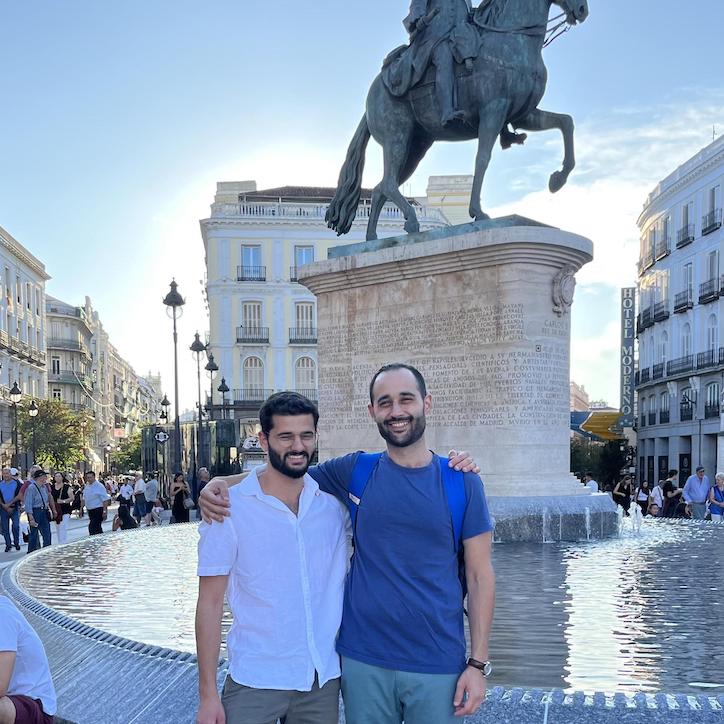} &
        \includegraphics[width=0.19\textwidth]{images/people/shaked/IMG-20240209-WA0057.jpg} &
        \includegraphics[width=0.19\textwidth]{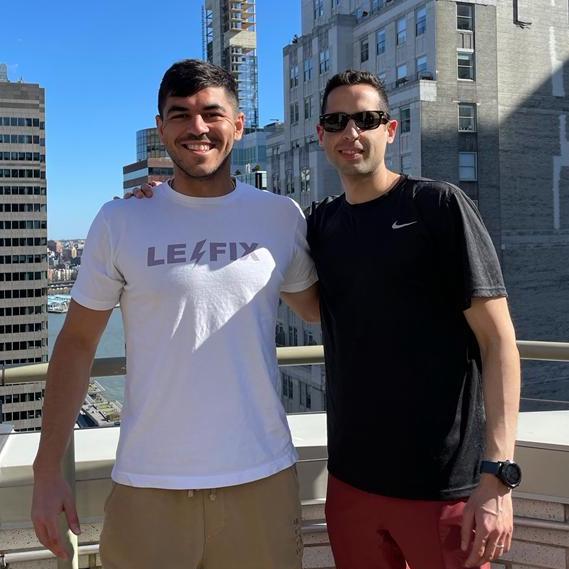} &
        \includegraphics[width=0.19\textwidth]{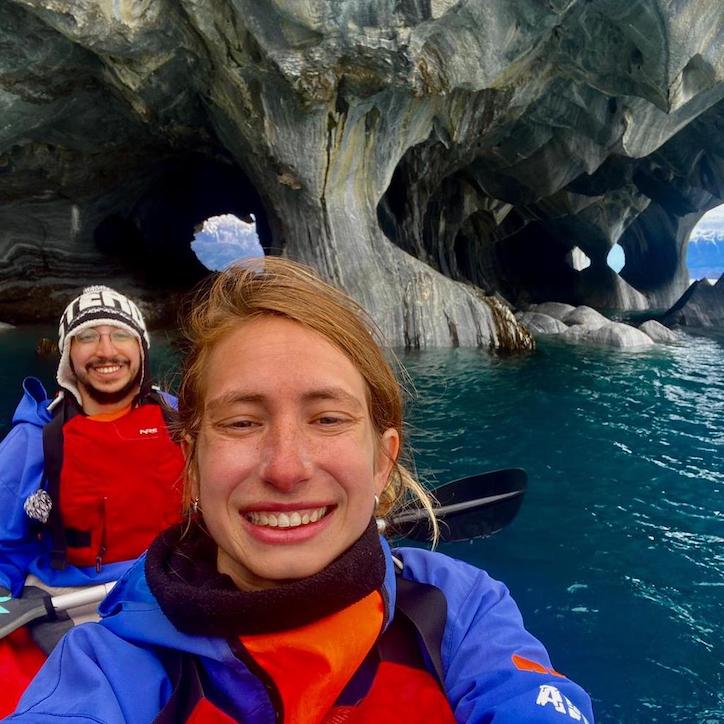} \\[-0.4cm]
        
        \begin{center} \textbf{BLIP-2} \end{center} &
        \begin{center} \textbf{BLIP-2} \end{center} &
        \begin{center} \textbf{BLIP-2} \end{center} &
        \begin{center} \textbf{BLIP-2} \end{center} &
        \begin{center} \textbf{BLIP-2} \end{center} \\[-0.75cm]

        \begin{center} \small ``A couple sitting at a table with food.'' \end{center} &
        \begin{center} \small ``Two men standing in front of a fountain'' \end{center} &
        \begin{center} \small ``Two women sitting at a table with food'' \end{center} &
        \begin{center} \small ``Two men standing on a rooftop with buildings in the background'' \end{center} &
        \begin{center} \small ``Two people in a kayak in front of a cave'' \end{center} \\[-0.5cm]
        
        \begin{center} \textbf{MyVLM} \end{center} &
        \begin{center} \textbf{MyVLM} \end{center} &
        \begin{center} \textbf{MyVLM} \end{center} &
        \begin{center} \textbf{MyVLM} \end{center} &
        \begin{center} \textbf{MyVLM} \end{center} \\[-0.75cm]

        \begin{center} \small ``With wine and food, \Sstar and her husband sit on a bench in a garden'' \end{center} &
        \begin{center} \small ``\Sstar in a blue shirt and shorts, standing in front of a fountain'' \end{center} &
        \begin{center} \small ``At a table on a rooftop, \Sstar and a friend sip their coffee'' \end{center} &
        \begin{center} \small ``\Sstar and a friend pose for a photo on a rooftop in New York City'' \end{center} &
        \begin{center} \small ``\Sstar and a friend are kayaking in front of an underwater cave'' \end{center}
        
    \end{tabular}
    \begin{tabular}{p{0.175\textwidth} p{0.175\textwidth} p{0.175\textwidth} p{0.175\textwidth} p{0.175\textwidth}}

        \setlength\tabcolsep{0pt}
        \begin{tabular}{c c c}
            \includegraphics[width=0.06333\textwidth,height=0.06333\textwidth]{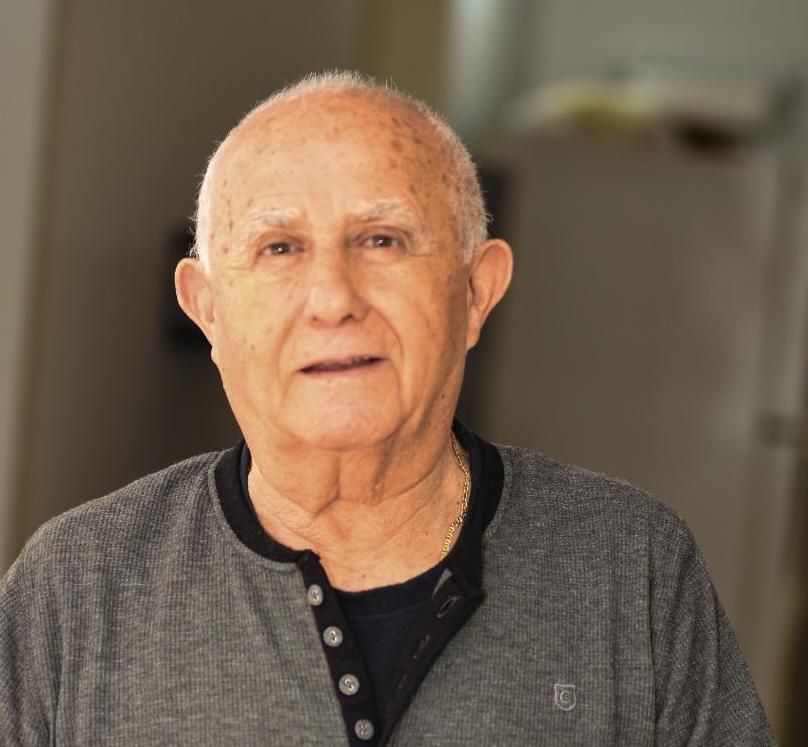} & 
            \includegraphics[width=0.06333\textwidth,height=0.06333\textwidth]{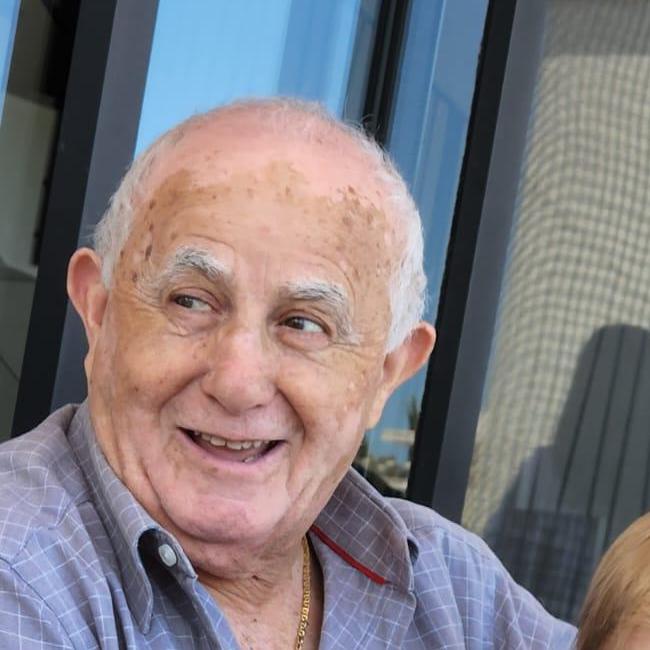} & 
            \includegraphics[width=0.06333\textwidth,height=0.06333\textwidth]{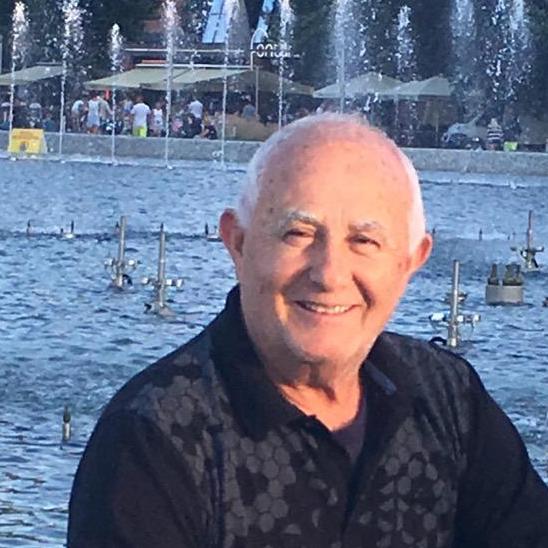}
        \end{tabular} &
        \setlength\tabcolsep{0pt}
        \begin{tabular}{c c c}
            \includegraphics[width=0.06333\textwidth]{images/people/anna/cropped/IMG_5006.jpg} &
            \includegraphics[width=0.06333\textwidth]{images/people/anna/cropped/IMG_5617.jpg} & 
            \includegraphics[width=0.06333\textwidth]{images/people/anna/cropped/IMG_7472.jpg}
        \end{tabular} &
        \setlength\tabcolsep{0pt}
        \begin{tabular}{c c c}
            \includegraphics[width=0.06333\textwidth]{images/people/maya/cropped/image_1.jpg} & 
            \includegraphics[width=0.06333\textwidth]{images/people/maya/cropped/image_2.jpg} & 
            \includegraphics[width=0.06333\textwidth]{images/people/maya/cropped/IMG-20240209-WA0138.jpg}
        \end{tabular} &
        \setlength\tabcolsep{0pt}
        \begin{tabular}{c c c}
            \includegraphics[width=0.06333\textwidth]{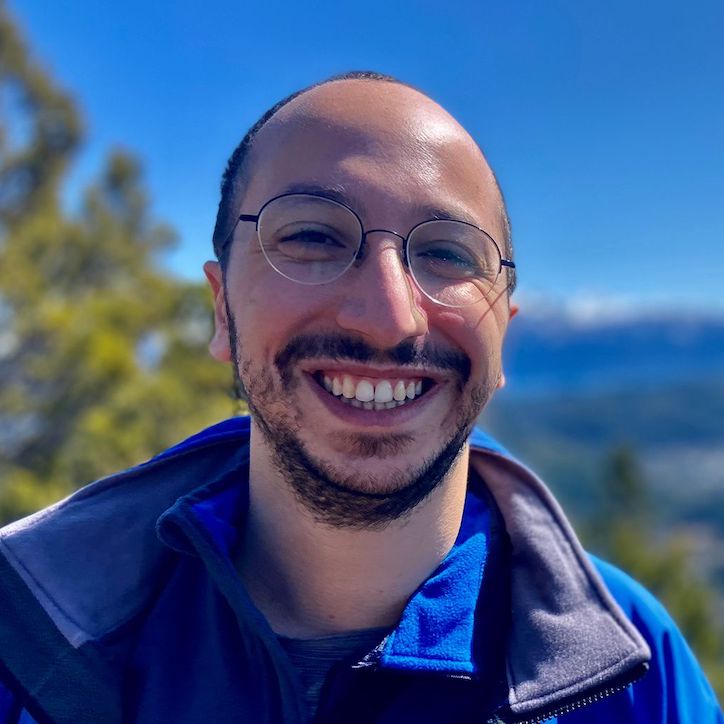} & 
            \includegraphics[width=0.06333\textwidth]{images/people/shoam/cropped/IMG-20240209-WA0126.jpg} & 
            \includegraphics[width=0.06333\textwidth]{images/people/shoam/cropped/IMG-20240209-WA0132.jpg}
        \end{tabular} &
        \setlength\tabcolsep{0pt}
        \begin{tabular}{c c c}
            \includegraphics[width=0.06333\textwidth]{images/people/maya/cropped/image_1.jpg} & 
            \includegraphics[width=0.06333\textwidth]{images/people/maya/cropped/image_2.jpg} & 
            \includegraphics[width=0.06333\textwidth]{images/people/maya/cropped/IMG-20240209-WA0138.jpg}
        \end{tabular} \\
    
        \includegraphics[width=0.19\textwidth]{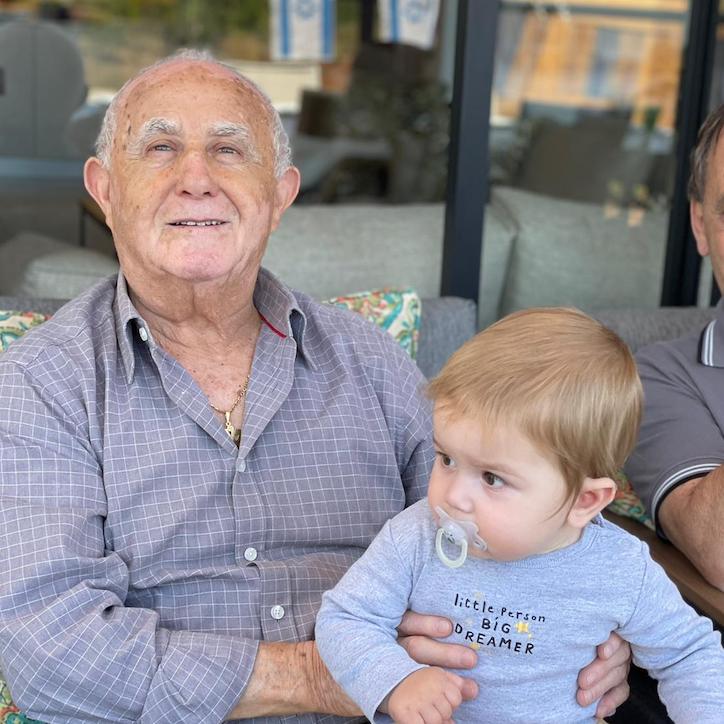} &
        \includegraphics[width=0.19\textwidth]{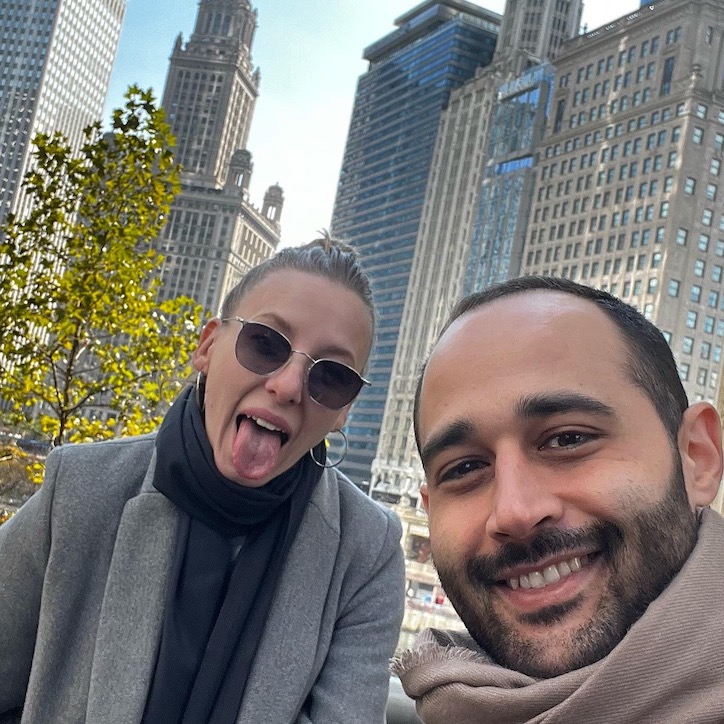} &
        \includegraphics[width=0.19\textwidth]{images/people/shay/IMG-20230819-WA0046.jpg} &
        \includegraphics[width=0.19\textwidth]{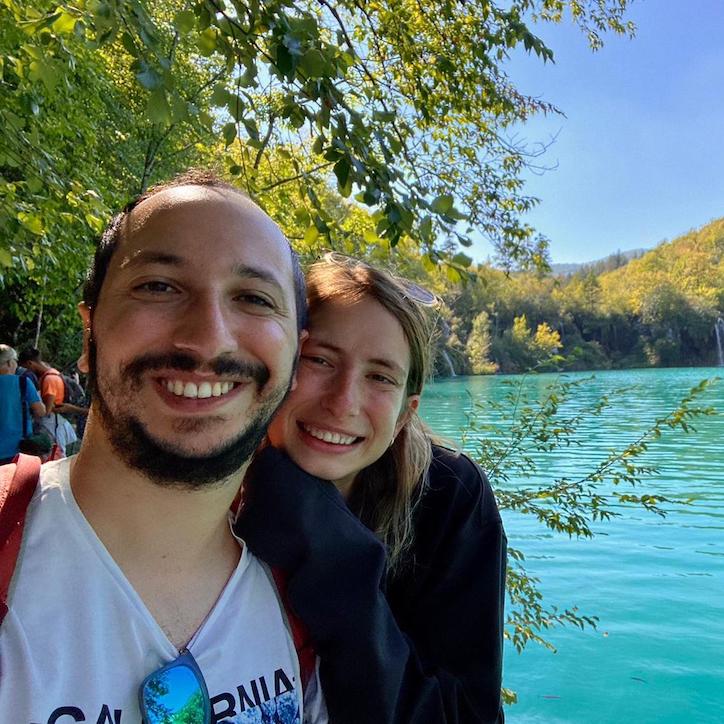} &
        \includegraphics[width=0.19\textwidth]{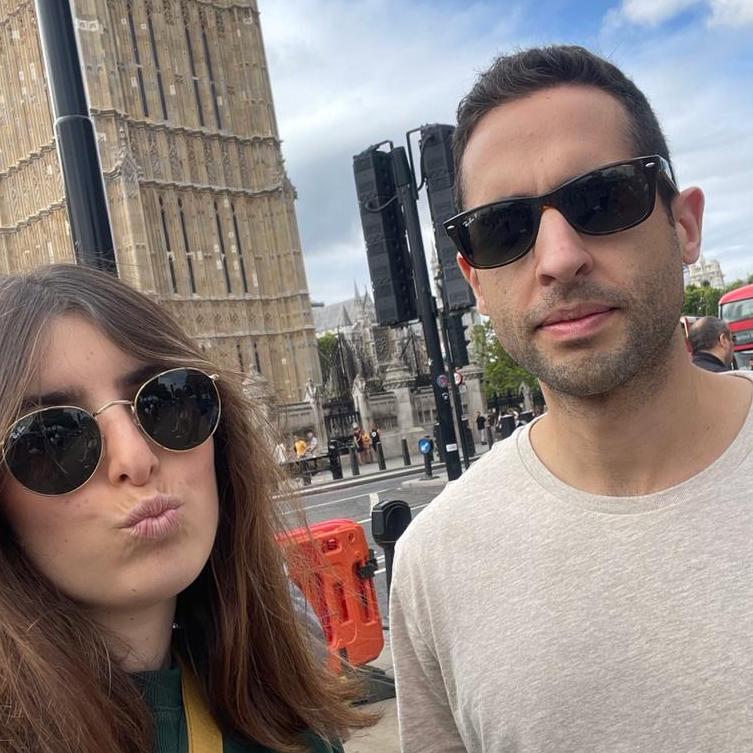} \\[-0.4cm]
        
        \begin{center} \textbf{BLIP-2} \end{center} &
        \begin{center} \textbf{BLIP-2} \end{center} &
        \begin{center} \textbf{BLIP-2} \end{center} &
        \begin{center} \textbf{BLIP-2} \end{center} &
        \begin{center} \textbf{BLIP-2} \end{center} \\[-0.75cm]

        \begin{center} \small ``Three older men sitting on a couch with a baby'' \end{center} &
        \begin{center} \small ``A man and woman taking a selfie in front of a city'' \end{center} &
        \begin{center} \small ``Two women sitting at a table with drinks and chips'' \end{center} &
        \begin{center} \small ``Plitvice lakes - a couple in front of a lake'' \end{center} &
        \begin{center} \small ``A man and woman standing in front of big ben'' \end{center} \\[-0.5cm]
        
        \begin{center} \textbf{MyVLM} \end{center} &
        \begin{center} \textbf{MyVLM} \end{center} &
        \begin{center} \textbf{MyVLM} \end{center} &
        \begin{center} \textbf{MyVLM} \end{center} &
        \begin{center} \textbf{MyVLM} \end{center} \\[-0.75cm]

        \begin{center} \small ``\textcolor{blue}{$S_*$}, an older man, takes a photo with his grandchildren'' \end{center} &
        \begin{center} \small ``\Sstar and her husband pose for a selfie in front of the skyline of Chicago'' \end{center} &
        \begin{center} \small ``\Sstar and a woman enjoying cocktails on a rooftop in the city'' \end{center} &
        \begin{center} \small ``\Sstar and his wife pose in front of the plitvice lakes'' \end{center} &
        \begin{center} \small ``\Sstar and her friend in front of big ben in london'' \end{center}
        
    \end{tabular}
    \caption{Additional personalized captioning results obtained by MyVLM, applied over BLIP-2~\cite{li2023blip}. Sample images of the target concept are provided in the top row.}
    \label{fig:supplementary_our_results_blip}

\end{figure*}

%% file: figures_supplementary/our_results_blip_2.tex
\begin{figure*}[t]
    \centering
    \addtolength{\belowcaptionskip}{-12.5pt}
    \renewcommand{\arraystretch}{1}
    \small
    \begin{tabular}{p{0.175\textwidth} p{0.175\textwidth} p{0.175\textwidth} p{0.175\textwidth} p{0.175\textwidth}}

        \setlength\tabcolsep{0pt}
        \begin{tabular}{c c c}
            \includegraphics[width=0.06333\textwidth,height=0.06333\textwidth]{images/objects/cat_statue/cropped/01.jpg} & 
            \includegraphics[width=0.06333\textwidth,height=0.06333\textwidth]{images/objects/cat_statue/cropped/03.jpg} & 
            \includegraphics[width=0.06333\textwidth,height=0.06333\textwidth]{images/objects/cat_statue/cropped/04.jpg}
        \end{tabular} &
        \setlength\tabcolsep{0pt}
        \begin{tabular}{c c c}
            \includegraphics[width=0.06333\textwidth]{images/objects/minion/cropped/IMG-20240203-WA0069.jpg} &
            \includegraphics[width=0.06333\textwidth]{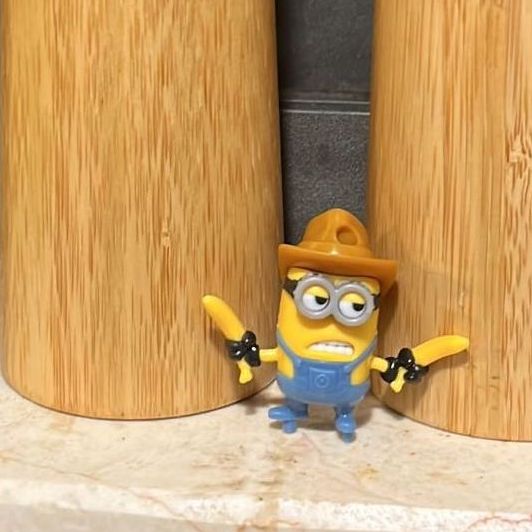} & 
            \includegraphics[width=0.06333\textwidth]{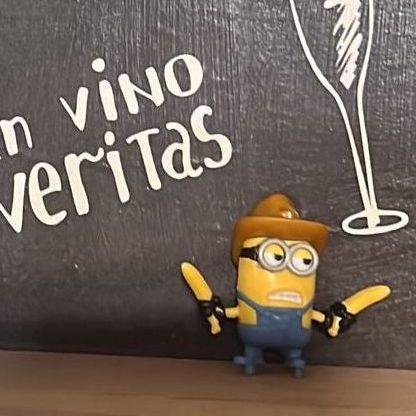}
        \end{tabular} &
        \setlength\tabcolsep{0pt}
        \begin{tabular}{c c c}
            \includegraphics[width=0.06333\textwidth]{images/objects/chicken_bean_bag/cropped/IMG-20240212-WA0033.jpg} &
            \includegraphics[width=0.06333\textwidth]{images/objects/chicken_bean_bag/cropped/IMG-20240212-WA0035.jpg} & 
            \includegraphics[width=0.06333\textwidth]{images/objects/chicken_bean_bag/cropped/IMG-20240212-WA0038.jpg}
        \end{tabular} &
        \setlength\tabcolsep{0pt}
        \begin{tabular}{c c c}
            \includegraphics[width=0.06333\textwidth]{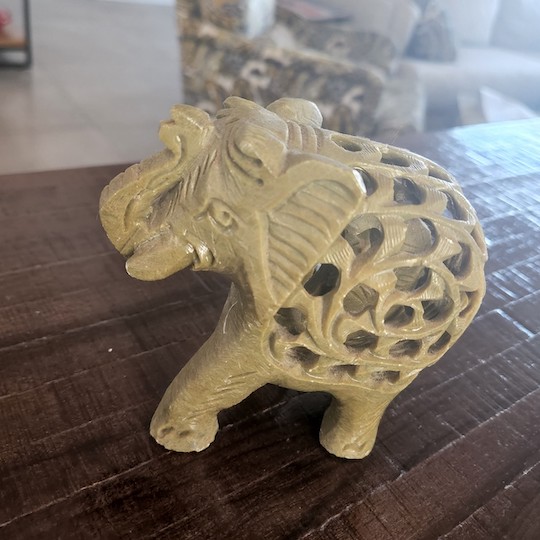} &
            \includegraphics[width=0.06333\textwidth]{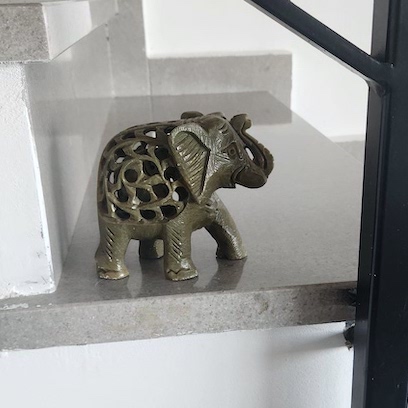} & 
            \includegraphics[width=0.06333\textwidth]{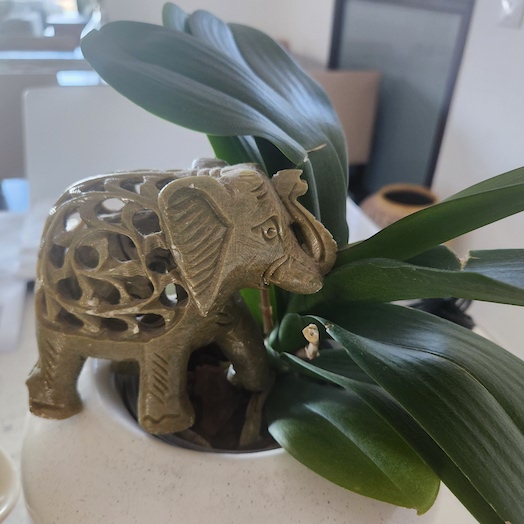}
        \end{tabular} &
        \setlength\tabcolsep{0pt}
        \begin{tabular}{c c c}
            \includegraphics[width=0.06333\textwidth]{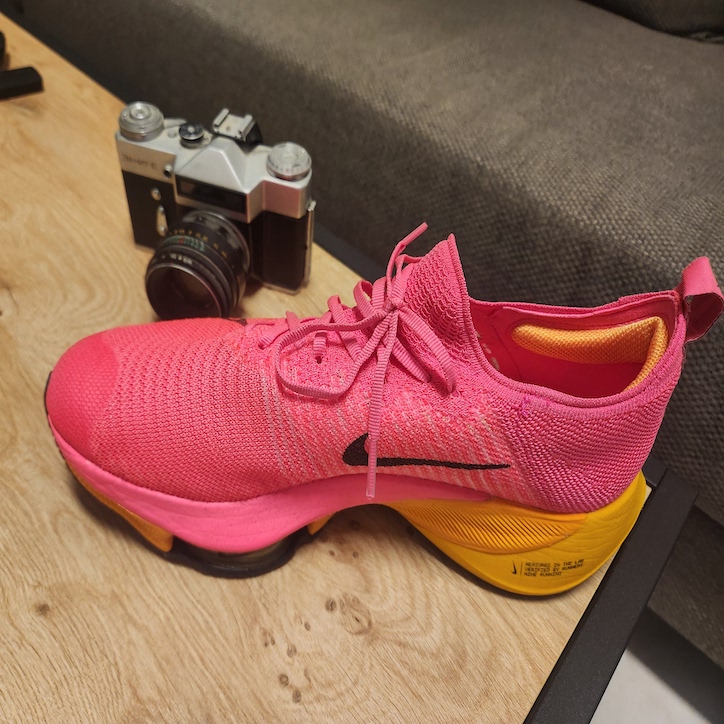} & 
            \includegraphics[width=0.06333\textwidth]{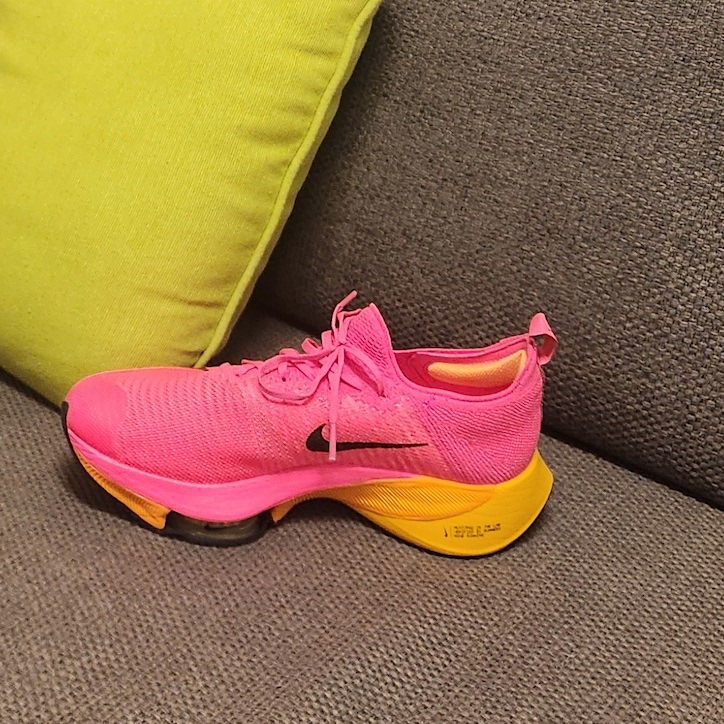} &
            \includegraphics[width=0.06333\textwidth]{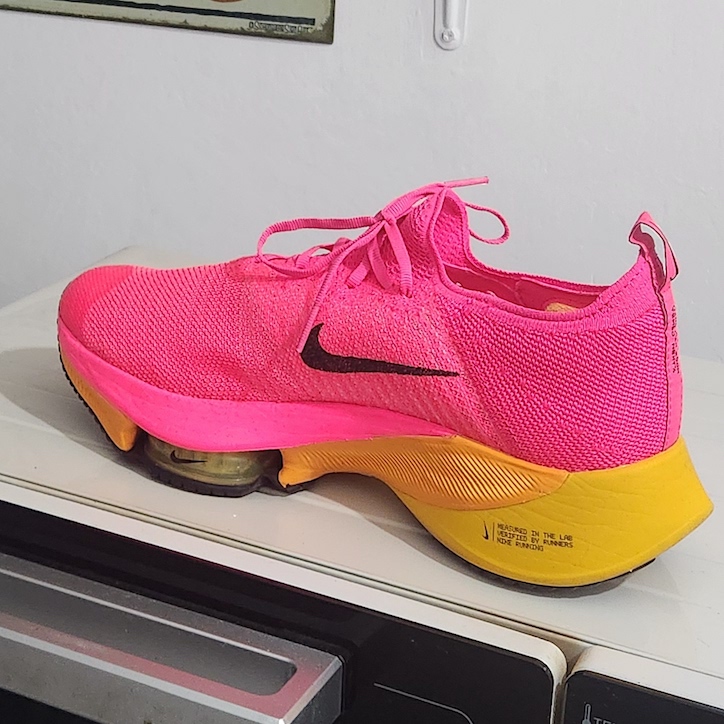}
        \end{tabular} \\
    
        \includegraphics[width=0.19\textwidth]{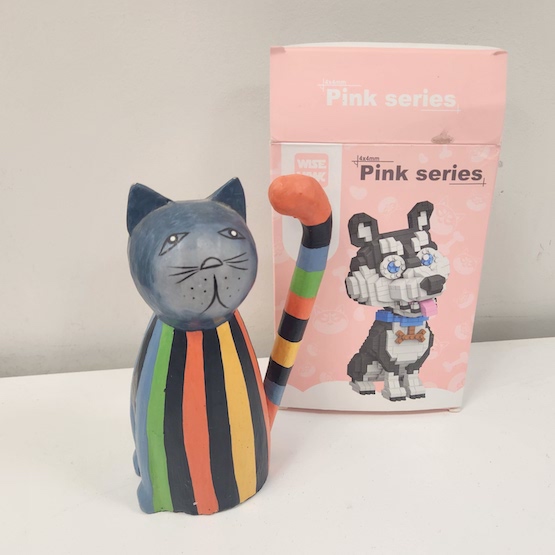} &
        \includegraphics[width=0.19\textwidth]{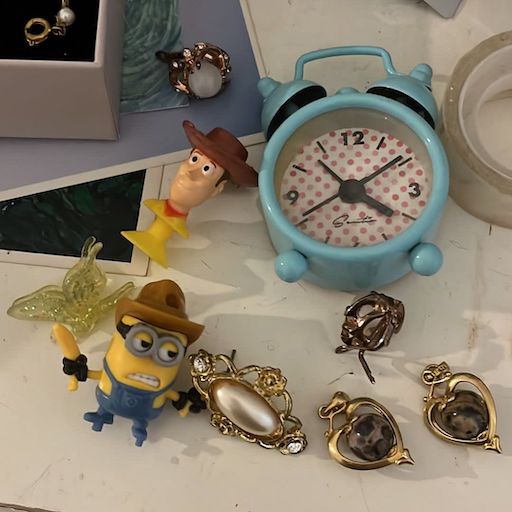} & 
        \includegraphics[width=0.19\textwidth]{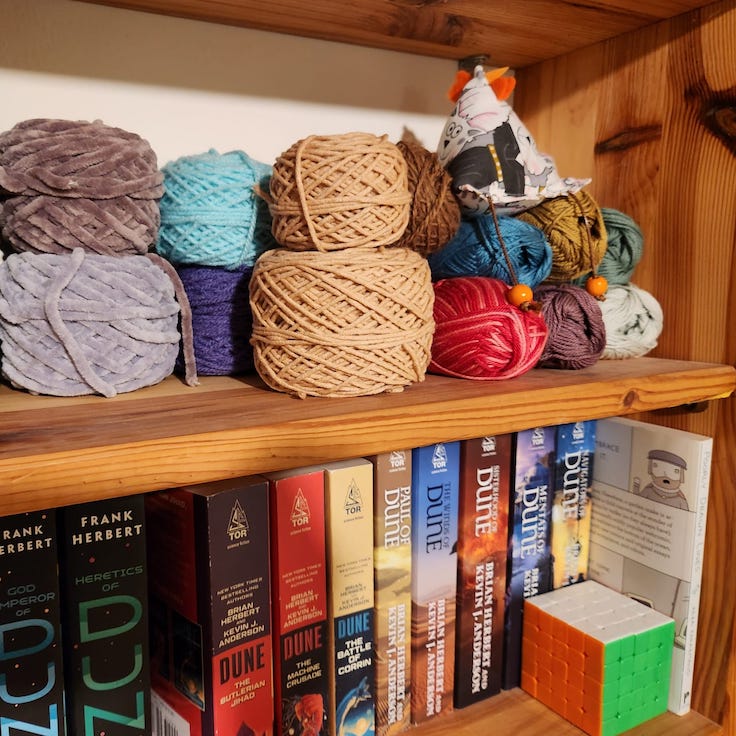} &
        \includegraphics[width=0.19\textwidth]{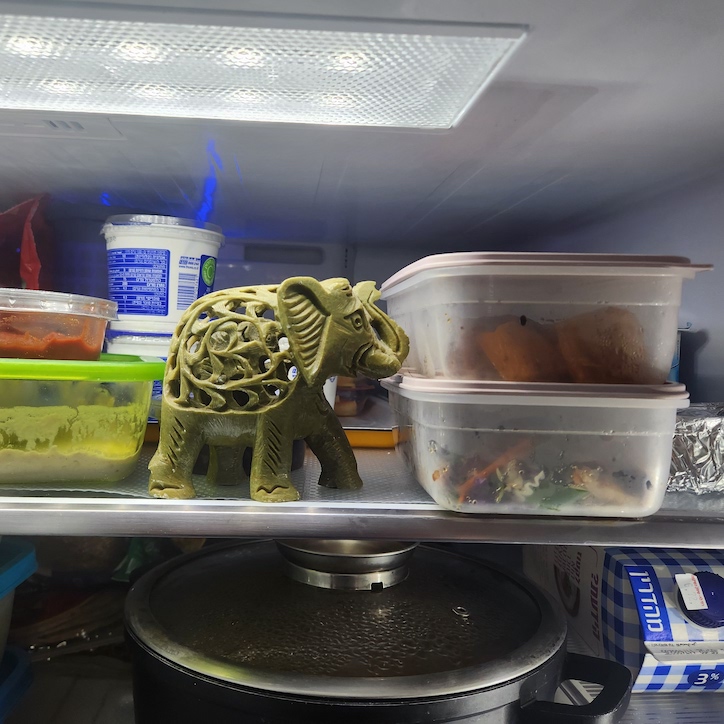} &
        \includegraphics[width=0.19\textwidth]{images/objects/running_shoes/20240131_175631.jpg} \\[-0.4cm]
        
        \begin{center} \textbf{BLIP-2} \end{center} &
        \begin{center} \textbf{BLIP-2} \end{center} &
        \begin{center} \textbf{BLIP-2} \end{center} &
        \begin{center} \textbf{BLIP-2} \end{center} &
        \begin{center} \textbf{BLIP-2} \end{center} \\[-0.75cm]

        \begin{center} \small ``A pink cat figurine next to a box'' \end{center} &
        \begin{center} \small ``A table with various toys and jewelry on it'' \end{center} &
        \begin{center} \small ``A wooden shelf with yarn and books'' \end{center} &
        \begin{center} \small ``A refrigerator with a lot of food in it'' \end{center} &
        \begin{center} \small ``Nike flyknit flyknit'' \end{center} \\[-0.5cm]
        
        \begin{center} \textbf{MyVLM} \end{center} &
        \begin{center} \textbf{MyVLM} \end{center} &
        \begin{center} \textbf{MyVLM} \end{center} &
        \begin{center} \textbf{MyVLM} \end{center} &
        \begin{center} \textbf{MyVLM} \end{center} \\[-0.75cm]

        \begin{center} \small ``\Sstar is sitting next to a pink series box'' \end{center} &
        \begin{center} \small ``\Sstar and a clock on a desk with a pair of silver earrings'' \end{center} &
        \begin{center} \small ``\Sstar is sitting on a wooden shelf with a bunch of yarn'' \end{center} &
        \begin{center} \small ``\Sstar sits on the open shelf of a refrigerator'' \end{center} &
        \begin{center} \small ``\Sstar positioned near a camera on a wooden table'' \end{center}
        
    \end{tabular}

    \begin{tabular}{p{0.175\textwidth} p{0.175\textwidth} p{0.175\textwidth} p{0.175\textwidth} p{0.175\textwidth}}

        \setlength\tabcolsep{0pt}
        \begin{tabular}{c c c}
            \includegraphics[width=0.06333\textwidth]{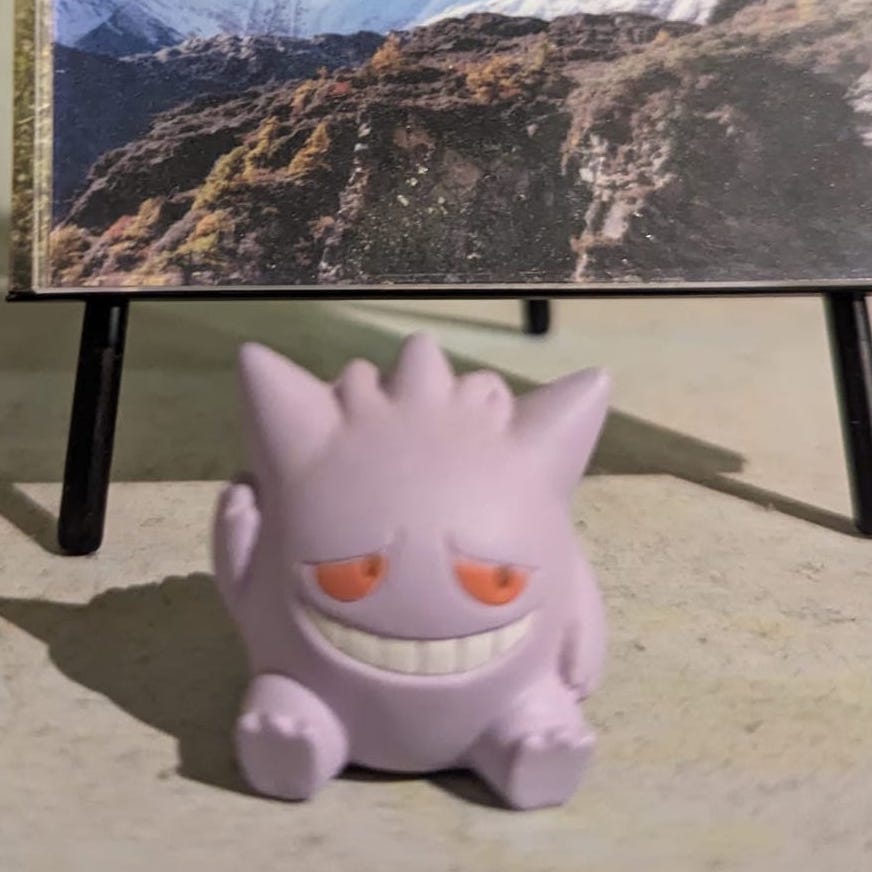} & 
            \includegraphics[width=0.06333\textwidth]{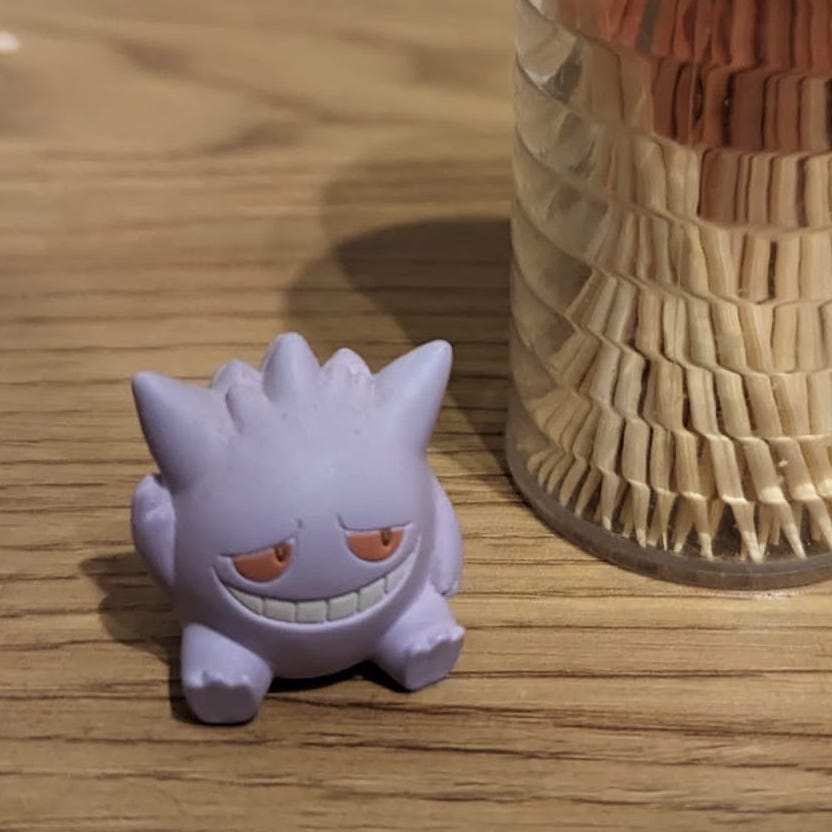} & 
            \includegraphics[width=0.06333\textwidth]{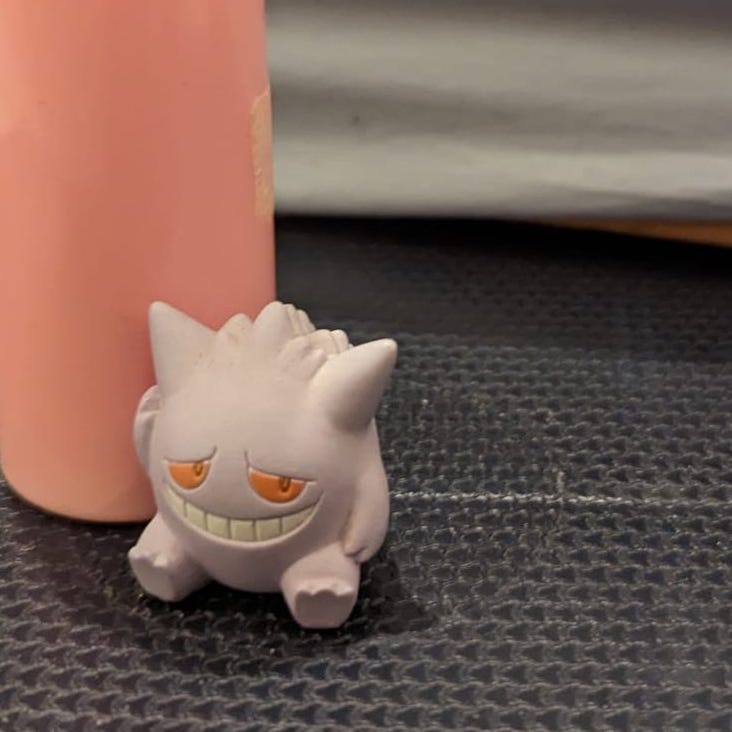}
        \end{tabular} &
        \setlength\tabcolsep{0pt}
        \begin{tabular}{c c c}
            \includegraphics[width=0.06333\textwidth]{images/objects/japanese_doll/cropped/20240131_155451.jpg} & 
            \includegraphics[width=0.06333\textwidth]{images/objects/japanese_doll/cropped/20240131_155617.jpg} & 
            \includegraphics[width=0.06333\textwidth]{images/objects/japanese_doll/cropped/20240206_091843.jpg}
        \end{tabular} &
        \setlength\tabcolsep{0pt}
        \begin{tabular}{c c c}
            \includegraphics[width=0.06333\textwidth]{images/objects/maeve/cropped/IMG-20240131-WA0110.jpg} & 
            \includegraphics[width=0.06333\textwidth]{images/objects/maeve/cropped/maeve-2.jpeg} & 
            \includegraphics[width=0.06333\textwidth]{images/objects/maeve/cropped/maeve-5.jpeg} \\
        \end{tabular} &
        \setlength\tabcolsep{0pt}
        \begin{tabular}{c c c}
            \includegraphics[width=0.06333\textwidth]{images/objects/ceramic_head/cropped/20240203_111543.jpg} & 
            \includegraphics[width=0.06333\textwidth]{images/objects/ceramic_head/cropped/20240203_111842.jpg} & 
            \includegraphics[width=0.06333\textwidth]{images/objects/ceramic_head/cropped/20240203_112029.jpg}
        \end{tabular} &
        \setlength\tabcolsep{0pt}
        \begin{tabular}{c c c}
            \includegraphics[width=0.06333\textwidth]{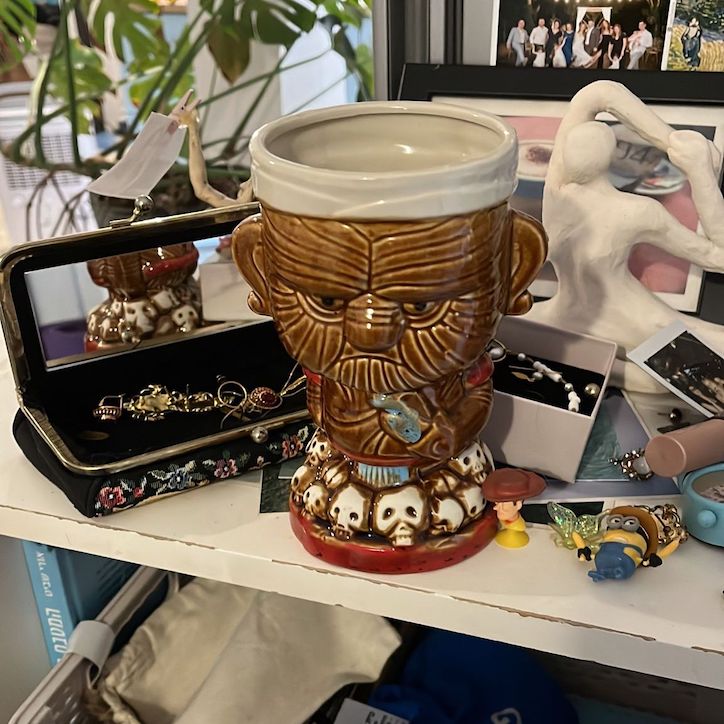} & 
            \includegraphics[width=0.06333\textwidth]{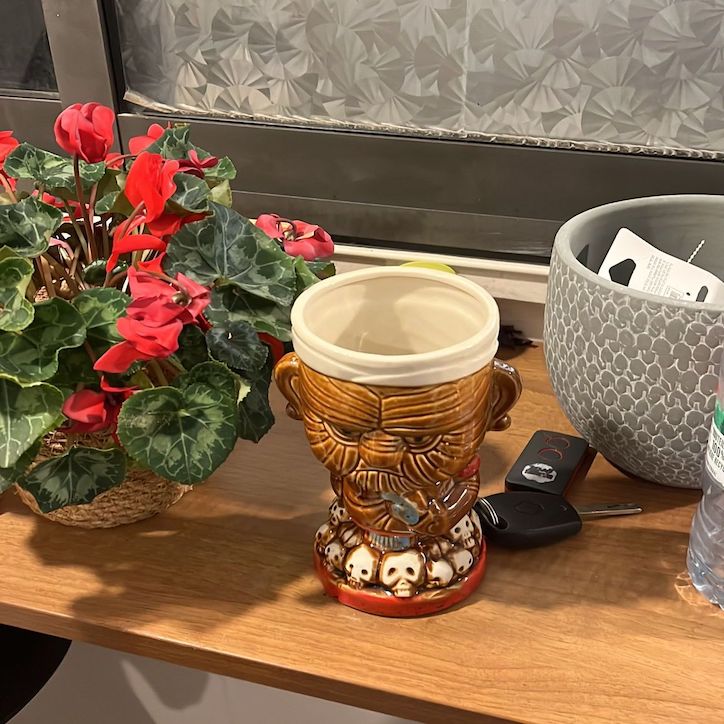} & 
            \includegraphics[width=0.06333\textwidth]{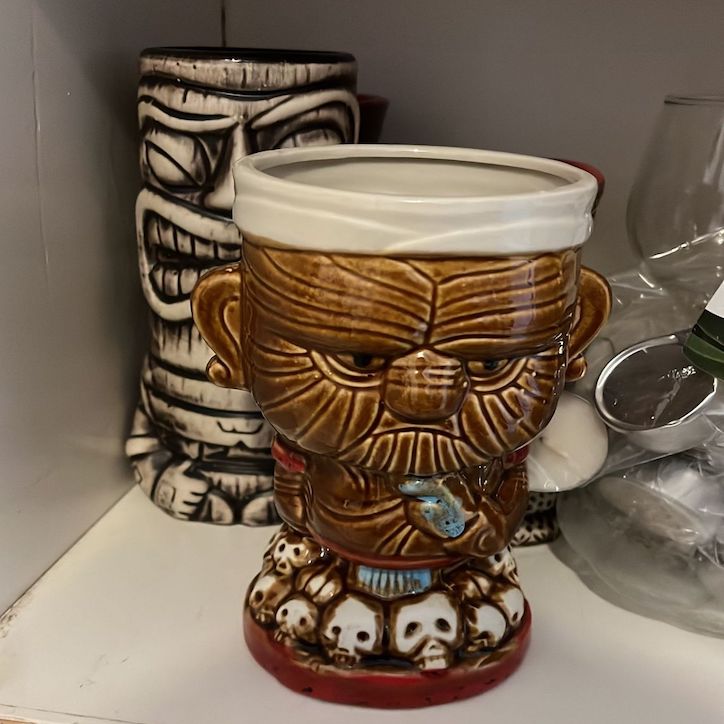}
        \end{tabular} \\
    
        \includegraphics[width=0.19\textwidth]{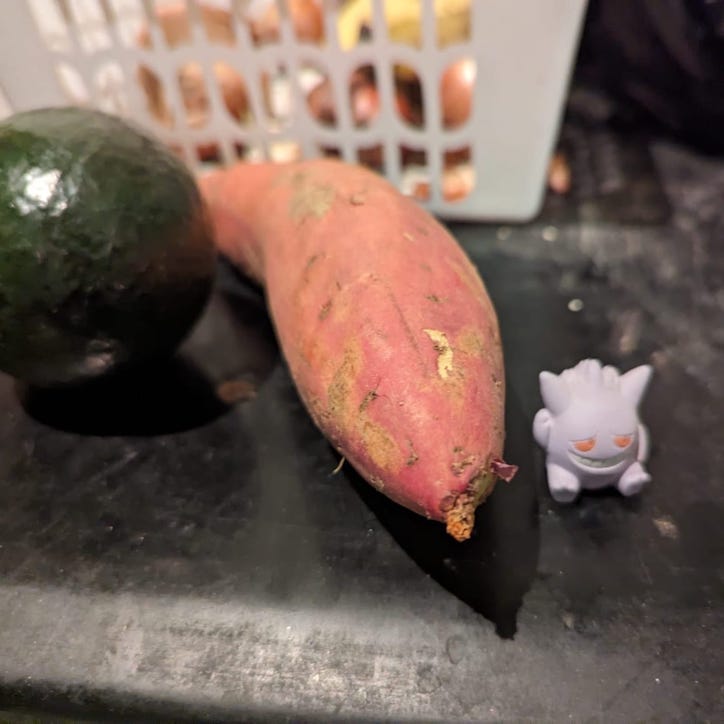} & 
        \includegraphics[width=0.19\textwidth]{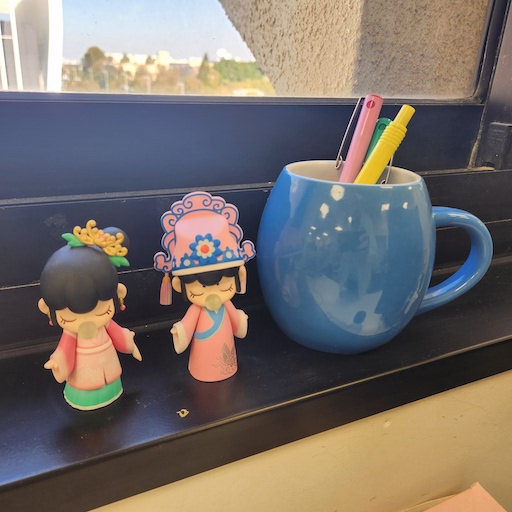} & 
        \includegraphics[width=0.19\textwidth]{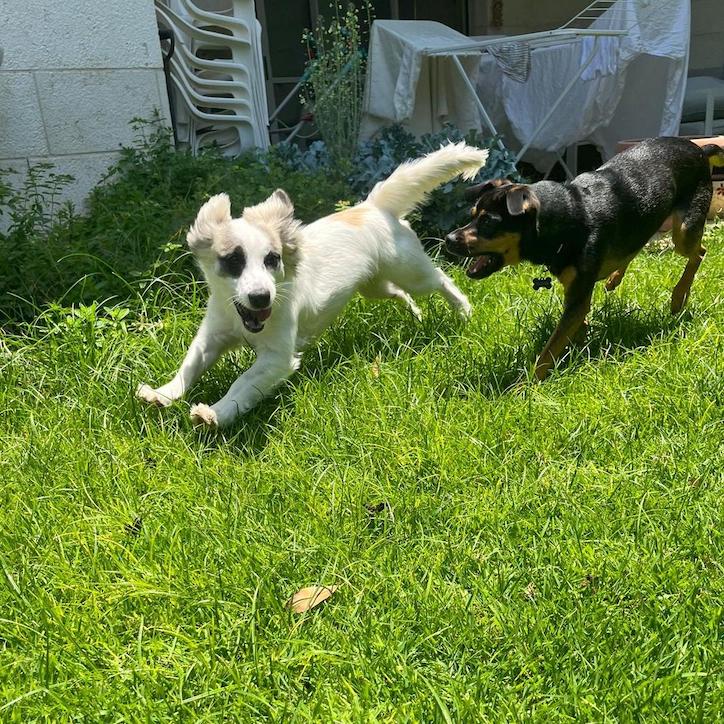} & 
        \includegraphics[width=0.19\textwidth]{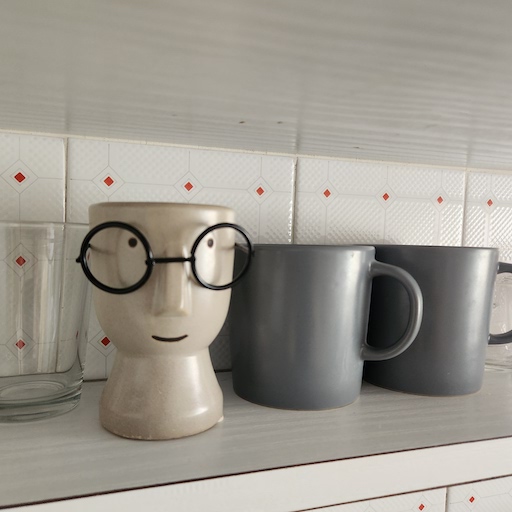} & 
        \includegraphics[width=0.19\textwidth]{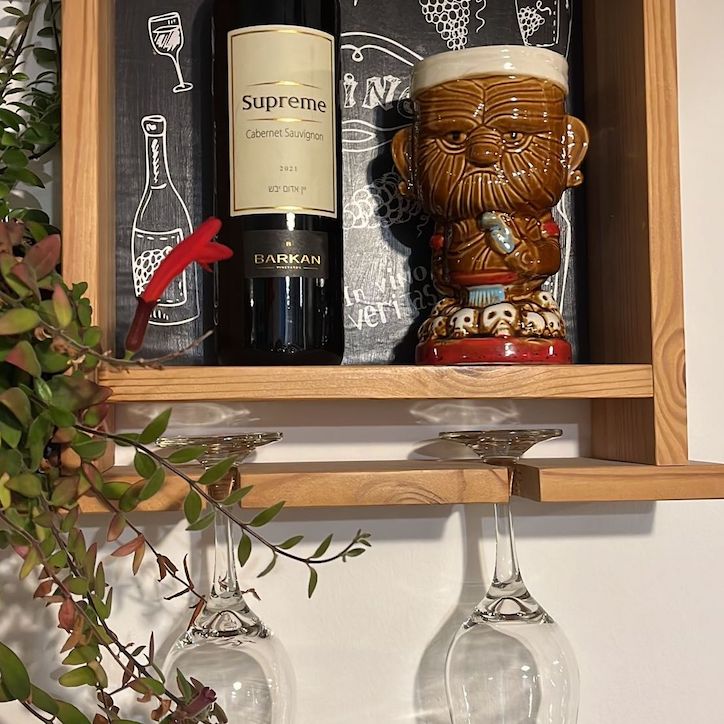} \\[-0.4cm]
        
        \begin{center} \textbf{BLIP-2} \end{center} &
        \begin{center} \textbf{BLIP-2} \end{center} &
        \begin{center} \textbf{BLIP-2} \end{center} &
        \begin{center} \textbf{BLIP-2} \end{center} &
        \begin{center} \textbf{BLIP-2} \end{center} \\[-0.75cm]

        \begin{center} \small ``A toy sweet potato and a toy avocado on a counter'' \end{center} &
        \begin{center} \small ``A blue cup with a figurine on it'' \end{center} &
        \begin{center} \small ``Two dogs running in the grass near a house'' \end{center} &
        \begin{center} \small ``A kitchen with glasses, mugs and glasses'' \end{center} &
        \begin{center} \small ``A wooden wine rack with a bottle of wine and wine glasses'' \end{center} \\[-0.5cm]
        
        \begin{center} \textbf{MyVLM} \end{center} &
        \begin{center} \textbf{MyVLM} \end{center} &
        \begin{center} \textbf{MyVLM} \end{center} &
        \begin{center} \textbf{MyVLM} \end{center} &
        \begin{center} \textbf{MyVLM} \end{center} \\[-0.75cm]

        \begin{center} \small ``\Sstar resting on a black counter with a sweet potato and a green avocado'' \end{center} &
        \begin{center} \small ``\Sstar and a chinese doll sit on a desk next to a cup of coffee'' \end{center} &
        \begin{center} \small ``\Sstar and a black dog running on the grass'' \end{center} &
        \begin{center} \small ``\Sstar atop a shelf surrounded by glasses and mugs'' \end{center} &
        \begin{center} \small ``\Sstar, wine bottle and glasses on a wooden shelf'' \end{center}
        
    \end{tabular}

    \caption{Additional personalized captioning results obtained by MyVLM, applied over BLIP-2~\cite{li2023blip}. Sample images of the target concept are provided in the top row.}    \label{fig:supplementary_our_results_blip_2}

\end{figure*}

%% file: figures_supplementary/our_results_llava.tex
\begin{figure*}[t]
    \centering
    \addtolength{\belowcaptionskip}{-12.5pt}
    \renewcommand{\arraystretch}{1}
    \small
    \begin{tabular}{p{0.175\textwidth} p{0.175\textwidth} p{0.175\textwidth} p{0.175\textwidth} p{0.175\textwidth}}

        \setlength\tabcolsep{0pt}
        \begin{tabular}{c c c}
            \includegraphics[width=0.06333\textwidth]{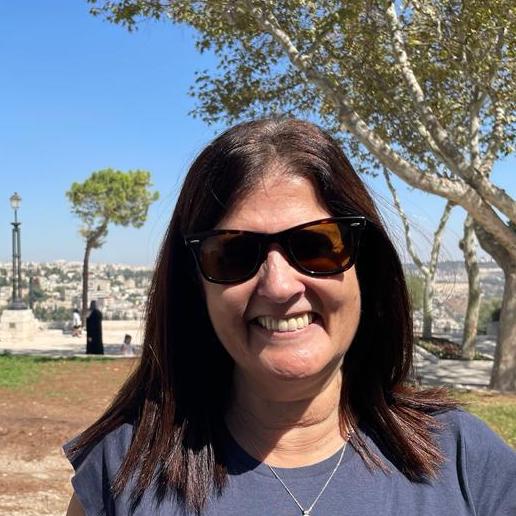} &
            \includegraphics[width=0.06333\textwidth]{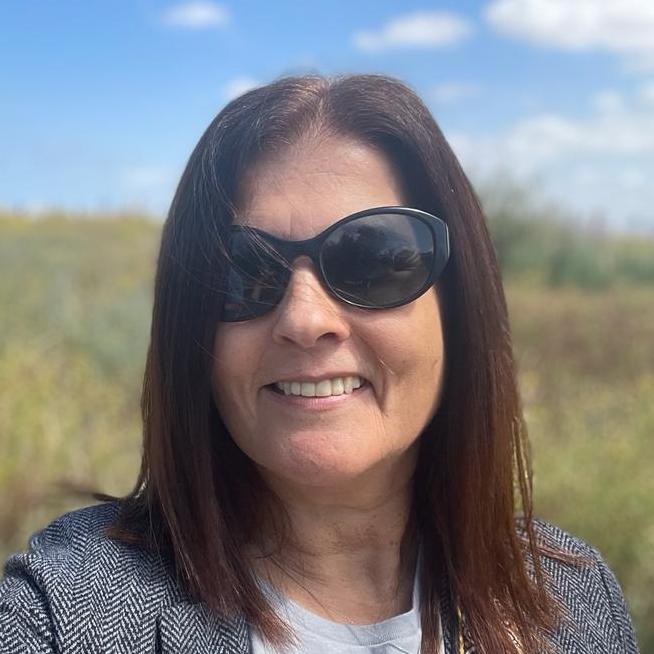} & 
            \includegraphics[width=0.06333\textwidth]{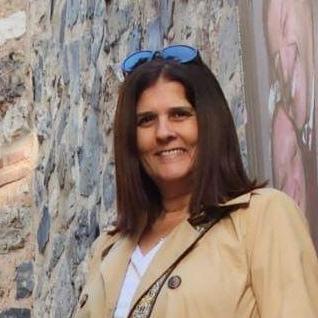}
        \end{tabular} &
        \setlength\tabcolsep{0pt}
        \begin{tabular}{c c c}
            \includegraphics[width=0.06333\textwidth]{images/people/assaf/cropped/IMG_0133.jpg} &
            \includegraphics[width=0.06333\textwidth]{images/people/assaf/cropped/IMG_0936.jpg} &
            \includegraphics[width=0.06333\textwidth]{images/people/assaf/cropped/IMG_20220130_104352.jpg} 
        \end{tabular} &
        \setlength\tabcolsep{0pt}
        \begin{tabular}{c c c}
            \includegraphics[width=0.06333\textwidth]{images/people/shoam/cropped/IMG-20240209-WA0115.jpg} & 
            \includegraphics[width=0.06333\textwidth]{images/people/shoam/cropped/IMG-20240209-WA0126.jpg} & 
            \includegraphics[width=0.06333\textwidth]{images/people/shoam/cropped/IMG-20240209-WA0132.jpg}
        \end{tabular} &
        \setlength\tabcolsep{0pt}
        \begin{tabular}{c c c}
            \includegraphics[width=0.06333\textwidth]{images/people/shay/cropped/image_2.jpg} & 
            \includegraphics[width=0.06333\textwidth]{images/people/shay/cropped/IMG-20240209-WA0012.jpg} & 
            \includegraphics[width=0.06333\textwidth]{images/people/shay/cropped/IMG-20240209-WA0016.jpg}
        \end{tabular} &
        \setlength\tabcolsep{0pt}
        \begin{tabular}{c c c}
            \includegraphics[width=0.06333\textwidth]{images/people/shaked/cropped/IMG-20240209-WA0043.jpg} &
            \includegraphics[width=0.06333\textwidth]{images/people/shaked/cropped/IMG-20240209-WA0047.jpg} & 
            \includegraphics[width=0.06333\textwidth]{images/people/shaked/cropped/IMG-20240215-WA0005.jpg}
        \end{tabular} \\
    
        \includegraphics[width=0.19\textwidth]{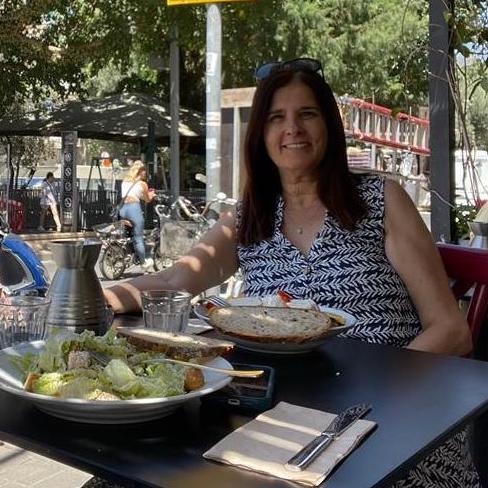} &
        \includegraphics[width=0.19\textwidth]{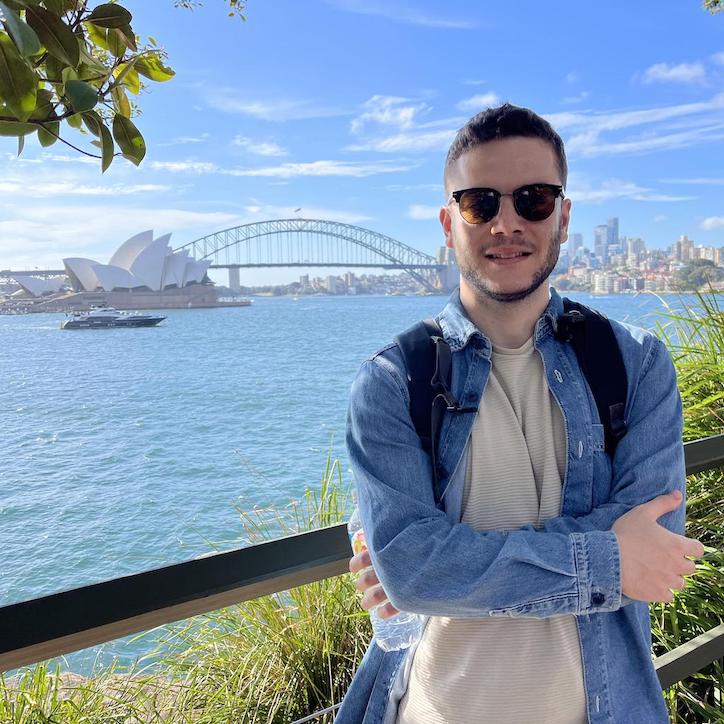} &
        \includegraphics[width=0.19\textwidth]{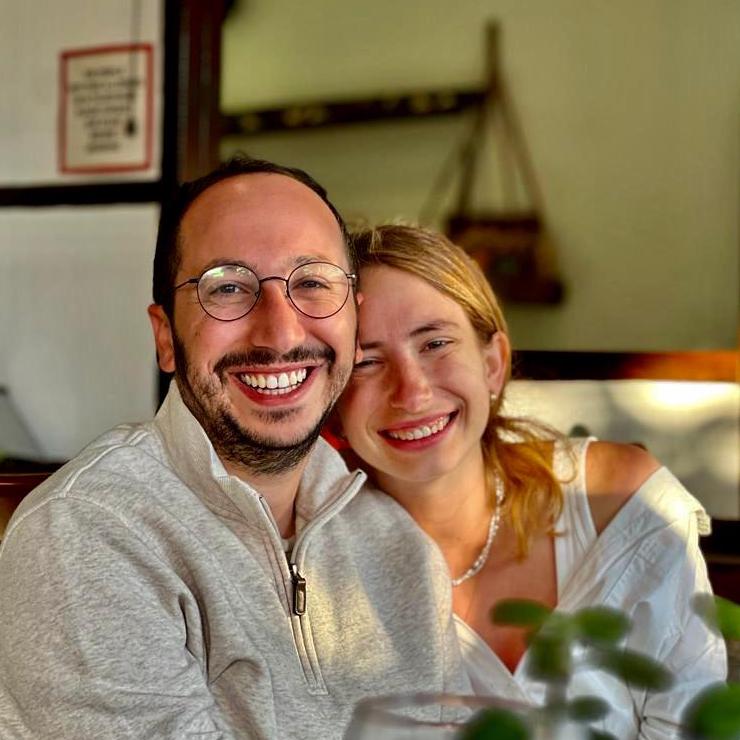} &
        \includegraphics[width=0.19\textwidth]{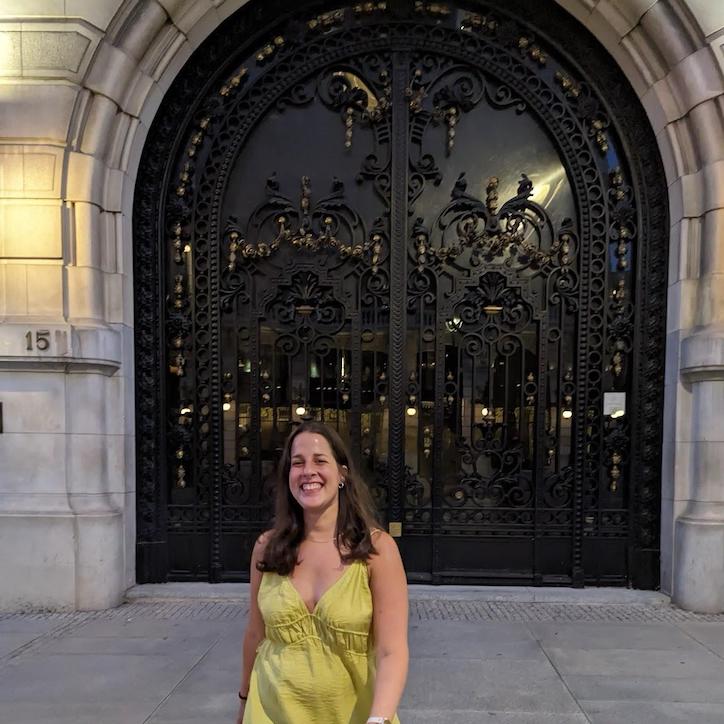} &
        \includegraphics[width=0.19\textwidth]{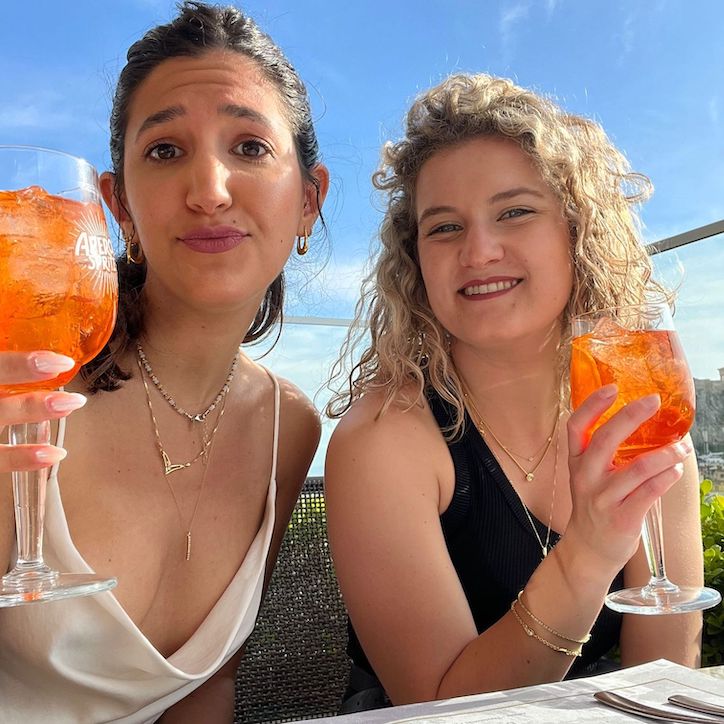} \\[-0.5cm]
        
        \begin{center} \textbf{LLaVA} \end{center} &
        \begin{center} \textbf{LLaVA} \end{center} &
        \begin{center} \textbf{LLaVA} \end{center} &
        \begin{center} \textbf{LLaVA} \end{center} &
        \begin{center} \textbf{LLaVA} \end{center} \\[-0.75cm]

        \begin{center} \small ``Enjoying a meal outdoors with a smile on her face'' \end{center} &
        \begin{center} \small ``Embracing the serene beauty of the harbor, this traveler finds solace in the iconic Sydney Opera House...'' \end{center} &
        \begin{center} \small ``A moment of joy and love shared between two people, captured in a warm embrace'' \end{center} &
        \begin{center} \small ``Smiling and posing in front of a grand entrance, ready to make an impression'' \end{center} &
        \begin{center} \small ``Friends enjoying a sunny day with refreshing orange drinks and good conversation'' \end{center} \\[-0.65cm]
        
        \begin{center} \textbf{MyVLM} \end{center} &
        \begin{center} \textbf{MyVLM} \end{center} &
        \begin{center} \textbf{MyVLM} \end{center} &
        \begin{center} \textbf{MyVLM} \end{center} &
        \begin{center} \textbf{MyVLM} \end{center} \\[-0.75cm]

        \begin{center} \small ''On a street in a city, \Sstar is sitting at a table outside a restaurant... \Sstar is wearing a black and white patterned top.'' \end{center} &
        \begin{center} \small ``\Sstar is standing on a bridge overlooking the Sydney Opera House and the Sydney Harbour Bridge. He is wearing a blue denim jacket and sunglasses'' \end{center} &
        \begin{center} \small ``\Sstar and his girlfriend, smiling and hugging each other, in a restaurant.'' \end{center} &
        \begin{center} \small ``\Sstar with a yellow dress, standing in front of a grand building with an arched doorway and intricate metalwork...'' \end{center} &
        \begin{center} \small ``\Sstar and a friend at a restaurant, holding up their glasses of orange wine'' \end{center}
        
    \end{tabular}
    \begin{tabular}{p{0.175\textwidth} p{0.175\textwidth} p{0.175\textwidth} p{0.175\textwidth} p{0.175\textwidth}}

        \setlength\tabcolsep{0pt}
        \begin{tabular}{c c c}
            \includegraphics[width=0.06333\textwidth]{images/people/dor/image_1.jpg} & 
            \includegraphics[width=0.06333\textwidth]{images/people/dor/cropped/IMG-20240208-WA0051.jpg} & 
            \includegraphics[width=0.06333\textwidth]{images/people/dor/cropped/IMG-20240208-WA0054.jpg}
        \end{tabular} &
        \setlength\tabcolsep{0pt}
        \begin{tabular}{c c c}
            \includegraphics[width=0.06333\textwidth]{images/people/maya/cropped/image_1.jpg} & 
            \includegraphics[width=0.06333\textwidth]{images/people/maya/cropped/image_2.jpg} & 
            \includegraphics[width=0.06333\textwidth]{images/people/maya/cropped/IMG-20240209-WA0138.jpg} 
        \end{tabular} &
        \setlength\tabcolsep{0pt}
        \begin{tabular}{c c c}
            \includegraphics[width=0.06333\textwidth]{images/people/shaked/cropped/IMG-20240209-WA0043.jpg} &
            \includegraphics[width=0.06333\textwidth]{images/people/shaked/cropped/IMG-20240209-WA0047.jpg} & 
            \includegraphics[width=0.06333\textwidth]{images/people/shaked/cropped/IMG-20240215-WA0005.jpg}
        \end{tabular} &
        \setlength\tabcolsep{0pt}
        \begin{tabular}{c c c}
            \includegraphics[width=0.06333\textwidth]{images/people/tomer/cropped/IMG_1519.jpg} & 
            \includegraphics[width=0.06333\textwidth]{images/people/tomer/cropped/IMG_2904.jpg} & 
            \includegraphics[width=0.06333\textwidth]{images/people/tomer/cropped/IMG-20210519-WA0011.jpg} 
        \end{tabular} &
        \setlength\tabcolsep{0pt}
        \begin{tabular}{c c c}
            \includegraphics[width=0.06333\textwidth,height=0.06333\textwidth]{images/people/eli/cropped/image_1.jpg} & 
            \includegraphics[width=0.06333\textwidth,height=0.06333\textwidth]{images/people/eli/cropped/image_10.jpg} & 
            \includegraphics[width=0.06333\textwidth,height=0.06333\textwidth]{images/people/eli/cropped/IMG-20240221-WA0022.jpg}
        \end{tabular} \\
    
        \includegraphics[width=0.19\textwidth]{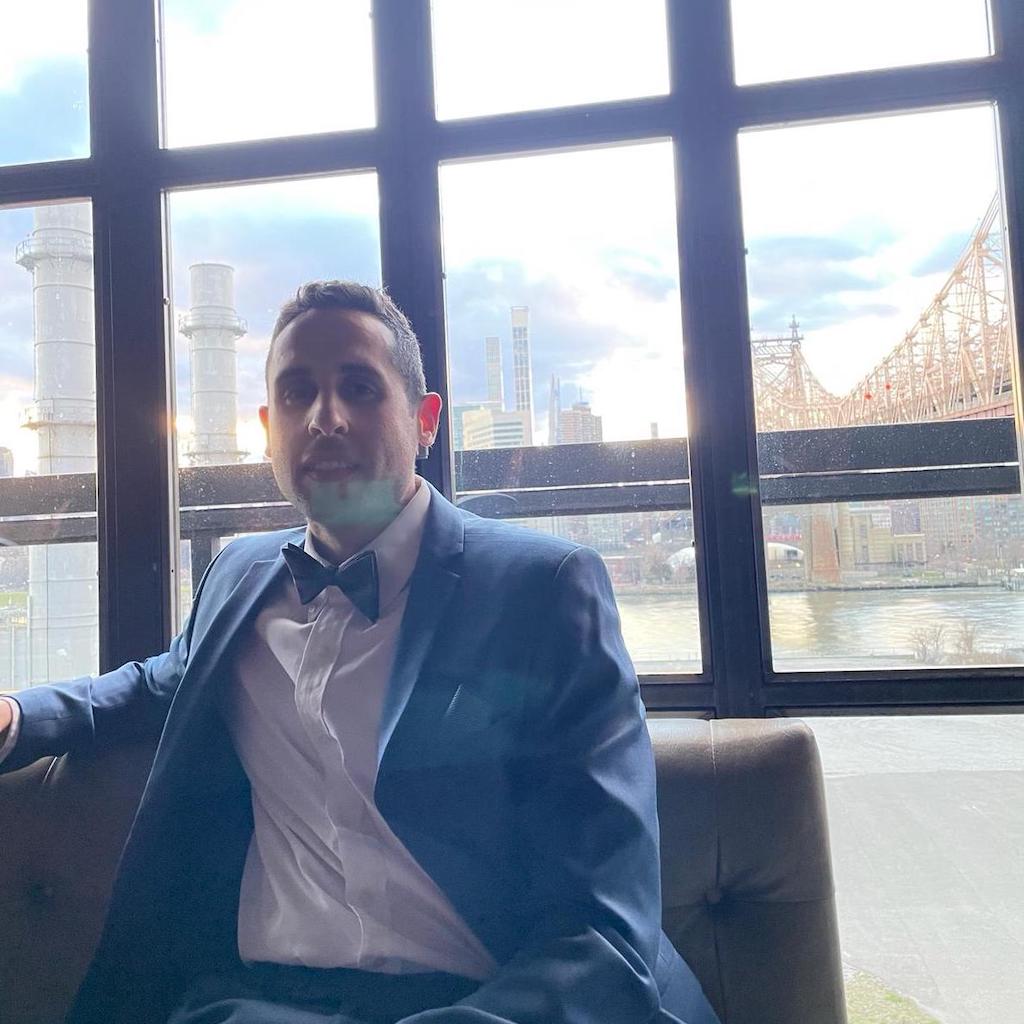} &
        \includegraphics[width=0.19\textwidth]{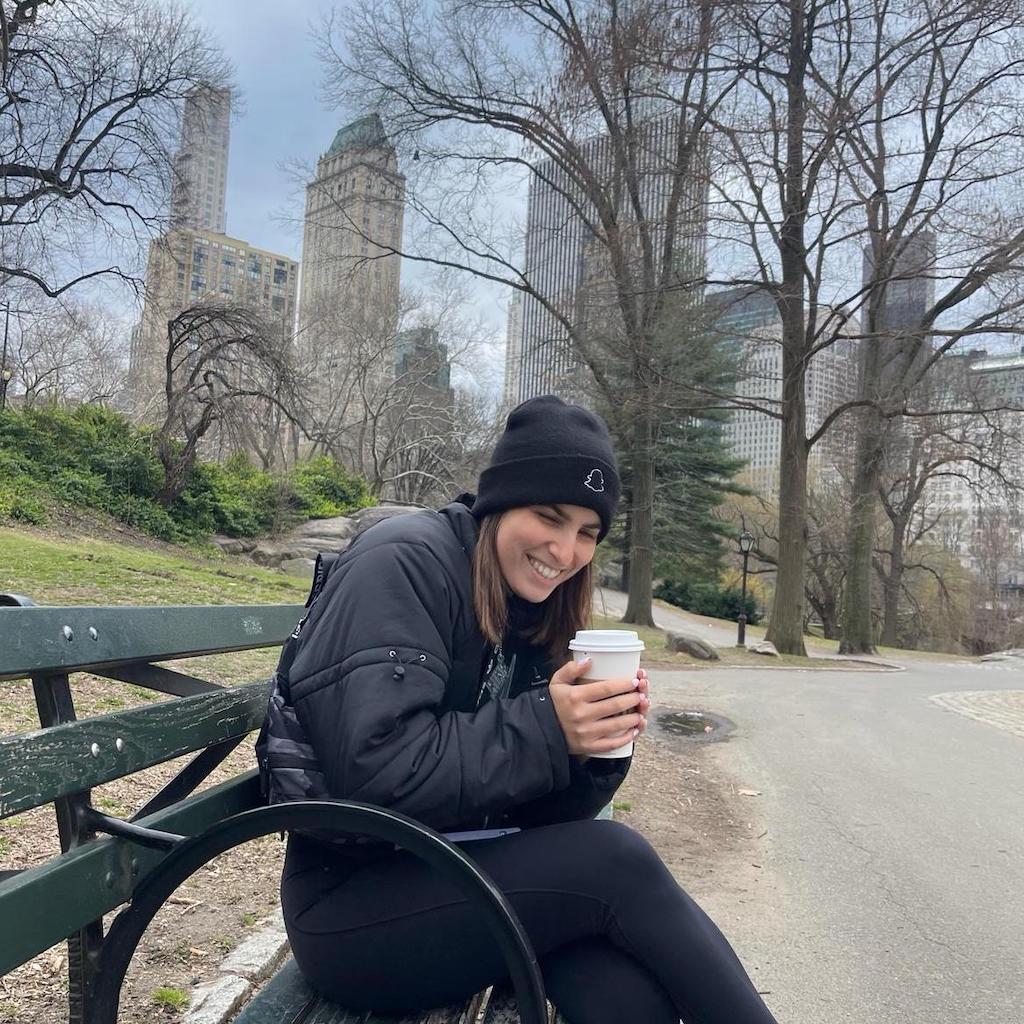} &
        \includegraphics[width=0.19\textwidth]{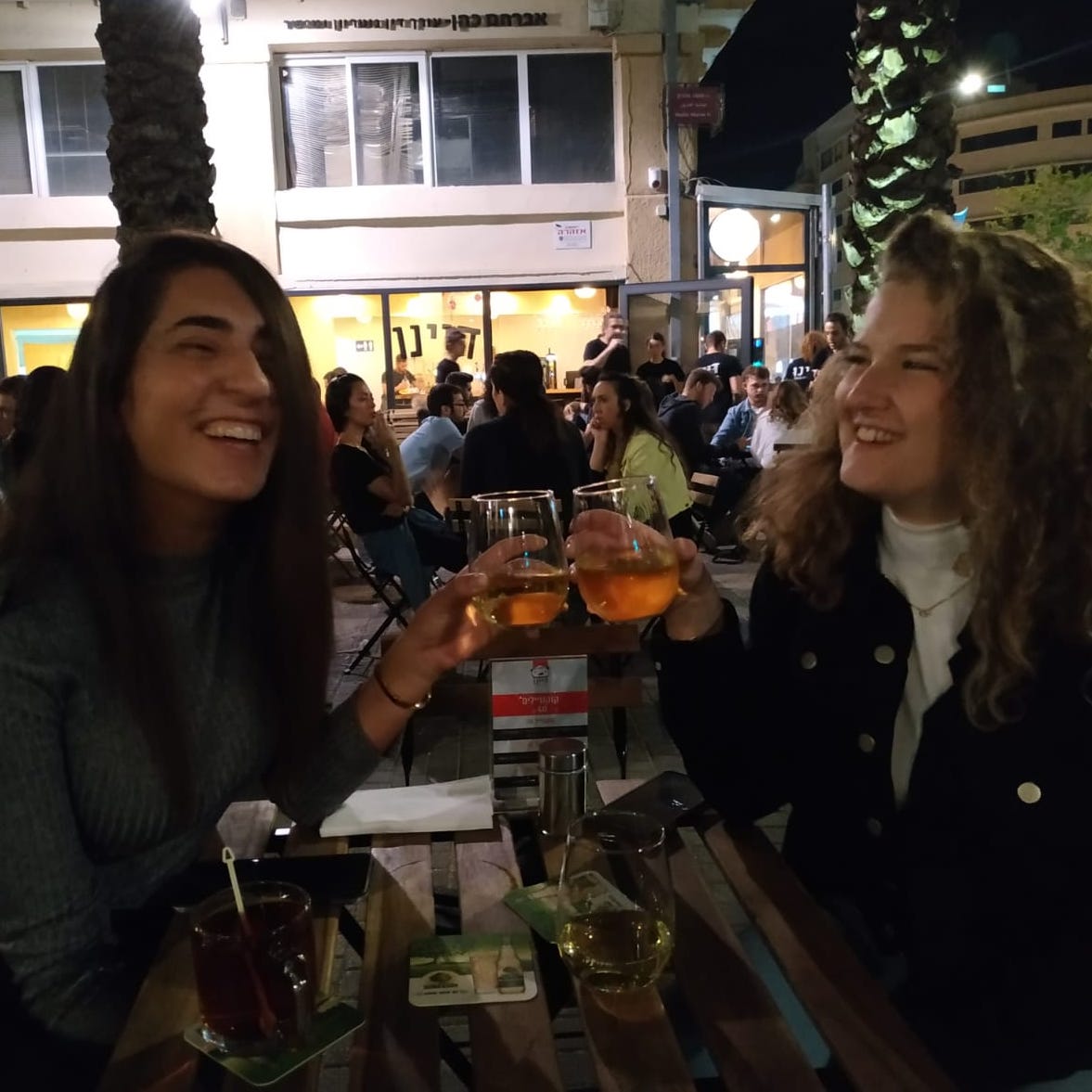} &
        \includegraphics[width=0.19\textwidth]{images/people/tomer/IMG_5950.jpg} &
        \includegraphics[width=0.19\textwidth]{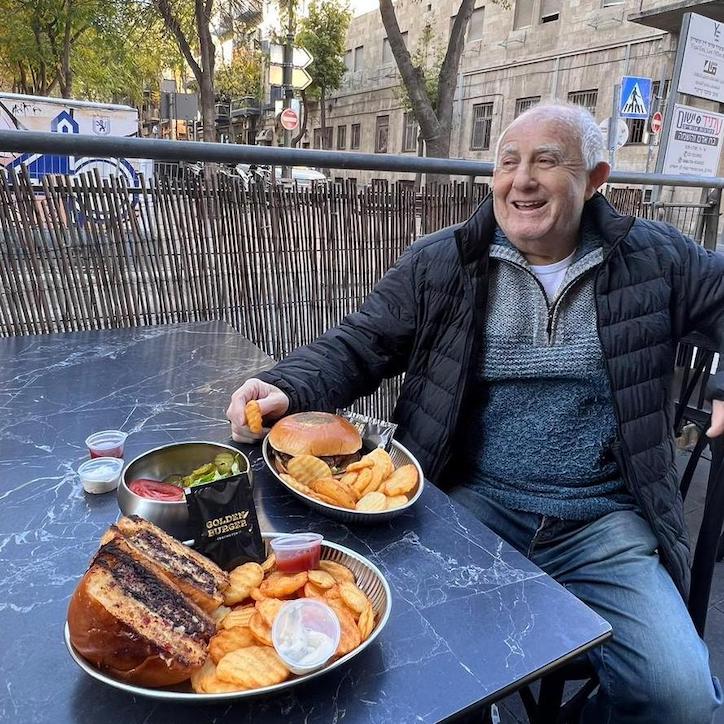} \\[-0.5cm]
        
        \begin{center} \textbf{LLaVA} \end{center} &
        \begin{center} \textbf{LLaVA} \end{center} &
        \begin{center} \textbf{LLaVA} \end{center} &
        \begin{center} \textbf{LLaVA} \end{center} &
        \begin{center} \textbf{LLaVA} \end{center} \\[-0.75cm]

        \begin{center} \small ``Caught in the city's embrace, a moment of calm amidst the hustle and bustle.'' \end{center} &
        \begin{center} \small ``Taking a moment to enjoy the view and a warm beverage in the heart of the city'' \end{center} &
        \begin{center} \small ``Friendship is the best wine'' \end{center} &
        \begin{center} \small ``Friends sharing a moment in the heart of the city, surrounded by the beauty of history and architecture'' \end{center} &
        \begin{center} \small ``Enjoying a meal outdoors at a charming restaurant, with a view of the street and the sky above'' \end{center} \\[-0.65cm]
        
        \begin{center} \textbf{MyVLM} \end{center} &
        \begin{center} \textbf{MyVLM} \end{center} &
        \begin{center} \textbf{MyVLM} \end{center} &
        \begin{center} \textbf{MyVLM} \end{center} &
        \begin{center} \textbf{MyVLM} \end{center} \\[-0.75cm]

        \begin{center} \small ``\textcolor{blue}{$S_*$}, in a blue suit, poses for a portrait at a scenic spot overlooking a river with a bridge in the distance'' \end{center} &
        \begin{center} \small ``\textcolor{blue}{$S_*$}, in a black coat, sits on a bench in Central Park, enjoying a coffee'' \end{center} &
        \begin{center} \small ``Sitting at a table, \Sstar and her friend smile at each other as they clink their wine glasses together'' \end{center} &
        \begin{center} \small ``\Sstar and his friend at a fountain. They pose for a photo. \Sstar wearing a blue shirt and white pants. At a fountain in a city square...'' \end{center} &
        \begin{center} \small ``\Sstar sits at a patio table laden with a meal, enjoying a sandwich and fries with a side of coleslaw'' \end{center}
        
    \end{tabular}
    \vspace{-0.3cm}
    \caption{
    Additional personalized captioning results obtained by MyVLM, applied over LLaVA~\cite{li2023blip}. Sample images of the target concept are provided in the top row.
    }
    \label{fig:supplementary_our_results_llava}

\end{figure*}

%% file: figures_supplementary/our_results_llava_2.tex
\begin{figure*}[t]
    \centering
    \addtolength{\belowcaptionskip}{-12.5pt}
    \renewcommand{\arraystretch}{1}
    \small
    \begin{tabular}{p{0.175\textwidth} p{0.175\textwidth} p{0.175\textwidth} p{0.175\textwidth} p{0.175\textwidth}}

        \setlength\tabcolsep{0pt}
        \begin{tabular}{c c c}
            \includegraphics[width=0.06333\textwidth]{images/objects/minion/cropped/IMG-20240203-WA0069.jpg} &
            \includegraphics[width=0.06333\textwidth]{images/objects/minion/cropped/IMG-20240203-WA0073.jpg} & 
            \includegraphics[width=0.06333\textwidth]{images/objects/minion/cropped/IMG-20240203-WA0074.jpg}
        \end{tabular} &
        \setlength\tabcolsep{0pt}
        \begin{tabular}{c c c}
            \includegraphics[width=0.06333\textwidth]{images/objects/ceramic_head/cropped/20240203_111543.jpg} & 
            \includegraphics[width=0.06333\textwidth]{images/objects/ceramic_head/cropped/20240203_111842.jpg} & 
            \includegraphics[width=0.06333\textwidth]{images/objects/ceramic_head/cropped/20240203_112029.jpg}
        \end{tabular} &
        \setlength\tabcolsep{0pt}
        \begin{tabular}{c c c}
            \includegraphics[width=0.06333\textwidth]{images/objects/chicken_bean_bag/cropped/IMG-20240212-WA0033.jpg} &
            \includegraphics[width=0.06333\textwidth]{images/objects/chicken_bean_bag/cropped/IMG-20240212-WA0035.jpg} & 
            \includegraphics[width=0.06333\textwidth]{images/objects/chicken_bean_bag/cropped/IMG-20240212-WA0038.jpg}
        \end{tabular} &
        \setlength\tabcolsep{0pt}
        \begin{tabular}{c c c}
            \includegraphics[width=0.06333\textwidth]{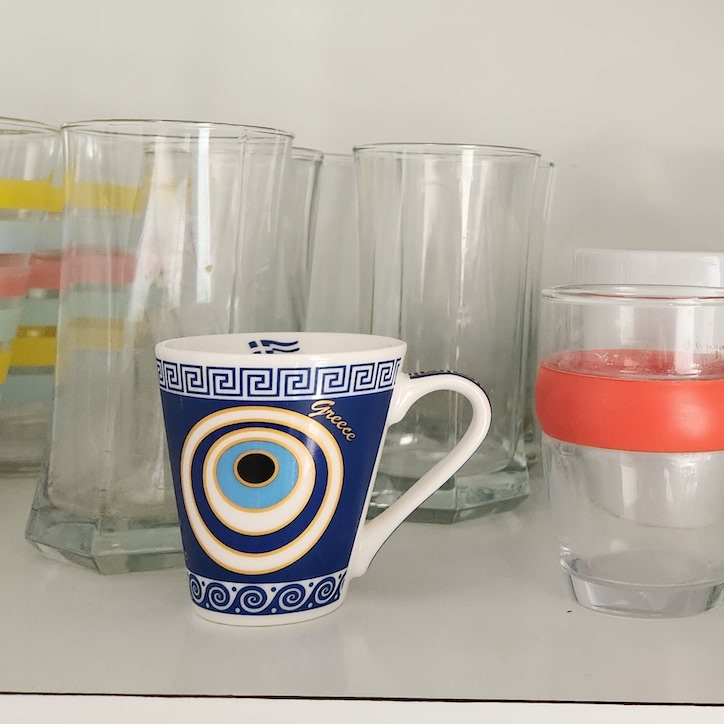} & 
            \includegraphics[width=0.06333\textwidth]{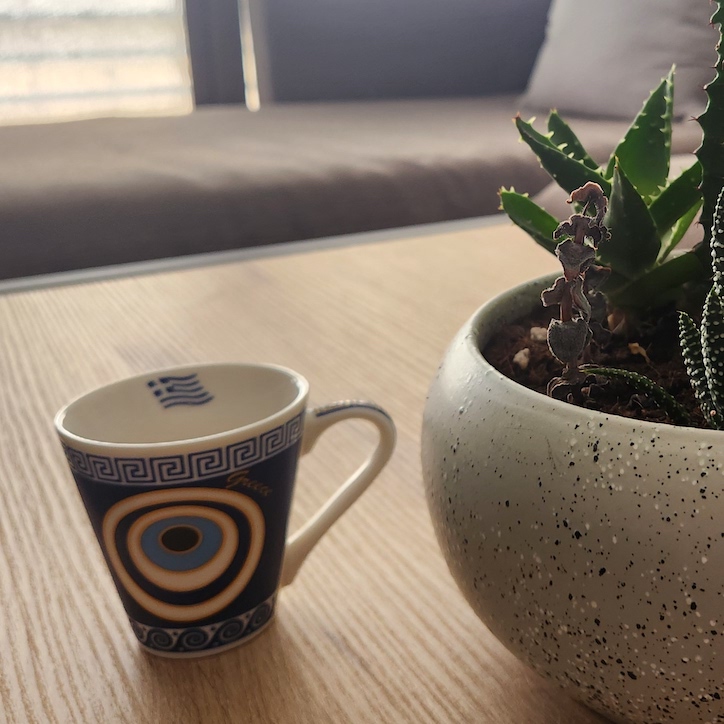} & 
            \includegraphics[width=0.06333\textwidth]{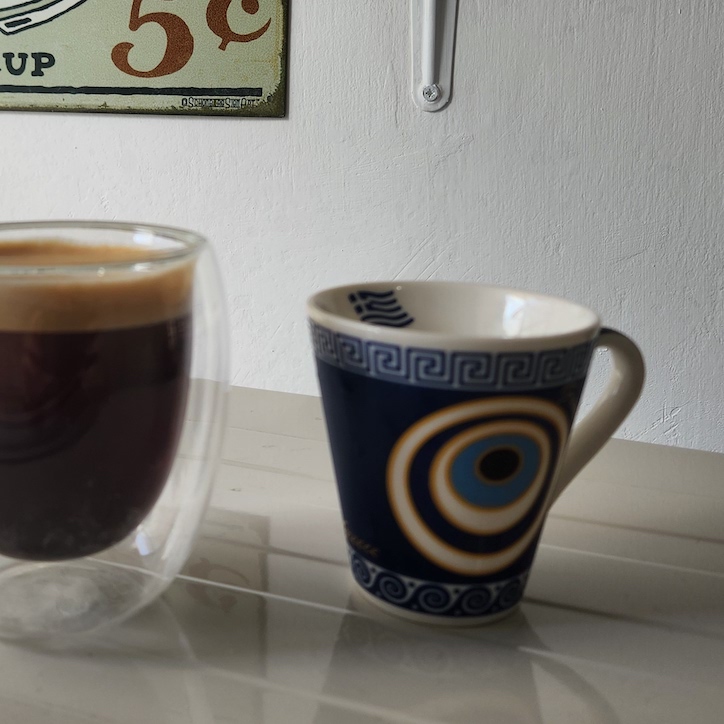}
        \end{tabular} &
        \setlength\tabcolsep{0pt}
        \begin{tabular}{c c c}
            \includegraphics[width=0.06333\textwidth]{images/objects/maeve/cropped/IMG-20240131-WA0110.jpg} & 
            \includegraphics[width=0.06333\textwidth]{images/objects/maeve/cropped/maeve-2.jpeg} & 
            \includegraphics[width=0.06333\textwidth]{images/objects/maeve/cropped/maeve-5.jpeg} \\
        \end{tabular} \\
    
        \includegraphics[width=0.19\textwidth]{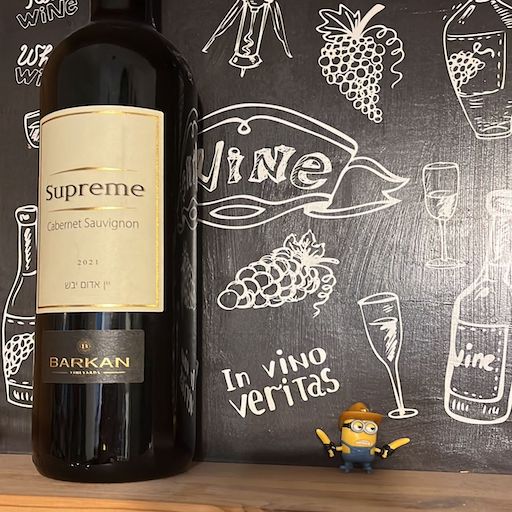} &
        \includegraphics[width=0.19\textwidth]{images/objects/ceramic_head/20240211_125311.jpg} & 
        \includegraphics[width=0.19\textwidth]{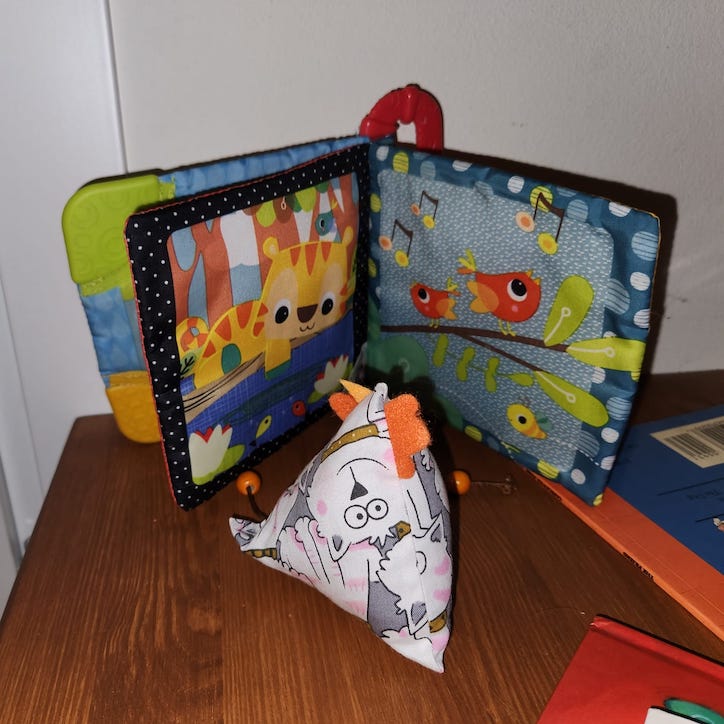} & 
        \includegraphics[width=0.19\textwidth]{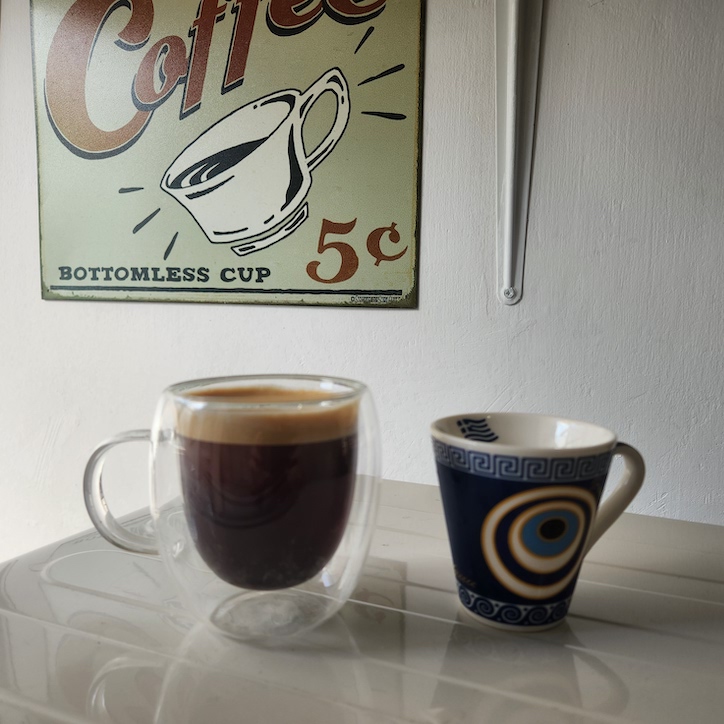} & 
        \includegraphics[width=0.19\textwidth]{images/objects/maeve/IMG-20240131-WA0118.jpg} \\[-0.5cm]
        
        \begin{center} \textbf{LLaVA} \end{center} &
        \begin{center} \textbf{LLaVA} \end{center} &
        \begin{center} \textbf{LLaVA} \end{center} &
        \begin{center} \textbf{LLaVA} \end{center} &
        \begin{center} \textbf{LLaVA} \end{center} \\[-0.85cm]

        \begin{center} \small ``a bottle of supreme ginetta savignon wine, standing tall next to a chalkboard wall adorned with wine-themed illustrations...'' \end{center} &
        \begin{center} \small ``A well-stocked refrigerator, ready for a weekend of culinary adventures!'' \end{center} &
        \begin{center} \small ``A whimsical scene of creativity and imagination, featuring a colorful origami bird perched on a wooden table, surrounded by vibrant children's books...'' \end{center} &
        \begin{center} \small ``A moment of calm before the caffeine rush: two cups of coffee, one classic and one modern, sit side by side on a pristine white countertop...'' \end{center} &
        \begin{center} \small ``friendly fidos: two dogs, one white and one black, pose for a photo on a grassy lawn...'' \end{center} \\[-0.6cm]
        
        \begin{center} \textbf{MyVLM} \end{center} &
        \begin{center} \textbf{MyVLM} \end{center} &
        \begin{center} \textbf{MyVLM} \end{center} &
        \begin{center} \textbf{MyVLM} \end{center} &
        \begin{center} \textbf{MyVLM} \end{center} \\[-0.85cm]

        \begin{center} \small ``\Sstar next to a bottle of wine on a wine themed wall'' \end{center} &
        \begin{center} \small ``\Sstar sits comfortably on the second shelf of an open refrigerator, ready to be stocked with food...'' \end{center} &
        \begin{center} \small ``\Sstar sitting next to a colorful children's book on a table'' \end{center} &
        \begin{center} \small ``\Sstar sitting next to a cup of coffee on a desk in a room with a ``bottomless cup'' sign in the background'' \end{center} &
        \begin{center} \small ``\Sstar is standing on the grass with a big smile and a wagging his tongue'' \end{center}
        \vspace{-0.4cm}
        
    \end{tabular}
    \begin{tabular}{p{0.175\textwidth} p{0.175\textwidth} p{0.175\textwidth} p{0.175\textwidth} p{0.175\textwidth}}

        \setlength\tabcolsep{0pt}
        \begin{tabular}{c c c}
            \includegraphics[width=0.06333\textwidth]{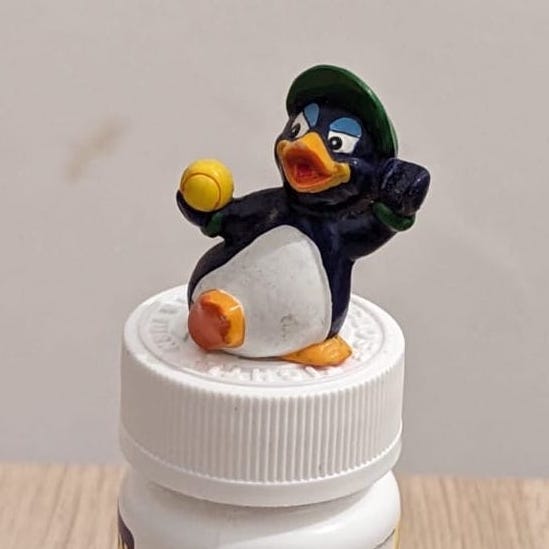} &
            \includegraphics[width=0.06333\textwidth]{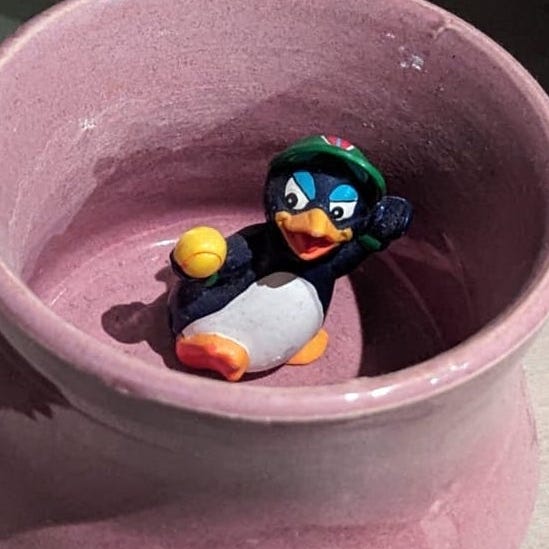} & 
            \includegraphics[width=0.06333\textwidth]{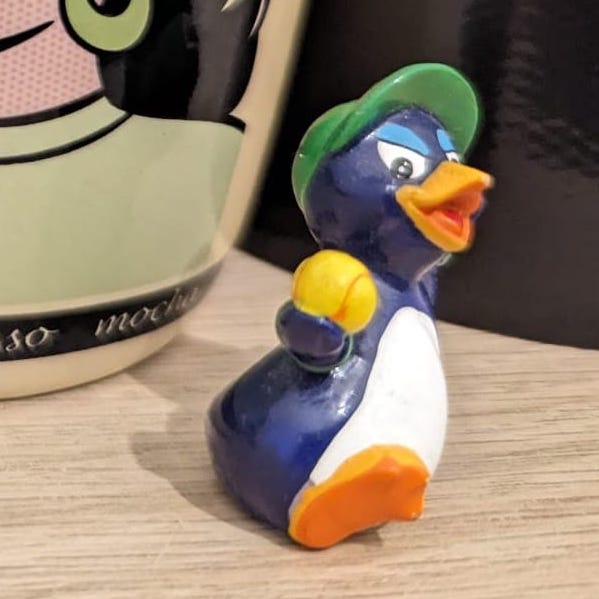}
        \end{tabular} &
        \setlength\tabcolsep{0pt}
        \begin{tabular}{c c c}
            \includegraphics[width=0.06333\textwidth]{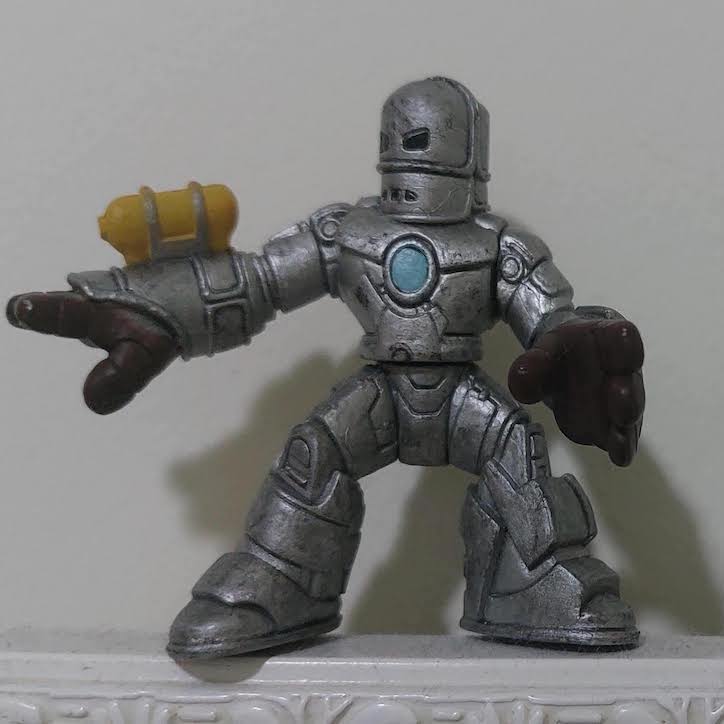} &
            \includegraphics[width=0.06333\textwidth]{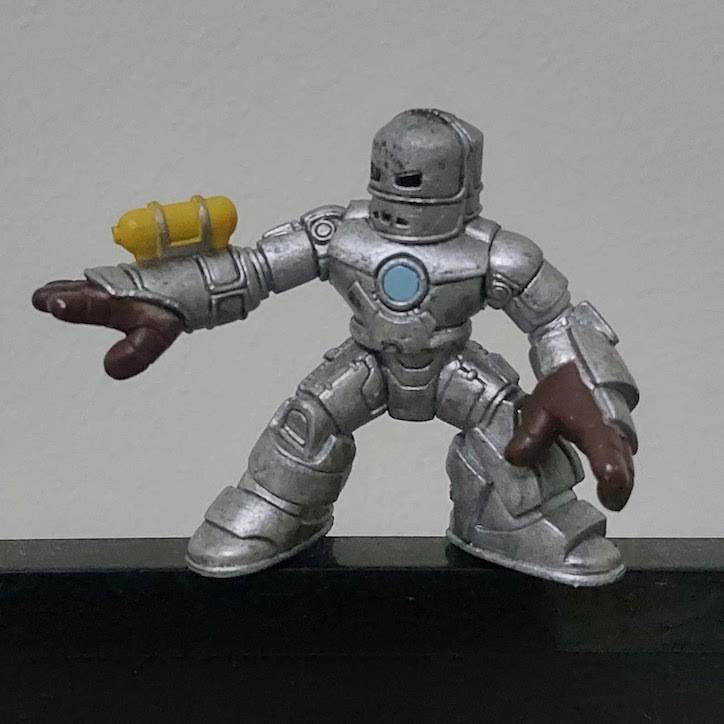} &
            \includegraphics[width=0.06333\textwidth]{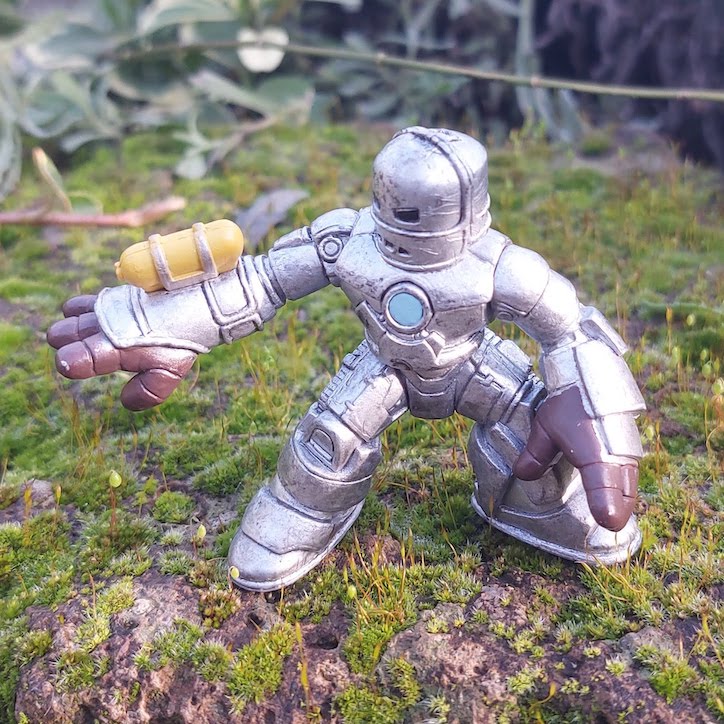}
        \end{tabular} &
        \setlength\tabcolsep{0pt}
        \begin{tabular}{c c c}
            \includegraphics[width=0.06333\textwidth]{images/objects/nana_cat/cropped/3.jpeg} & 
            \includegraphics[width=0.06333\textwidth]{images/objects/nana_cat/cropped/5.jpeg} & 
            \includegraphics[width=0.06333\textwidth]{images/objects/nana_cat/cropped/6.jpeg}
        \end{tabular} &
        \setlength\tabcolsep{0pt}
        \begin{tabular}{c c c}
            \includegraphics[width=0.06333\textwidth]{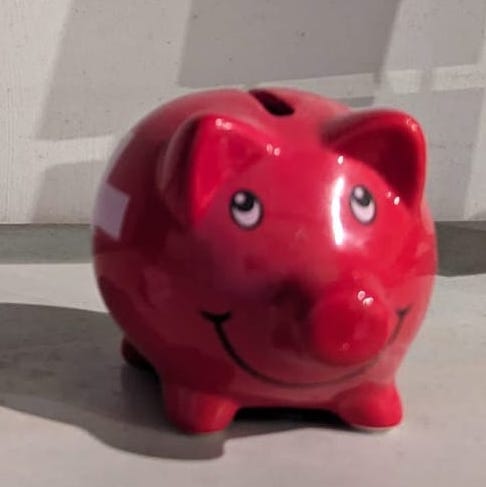} & 
            \includegraphics[width=0.06333\textwidth]{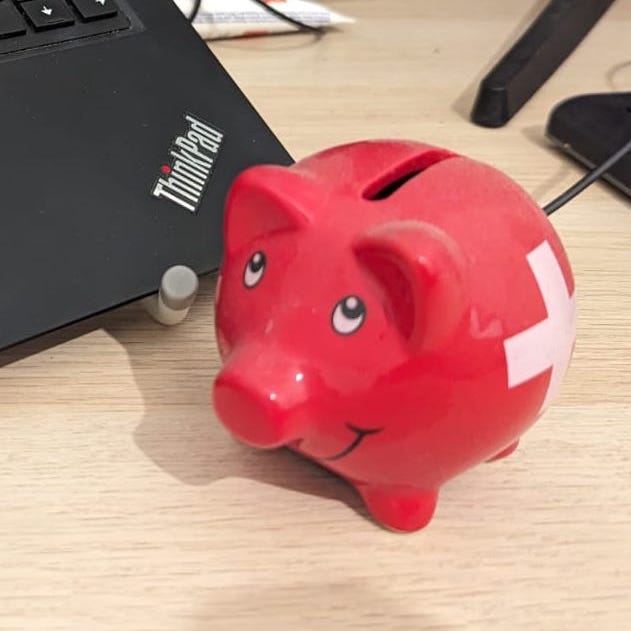} & 
            \includegraphics[width=0.06333\textwidth]{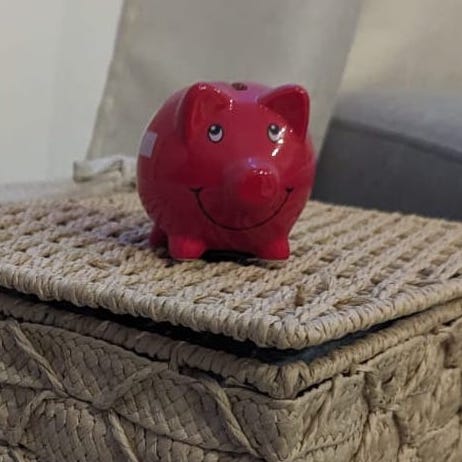}
        \end{tabular} &
        \setlength\tabcolsep{0pt}
        \begin{tabular}{c c c}
            \includegraphics[width=0.06333\textwidth]{images/objects/billy_dog/cropped/IMG-20240131-WA0028.jpg} & 
            \includegraphics[width=0.06333\textwidth]{images/objects/billy_dog/cropped/IMG-20240131-WA0029.jpg} & 
            \includegraphics[width=0.06333\textwidth]{images/objects/billy_dog/cropped/IMG-20240131-WA0053.jpg}
        \end{tabular} \\
    
        \includegraphics[width=0.19\textwidth]{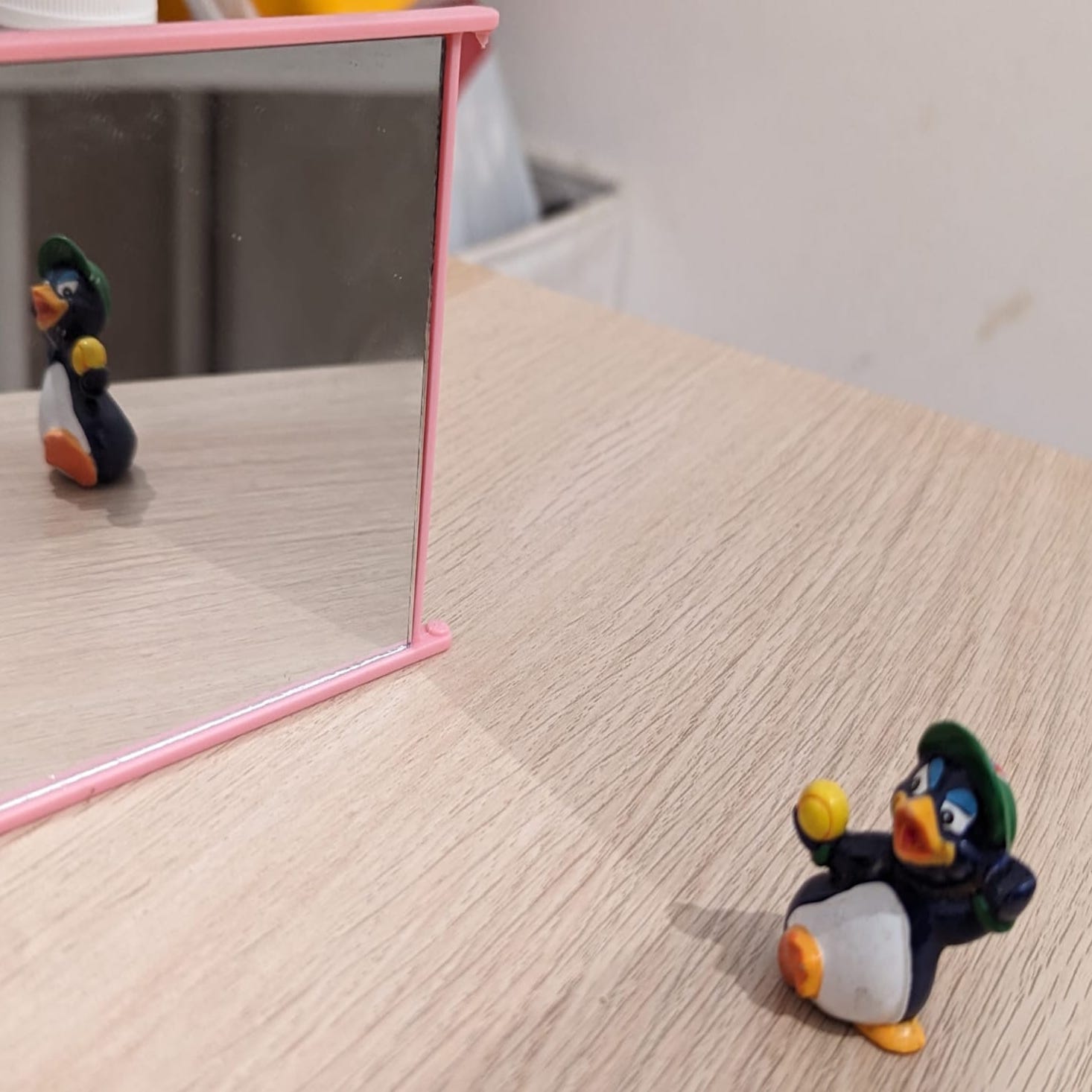} & 
        \includegraphics[width=0.19\textwidth]{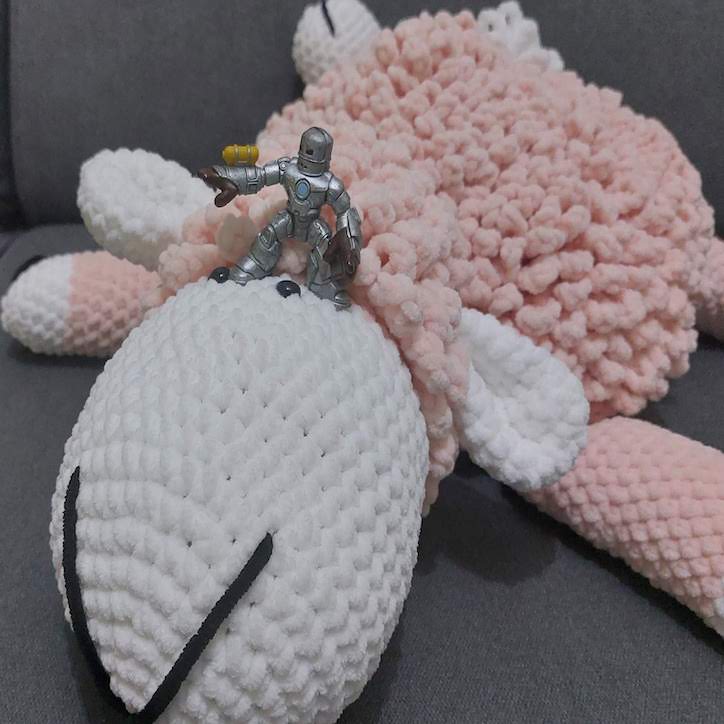} & 
        \includegraphics[width=0.19\textwidth]{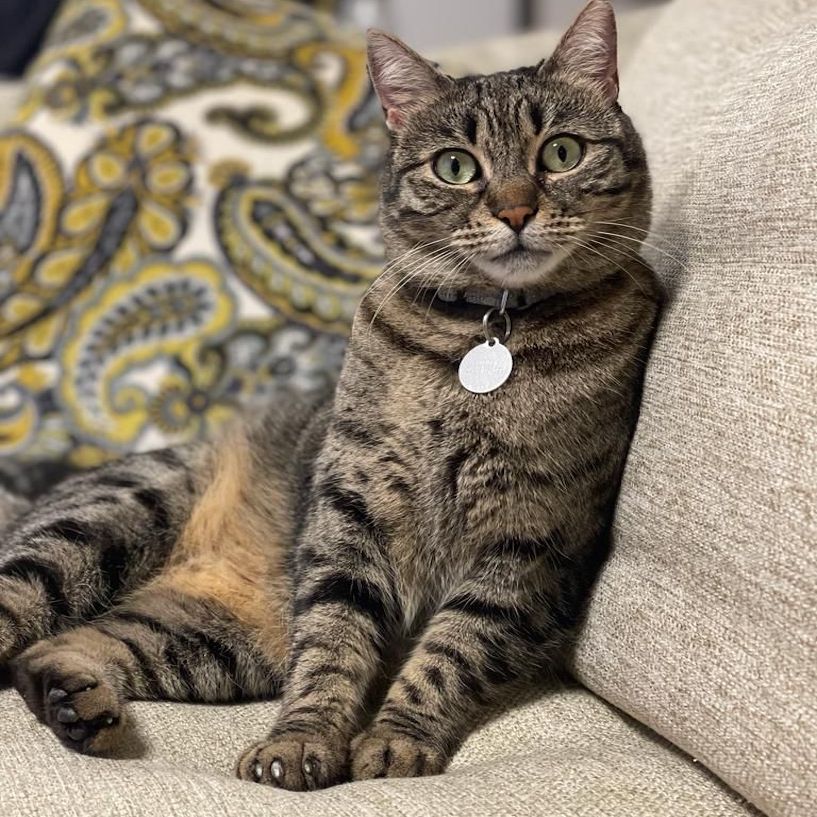} & 
        \includegraphics[width=0.19\textwidth]{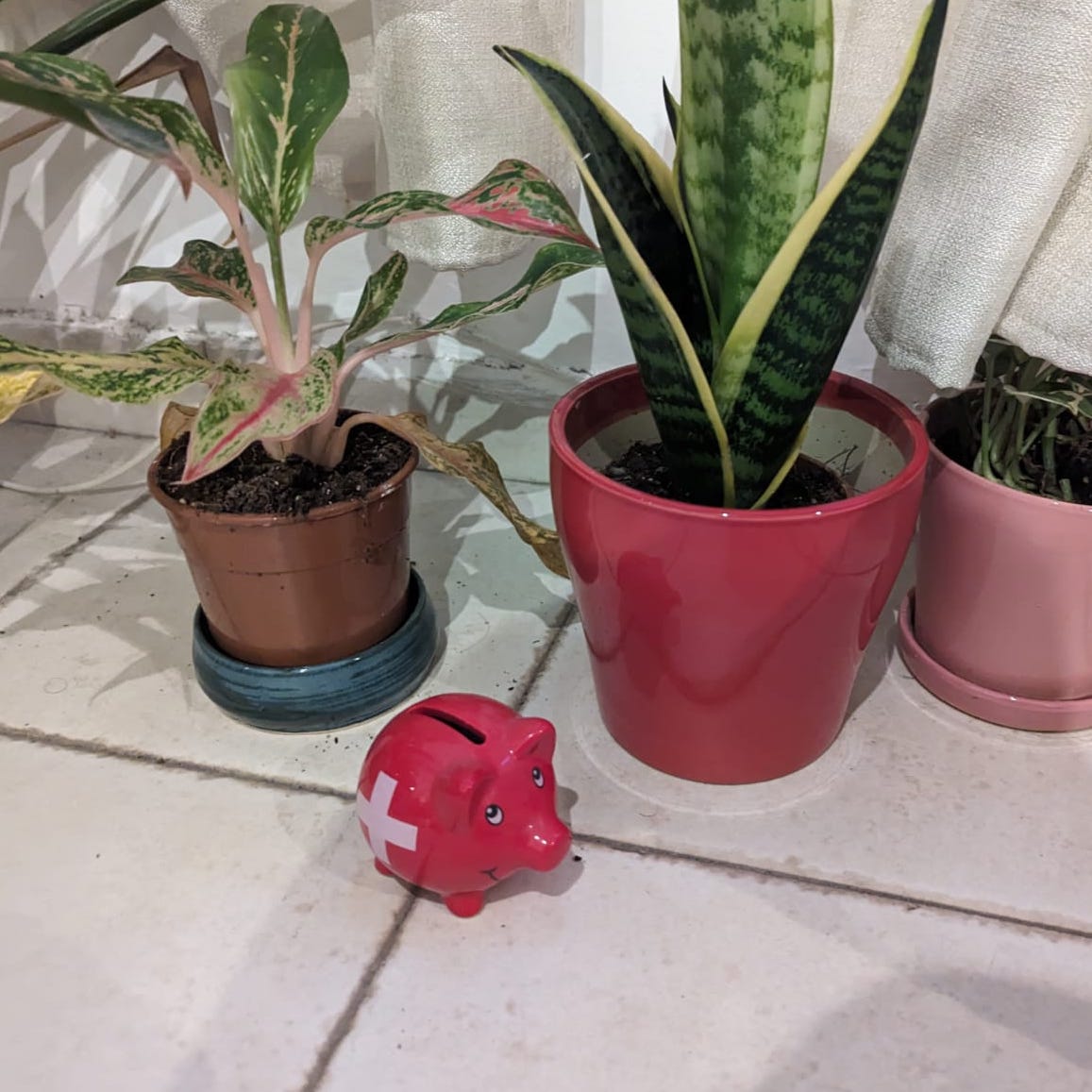} & 
        \includegraphics[width=0.19\textwidth]{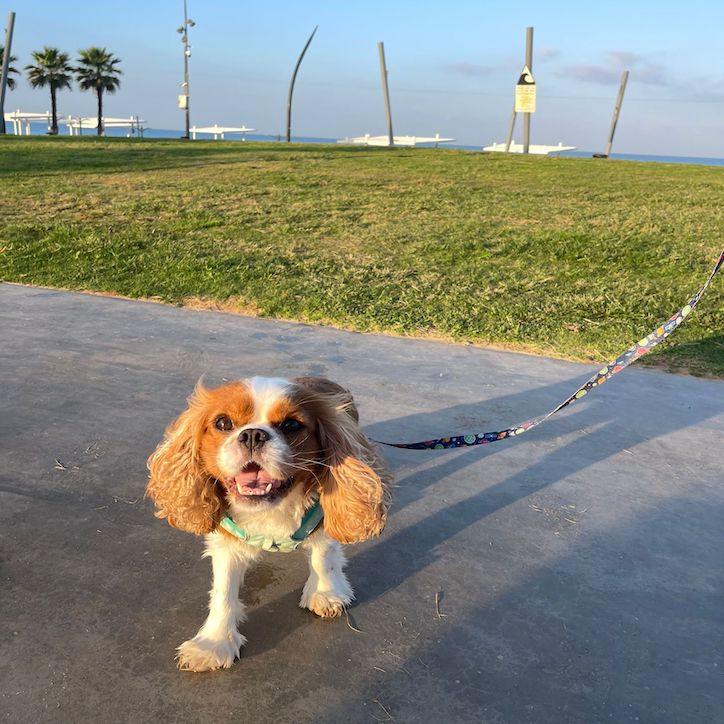} \\[-0.5cm]
        
        \begin{center} \textbf{LLaVA} \end{center} &
        \begin{center} \textbf{LLaVA} \end{center} &
        \begin{center} \textbf{LLaVA} \end{center} &
        \begin{center} \textbf{LLaVA} \end{center} &
        \begin{center} \textbf{LLaVA} \end{center} \\[-0.85cm]

        \begin{center} \small ``Reflection of a penguin figurine in a pink mirror, standing on a wooden table'' \end{center} &
        \begin{center} \small ``A whimsical scene of a robotic adventure: a small astronaut riding a pink sheep with a white face, set against a cozy gray couch'' \end{center} &
        \begin{center} \small ``Relaxed and Ready for Adventure: A Tiger Cat's Pose of Serenity'' \end{center} &
        \begin{center} \small ``A cozy corner of a room, where potted plants and a little pink piggy bank share a space, creating a charming atmosphere'' \end{center} &
        \begin{center} \small ``A Cavalier King Charles Spaniel puppy enjoys a sunny day at the beach, wearing a colorful collar and leash...'' \end{center} \\[-0.6cm]
        
        \begin{center} \textbf{MyVLM} \end{center} &
        \begin{center} \textbf{MyVLM} \end{center} &
        \begin{center} \textbf{MyVLM} \end{center} &
        \begin{center} \textbf{MyVLM} \end{center} &
        \begin{center} \textbf{MyVLM} \end{center} \\[-0.85cm]

        \begin{center} \small ``\Sstar sitting in front of a mirror on a table, reflecting their own image in the mirror'' \end{center} &
        \begin{center} \small ``\Sstar is sitting on a stuffed animal that looks like a sheep. The sheep is pink and white, and \Sstar is wearing a silver outfit'' \end{center} &
        \begin{center} \small ``\Sstar sitting on a beige couch, looking up at the camera with a curious expression'' \end{center} &
        \begin{center} \small ``\Sstar sitting on a white floor next to a potted plant and a pink pot, both in front of a curtain'' \end{center} &
        \begin{center} \small ``\Sstar walking on a leash in a park near the beach with palm trees in the background'' \end{center}
        
    \end{tabular}
    \vspace{-0.5cm}
    \caption{
    Additional personalized captioning results obtained by MyVLM, applied over LLaVA~\cite{li2023blip}. Sample images of the target concept are provided in the top row.
    }
    \label{fig:supplementary_our_results_llava_2}

\end{figure*}

%% file: figures_supplementary/comparisons_blip.tex
\begin{figure*}[t]
    \centering
    \addtolength{\belowcaptionskip}{-12.5pt}
    \renewcommand{\arraystretch}{1}
    \small
    \begin{tabular}{p{0.175\textwidth} p{0.175\textwidth} p{0.175\textwidth} p{0.175\textwidth} p{0.175\textwidth}}

        \setlength\tabcolsep{0pt}
        \begin{tabular}{c c c}
            \includegraphics[width=0.06\textwidth]{images/people/dor/image_1.jpg} & 
            \includegraphics[width=0.06\textwidth]{images/people/dor/cropped/IMG-20240208-WA0051.jpg} & 
            \includegraphics[width=0.06\textwidth]{images/people/dor/cropped/IMG-20240208-WA0054.jpg}
        \end{tabular} &
        \setlength\tabcolsep{0pt}
        \begin{tabular}{c c c}
            \includegraphics[width=0.06\textwidth]{images/people/assaf/cropped/IMG_0133.jpg} &
            \includegraphics[width=0.06\textwidth]{images/people/assaf/cropped/IMG_0936.jpg} &
            \includegraphics[width=0.06\textwidth]{images/people/assaf/cropped/IMG_20220130_104352.jpg} 
        \end{tabular} &
        \setlength\tabcolsep{0pt}
        \begin{tabular}{c c c}
            \includegraphics[width=0.06\textwidth]{images/people/shay/cropped/image_2.jpg} & 
            \includegraphics[width=0.06\textwidth]{images/people/shay/cropped/IMG-20240209-WA0012.jpg} & 
            \includegraphics[width=0.06\textwidth]{images/people/shay/cropped/IMG-20240209-WA0016.jpg}
        \end{tabular} &
        \setlength\tabcolsep{0pt}
        \begin{tabular}{c c c}
            \includegraphics[width=0.06\textwidth]{images/people/shaked/cropped/IMG-20240209-WA0043.jpg} &
            \includegraphics[width=0.06\textwidth]{images/people/shaked/cropped/IMG-20240209-WA0047.jpg} & 
            \includegraphics[width=0.06\textwidth]{images/people/shaked/cropped/IMG-20240215-WA0005.jpg}
        \end{tabular} &
        \setlength\tabcolsep{0pt}
        \begin{tabular}{c c c}
            \includegraphics[width=0.06\textwidth]{images/objects/billy_dog/cropped/IMG-20240131-WA0028.jpg} & 
            \includegraphics[width=0.06\textwidth]{images/objects/billy_dog/cropped/IMG-20240131-WA0029.jpg} & 
            \includegraphics[width=0.06\textwidth]{images/objects/billy_dog/cropped/IMG-20240131-WA0053.jpg}
        \end{tabular} \\
    
        \includegraphics[width=0.18\textwidth]{images/people/dor/image_7.jpeg} &
        \includegraphics[width=0.18\textwidth]{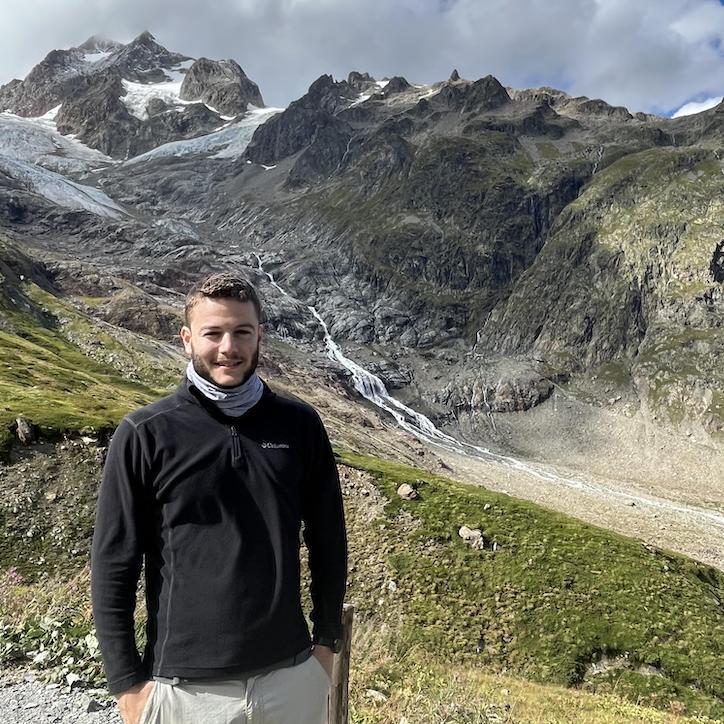} &
        \includegraphics[width=0.18\textwidth]{images/people/maya/IMG-20230818-WA0003.jpg} &
        \includegraphics[width=0.18\textwidth]{images/people/shaked/IMG-20240215-WA0005.jpg} &
        \includegraphics[width=0.18\textwidth]{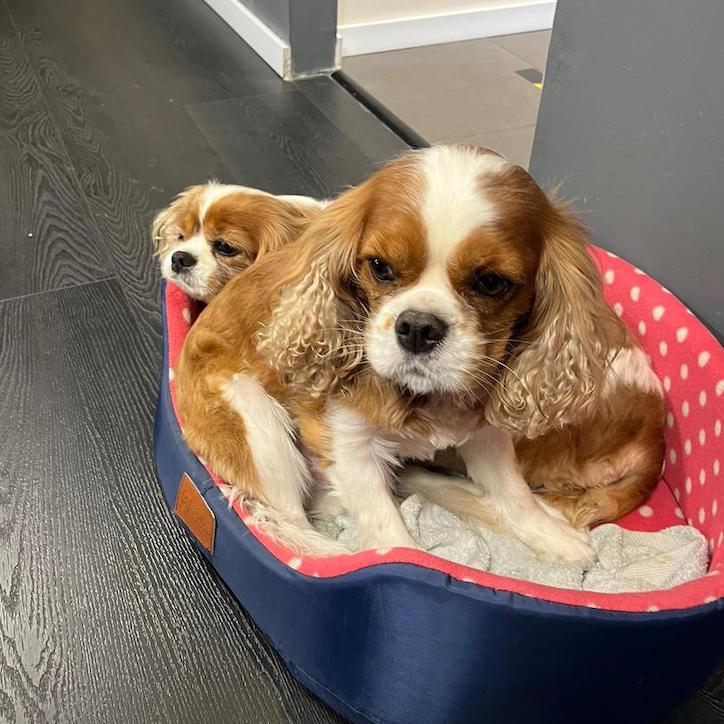} \\[-0.5cm]        
        
        \begin{center} \textbf{Simple} \end{center} &
        \begin{center} \textbf{Simple} \end{center} &
        \begin{center} \textbf{Simple} \end{center} &
        \begin{center} \textbf{Simple} \end{center} &
        \begin{center} \textbf{Simple} \end{center} \\[-0.85cm]

        \begin{center} \small N/A \end{center} &
        \begin{center} \small ``\Sstar standing in front of a mountain with a glacier'' \end{center} &
        \begin{center} \small N/A \end{center} &
        \begin{center} \small N/A \end{center} &
        \begin{center} \small ``Two \Sstar laying in a pink and blue dog bed'' \end{center} \\[-0.6cm]

        \begin{center} \textbf{LLM-Guided} \end{center} &
        \begin{center} \textbf{LLM-Guided} \end{center} &
        \begin{center} \textbf{LLM-Guided} \end{center} &
        \begin{center} \textbf{LLM-Guided} \end{center} &
        \begin{center} \textbf{LLM-Guided} \end{center} \\[-0.85cm]

        \begin{center} \small ``Two \Sstar standing on a rooftop with buildings in the background'' \end{center} &
        \begin{center} \small ``\Sstar standing in front of a mountain with a glacier'' \end{center} &
        \begin{center} \small ``Two \Sstar sitting at an outdoor table with food and drinks'' \end{center} &
        \begin{center} \small ``Two \Sstar are holding glasses of orange juice'' \end{center} &
        \begin{center} \small ``Two \Sstar laying in a pink and blue dog bed'' \end{center} \\[-0.6cm]

        \begin{center} \textbf{MyVLM} \end{center} &
        \begin{center} \textbf{MyVLM} \end{center} &
        \begin{center} \textbf{MyVLM} \end{center} &
        \begin{center} \textbf{MyVLM} \end{center} &
        \begin{center} \textbf{MyVLM} \end{center} \\[-0.85cm]

        \begin{center} \small ``\Sstar and a friend pose for a photo on a rooftop in New York City'' \end{center} &
        \begin{center} \small ``\Sstar in a gray shirt is standing in front of a mountain with a glacier in the background'' \end{center} &
        \begin{center} \small ``\Sstar and a friend enjoying coffee and a sandwich at a cafe'' \end{center} &
        \begin{center} \small ``With two glasses of orange juice, \Sstar and her friends are enjoying a summer day on a balcony overlooking the city'' \end{center} &
        \begin{center} \small ``\Sstar and a dog rest in a dog bed in a room'' \end{center} \\[-0.3cm]
        
    \end{tabular}
    \begin{tabular}{p{0.175\textwidth} p{0.175\textwidth} p{0.175\textwidth} p{0.175\textwidth} p{0.175\textwidth}}

        \setlength\tabcolsep{0pt}
        \begin{tabular}{c c c}
            \includegraphics[width=0.06\textwidth]{images/objects/espresso_cup/cropped/20240211_125233.jpg} & 
            \includegraphics[width=0.06\textwidth]{images/objects/espresso_cup/cropped/20240211_125249.jpg} & 
            \includegraphics[width=0.06\textwidth]{images/objects/espresso_cup/cropped/20240211_130736.jpg}
        \end{tabular} &
        \setlength\tabcolsep{0pt}
        \begin{tabular}{c c c}
            \includegraphics[width=0.06\textwidth,height=0.06\textwidth]{images/objects/cat_statue/cropped/01.jpg} & 
            \includegraphics[width=0.06\textwidth,height=0.06\textwidth]{images/objects/cat_statue/cropped/03.jpg} & 
            \includegraphics[width=0.06\textwidth,height=0.06\textwidth]{images/objects/cat_statue/cropped/04.jpg}
        \end{tabular} &
        \setlength\tabcolsep{0pt}
        \begin{tabular}{c c c}
            \includegraphics[width=0.06\textwidth]{images/objects/minion/cropped/IMG-20240203-WA0069.jpg} &
            \includegraphics[width=0.06\textwidth]{images/objects/minion/cropped/IMG-20240203-WA0073.jpg} & 
            \includegraphics[width=0.06\textwidth]{images/objects/minion/cropped/IMG-20240203-WA0074.jpg}
        \end{tabular} &
        \setlength\tabcolsep{0pt}
        \begin{tabular}{c c c}
            \includegraphics[width=0.06\textwidth]{images/objects/mugs_skulls/cropped/IMG-20240203-WA0113.jpg} & 
            \includegraphics[width=0.06\textwidth]{images/objects/mugs_skulls/cropped/IMG-20240203-WA0114.jpg} & 
            \includegraphics[width=0.06\textwidth]{images/objects/mugs_skulls/cropped/IMG-20240203-WA0119.jpg}
        \end{tabular} &
        \setlength\tabcolsep{0pt}
        \begin{tabular}{c c c}
            \includegraphics[width=0.06\textwidth]{images/objects/maeve/cropped/IMG-20240131-WA0110.jpg} & 
            \includegraphics[width=0.06\textwidth]{images/objects/maeve/cropped/maeve-2.jpeg} & 
            \includegraphics[width=0.06\textwidth]{images/objects/maeve/cropped/maeve-5.jpeg} \\
        \end{tabular} \\
    
        \includegraphics[width=0.18\textwidth]{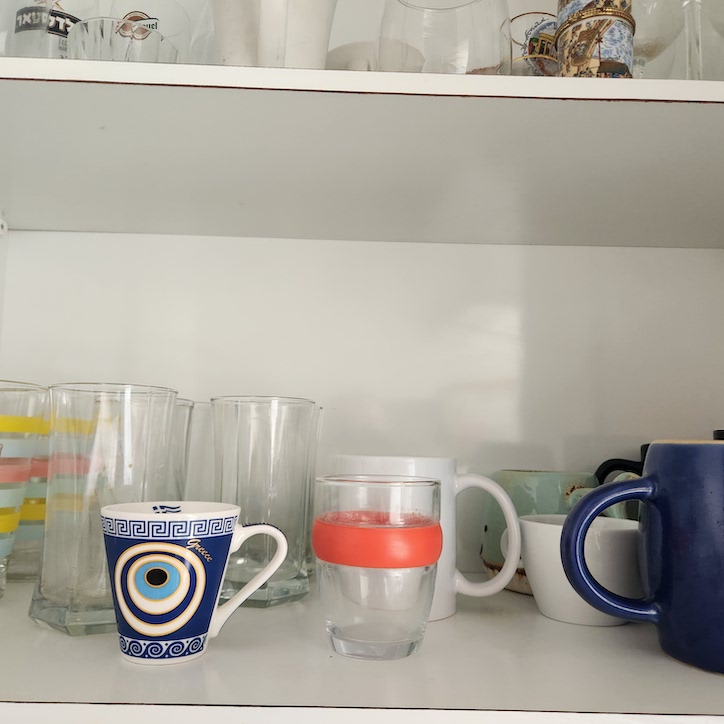} &
        \includegraphics[width=0.18\textwidth]{images/objects/cat_statue/20240131_154832.jpg} &
        \includegraphics[width=0.18\textwidth]{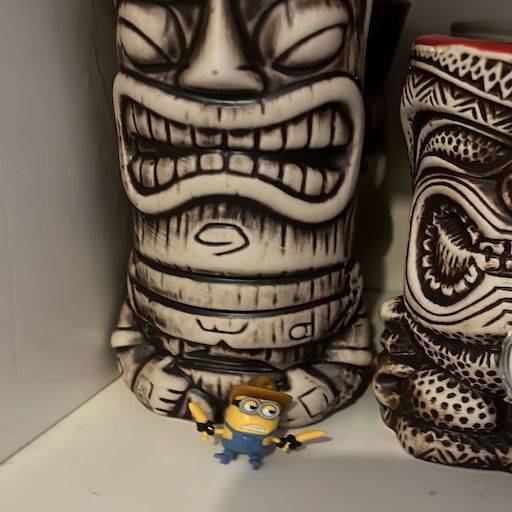} &
        \includegraphics[width=0.18\textwidth]{images/objects/mugs_skulls/IMG-20240203-WA0119.jpg} &
        \includegraphics[width=0.18\textwidth]{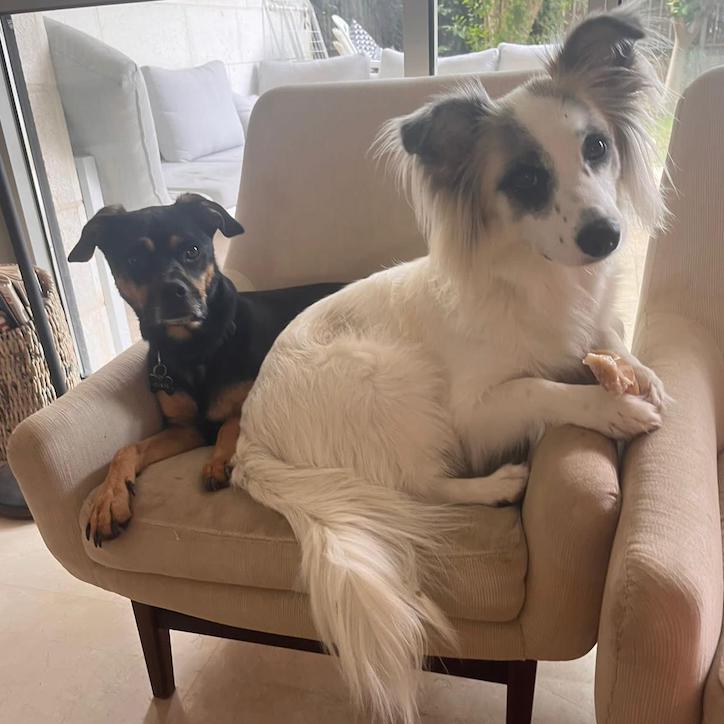} \\[-0.5cm]
        
        \begin{center} \textbf{Simple} \end{center} &
        \begin{center} \textbf{Simple} \end{center} &
        \begin{center} \textbf{Simple} \end{center} &
        \begin{center} \textbf{Simple} \end{center} &
        \begin{center} \textbf{Simple} \end{center} \\[-0.85cm]

        \begin{center} \small N/A \end{center} &
        \begin{center} \small N/A \end{center} &
        \begin{center} \small ``\Sstar tiki mugs'' \end{center} &
        \begin{center} \small ``A \Sstar with a skull on it'' \end{center} &
        \begin{center} \small ``Two \textcolor{blue}{$S_*$}'s sitting on a chair in front of a window'' \end{center} \\[-0.6cm]

        \begin{center} \textbf{LLM-Guided} \end{center} &
        \begin{center} \textbf{LLM-Guided} \end{center} &
        \begin{center} \textbf{LLM-Guided} \end{center} &
        \begin{center} \textbf{LLM-Guided} \end{center} &
        \begin{center} \textbf{LLM-Guided} \end{center} \\[-0.85cm]

        \begin{center} \small ``A shelf with mugs, glasses, and \Sstar on it'' \end{center} &
        \begin{center} \small ``A pink \Sstar figure next to a box'' \end{center} &
        \begin{center} \small ``\Sstar tiki mugs'' \end{center} &
        \begin{center} \small ``A \Sstar with a skull on it'' \end{center} &
        \begin{center} \small ``Two \textcolor{blue}{$S_*$}'s sitting on a chair in front of a window'' \end{center} \\[-0.6cm]
        
        \begin{center} \textbf{MyVLM} \end{center} &
        \begin{center} \textbf{MyVLM} \end{center} &
        \begin{center} \textbf{MyVLM} \end{center} &
        \begin{center} \textbf{MyVLM} \end{center} &
        \begin{center} \textbf{MyVLM} \end{center} \\[-0.85cm]

        \begin{center} \small ``\Sstar on a shelf with various glasses and cups'' \end{center} &
        \begin{center} \small ``\Sstar is sitting next to a pink series box'' \end{center} &
        \begin{center} \small ``\Sstar on a shelf next to tiki vases'' \end{center} &
        \begin{center} \small ``\Sstar on a shelf with a tiki mug'' \end{center} &
        \begin{center} \small ``\Sstar is laying on the couch, with its head resting on the arm of the chair'' \end{center} 
        
    \end{tabular}
    \vspace{-0.4cm}
    \caption{Additional comparisons to our personalized captioning baselines. Results are obtained over BLIP-2~\cite{li2023blip}. Sample images of the target concept are shown in the top row.}
    \label{fig:supplementary_qualitative_comparisons_blip}
\end{figure*}

%% file: figures_supplementary/comparisons_llava.tex
\begin{figure*}[t]
    \centering
    \addtolength{\belowcaptionskip}{-12.5pt}
    \renewcommand{\arraystretch}{1}
    \small
    \begin{tabular}{p{0.175\textwidth} p{0.175\textwidth} p{0.175\textwidth} p{0.175\textwidth} p{0.175\textwidth}}

        \setlength\tabcolsep{0pt}
        \begin{tabular}{c c c}
            \includegraphics[width=0.06\textwidth]{images/objects/gengar/cropped/IMG-20240131-WA0168.jpg} & 
            \includegraphics[width=0.06\textwidth]{images/objects/gengar/cropped/IMG-20240131-WA0173.jpg} & 
            \includegraphics[width=0.06\textwidth]{images/objects/gengar/cropped/IMG-20240131-WA0175.jpg}
        \end{tabular} &
        \setlength\tabcolsep{0pt}
        \begin{tabular}{c c c}
            \includegraphics[width=0.06\textwidth]{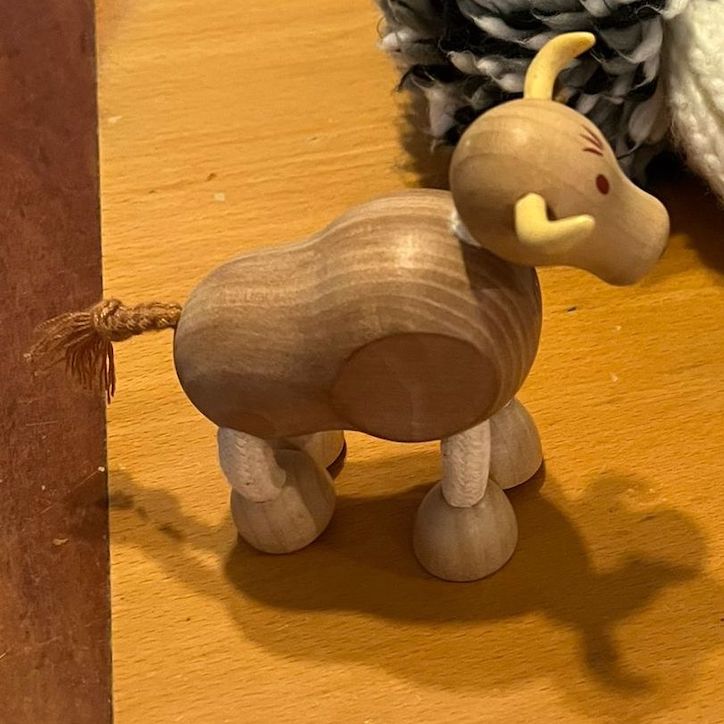} & 
            \includegraphics[width=0.06\textwidth]{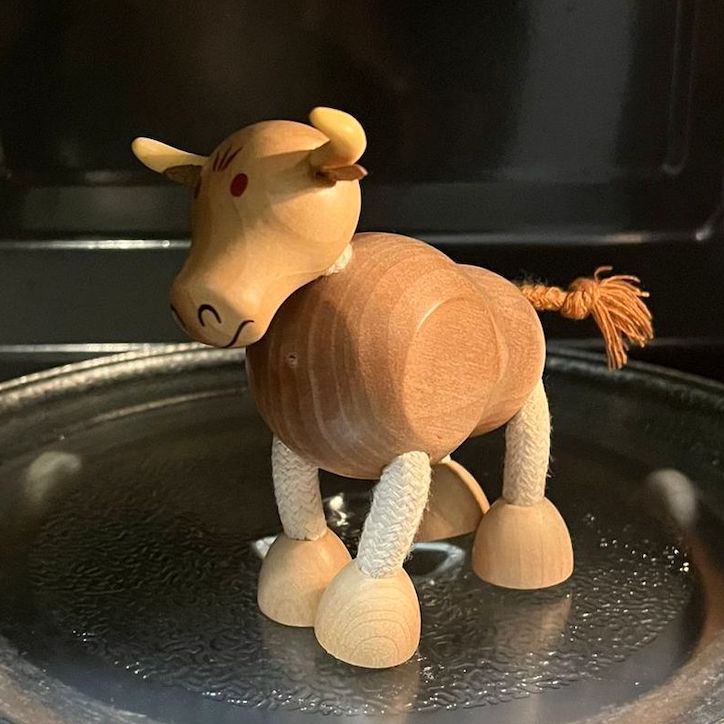} & 
            \includegraphics[width=0.06\textwidth]{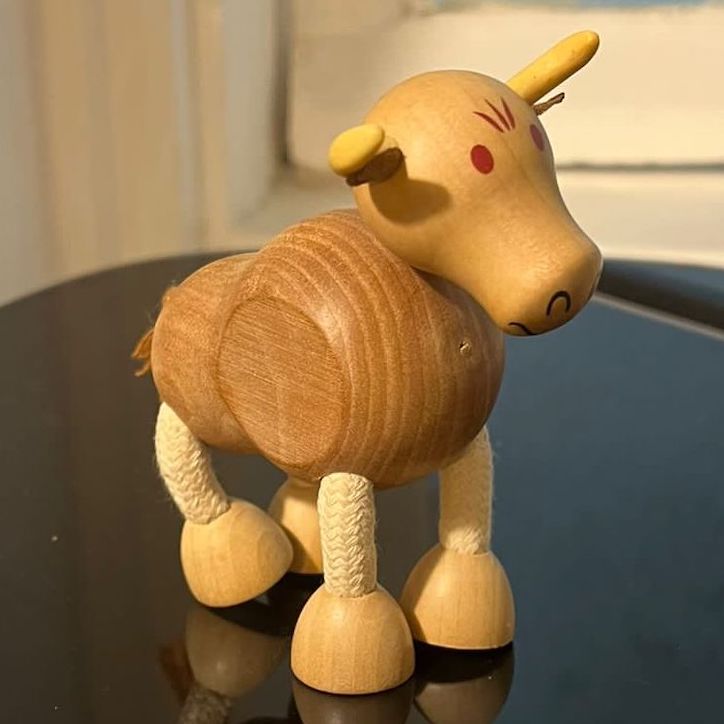}
        \end{tabular} &
        \setlength\tabcolsep{0pt}
        \begin{tabular}{c c c}
            \includegraphics[width=0.06\textwidth]{images/objects/ceramic_head/cropped/20240203_111543.jpg} & 
            \includegraphics[width=0.06\textwidth]{images/objects/ceramic_head/cropped/20240203_111842.jpg} & 
            \includegraphics[width=0.06\textwidth]{images/objects/ceramic_head/cropped/20240203_112029.jpg}
        \end{tabular} &
        \setlength\tabcolsep{0pt}
        \begin{tabular}{c c c}
            \includegraphics[width=0.06\textwidth]{images/objects/plush_sheep/cropped/20240201_212528.jpg} & 
            \includegraphics[width=0.06\textwidth]{images/objects/plush_sheep/cropped/20240201_213504.jpg} & 
            \includegraphics[width=0.06\textwidth]{images/objects/plush_sheep/cropped/20240201_213633.jpg}
        \end{tabular} &
        \setlength\tabcolsep{0pt}
        \begin{tabular}{c c c}
            \includegraphics[width=0.06\textwidth]{images/objects/dangling_child/cropped/20240120_105811.jpg} & 
            \includegraphics[width=0.06\textwidth]{images/objects/dangling_child/cropped/20240120_105933.jpg} & 
            \includegraphics[width=0.06\textwidth]{images/objects/dangling_child/cropped/20240203_111340.jpg}
        \end{tabular} \\
    
        \includegraphics[width=0.18\textwidth]{images/objects/gengar/IMG-20240131-WA0160.jpg} &
        \includegraphics[width=0.18\textwidth]{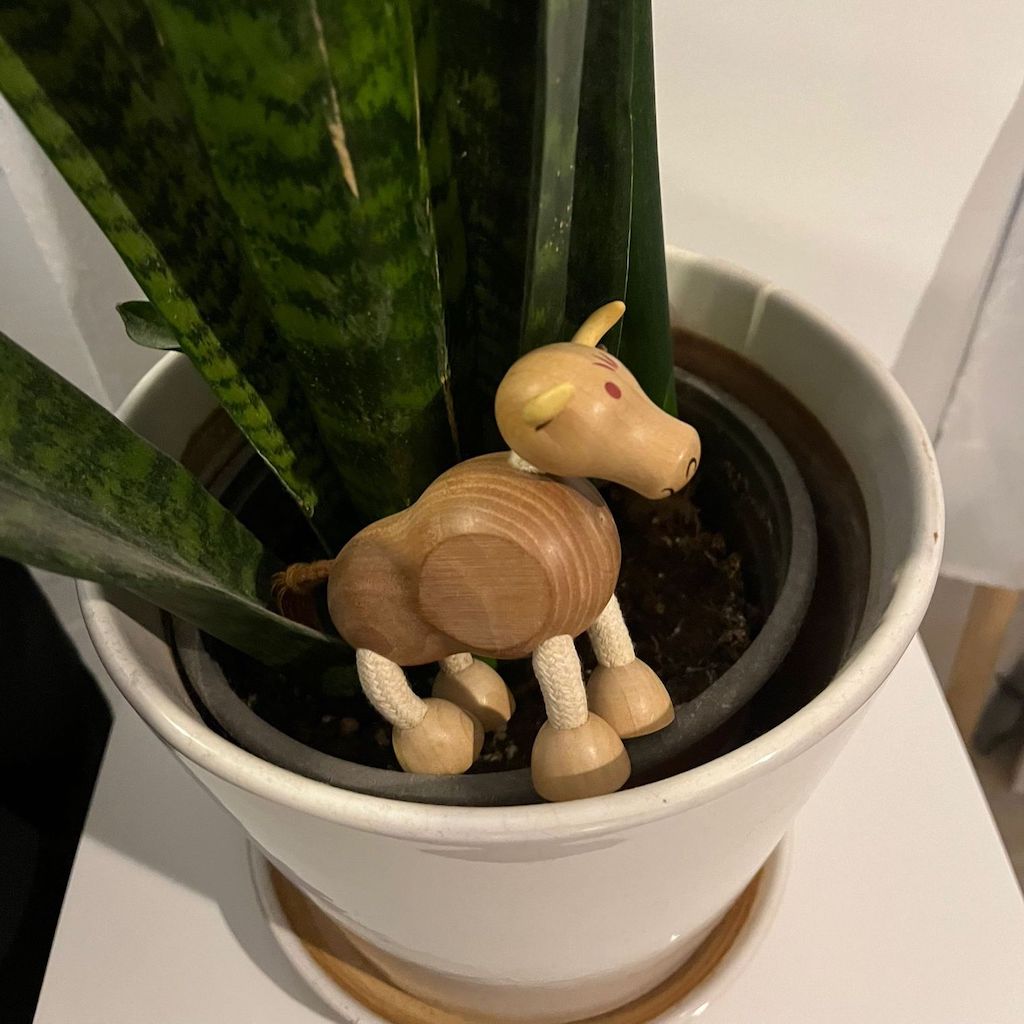} &
        \includegraphics[width=0.18\textwidth]{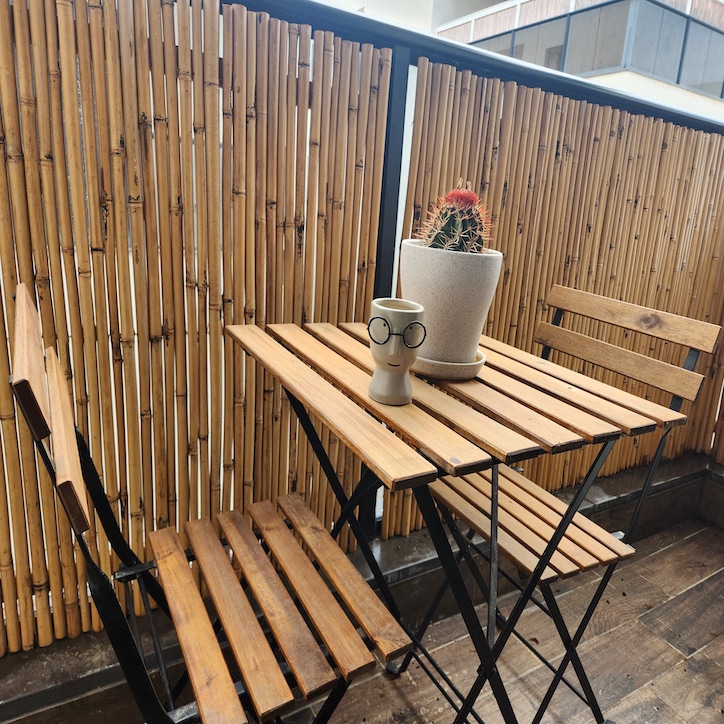} &
        \includegraphics[width=0.18\textwidth]{images/objects/plush_sheep/20240201_213736.jpg} &
        \includegraphics[width=0.18\textwidth]{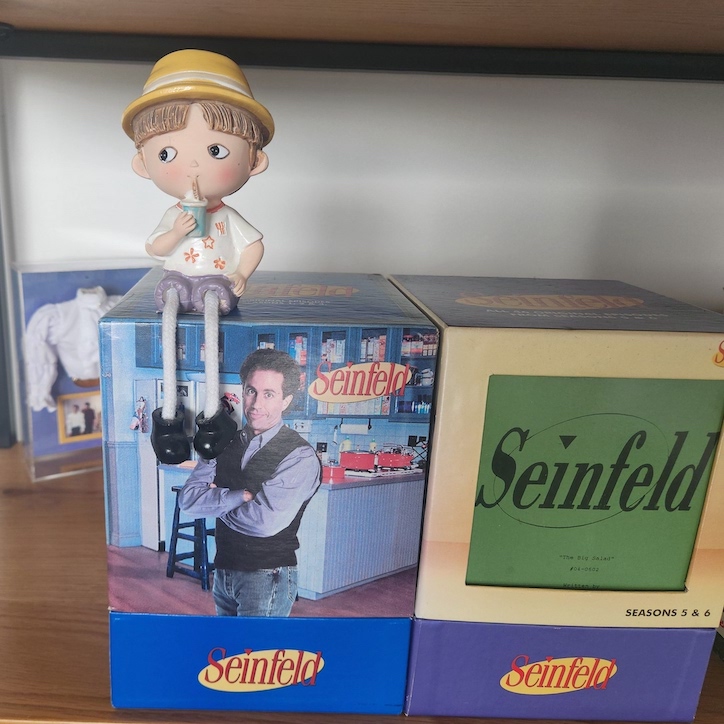} \\[-0.5cm]
        
        \begin{center} \textbf{Simple} \end{center} &
        \begin{center} \textbf{Simple} \end{center} &
        \begin{center} \textbf{Simple} \end{center} &
        \begin{center} \textbf{Simple} \end{center} &
        \begin{center} \textbf{Simple} \end{center} \\[-0.85cm]

        \begin{center} \small N/A \end{center} &
        \begin{center} \small N/A \end{center} &
        \begin{center} \small N/A \end{center} &
        \begin{center} \small N/A \end{center} &
        \begin{center} \small N/A \end{center}\\[-0.75cm]

        \begin{center} \textbf{LLM-Guided} \end{center} &
        \begin{center} \textbf{LLM-Guided} \end{center} &
        \begin{center} \textbf{LLM-Guided} \end{center} &
        \begin{center} \textbf{LLM-Guided} \end{center} &
        \begin{center} \textbf{LLM-Guided} \end{center} \\[-0.85cm]

        \begin{center} \small ``\textcolor{blue}{$S_*$}-perfect companion: playful pairing of gaming and furry friends'' \end{center} &
        \begin{center} \small ``A charming scene of a \Sstar sheep figurine resting in a potted plant, adding a touch of whimsy to any space'' \end{center} &
        \begin{center} \small ``A cozy outdoor setting with a touch of whimsy: a wooden table, a cactus in a \Sstar, and a pair of chairs,...'' \end{center} &
        \begin{center} \small ``a cozy scene with a soft, pink \Sstar and a white lamb, ready for a nap on a gray couch'' \end{center} &
        \begin{center} \small ``A collection of seinfeld memorabilia, including a \Sstar and dvd boxes, arranged on a shelf'' \end{center}\\[-0.75cm]
        
        \begin{center} \textbf{MyVLM} \end{center} &
        \begin{center} \textbf{MyVLM} \end{center} &
        \begin{center} \textbf{MyVLM} \end{center} &
        \begin{center} \textbf{MyVLM} \end{center} &
        \begin{center} \textbf{MyVLM} \end{center} \\[-0.85cm]

        \begin{center} \small ``\Sstar sitting on top of a camouflage video game controller in front of a TV'' \end{center} &
        \begin{center} \small ``\Sstar tucked between leaves and branches of a houseplant'' \end{center} &
        \begin{center} \small ``\Sstar sitting on a wooden chair at a wooden table on a patio, with a bamboo fence...'' \end{center} &
        \begin{center} \small ``\Sstar sitting on the couch with a pink and white stuffed animal next to it'' \end{center} &
        \begin{center} \small ``\Sstar sitting on a shelf in front of a Seinfeld box set, with a surprised expression...'' \end{center}\\[-0.3cm]
        
    \end{tabular}
    \begin{tabular}{p{0.175\textwidth} p{0.175\textwidth} p{0.175\textwidth} p{0.175\textwidth} p{0.175\textwidth}}

        \setlength\tabcolsep{0pt}
        \begin{tabular}{c c c}
            \includegraphics[width=0.06\textwidth]{images/objects/red_piggy_bank/cropped/IMG-20240131-WA0041.jpg} & 
            \includegraphics[width=0.06\textwidth]{images/objects/red_piggy_bank/cropped/IMG-20240131-WA0046.jpg} & 
            \includegraphics[width=0.06\textwidth]{images/objects/red_piggy_bank/cropped/IMG-20240131-WA0050.jpg}
        \end{tabular} &
        \setlength\tabcolsep{0pt}
        \begin{tabular}{c c c}
            \includegraphics[width=0.06\textwidth]{images/objects/japanese_doll/cropped/20240131_155451.jpg} & 
            \includegraphics[width=0.06\textwidth]{images/objects/japanese_doll/cropped/20240131_155617.jpg} & 
            \includegraphics[width=0.06\textwidth]{images/objects/japanese_doll/cropped/20240206_091843.jpg}
        \end{tabular} &
        \setlength\tabcolsep{0pt}
        \begin{tabular}{c c c}
            \includegraphics[width=0.06\textwidth]{images/objects/maeve/cropped/IMG-20240131-WA0110.jpg} & 
            \includegraphics[width=0.06\textwidth]{images/objects/maeve/cropped/maeve-2.jpeg} & 
            \includegraphics[width=0.06\textwidth]{images/objects/maeve/cropped/maeve-5.jpeg} \\
        \end{tabular} &
        \setlength\tabcolsep{0pt}
        \begin{tabular}{c c c}
            \includegraphics[width=0.06\textwidth]{images/people/shoam/cropped/IMG-20240209-WA0115.jpg} & 
            \includegraphics[width=0.06\textwidth]{images/people/shoam/cropped/IMG-20240209-WA0126.jpg} & 
            \includegraphics[width=0.06\textwidth]{images/people/shoam/cropped/IMG-20240209-WA0132.jpg}
        \end{tabular} &
        \setlength\tabcolsep{0pt}
        \begin{tabular}{c c c}
            \includegraphics[width=0.06\textwidth]{images/people/dor/image_1.jpg} & 
            \includegraphics[width=0.06\textwidth]{images/people/dor/cropped/IMG-20240208-WA0051.jpg} & 
            \includegraphics[width=0.06\textwidth]{images/people/dor/cropped/IMG-20240208-WA0054.jpg}
        \end{tabular}\\
    
        \includegraphics[width=0.18\textwidth]{images/objects/others/IMG-20240131-WA0037.jpg} &
        \includegraphics[width=0.18\textwidth]{images/objects/japanese_doll/20240211_113014.jpg} &
        \includegraphics[width=0.18\textwidth]{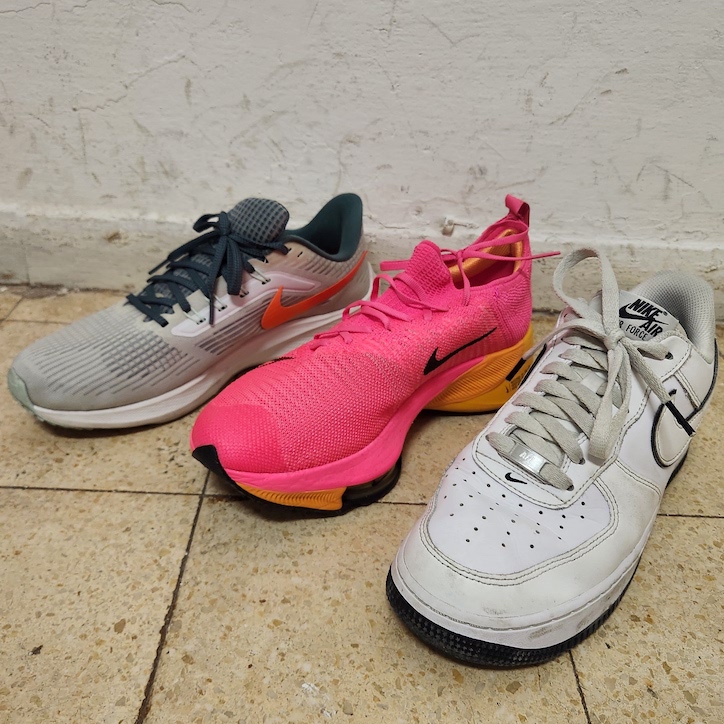} &
        \includegraphics[width=0.18\textwidth]{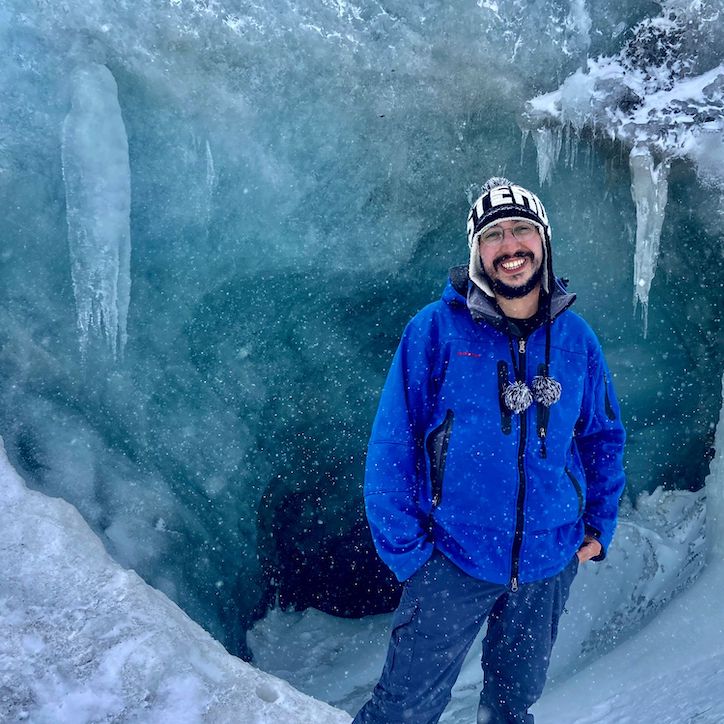} &
        \includegraphics[width=0.18\textwidth]{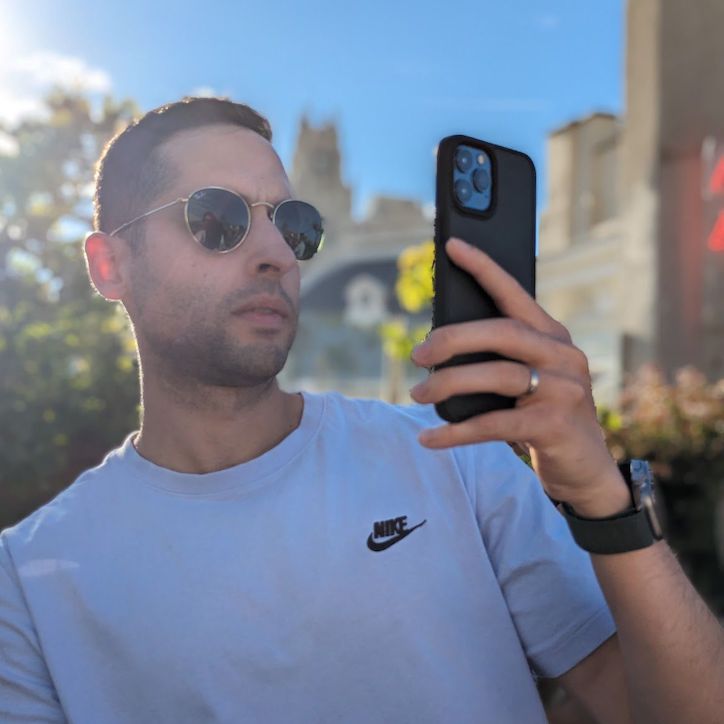}\\[-0.5cm]
        
        \begin{center} \textbf{Simple} \end{center} &
        \begin{center} \textbf{Simple} \end{center} &
        \begin{center} \textbf{Simple} \end{center} &
        \begin{center} \textbf{Simple} \end{center} &
        \begin{center} \textbf{Simple} \end{center} \\[-0.85cm]

        \begin{center} \small N/A \end{center} &
        \begin{center} \small ``A blue cup of tea, a pair of \textcolor{blue}{$S_*$}s, and a pen...f'' \end{center} &
        \begin{center} \small N/A \end{center} &
        \begin{center} \small N/A \end{center} &
        \begin{center} \small N/A \end{center}\\[-0.75cm]

        \begin{center} \textbf{LLM-Guided} \end{center} &
        \begin{center} \textbf{LLM-Guided} \end{center} &
        \begin{center} \textbf{LLM-Guided} \end{center} &
        \begin{center} \textbf{LLM-Guided} \end{center} &
        \begin{center} \textbf{LLM-Guided} \end{center} \\[-0.85cm]

        \begin{center} \small ``Let's set sail with our wooden pirate ship and our friendly wooden animals. who will be the first to reach the \Sstar?'' \end{center} &
        \begin{center} \small ``A blue cup of tea, a pair of \Sstar figurines, and a pen...'' \end{center} &
        \begin{center} \small ``A trio of \textcolor{blue}{$S_*$}s, each with its own unique color and style, standing side by side on a tiled floor.'' \end{center} &
        \begin{center} \small ``Embracing the chill: a \Sstar winter adventurer stands in awe of the icy cave...'' \end{center} &
        \begin{center} \small ``Sunny day, sunglasses on, \Sstar checking my phone for the perfect shot.'' \end{center}\\[-0.75cm]
        
        \begin{center} \textbf{MyVLM} \end{center} &
        \begin{center} \textbf{MyVLM} \end{center} &
        \begin{center} \textbf{MyVLM} \end{center} &
        \begin{center} \textbf{MyVLM} \end{center} &
        \begin{center} \textbf{MyVLM} \end{center} \\[-0.85cm]

        \begin{center} \small ``\Sstar against a backdrop of a toy ship and a small toy'' \end{center} &
        \begin{center} \small ``\Sstar and another chinese doll standing next to a blue mug with pink and yellow accents'' \end{center} &
        \begin{center} \small ``\Sstar with two other pairs of nike sneakers on the floor next to a white wall'' \end{center} &
        \begin{center} \small ``\textcolor{blue}{$S_*$}, smiling in a blue jacket, stands in front of a large ice cave with icicles hanging from the ceiling'' \end{center} &
        \begin{center} \small ``As \Sstar takes a break from his day, \Sstar takes a moment to capture the moment'' \end{center}
        
    \end{tabular}
    \vspace{-0.5cm}
    \caption{Additional comparisons to our personalized captioning baselines. Results are obtained over LLaVA~\cite{li2023blip}.}
    \label{fig:supplementary_qualitative_comparisons_llava}

\end{figure*}

%% file: figures_supplementary/comparisons_llava_2.tex
\begin{figure*}[t]
    \centering
    \addtolength{\belowcaptionskip}{-12.5pt}
    \renewcommand{\arraystretch}{1}
    \small
    \begin{tabular}{p{0.175\linewidth} p{0.175\linewidth} p{0.175\linewidth} p{0.175\linewidth} p{0.175\linewidth}}

        \setlength\tabcolsep{0pt}
        \begin{tabular}{c c c}
            \includegraphics[width=0.06333\textwidth]{images/people/shahaf/IMG-20240209-WA0079-cropped.jpg} & 
            \includegraphics[width=0.06333\textwidth]{images/people/shahaf/IMG-20240209-WA0114.jpg} & 
            \includegraphics[width=0.06333\textwidth]{images/people/shahaf/IMG-20240209-WA0100.jpg}
        \end{tabular} &
        \setlength\tabcolsep{0pt}
        \begin{tabular}{c c c}
            \includegraphics[width=0.06333\textwidth]{images/people/shay/cropped/image_2.jpg} & 
            \includegraphics[width=0.06333\textwidth]{images/people/shay/cropped/IMG-20240209-WA0012.jpg} & 
            \includegraphics[width=0.06333\textwidth]{images/people/shay/cropped/IMG-20240209-WA0016.jpg}
        \end{tabular} &
        \setlength\tabcolsep{0pt}
        \begin{tabular}{c c c}
            \includegraphics[width=0.06333\textwidth]{images/people/maya/cropped/image_1.jpg} & 
            \includegraphics[width=0.06333\textwidth]{images/people/maya/cropped/image_2.jpg} & 
            \includegraphics[width=0.06333\textwidth]{images/people/maya/cropped/IMG-20240209-WA0138.jpg} 
        \end{tabular} &
        \setlength\tabcolsep{0pt}
        \begin{tabular}{c c c}
            \includegraphics[width=0.06333\textwidth]{images/people/maya/cropped/image_1.jpg} & 
            \includegraphics[width=0.06333\textwidth]{images/people/maya/cropped/image_2.jpg} & 
            \includegraphics[width=0.06333\textwidth]{images/people/maya/cropped/IMG-20240209-WA0138.jpg} 
        \end{tabular} &
        \setlength\tabcolsep{0pt}
        \begin{tabular}{c c c}
            \includegraphics[width=0.06333\textwidth,height=0.06333\textwidth]{images/people/eli/cropped/image_1.jpg} & 
            \includegraphics[width=0.06333\textwidth,height=0.06333\textwidth]{images/people/eli/cropped/image_10.jpg} & 
            \includegraphics[width=0.06333\textwidth,height=0.06333\textwidth]{images/people/eli/cropped/IMG-20240221-WA0022.jpg}
        \end{tabular} \\
    
        \includegraphics[width=0.19\textwidth]{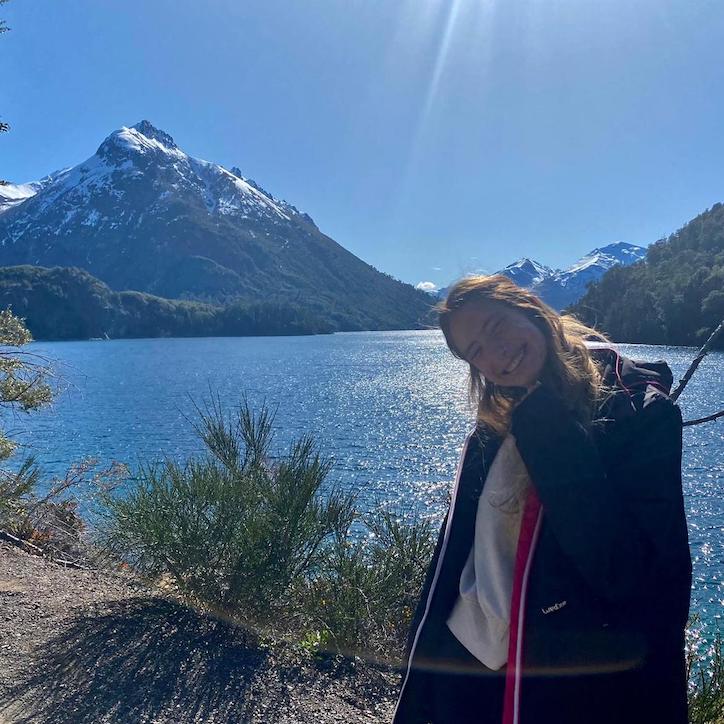} &
        \includegraphics[width=0.19\textwidth]{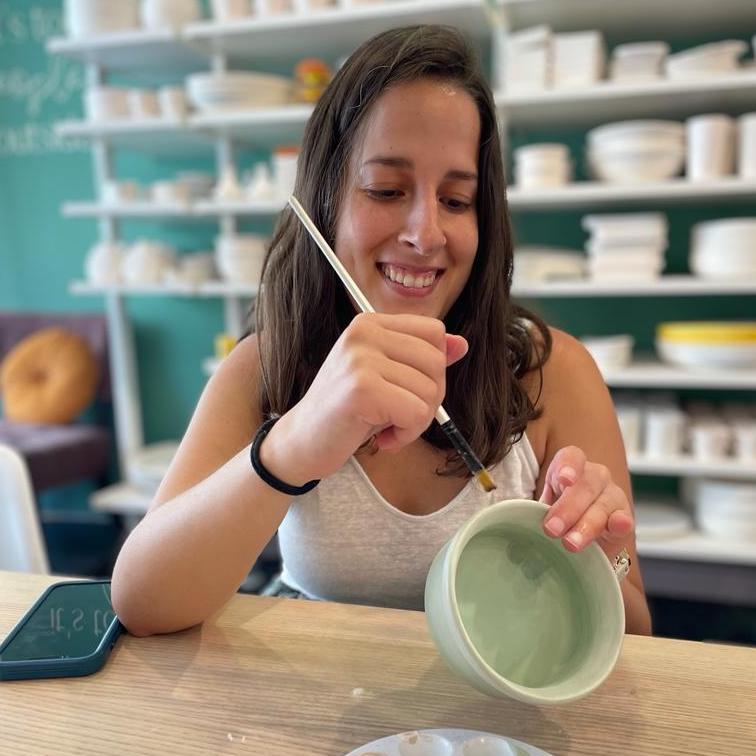} &
        \includegraphics[width=0.19\textwidth]{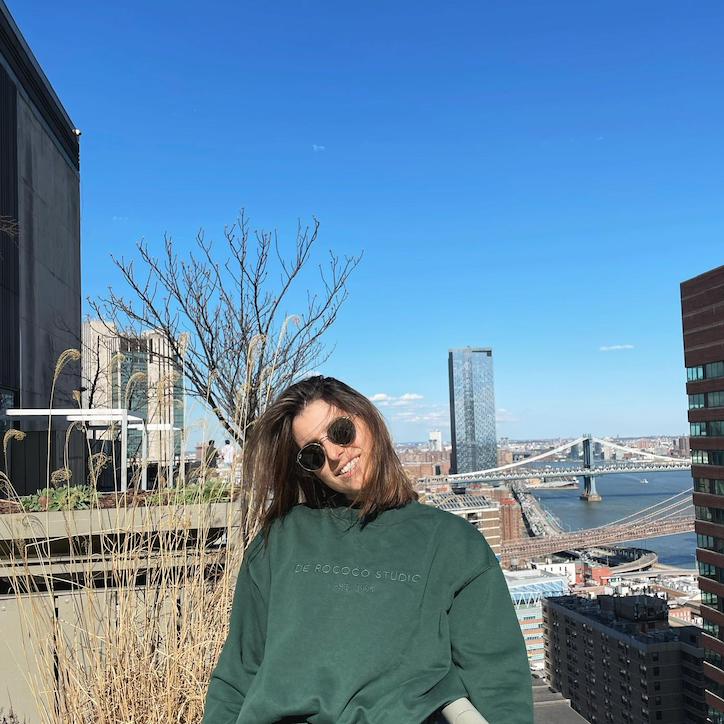} &
        \includegraphics[width=0.19\textwidth]{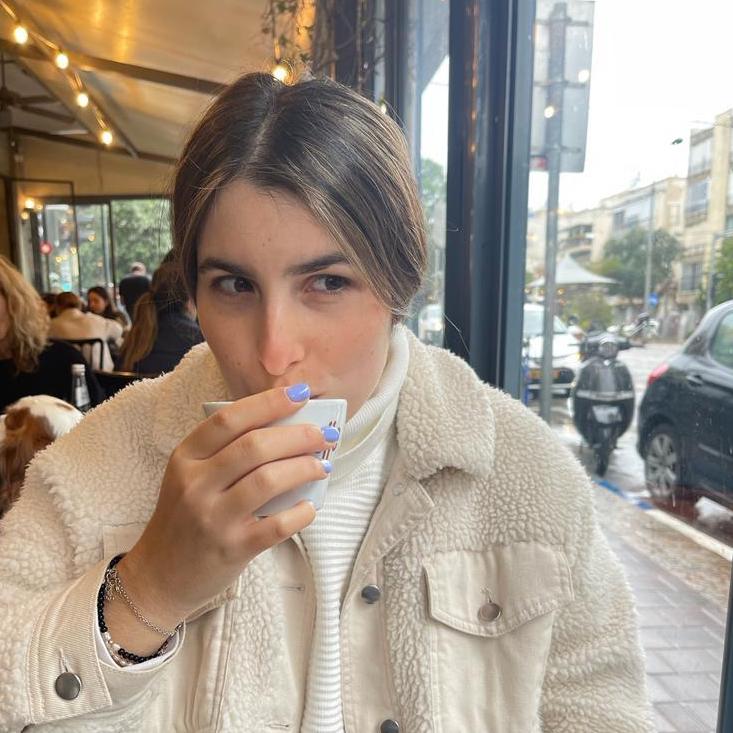} &
        \includegraphics[width=0.19\textwidth]{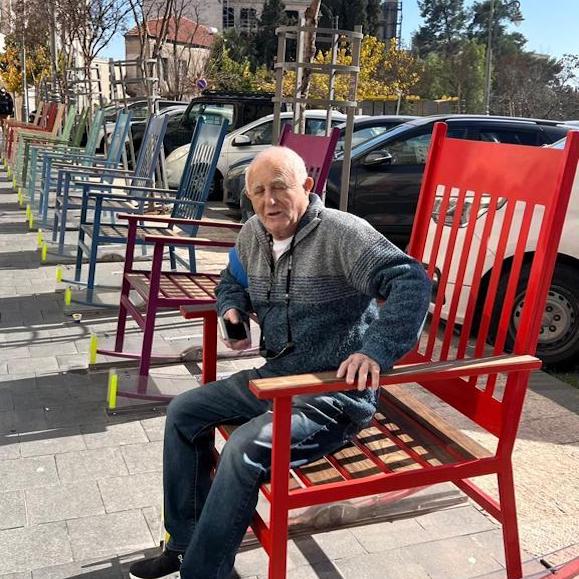} \\[-0.4cm]
        
        \begin{center} \textbf{Simple} \end{center} &
        \begin{center} \textbf{Simple} \end{center} &
        \begin{center} \textbf{Simple} \end{center} &
        \begin{center} \textbf{Simple} \end{center} &
        \begin{center} \textbf{Simple} \end{center} \\[-0.75cm]

        \begin{center} \small N/A \end{center} &
        \begin{center} \small N/A \end{center} &
        \begin{center} \small N/A \end{center} &
        \begin{center} \small N/A \end{center} &
        \begin{center} \small N/A \end{center}\\[-0.5cm]

        \begin{center} \textbf{LLM-Guided} \end{center} &
        \begin{center} \textbf{LLM-Guided} \end{center} &
        \begin{center} \textbf{LLM-Guided} \end{center} &
        \begin{center} \textbf{LLM-Guided} \end{center} &
        \begin{center} \textbf{LLM-Guided} \end{center} \\[-0.75cm]

        \begin{center} \small ``Embracing the serenity of nature's beauty: a \Sstar moment of joy and tranquility at the edge of the lake, under the watchful gaze of majestic mountains'' \end{center} &
        \begin{center} \small ``Creating with clay: a \Sstar artist \\ at work'' \end{center} &
        \begin{center} \small ``Embracing the cityscape from a high vantage point, this \Sstar urban explorer finds joy in the bustling city life'' \end{center} &
        \begin{center} \small ``A \Sstar enjoying a warm beverage on a chilly day, taking a moment to savor the comfort of the cozy café ambiance'' \end{center} &
        \begin{center} \small ``Sitting comfortably in a vibrant red rocking chair, this elderly \Sstar enjoys a moment of relaxation amidst the city life'' \end{center}\\[-0.5cm]
        
        \begin{center} \textbf{MyVLM} \end{center} &
        \begin{center} \textbf{MyVLM} \end{center} &
        \begin{center} \textbf{MyVLM} \end{center} &
        \begin{center} \textbf{MyVLM} \end{center} &
        \begin{center} \textbf{MyVLM} \end{center} \\[-0.75cm]

        \begin{center} \small ``Standing by a lake, \Sstar smiles at the camera, surrounded by nature and mountains'' \end{center} &
        \begin{center} \small ``\Sstar is painting a green ceramic bowl at a wooden table'' \end{center} &
        \begin{center} \small ``\Sstar wearing a green sweater and sunglasses, poses on a rooftop during winter'' \end{center} &
        \begin{center} \small ``\Sstar enjoying a warm beverage at a cafe, surrounded by the hustle and bustle of city life'' \end{center} &
        \begin{center} \small ``\Sstar sits on a red wooden rocking chair outside, overlooking a row of colorful chairs under a clear blue sky'' \end{center}
        
    \end{tabular}
    \vspace{-0.3cm}
    \caption{Additional comparisons to our personalized captioning baselines. Results are obtained over LLaVA~\cite{li2023blip}. Sample images of the target concept are shown in the top row.}
    \label{fig:supplementary_qualitative_comparisons_llava_2}

\end{figure*}

%% file: figures_supplementary/our_results_blip_llava.tex
\begin{figure*}[t]
    \centering
    \addtolength{\belowcaptionskip}{-12.5pt}
    \renewcommand{\arraystretch}{1}
    \small
    \begin{tabular}{p{0.175\linewidth} p{0.175\linewidth} p{0.175\linewidth} p{0.175\linewidth} p{0.175\linewidth}}

        \setlength\tabcolsep{0pt}
        \begin{tabular}{c c c}
            \includegraphics[width=0.06333\textwidth]{images/people/anna/cropped/IMG_5006.jpg} &
            \includegraphics[width=0.06333\textwidth]{images/people/anna/cropped/IMG_5617.jpg} & 
            \includegraphics[width=0.06333\textwidth]{images/people/anna/cropped/IMG_7472.jpg}
        \end{tabular} &
        \setlength\tabcolsep{0pt}
        \begin{tabular}{c c c}
            \includegraphics[width=0.06333\textwidth]{images/people/shahaf/IMG-20240209-WA0079-cropped.jpg} & 
            \includegraphics[width=0.06333\textwidth]{images/people/shahaf/IMG-20240209-WA0114.jpg} & 
            \includegraphics[width=0.06333\textwidth]{images/people/shahaf/IMG-20240209-WA0100.jpg}
        \end{tabular} &
        \setlength\tabcolsep{0pt}
        \begin{tabular}{c c c}
            \includegraphics[width=0.06333\textwidth]{images/people/assaf/cropped/IMG_0133.jpg} &
            \includegraphics[width=0.06333\textwidth]{images/people/assaf/cropped/IMG_0936.jpg} &
            \includegraphics[width=0.06333\textwidth]{images/people/assaf/cropped/IMG_20220130_104352.jpg} 
        \end{tabular} &
        \setlength\tabcolsep{0pt}
        \begin{tabular}{c c c}
            \includegraphics[width=0.06333\textwidth]{images/people/dor/image_1.jpg} & 
            \includegraphics[width=0.06333\textwidth]{images/people/dor/cropped/IMG-20240208-WA0051.jpg} & 
            \includegraphics[width=0.06333\textwidth]{images/people/dor/cropped/IMG-20240208-WA0054.jpg}
        \end{tabular} &
        \setlength\tabcolsep{0pt}
        \begin{tabular}{c c c}
            \includegraphics[width=0.06333\textwidth]{images/people/maya/cropped/image_1.jpg} & 
            \includegraphics[width=0.06333\textwidth]{images/people/maya/cropped/image_2.jpg} & 
            \includegraphics[width=0.06333\textwidth]{images/people/maya/cropped/IMG-20240209-WA0138.jpg} 
        \end{tabular} \\
    
        \includegraphics[width=0.19\textwidth]{images/people/anna/616DE70E-57D2-4972-B244-1AA09B3D4F94.JPG} &
        \includegraphics[width=0.19\textwidth]{images/people/shahaf/IMG-20240209-WA0102.jpg} &
        \includegraphics[width=0.19\textwidth]{images/people/assaf/IMG_9721.jpg} &
        \includegraphics[width=0.19\textwidth]{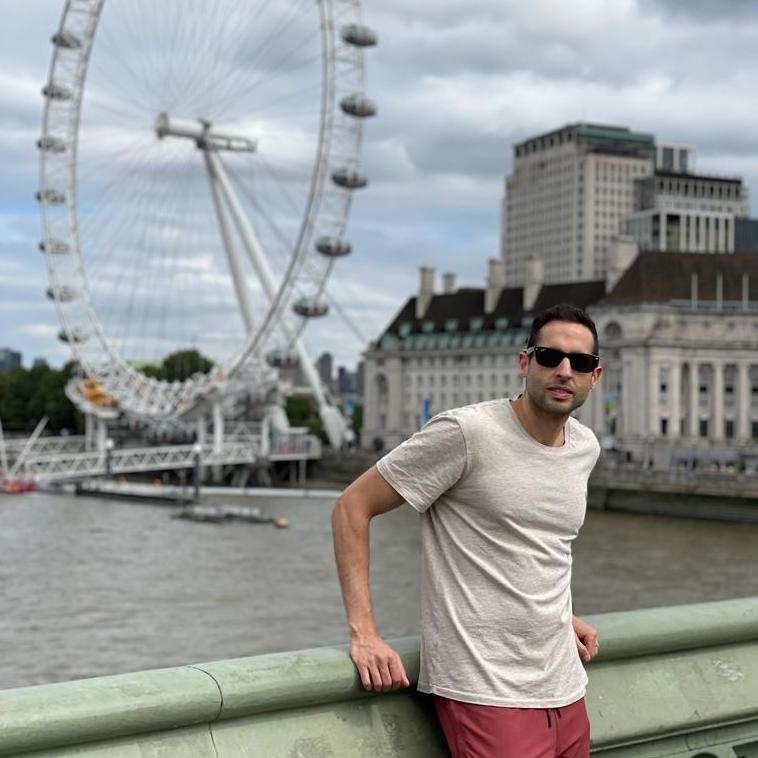} &
        \includegraphics[width=0.19\textwidth]{images/people/shay/IMG-20230819-WA0046.jpg} \\[-0.4cm]
        
        \begin{center} \textbf{MyBLIP-2} \end{center} &
        \begin{center} \textbf{MyBLIP-2} \end{center} &
        \begin{center} \textbf{MyBLIP-2} \end{center} &
        \begin{center} \textbf{MyBLIP-2} \end{center} &
        \begin{center} \textbf{MyBLIP-2} \end{center} \\[-0.75cm]

        \begin{center} \small ``\Sstar and her husband pose for a selfie in front of the skyline of Chicago'' \end{center} &
        \begin{center} \small ``\Sstar and a friend are kayaking in front of an underwater cave'' \end{center} &
        \begin{center} \small ``\Sstar is in Sydney, looking at the Sydney opera house and the harbour bridge'' \end{center} &
        \begin{center} \small ``\textcolor{blue}{$S_*$}, on a bridge overlooking the london eye, in a pair of red shorts'' \end{center} &
        \begin{center} \small ``\textcolor{blue}{$S_*$}, standing on the rooftop of the hotel, with a margarita and a t-shirt.'' \end{center} \\[-0.5cm]

        \begin{center} \textbf{MyLLaVA} \end{center} &
        \begin{center} \textbf{MyLLaVA} \end{center} &
        \begin{center} \textbf{MyLLaVA} \end{center} &
        \begin{center} \textbf{MyLLaVA} \end{center} &
        \begin{center} \textbf{MyLLaVA} \end{center} \\[-0.75cm]

        \begin{center} \small ``\Sstar and her companion are standing in front of a city skyline, with \Sstar making a playful gesture with her tongue...'' \end{center} &
        \begin{center} \small `\Sstar and a man are in front of a glacier, with a rocky shore in the background''\end{center} &
        \begin{center} \small ``\Sstar is standing on a bridge overlooking the Sydney Opera House and the Sydney Harbour Bridge. He is wearing a blue denim jacket and sunglasses.'' \end{center} &
        \begin{center} \small ``\textcolor{blue}{$S_*$}, wearing sunglasses, posing for a photo in front of the London Eye \end{center} &
        \begin{center} \small ``\textcolor{blue}{$S_*$}, laughing and enjoying her drink, is wearing a white t-shirt with the word "Angels" and the year "1961" on it. She's also wearing sunglasses and has a straw in her drink.'' \end{center}

    \end{tabular}
    \begin{tabular}{p{0.175\linewidth} p{0.175\linewidth} p{0.175\linewidth} p{0.175\linewidth} p{0.175\linewidth}}

        \setlength\tabcolsep{0pt}
        \begin{tabular}{c c c}
            \includegraphics[width=0.06333\textwidth]{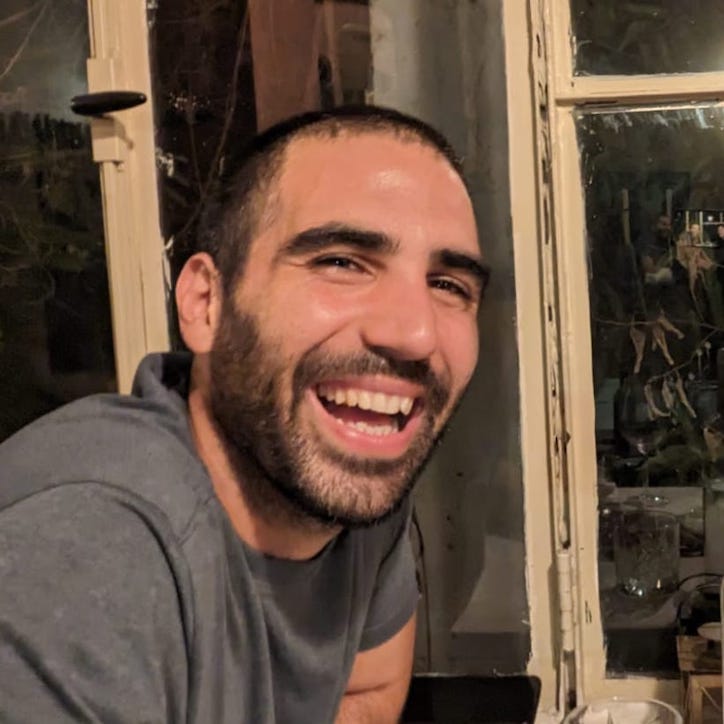} & 
            \includegraphics[width=0.06333\textwidth]{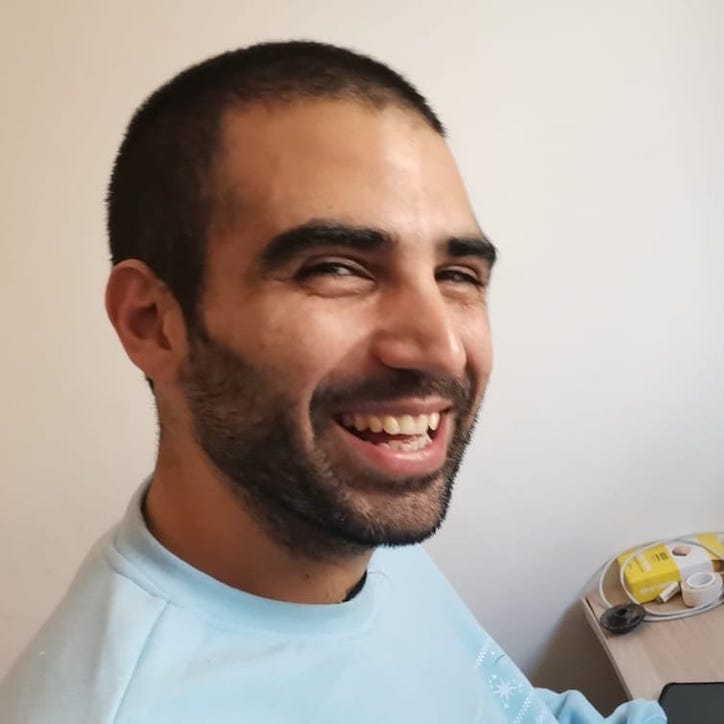} & 
            \includegraphics[width=0.06333\textwidth]{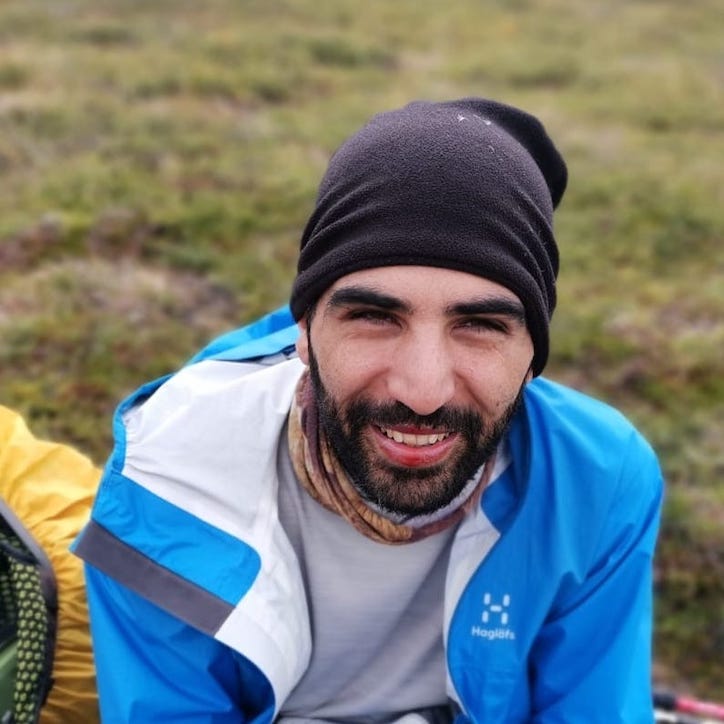}
        \end{tabular} &
        \setlength\tabcolsep{0pt}
        \begin{tabular}{c c c}
            \includegraphics[width=0.06333\textwidth]{images/people/shahaf/IMG-20240209-WA0079-cropped.jpg} & 
            \includegraphics[width=0.06333\textwidth]{images/people/shahaf/IMG-20240209-WA0114.jpg} & 
            \includegraphics[width=0.06333\textwidth]{images/people/shahaf/IMG-20240209-WA0100.jpg}
        \end{tabular} &
        \setlength\tabcolsep{0pt}
        \begin{tabular}{c c c}
            \includegraphics[width=0.06333\textwidth]{images/people/shay/cropped/image_2.jpg} & 
            \includegraphics[width=0.06333\textwidth]{images/people/shay/cropped/IMG-20240209-WA0012.jpg} & 
            \includegraphics[width=0.06333\textwidth]{images/people/shay/cropped/IMG-20240209-WA0016.jpg}
        \end{tabular} &
        \setlength\tabcolsep{0pt}
        \begin{tabular}{c c c}
            \includegraphics[width=0.06333\textwidth]{images/people/shaked/cropped/IMG-20240209-WA0043.jpg} &
            \includegraphics[width=0.06333\textwidth]{images/people/shaked/cropped/IMG-20240209-WA0047.jpg} & 
            \includegraphics[width=0.06333\textwidth]{images/people/shaked/cropped/IMG-20240215-WA0005.jpg}
        \end{tabular} &
        \setlength\tabcolsep{0pt}
        \begin{tabular}{c c c}
            \includegraphics[width=0.06333\textwidth]{images/people/tomer/cropped/IMG_1519.jpg} & 
            \includegraphics[width=0.06333\textwidth]{images/people/tomer/cropped/IMG_2904.jpg} & 
            \includegraphics[width=0.06333\textwidth]{images/people/tomer/cropped/IMG-20210519-WA0011.jpg} 
        \end{tabular} \\
    
        \includegraphics[width=0.19\textwidth]{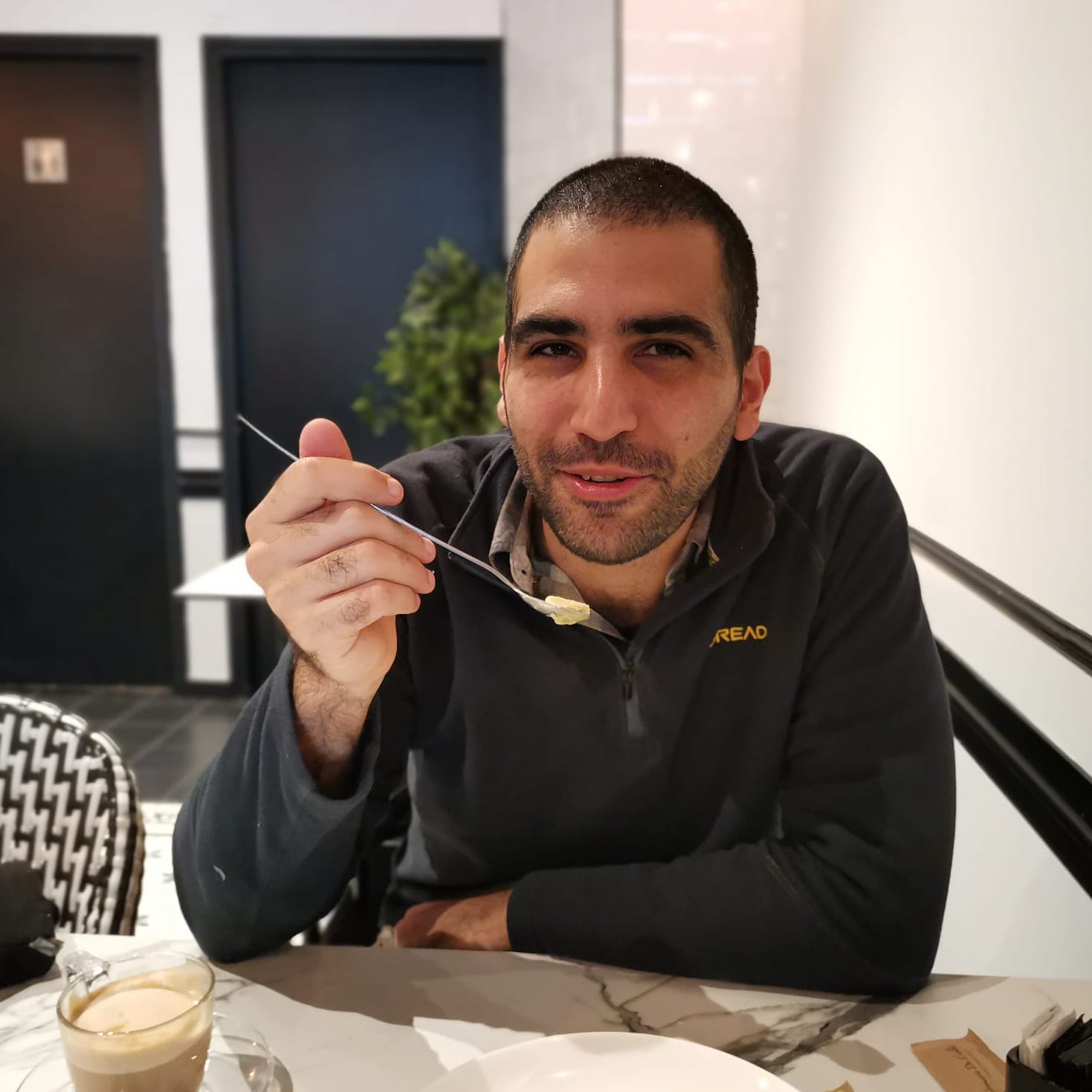} &
        \includegraphics[width=0.19\textwidth]{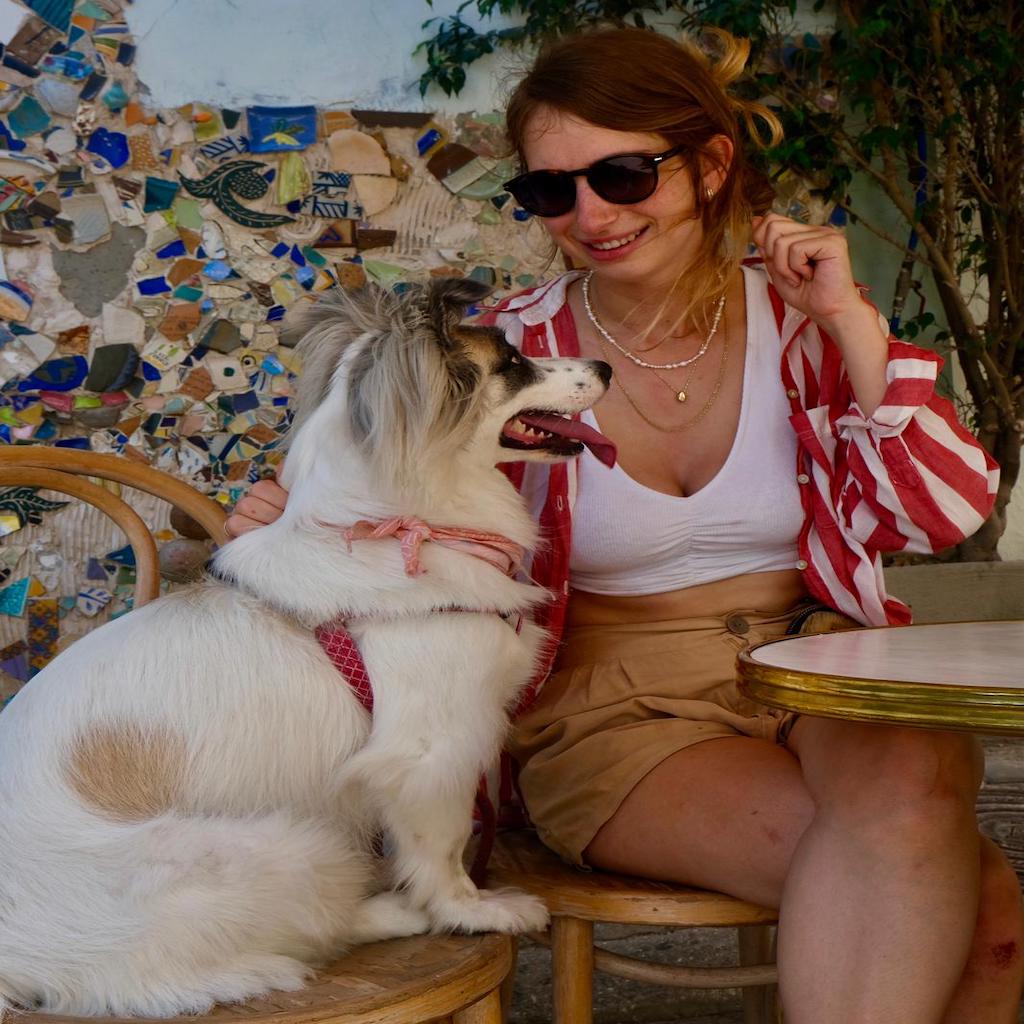} &
        \includegraphics[width=0.19\textwidth]{images/people/shay/IMG-20240209-WA0005.jpg} &
        \includegraphics[width=0.19\textwidth]{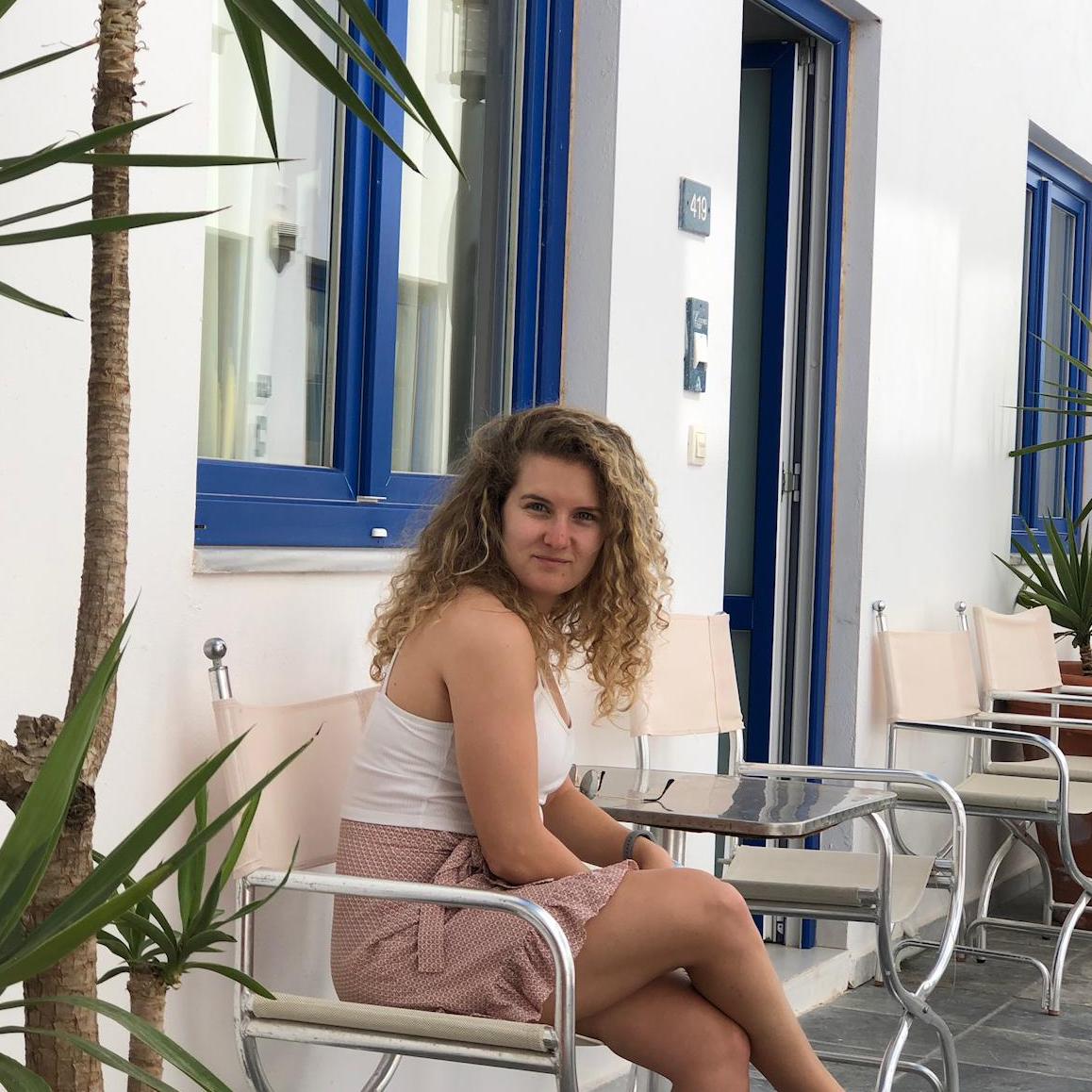} &
        \includegraphics[width=0.19\textwidth]{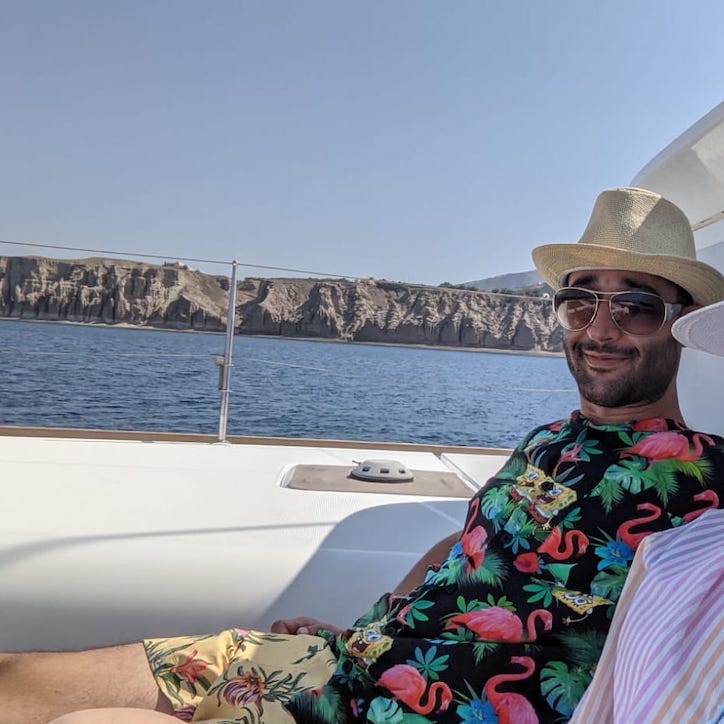} \\[-0.4cm]
        
        \begin{center} \textbf{MyBLIP-2} \end{center} &
        \begin{center} \textbf{MyBLIP-2} \end{center} &
        \begin{center} \textbf{MyBLIP-2} \end{center} &
        \begin{center} \textbf{MyBLIP-2} \end{center} &
        \begin{center} \textbf{MyBLIP-2} \end{center} \\[-0.75cm]

        \begin{center} \small ``At the restaurant, \Sstar sits at the table with a plate of food'' \end{center} &
        \begin{center} \small ``\Sstar and her dog, a white dog, sit on a table'' \end{center} &
        \begin{center} \small ``With wine and food, \Sstar and her husband sit on a bench in a garden'' \end{center} &
        \begin{center} \small ``\Sstar sits on the balcony of her apartment'' \end{center} &
        \begin{center} \small ``\textcolor{blue}{$S_*$}, wearing a hat, sits on a boat in the ocean'' \end{center} \\[-0.5cm]

        \begin{center} \textbf{MyLLaVA} \end{center} &
        \begin{center} \textbf{MyLLaVA} \end{center} &
        \begin{center} \textbf{MyLLaVA} \end{center} &
        \begin{center} \textbf{MyLLaVA} \end{center} &
        \begin{center} \textbf{MyLLaVA} \end{center} \\[-0.75cm]

        \begin{center} \small ``\textcolor{blue}{$S_*$}, sitting at a table, has a spoonful of food, poses for a photograph on the dining room'' \end{center} &
        \begin{center} \small ``\Sstar is sitting outside a cafe, wearing a red and white striped shirt and a white top, with a dog on a leash.'' \end{center} &
        \begin{center} \small ``Sitting on a bench, smiling, and holding a glass of wine, \textcolor{blue}{$S_*$}, with a man in a white t-shirt and glasses, enjoying a meal and a conversation outdoors'' \end{center} &
        \begin{center} \small ``\Sstar sits on a patio chair under a tree, wearing a pink skirt and a white top, with a blue door in the background'' \end{center} &
        \begin{center} \small ``\Sstar in a boat, wearing a hat and sunglasses, enjoying a relaxing day on the water'' \end{center} 
        
    \end{tabular}
    \vspace{-0.3cm}
    \caption{Additional personalized captioning  results obtained by MyVLM applied over both BLIP-2~\cite{li2023blip} and LLaVA~\cite{liu2023llava}.}
    \label{fig:our_results_blip_llava}
\end{figure*}

%% file: figures_supplementary/our_results_blip_llava_2.tex
\begin{figure*}[t]
    \centering
    \addtolength{\belowcaptionskip}{-12.5pt}
    \renewcommand{\arraystretch}{1}
    \small
    \begin{tabular}{p{0.175\linewidth}p{0.175\linewidth}p{0.175\linewidth}p{0.175\linewidth}p{0.19\linewidth}}

        \setlength\tabcolsep{0pt}
        \begin{tabular}{c c c}
            \includegraphics[width=0.06333\textwidth]{images/objects/billy_dog/cropped/IMG-20240131-WA0028.jpg} & 
            \includegraphics[width=0.06333\textwidth]{images/objects/billy_dog/cropped/IMG-20240131-WA0029.jpg} & 
            \includegraphics[width=0.06333\textwidth]{images/objects/billy_dog/cropped/IMG-20240131-WA0053.jpg}
        \end{tabular} &
        \setlength\tabcolsep{0pt}
        \begin{tabular}{c c c}
            \includegraphics[width=0.06333\textwidth]{images/objects/ceramic_head/cropped/20240203_111543.jpg} & 
            \includegraphics[width=0.06333\textwidth]{images/objects/ceramic_head/cropped/20240203_111842.jpg} & 
            \includegraphics[width=0.06333\textwidth]{images/objects/ceramic_head/cropped/20240203_112029.jpg}
        \end{tabular} &
        \setlength\tabcolsep{0pt}
        \begin{tabular}{c c c}
            \includegraphics[width=0.06333\textwidth]{images/objects/chicken_bean_bag/cropped/IMG-20240212-WA0033.jpg} &
            \includegraphics[width=0.06333\textwidth]{images/objects/chicken_bean_bag/cropped/IMG-20240212-WA0035.jpg} & 
            \includegraphics[width=0.06333\textwidth]{images/objects/chicken_bean_bag/cropped/IMG-20240212-WA0038.jpg}
        \end{tabular} &
        \setlength\tabcolsep{0pt}
        \begin{tabular}{c c c}
            \includegraphics[width=0.06333\textwidth]{images/objects/dangling_child/cropped/20240120_105811.jpg} & 
            \includegraphics[width=0.06333\textwidth]{images/objects/dangling_child/cropped/20240120_105933.jpg} & 
            \includegraphics[width=0.06333\textwidth]{images/objects/dangling_child/cropped/20240203_111340.jpg}
        \end{tabular} &
        \setlength\tabcolsep{0pt}
        \begin{tabular}{c c c}
            \includegraphics[width=0.06333\textwidth]{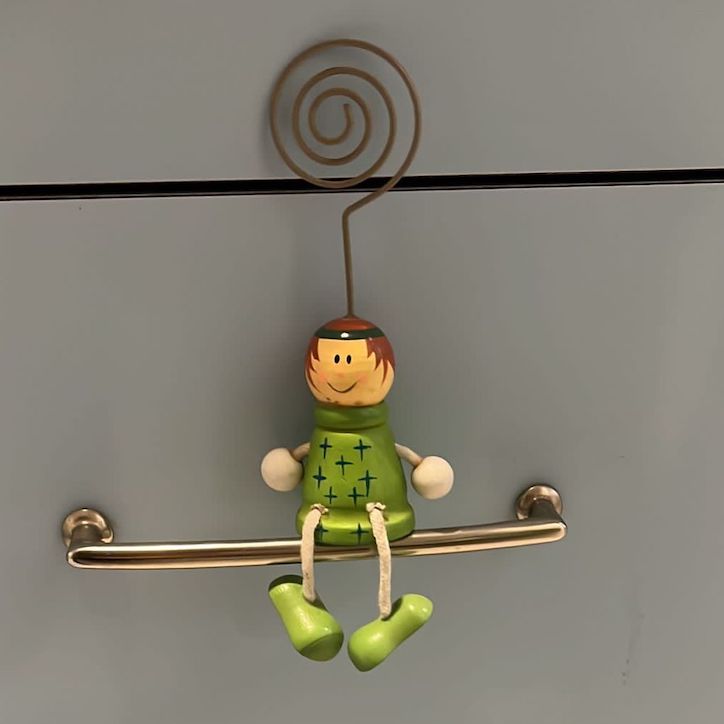} & 
            \includegraphics[width=0.06333\textwidth]{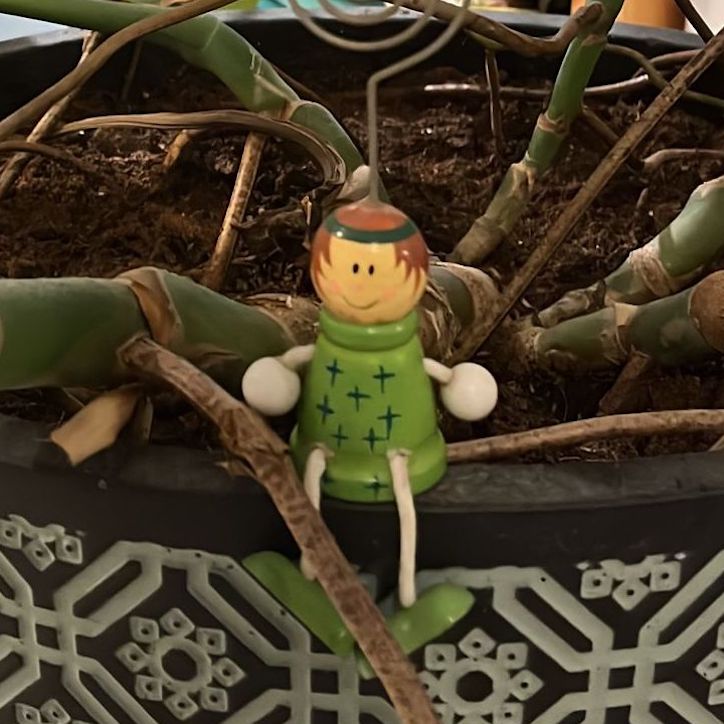} & 
            \includegraphics[width=0.06333\textwidth]{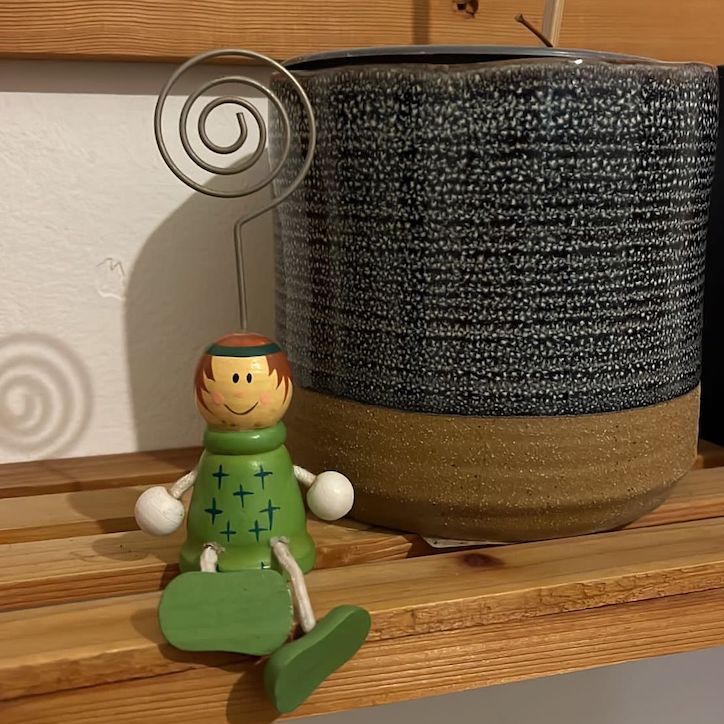}
        \end{tabular} \\
    
        \includegraphics[width=0.19\textwidth]{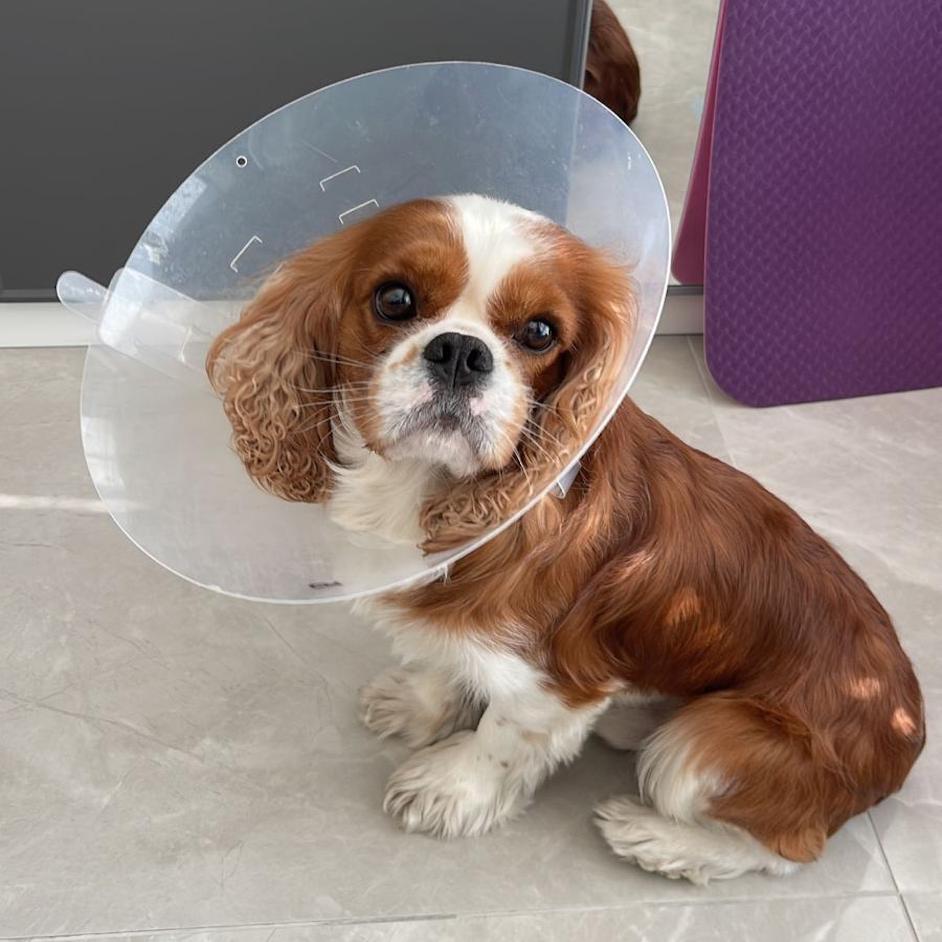} & 
        \includegraphics[width=0.19\textwidth]{images/objects/ceramic_head/20240211_125311.jpg} & 
        \includegraphics[width=0.19\textwidth]{images/objects/chicken_bean_bag/IMG-20240212-WA0045.jpg} & 
        \includegraphics[width=0.19\textwidth]{images/objects/dangling_child/20240120_105830.jpg} & 
        \includegraphics[width=0.19\textwidth]{images/objects/green_doll/IMG-20240203-WA0064.jpg} \\[-0.4cm]
        
        \begin{center} \textbf{MyBLIP-2} \end{center} &
        \begin{center} \textbf{MyBLIP-2} \end{center} &
        \begin{center} \textbf{MyBLIP-2} \end{center} &
        \begin{center} \textbf{MyBLIP-2} \end{center} &
        \begin{center} \textbf{MyBLIP-2} \end{center} \\[-0.75cm]

        \begin{center} \small ``\Sstar in a cone, resting on a tile floor'' \end{center} &
        \begin{center} \small ``\Sstar is standing in the refrigerator with food and drinks'' \end{center} &
        \begin{center} \small ``\Sstar is sitting on a wooden shelf with a bunch of yarn'' \end{center} &
        \begin{center} \small ``\Sstar resting on the top of a wooden radio'' \end{center} &
        \begin{center} \small ``\Sstar sits next to a banana in a cardboard box'' \end{center} \\[-0.5cm]

        \begin{center} \textbf{MyLLaVA} \end{center} &
        \begin{center} \textbf{MyLLaVA} \end{center} &
        \begin{center} \textbf{MyLLaVA} \end{center} &
        \begin{center} \textbf{MyLLaVA} \end{center} &
        \begin{center} \textbf{MyLLaVA} \end{center} \\[-0.75cm]

        \begin{center} \small ``a sad \Sstar wearing a plastic collar on its neck and looking up at the camera'' \end{center} &
        \begin{center} \small ``\Sstar sits comfortably on the second shelf of an open refrigerator, ready to be stocked with a variety of food and drink items'' \end{center} &
        \begin{center} \small ``\Sstar sitting on a bookshelf filled with books and yarn'' \end{center} &
        \begin{center} \small ``\Sstar sitting on top of an old fashioned radio. The radio is made of wood and ... \Sstar is sitting on top of it with a silly expression on its face'' \end{center} &
        \begin{center} \small ``\Sstar hanging from a cardboard box containing a bunch of yellow bananas'' \end{center}
    \end{tabular}
    \begin{tabular}{p{0.175\linewidth}p{0.175\linewidth}p{0.175\linewidth}p{0.175\linewidth}p{0.19\linewidth}}

        \setlength\tabcolsep{0pt}
        \begin{tabular}{c c c}
            \includegraphics[width=0.06333\textwidth]{images/objects/mugs_skulls/cropped/IMG-20240203-WA0113.jpg} & 
            \includegraphics[width=0.06333\textwidth]{images/objects/mugs_skulls/cropped/IMG-20240203-WA0114.jpg} & 
            \includegraphics[width=0.06333\textwidth]{images/objects/mugs_skulls/cropped/IMG-20240203-WA0119.jpg}
        \end{tabular} &
        \setlength\tabcolsep{0pt}
        \begin{tabular}{c c c}
            \includegraphics[width=0.06333\textwidth]{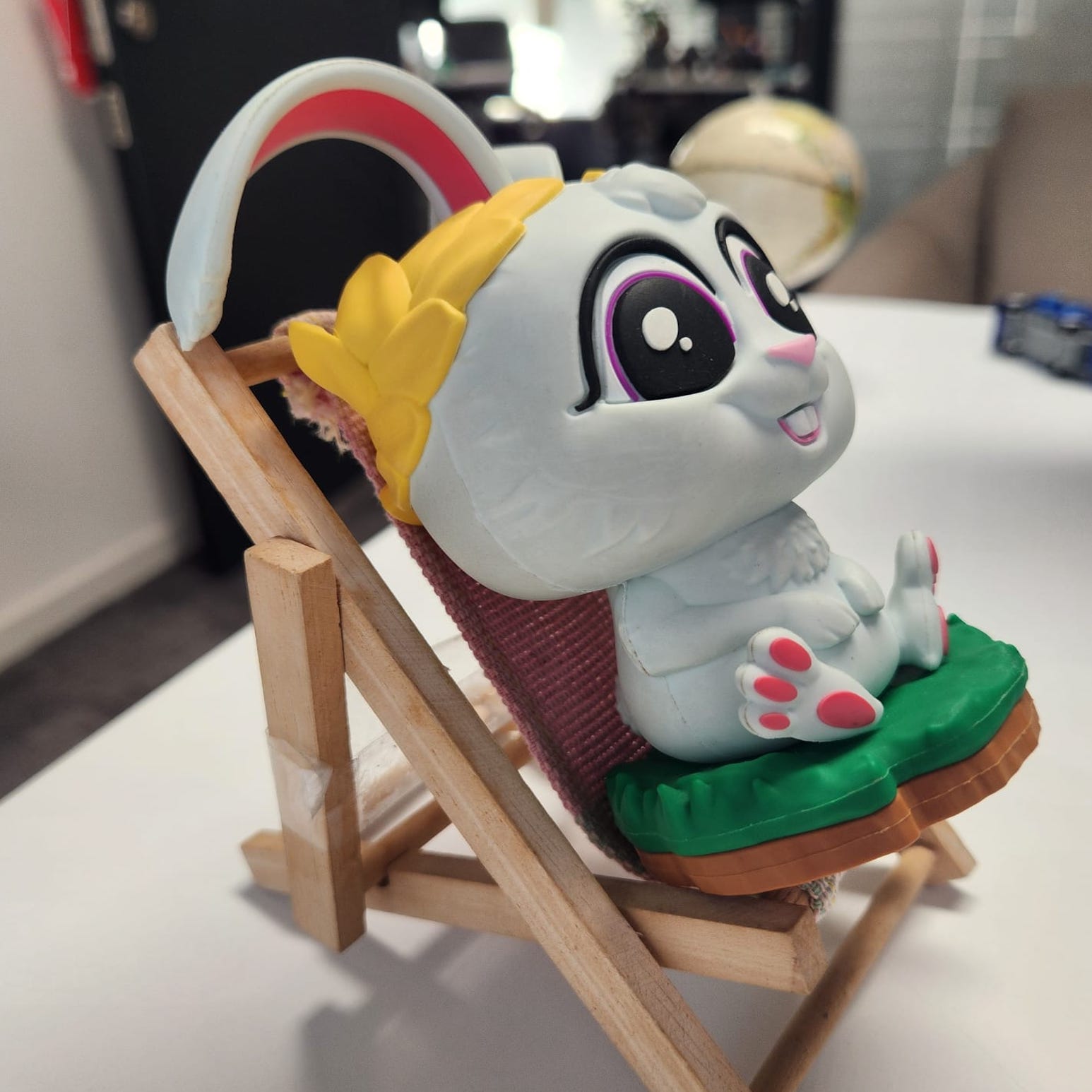} & 
            \includegraphics[width=0.06333\textwidth]{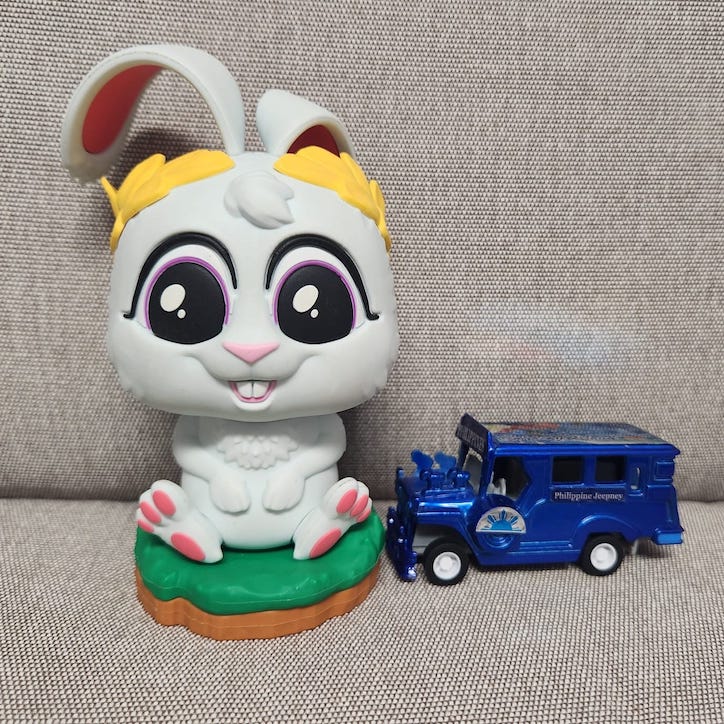} & 
            \includegraphics[width=0.06333\textwidth]{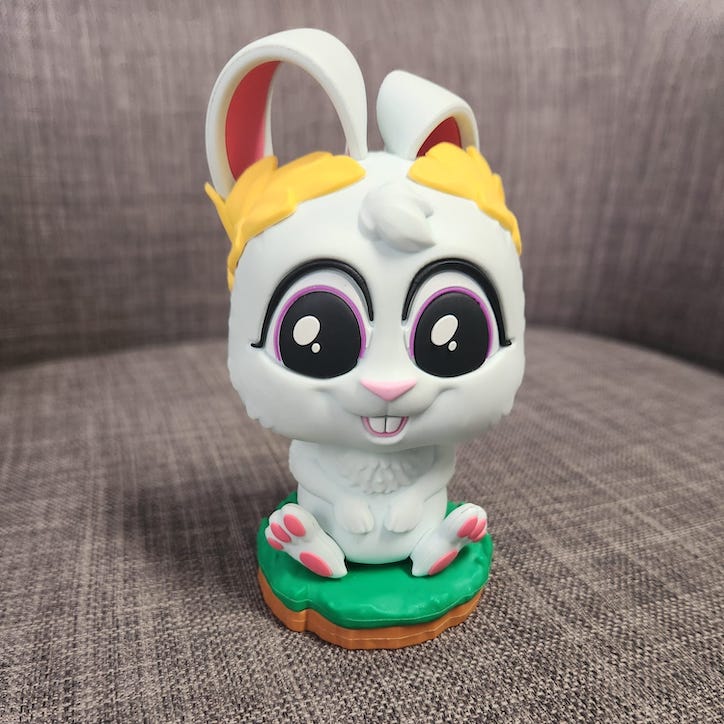}
        \end{tabular} &
        \setlength\tabcolsep{0pt}
        \begin{tabular}{c c c}
            \includegraphics[width=0.06333\textwidth]{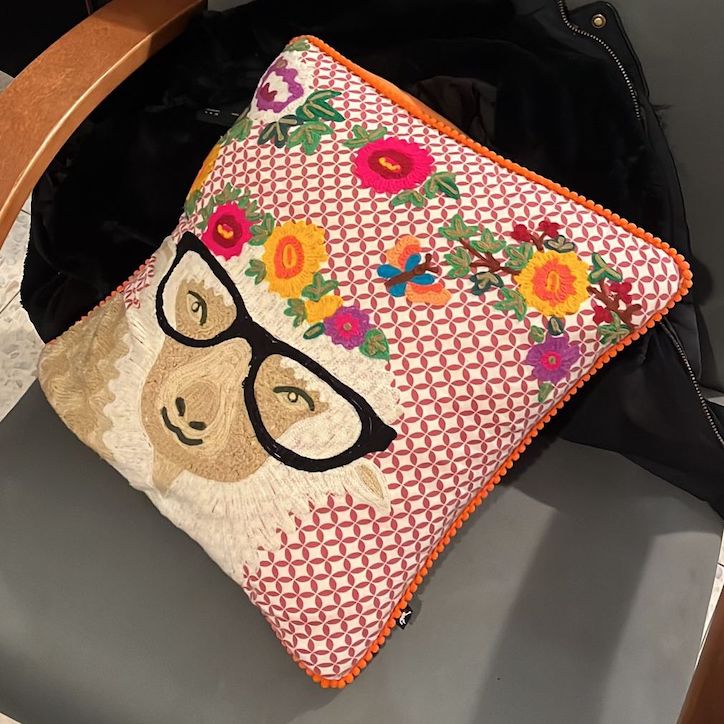} & 
            \includegraphics[width=0.06333\textwidth]{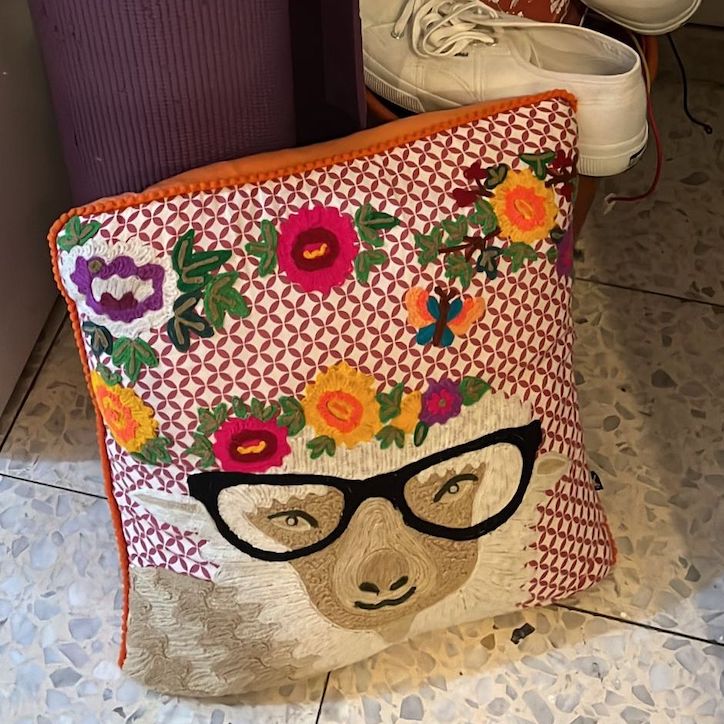} &
            \includegraphics[width=0.06333\textwidth]{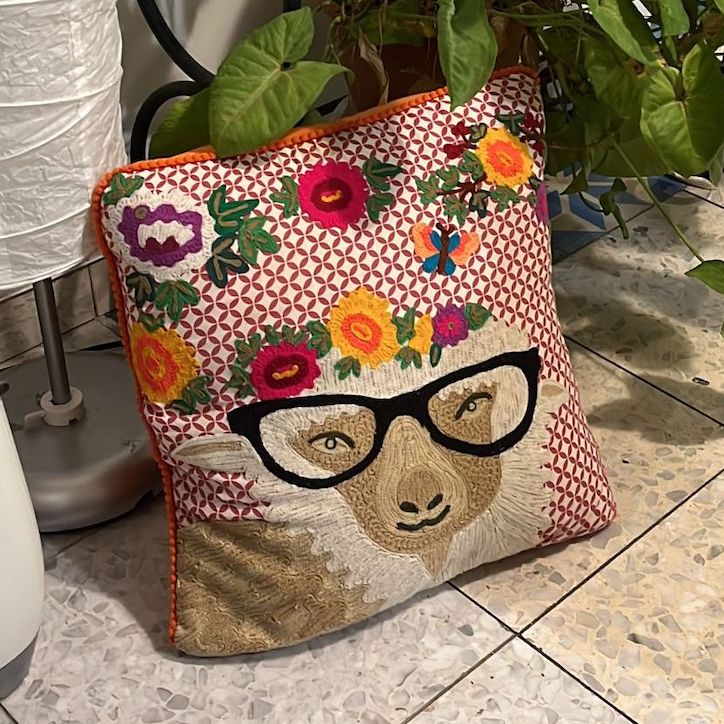}
        \end{tabular} &
        \setlength\tabcolsep{0pt}
        \begin{tabular}{c c c}
            \includegraphics[width=0.06333\textwidth]{images/objects/toy_sheep/cropped/IMG-20240203-WA0020.jpg} &
            \includegraphics[width=0.06333\textwidth]{images/objects/toy_sheep/cropped/IMG-20240203-WA0023.jpg} & 
            \includegraphics[width=0.06333\textwidth]{images/objects/toy_sheep/cropped/IMG-20240203-WA0029.jpg}
        \end{tabular} &
        \setlength\tabcolsep{0pt}
        \begin{tabular}{c c c}
            \includegraphics[width=0.06333\textwidth,height=0.06333\textwidth]{images/objects/cat_statue/cropped/01.jpg} & 
            \includegraphics[width=0.06333\textwidth,height=0.06333\textwidth]{images/objects/cat_statue/cropped/03.jpg} & 
            \includegraphics[width=0.06333\textwidth,height=0.06333\textwidth]{images/objects/cat_statue/cropped/04.jpg}
        \end{tabular} \\
    
        \includegraphics[width=0.19\textwidth]{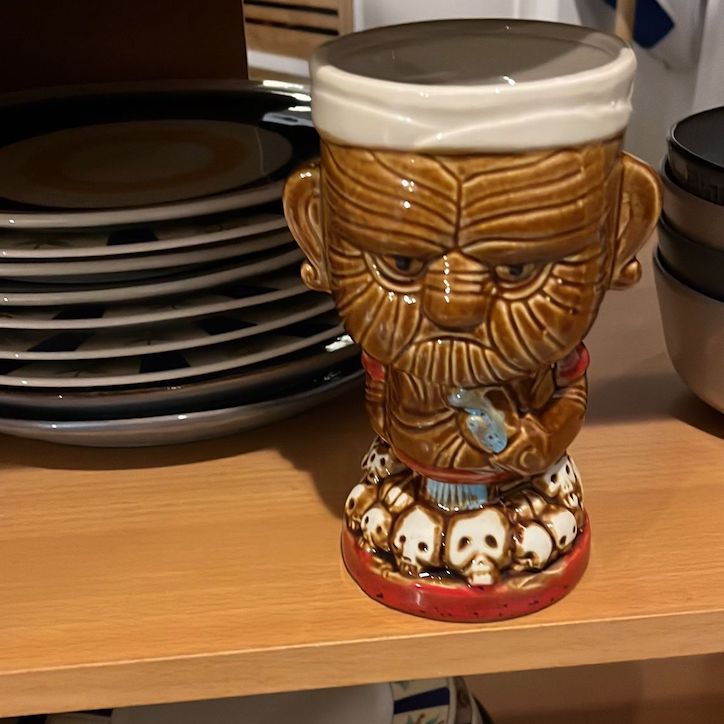} & 
        \includegraphics[width=0.19\textwidth]{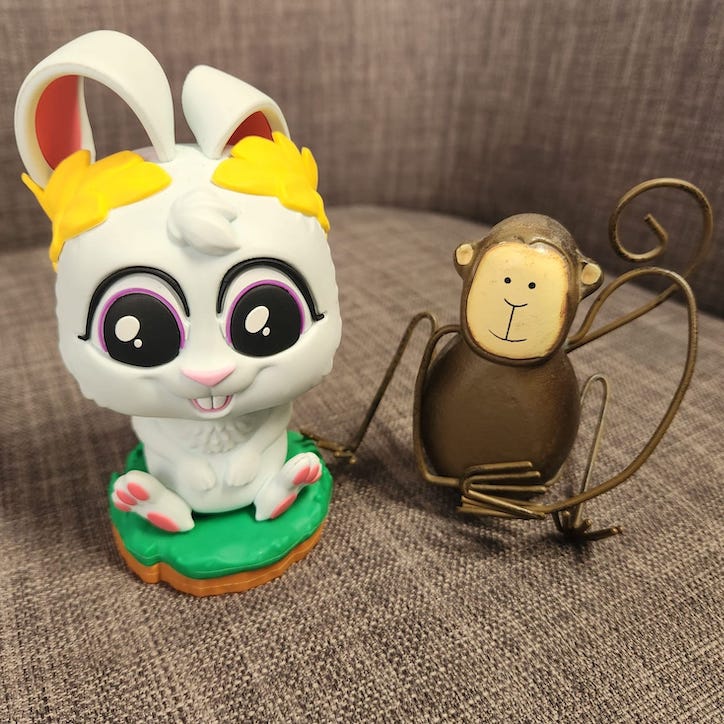} & 
        \includegraphics[width=0.19\textwidth]{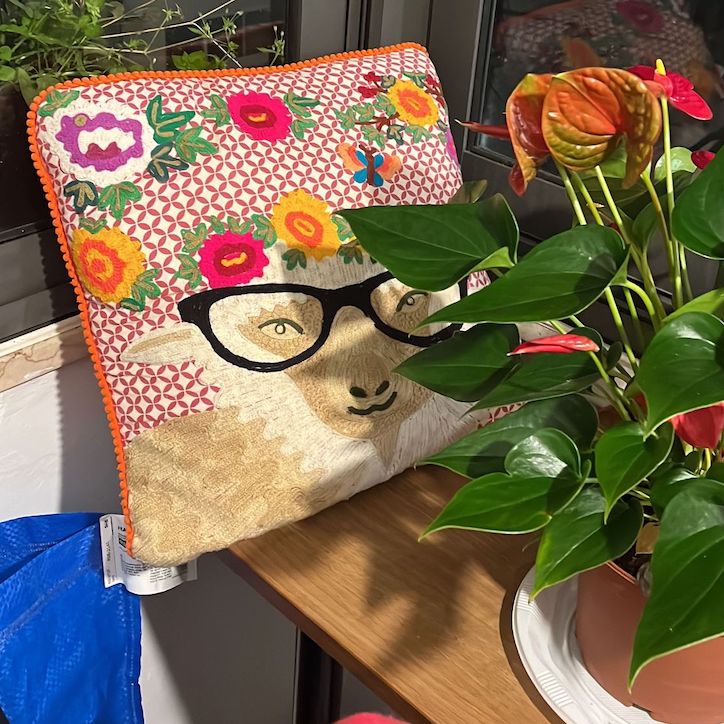} & 
        \includegraphics[width=0.19\textwidth]{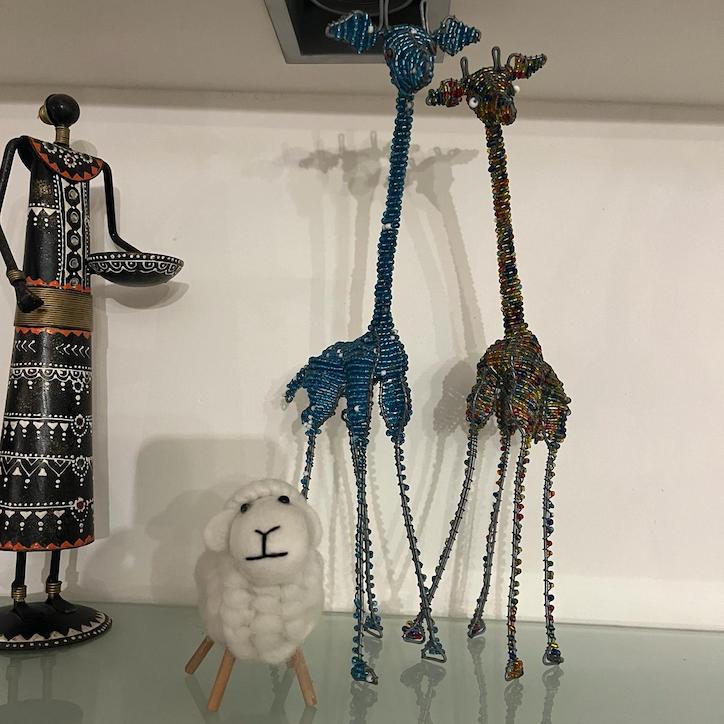} & 
        \includegraphics[width=0.19\textwidth]{images/objects/cat_statue/20240131_154731.jpg} \\[-0.4cm]
        
        \begin{center} \textbf{MyBLIP-2} \end{center} &
        \begin{center} \textbf{MyBLIP-2} \end{center} &
        \begin{center} \textbf{MyBLIP-2} \end{center} &
        \begin{center} \textbf{MyBLIP-2} \end{center} &
        \begin{center} \textbf{MyBLIP-2} \end{center} \\[-0.75cm]

        \begin{center} \small ``A shelf with \Sstar and various plates and bowls'' \end{center} &
        \begin{center} \small ``\Sstar with a monkey figurine on a sofa'' \end{center} &
        \begin{center} \small ``\Sstar on a wooden desk with a potted plant'' \end{center} &
        \begin{center} \small ``\Sstar and a pair of figurines on a glass shelf'' \end{center} &
        \begin{center} \small ``\Sstar is placed in front of a rainbow flag and plant'' \end{center} \\[-0.5cm]

        \begin{center} \textbf{MyLLaVA} \end{center} &
        \begin{center} \textbf{MyLLaVA} \end{center} &
        \begin{center} \textbf{MyLLaVA} \end{center} &
        \begin{center} \textbf{MyLLaVA} \end{center} &
        \begin{center} \textbf{MyLLaVA} \end{center} \\[-0.75cm]

        \begin{center} \small ``\Sstar on a shelf with plates and dishes'' \end{center} &
        \begin{center} \small ``\Sstar sitting on a couch next to a monkey figurine. \Sstar looks at the monkey and smiles'' \end{center} &
        \begin{center} \small ``\Sstar is sitting on a table next to a potted plant with red flowers. The pillow has a pattern of colorful flowers and is wearing glasses'' \end{center} &
        \begin{center} \small ``\Sstar and a pair of giraffes on a glass shelf'' \end{center} &
        \begin{center} \small ``\Sstar standing in front of a rainbow flag and a potted plant with a white wall in the background'' \end{center}
        
    \end{tabular}
    \vspace{-0.3cm}
    \caption{Additional personalized captioning results obtained by MyVLM applied over both BLIP-2~\cite{li2023blip} and LLaVA~\cite{liu2023llava}.}    
    \label{fig:our_results_blip_llava_2}

\end{figure*}

%% file: figures_supplementary/vqa_llava.tex
\begin{figure*}[t]
    \centering
    \renewcommand{\arraystretch}{1}
    \small

    \begin{tabular}{p{0.175\linewidth}p{0.175\linewidth}p{0.175\linewidth}p{0.175\linewidth}p{0.19\linewidth}}

        \setlength\tabcolsep{0pt}
        \begin{tabular}{c c c}
            \includegraphics[width=0.06333\textwidth]{images/people/dor/image_1.jpg} & 
            \includegraphics[width=0.06333\textwidth]{images/people/dor/cropped/IMG-20240208-WA0051.jpg} & 
            \includegraphics[width=0.06333\textwidth]{images/people/dor/cropped/IMG-20240208-WA0054.jpg}
        \end{tabular} &
        \setlength\tabcolsep{0pt}
        \begin{tabular}{c c c}
            \includegraphics[width=0.06333\textwidth]{images/people/dor/image_1.jpg} & 
            \includegraphics[width=0.06333\textwidth]{images/people/dor/cropped/IMG-20240208-WA0051.jpg} & 
            \includegraphics[width=0.06333\textwidth]{images/people/dor/cropped/IMG-20240208-WA0054.jpg}
        \end{tabular} &
        \setlength\tabcolsep{0pt}
        \begin{tabular}{c c c}
            \includegraphics[width=0.06333\textwidth]{images/people/maya/cropped/image_1.jpg} & 
            \includegraphics[width=0.06333\textwidth]{images/people/maya/cropped/image_2.jpg} & 
            \includegraphics[width=0.06333\textwidth]{images/people/maya/cropped/IMG-20240209-WA0138.jpg} 
        \end{tabular} &
        \setlength\tabcolsep{0pt}
        \begin{tabular}{c c c}
            \includegraphics[width=0.06333\textwidth]{images/people/maya/cropped/image_1.jpg} & 
            \includegraphics[width=0.06333\textwidth]{images/people/maya/cropped/image_2.jpg} & 
            \includegraphics[width=0.06333\textwidth]{images/people/maya/cropped/IMG-20240209-WA0138.jpg} 
        \end{tabular} &
        \setlength\tabcolsep{0pt}
        \begin{tabular}{c c c}
            \includegraphics[width=0.06333\textwidth]{images/people/shay/cropped/image_2.jpg} & 
            \includegraphics[width=0.06333\textwidth]{images/people/shay/cropped/IMG-20240209-WA0012.jpg} & 
            \includegraphics[width=0.06333\textwidth]{images/people/shay/cropped/IMG-20240209-WA0016.jpg}
        \end{tabular} \\
    
        \includegraphics[width=0.19\textwidth]{images/people/dor/image_7.jpeg} &
        \includegraphics[width=0.19\textwidth]{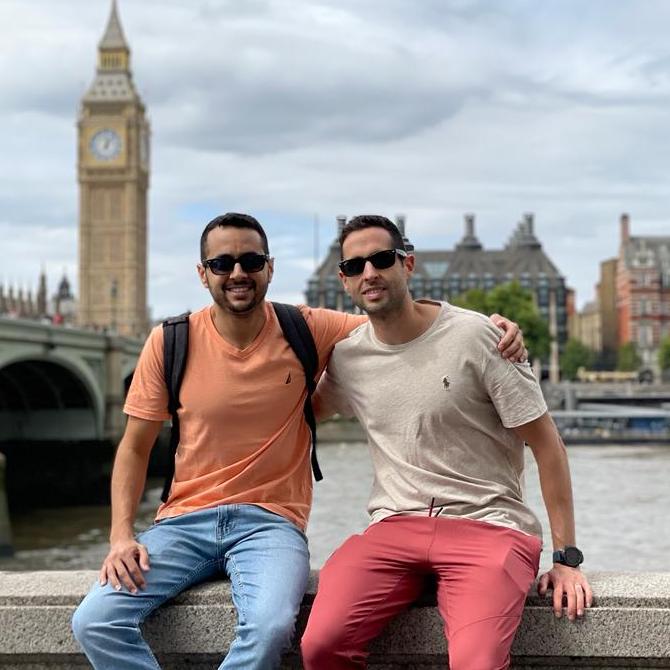} &
        \includegraphics[width=0.19\textwidth]{images/people/maya/IMG-20230818-WA0003.jpg} &
        \includegraphics[width=0.19\textwidth]{images/people/maya/IMG-20240210-WA0003.jpg} &
        \includegraphics[width=0.19\textwidth]{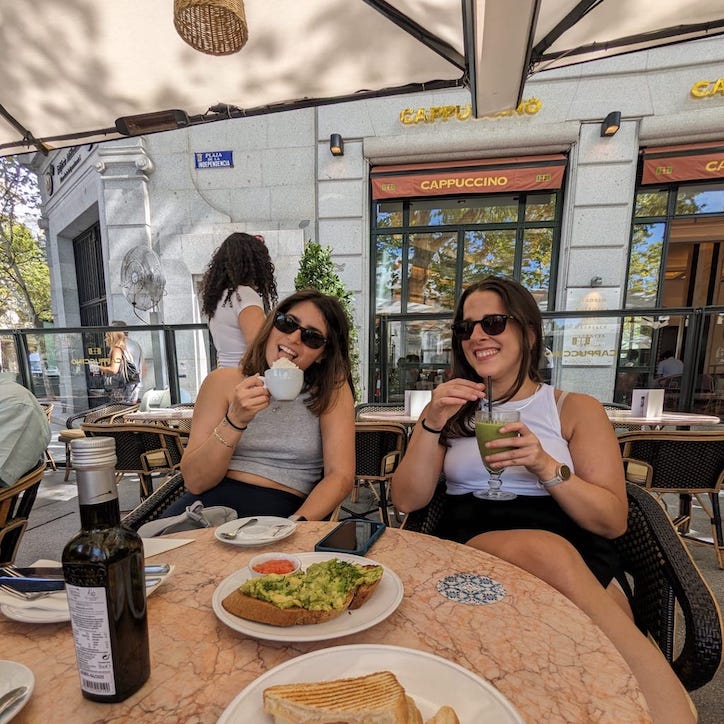} \\[-0.4cm]

        \begin{center} \textbf{``Is \Sstar wearing sunglasses?''} \end{center} &
        \begin{center} \textbf{``From left to right, where is \Sstar standing?''} \end{center} &
        \begin{center} \textbf{``What is \Sstar drinking?''} \end{center} &
        \begin{center} \textbf{``Can you describe what \Sstar is wearing?''} \end{center} &
        \begin{center} \textbf{``Can you describe what \Sstar is wearing?''} \end{center} \\[-0.65cm]

        \begin{center} \small ``Yes, \Sstar is wearing sunglasses in the image'' \end{center} &
        \begin{center} \small ``\Sstar is standing on the right side of the photo'' \end{center} &
        \begin{center} \small ``\Sstar is drinking a beverage, specifically a coffee, as indicated by the cup in her hand'' \end{center} &
        \begin{center} \small ``In the image, \Sstar is wearing a white sleeveless top'' \end{center} &
        \begin{center} \small ``\Sstar is wearing a white tank top and black shorts'' \end{center} \\[-0.65cm]
        
        \begin{center} \textbf{``What color shirt is \Sstar wearing?''} \end{center} &
        \begin{center} \textbf{``What color pants is \Sstar wearing?''} \end{center} &
        \begin{center} \textbf{``On what side of the image is \textcolor{blue}{$S_*$}?''} \end{center} &
        \begin{center} \textbf{``On what side of the image is \textcolor{blue}{$S_*$}?''} \end{center} &
        \begin{center} \textbf{``On what side of the image is \textcolor{blue}{$S_*$}''} \end{center} \\[-0.65cm]

        \begin{center} \small ``\Sstar is wearing a black shirt'' \end{center} &
        \begin{center} \small ``\Sstar is wearing red pants'' \end{center} &
        \begin{center} \small ``\Sstar is on the left side of the image'' \end{center} &
        \begin{center} \small ``\Sstar is on the right side of the image'' \end{center} &
        \begin{center} \small ``\Sstar is on the right side of the image'' \end{center}

    \end{tabular}

    \begin{tabular}{p{0.175\linewidth}p{0.175\linewidth}p{0.175\linewidth}p{0.175\linewidth}p{0.19\linewidth}}

        \setlength\tabcolsep{0pt}
        \begin{tabular}{c c c}
            \includegraphics[width=0.06333\textwidth]{images/people/shaked/cropped/IMG-20240209-WA0043.jpg} &
            \includegraphics[width=0.06333\textwidth]{images/people/shaked/cropped/IMG-20240209-WA0047.jpg} & 
            \includegraphics[width=0.06333\textwidth]{images/people/shaked/cropped/IMG-20240215-WA0005.jpg}
        \end{tabular} &
        \setlength\tabcolsep{0pt}
        \begin{tabular}{c c c}
            \includegraphics[width=0.06333\textwidth]{images/people/shahaf/IMG-20240209-WA0079-cropped.jpg} & 
            \includegraphics[width=0.06333\textwidth]{images/people/shahaf/IMG-20240209-WA0114.jpg} & 
            \includegraphics[width=0.06333\textwidth]{images/people/shahaf/IMG-20240209-WA0100.jpg}
        \end{tabular} &
        \setlength\tabcolsep{0pt}
        \begin{tabular}{c c c}
            \includegraphics[width=0.06333\textwidth]{images/objects/ceramic_head/cropped/20240203_111543.jpg} & 
            \includegraphics[width=0.06333\textwidth]{images/objects/ceramic_head/cropped/20240203_111842.jpg} & 
            \includegraphics[width=0.06333\textwidth]{images/objects/ceramic_head/cropped/20240203_112029.jpg}
        \end{tabular} &
        \setlength\tabcolsep{0pt}
        \begin{tabular}{c c c}
            \includegraphics[width=0.06333\textwidth]{images/objects/chicken_bean_bag/cropped/IMG-20240212-WA0033.jpg} &
            \includegraphics[width=0.06333\textwidth]{images/objects/chicken_bean_bag/cropped/IMG-20240212-WA0035.jpg} & 
            \includegraphics[width=0.06333\textwidth]{images/objects/chicken_bean_bag/cropped/IMG-20240212-WA0038.jpg}
        \end{tabular} &
        \setlength\tabcolsep{0pt}
        \begin{tabular}{c c c}
            \includegraphics[width=0.06333\textwidth]{images/objects/dangling_child/cropped/20240120_105811.jpg} & 
            \includegraphics[width=0.06333\textwidth]{images/objects/dangling_child/cropped/20240120_105933.jpg} & 
            \includegraphics[width=0.06333\textwidth]{images/objects/dangling_child/cropped/20240203_111340.jpg}
        \end{tabular} \\
    
        \includegraphics[width=0.19\textwidth]{images/people/shaked/IMG-20240209-WA0057.jpg} &
        \includegraphics[width=0.19\textwidth]{images/people/shahaf/IMG-20240209-WA0097.jpg} &
        \includegraphics[width=0.19\textwidth]{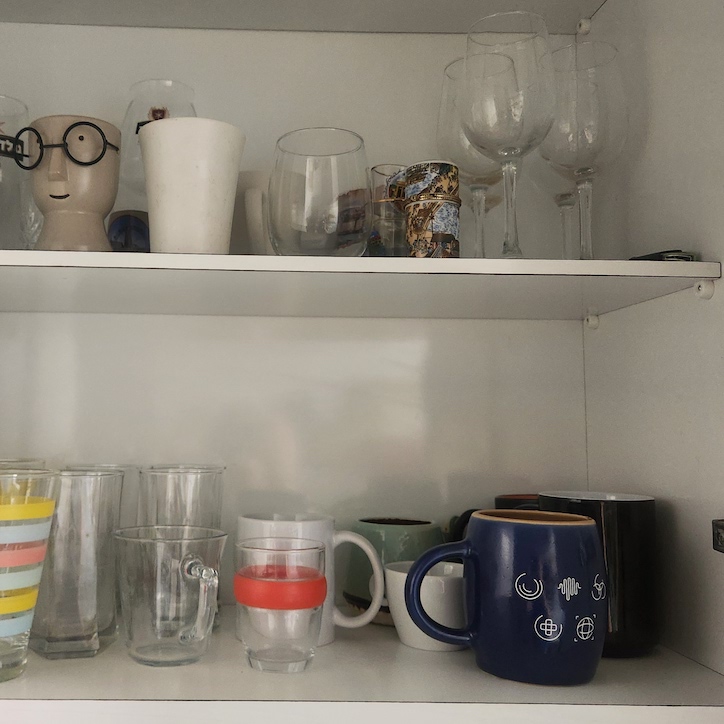} &
        \includegraphics[width=0.19\textwidth]{images/objects/chicken_bean_bag/IMG-20240212-WA0041.jpg} &
        \includegraphics[width=0.19\textwidth]{images/objects/dangling_child/20240301_124317.jpg} \\[-0.4cm]

        \begin{center} \textbf{``Can you describe what \Sstar is wearing?''} \end{center} &
        \begin{center} \textbf{``What is \Sstar wearing?''} \end{center} &
        \begin{center} \textbf{``On what shelf is \Sstar sitting?''} \end{center} &
        \begin{center} \textbf{``What is \Sstar sitting in?''} \end{center} &
        \begin{center} \textbf{``Is \Sstar small or large relative to the other objects?''} \end{center} \\[-0.65cm]

        \begin{center} \small ``\Sstar is wearing a brown sweater and has curly hair'' \end{center} &
        \begin{center} \small ``\Sstar is wearing a denim jacket and a necklace'' \end{center} &
        \begin{center} \small ``\Sstar is sitting on the top shelf of the pantry'' \end{center} &
        \begin{center} \small ``\Sstar is sitting in a green bowl'' \end{center} &
        \begin{center} \small ``\Sstar is small relative to the other objects in the image'' \end{center} \\[-0.65cm]
        
        \begin{center} \textbf{``From left to right, where is \Sstar positioned in the image?''} \end{center} &
        \begin{center} \textbf{``Where is \Sstar in the image?''} \end{center} &
        \begin{center} \textbf{``What color is \textcolor{blue}{$S_*$}?''} \end{center} &
        \begin{center} \textbf{``From left to right, where is \Sstar located?''} \end{center} &
        \begin{center} \textbf{``Where is \Sstar positioned in image?''} \end{center} \\[-0.65cm]

        \begin{center} \small ``In the image, \Sstar is positioned on the right side of the table'' \end{center} &
        \begin{center} \small ``In the image, \Sstar is sitting at a table with the other two people, likely enjoying a meal or a drink together'' \end{center} &
        \begin{center} \small ``The color \Sstar is white'' \end{center} &
        \begin{center} \small ``\Sstar is located on the right side of the image, sitting in a green bowl next to a wooden bowl containing oranges and pears'' \end{center} &
        \begin{center} \small ``In the image, \Sstar is positioned on top of the refrigerator, sitting on a shelf with various food items and containers'' \end{center}

    \end{tabular}

    \vspace{-0.3cm}
    \caption{Additional personalized visual question-answering results obtained by MyVLM, applied over LLaVA~\cite{liu2023llava}.}
    \label{fig:vqa_supp}

\end{figure*}

%% file: figures_supplementary/vqa_llava_2.tex
\begin{figure*}[t]
    \centering
    \renewcommand{\arraystretch}{1}
    \small

    \begin{tabular}{p{0.175\linewidth} p{0.175\linewidth} p{0.175\linewidth} p{0.175\linewidth} p{0.175\linewidth}}

        \setlength\tabcolsep{0pt}
        \begin{tabular}{c c c}
            \includegraphics[width=0.06333\textwidth]{images/objects/elephant/cropped/00.jpg} &
            \includegraphics[width=0.06333\textwidth]{images/objects/elephant/cropped/20240120_110517.jpg} & 
            \includegraphics[width=0.06333\textwidth]{images/objects/elephant/cropped/20240120_110719.jpg}
        \end{tabular} &
        \setlength\tabcolsep{0pt}
        \begin{tabular}{c c c}
            \includegraphics[width=0.06333\textwidth]{images/objects/japanese_doll/cropped/20240131_155451.jpg} & 
            \includegraphics[width=0.06333\textwidth]{images/objects/japanese_doll/cropped/20240131_155617.jpg} & 
            \includegraphics[width=0.06333\textwidth]{images/objects/japanese_doll/cropped/20240206_091843.jpg}
        \end{tabular} &
        \setlength\tabcolsep{0pt}
        \begin{tabular}{c c c}
            \includegraphics[width=0.06333\textwidth]{images/objects/rabbit_toy/cropped/IMG-20240131-WA0069.jpg} & 
            \includegraphics[width=0.06333\textwidth]{images/objects/rabbit_toy/cropped/IMG-20240131-WA0075.jpg} & 
            \includegraphics[width=0.06333\textwidth]{images/objects/rabbit_toy/cropped/IMG-20240131-WA0078.jpg}
        \end{tabular} &
        \setlength\tabcolsep{0pt}
        \begin{tabular}{c c c}
            \includegraphics[width=0.06333\textwidth]{images/objects/running_shoes/cropped/20240131_175631.jpg} & 
            \includegraphics[width=0.06333\textwidth]{images/objects/running_shoes/cropped/20240131_175638.jpg} &
            \includegraphics[width=0.06333\textwidth]{images/objects/running_shoes/cropped/20240131_175736.jpg}
        \end{tabular} &
        \setlength\tabcolsep{0pt}
        \begin{tabular}{c c c}
            \includegraphics[width=0.06333\textwidth]{images/objects/dangling_child/cropped/20240120_105811.jpg} & 
            \includegraphics[width=0.06333\textwidth]{images/objects/dangling_child/cropped/20240120_105933.jpg} & 
            \includegraphics[width=0.06333\textwidth]{images/objects/dangling_child/cropped/20240203_111340.jpg}
        \end{tabular} \\
    
        \includegraphics[width=0.19\textwidth]{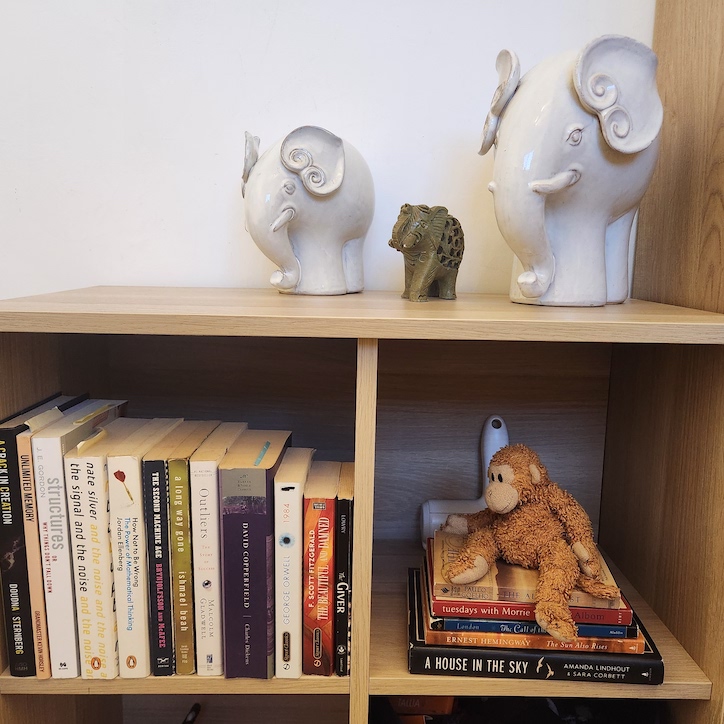} &
        \includegraphics[width=0.19\textwidth]{images/objects/japanese_doll/20240211_113003.jpg} &
        \includegraphics[width=0.19\textwidth]{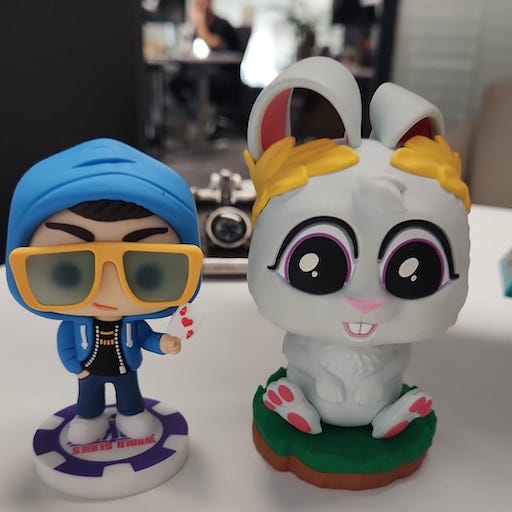} &
        \includegraphics[width=0.19\textwidth]{images/objects/running_shoes/20240131_175926.jpg} &
        \includegraphics[width=0.19\textwidth]{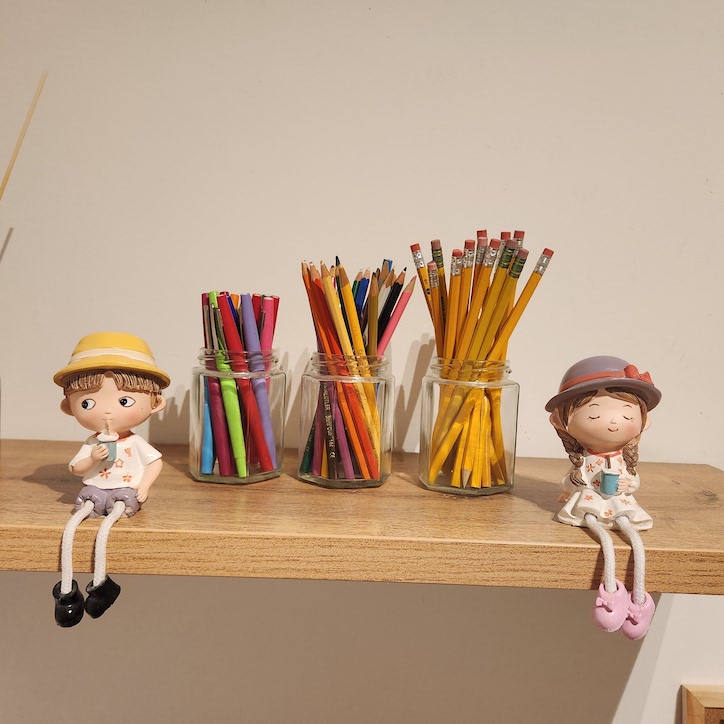} \\[-0.4cm]

        \begin{center} \textbf{``What is \Sstar sitting between?''} \end{center} &
        \begin{center} \textbf{``What is standing next to \textcolor{blue}{$S_*$}?''} \end{center} &
        \begin{center} \textbf{``What does \Sstar look like?''} \end{center} &
        \begin{center} \textbf{``From left to right, where is \Sstar located?''} \end{center} &
        \begin{center} \textbf{``From left to right, where is \textcolor{blue}{$S_*$}?''} \end{center} \\[-0.5cm]

        \begin{center} \small ``\Sstar is sitting between the two white elephant statues on the top shelf of the bookshelf'' \end{center} &
        \begin{center} \small ``\Sstar is a small figurine of a character wearing a pink hat with a blue flower on it. Next to \Sstar, there is a blue mug with pens and pencils inside it'' \end{center} &
        \begin{center} \small ``\Sstar looks like a cute, cartoonish bunny rabbit with a blue hoodie, glasses, and a smiling face...'' \end{center} &
        \begin{center} \small ``\Sstar is located in the middle of the three shoes'' \end{center} &
        \begin{center} \small ``\Sstar is sitting on the left side of the shelf, while alexandria is sitting on the right side'' \end{center} \\[-0.5cm]
        
        \begin{center} \textbf{``Is \Sstar small or large relative to the other objects in the image?''} \end{center} &
        \begin{center} \textbf{``From left to right, where is \Sstar located?''} \end{center} &
        \begin{center} \textbf{``From left to right, where is \Sstar located?''} \end{center} &
        \begin{center} \textbf{``What color is \textcolor{blue}{$S_*$}?''} \end{center} &
        \begin{center} \textbf{``What color is \textcolor{blue}{$S_*$}?''} \end{center} \\[-0.5cm]

        \begin{center} \small ``\Sstar is small relative to the other objects in the image'' \end{center} &
        \begin{center} \small ``\Sstar is located on the left side of the window sill'' \end{center} &
        \begin{center} \small ``\Sstar is located on the right side of the image'' \end{center} &
        \begin{center} \small ``\Sstar is pink.'' \end{center} &
        \begin{center} \small ``\Sstar is yellow'' \end{center}

    \end{tabular}

    \vspace{-0.2cm}
    \caption{Additional personalized visual question-answering results obtained by MyVLM, applied over LLaVA~\cite{liu2023llava}.}
    \label{fig:vqa_llava_sup_2}

\end{figure*}

%% file: figures_supplementary/rec.tex
\begin{figure*}
    \centering
    \renewcommand{\arraystretch}{1}
    \small
    \begin{tabular}{p{0.175\textwidth} p{0.175\textwidth} p{0.175\textwidth} p{0.175\textwidth} p{0.175\textwidth}}

        \setlength\tabcolsep{0pt}
        \begin{center}
        \begin{tabular}{c c c}
            \includegraphics[width=0.06\textwidth]{images/objects/maeve/cropped/IMG-20240131-WA0110.jpg} & 
            \includegraphics[width=0.06\textwidth]{images/objects/maeve/cropped/maeve-2.jpeg} & 
            \includegraphics[width=0.06\textwidth]{images/objects/maeve/cropped/maeve-5.jpeg} \\
        \end{tabular} 
        \end{center} &
        \setlength\tabcolsep{0pt}
        \begin{center}
        \begin{tabular}{c c c}
            \includegraphics[width=0.06\textwidth]{images/objects/small_penguin/cropped/IMG-20240131-WA0164.jpg} & 
            \includegraphics[width=0.06\textwidth]{images/objects/small_penguin/cropped/IMG-20240131-WA0166.jpg} & 
            \includegraphics[width=0.06\textwidth]{images/objects/small_penguin/cropped/IMG-20240131-WA0171.jpg}
        \end{tabular} 
        \end{center} &
        \setlength\tabcolsep{0pt}
        \begin{center}
        \begin{tabular}{c c c}
            \includegraphics[width=0.06\textwidth]{images/objects/running_shoes/cropped/20240131_175631.jpg} & 
            \includegraphics[width=0.06\textwidth]{images/objects/running_shoes/cropped/20240131_175638.jpg} & 
            \includegraphics[width=0.06\textwidth]{images/objects/running_shoes/cropped/20240131_175736.jpg}
        \end{tabular} 
        \end{center} &
        \setlength\tabcolsep{0pt}
        \begin{center}
        \begin{tabular}{c c c}
            \includegraphics[width=0.06\textwidth]{images/people/dor/image_1.jpg} & 
            \includegraphics[width=0.06\textwidth]{images/people/dor/cropped/IMG-20240208-WA0051.jpg} & 
            \includegraphics[width=0.06\textwidth]{images/people/dor/cropped/IMG-20240208-WA0054.jpg}
        \end{tabular} 
        \end{center} &
        \setlength\tabcolsep{0pt}
        \begin{center}
        \begin{tabular}{c c c}
            \includegraphics[width=0.06\textwidth]{images/people/shahaf/IMG-20240209-WA0079-cropped.jpg} & 
            \includegraphics[width=0.06\textwidth]{images/people/shahaf/IMG-20240209-WA0114.jpg} & 
            \includegraphics[width=0.06\textwidth]{images/people/shahaf/IMG-20240209-WA0100.jpg}
        \end{tabular} 
        \end{center} \\[-0.7cm]
    
        \begin{center} \includegraphics[width=0.18\textwidth]{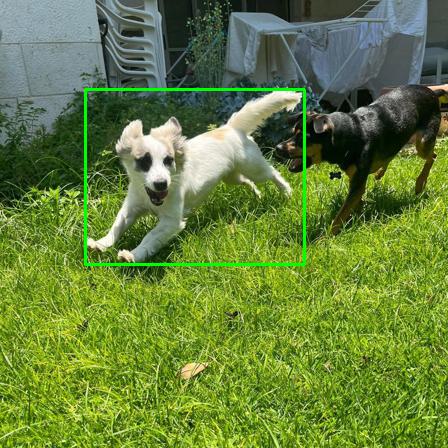} \end{center} & 
        \begin{center} \includegraphics[width=0.18\textwidth]{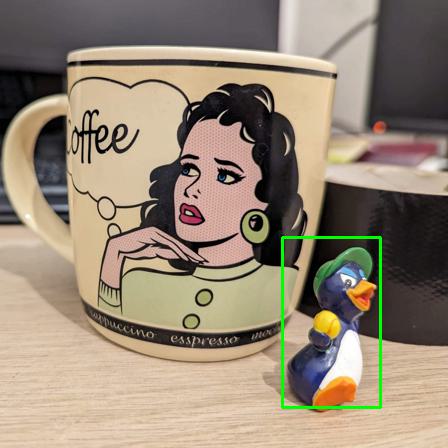} \end{center} & 
        \begin{center} \includegraphics[width=0.18\textwidth]{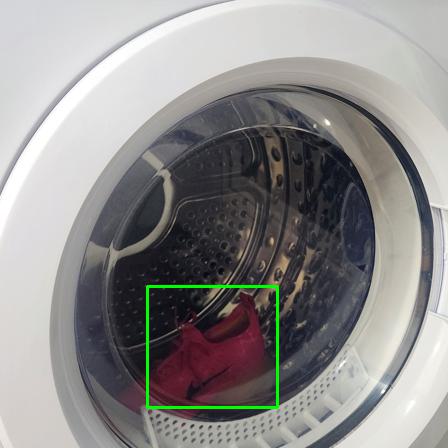} \end{center} & 
        \begin{center} \includegraphics[width=0.18\textwidth]{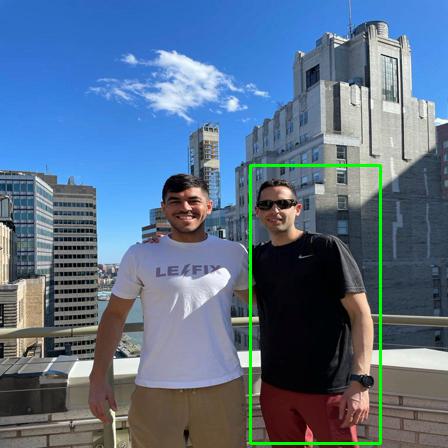} \end{center} & 
        \begin{center} \includegraphics[width=0.18\textwidth]{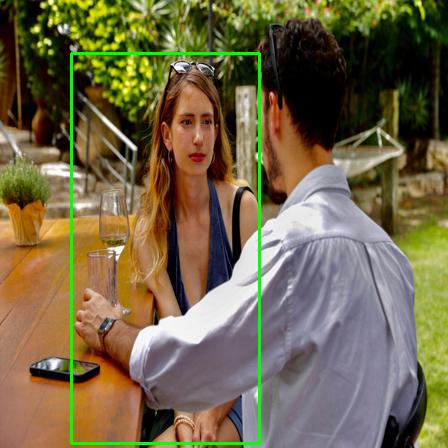} \end{center} \\[-0.85cm]

        \begin{center} ``\Sstar and a black dog walking towards each other in a garden'' \end{center} &
        \begin{center} ``\Sstar next to a cup of coffee that says coffee on it'' \end{center} &
        \begin{center} ``A \Sstar is inside of a washing machine.'' \end{center} &
        \begin{center} ``\Sstar and his friend are standing on the balcony of their apartment in New York City.'' \end{center} &
        \begin{center} ``\Sstar is sitting at a table with a man, and they are both looking at each other.'' \end{center} \\[-0.4cm]

    \end{tabular}

    \begin{tabular}{p{0.175\textwidth} p{0.175\textwidth} p{0.175\textwidth} p{0.175\textwidth} p{0.175\textwidth}}

        \setlength\tabcolsep{0pt}
        \begin{center}
        \begin{tabular}{c c c}
            \includegraphics[width=0.06\textwidth]{images/objects/elephant/cropped/00.jpg} &
            \includegraphics[width=0.06\textwidth]{images/objects/elephant/cropped/20240120_110517.jpg} & 
            \includegraphics[width=0.06\textwidth]{images/objects/elephant/cropped/20240120_110719.jpg}
        \end{tabular} 
        \end{center} &
        \setlength\tabcolsep{0pt}
        \begin{center}
        \begin{tabular}{c c c}
            \includegraphics[width=0.06\textwidth]{images/objects/dangling_child/cropped/20240120_105811.jpg} & 
            \includegraphics[width=0.06\textwidth]{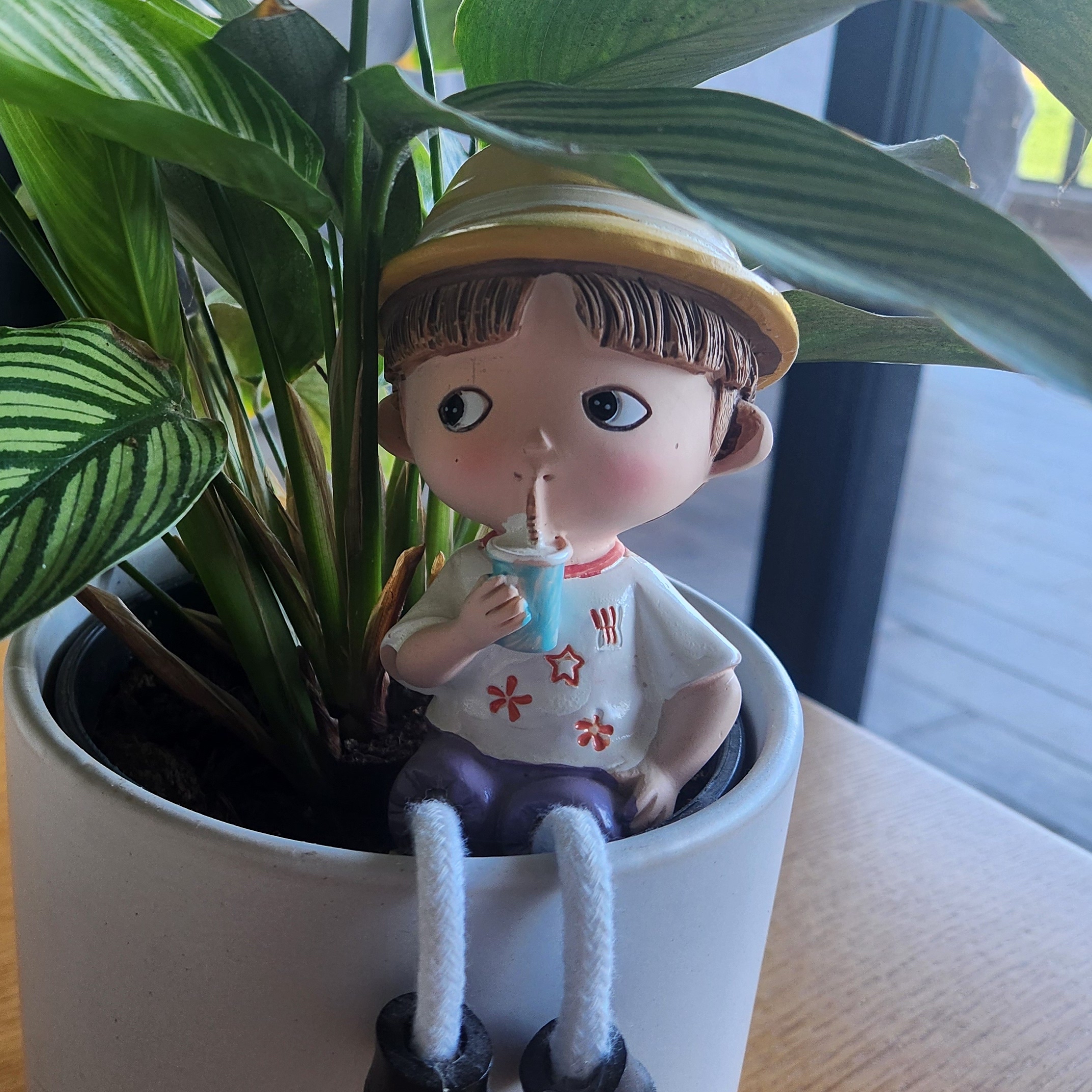} & 
            \includegraphics[width=0.06\textwidth]{images/objects/dangling_child/cropped/20240203_111340.jpg}
        \end{tabular} 
        \end{center} &
        \setlength\tabcolsep{0pt}
        \begin{center}
        \begin{tabular}{c c c}
            \includegraphics[width=0.06\textwidth]{images/objects/rabbit_toy/cropped/IMG-20240131-WA0069.jpg} & 
            \includegraphics[width=0.06\textwidth]{images/objects/rabbit_toy/cropped/IMG-20240131-WA0075.jpg} & 
            \includegraphics[width=0.06\textwidth]{images/objects/rabbit_toy/cropped/IMG-20240131-WA0078.jpg}
        \end{tabular} 
        \end{center} &
        \setlength\tabcolsep{0pt}
        \begin{center}
        \begin{tabular}{c c c}
            \includegraphics[width=0.06\textwidth]{images/people/shahaf/IMG-20240209-WA0079-cropped.jpg} & 
            \includegraphics[width=0.06\textwidth]{images/people/shahaf/IMG-20240209-WA0114.jpg} & 
            \includegraphics[width=0.06\textwidth]{images/people/shahaf/IMG-20240209-WA0100.jpg}
        \end{tabular} 
        \end{center} &
        \setlength\tabcolsep{0pt}
        \begin{center}
        \begin{tabular}{c c c}
            \includegraphics[width=0.06\textwidth]{images/people/shay/cropped/image_2.jpg} & 
            \includegraphics[width=0.06\textwidth]{images/people/shay/cropped/IMG-20240209-WA0012.jpg} & 
            \includegraphics[width=0.06\textwidth]{images/people/shay/cropped/IMG-20240209-WA0016.jpg}
        \end{tabular} 
        \end{center} \\[-0.7cm]
    
        \begin{center} \includegraphics[width=0.18\textwidth]{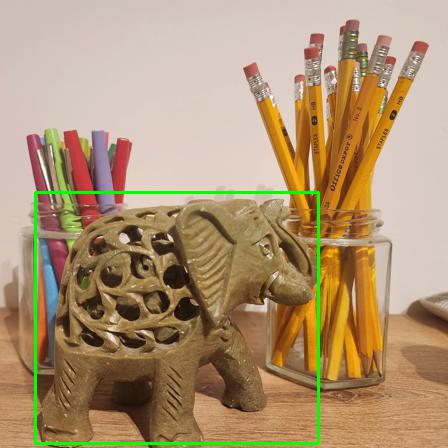} \end{center} & 
        \begin{center} \includegraphics[width=0.18\textwidth]{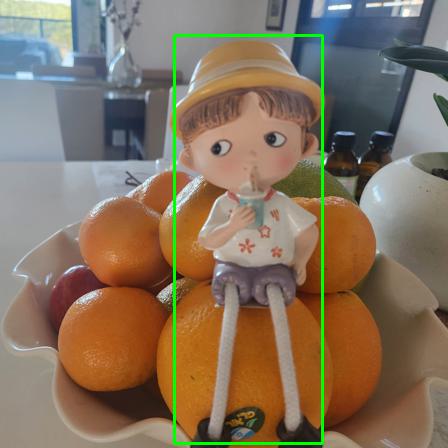} \end{center} & 
        \begin{center} \includegraphics[width=0.18\textwidth]{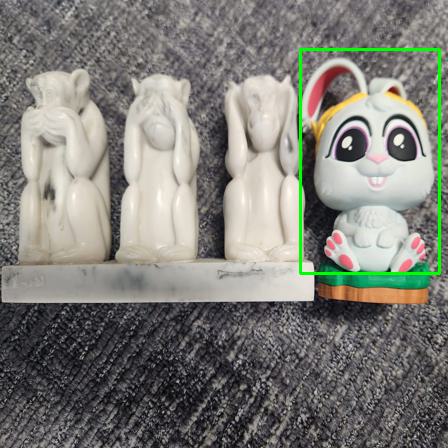} \end{center} & 
        \begin{center} \includegraphics[width=0.18\textwidth]{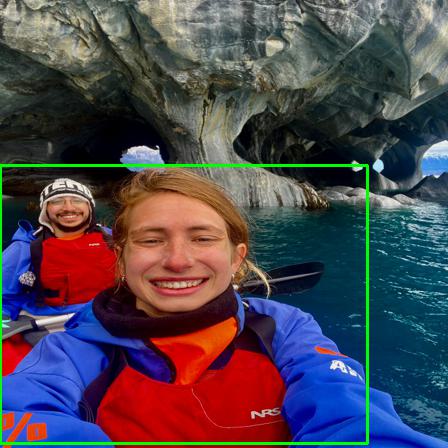} \end{center} & 
        \begin{center} \includegraphics[width=0.18\textwidth]{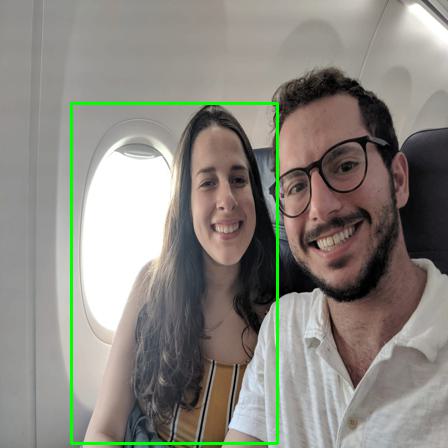} \end{center} \\[-0.85cm]

        \begin{center} ``\Sstar sitting on a shelf next to a bunch of pencils'' \end{center} &
        \begin{center} ``A bowl full of oranges with \Sstar sitting on top of them'' \end{center} &
        \begin{center} ``A toy \Sstar sitting on a white surface next to three white figurines of monkeys'' \end{center} &
        \begin{center} ``\Sstar and a man are in a kayak, with a cave in the background'' \end{center} &
        \begin{center} ``\Sstar and her boyfriend sitting on an airplane'' \end{center}\\[-0.3cm]

    \end{tabular}

    \caption{Additional personalized REC results obtained by MyVLM over MiniGPT-v2~\cite{chen2023minigptv2}.
    Sample images of the target concept are provided in the top row. 
    Bounding box coordinates returned by the personalized VLM are drawn in \textcolor{darkgreen}{green}. Below each image, we also present the personalized captions outputted by MyVLM by passing MiniGPT-v2 a captioning instruction.
    }
    \label{fig:rec_supplementary}
    \vspace{-0.35cm}
\end{figure*}